\documentclass[twoside,11pt]{article}

\usepackage{jmlr2e}

\usepackage[cmex10]{amsmath}
\usepackage{amsfonts}
\usepackage{amssymb}
\usepackage{mathtools}
\usepackage[utf8]{inputenc}
\usepackage[english]{babel}
\usepackage{array}
\usepackage{mdwmath}
\usepackage{mdwtab}
\usepackage[caption=false,font=footnotesize]{subfig}
\usepackage{floatrow}
\usepackage{multirow}
\usepackage{stfloats}
\usepackage{url}
\usepackage{wrapfig}
\usepackage{natbib}
\usepackage{bm}

\newcommand{\hypothesis}{\ensuremath{\mathcal{H}}}
\newcommand{\dataset}{\ensuremath{\mathcal{D}}}

\newcommand{\fbgeneral}{\ensuremath{\text{FB}_8}}
\newcommand{\fb}{\ensuremath{\text{FB}_5}}
\newcommand{\trans}{\ensuremath{^\mathsf{T}}}

\newcommand{\fisher}{\ensuremath{\mathcal{F}}(\boldsymbol{\Theta})}
\newcommand{\fisherone}{\ensuremath{\mathcal{F}}_1(\boldsymbol{\Theta})}
\newcommand{\fisherij}{\ensuremath{\mathcal{F}}}

\newcommand{\boldx}{\ensuremath{\mathbf{x}}}
\newcommand{\boldxt}{\ensuremath{\mathbf{x}^\mathsf{T}}}
\newcommand{\boldtheta}{\ensuremath{\boldsymbol{\Theta}}}
\newcommand{\boldphi}{\ensuremath{\boldsymbol{\Phi}}}

\newcommand{\mean}{\ensuremath{\pmb{\gamma}_1}}
\newcommand{\major}{\ensuremath{\pmb{\gamma}_2}}
\newcommand{\minor}{\ensuremath{\pmb{\gamma}_3}}
\newcommand{\tildemean}{\ensuremath{\widetilde{\pmb{\gamma}}_1}}
\newcommand{\tildemajor}{\ensuremath{\widetilde{\pmb{\gamma}}_2}}
\newcommand{\tildeminor}{\ensuremath{\widetilde{\pmb{\gamma}}_3}}
\newcommand{\meanj}{\ensuremath{\pmb{\gamma}_{1j}}}
\newcommand{\majorj}{\ensuremath{\pmb{\gamma}_{2j}}}
\newcommand{\minorj}{\ensuremath{\pmb{\gamma}_{3j}}}

\newcommand{\meana}{\ensuremath{\pmb{\gamma}_{a1}}}
\newcommand{\majora}{\ensuremath{\pmb{\gamma}_{a2}}}
\newcommand{\minora}{\ensuremath{\pmb{\gamma}_{a3}}}
\newcommand{\meanb}{\ensuremath{\pmb{\gamma}_{b1}}}
\newcommand{\majorb}{\ensuremath{\pmb{\gamma}_{b2}}}
\newcommand{\minorb}{\ensuremath{\pmb{\gamma}_{b3}}}
\newcommand{\gammai}{\ensuremath{\pmb{\gamma}_i}}
\newcommand{\gammam}{\ensuremath{\pmb{\gamma}_m}}

\newcommand{\expect}{\ensuremath{\mathbb{E}}}
\newcommand{\fancym}{\ensuremath{\mathcal{M}}}

\ShortHeadings{MML inference of Kent distributions}{Kasarapu}
\firstpageno{1}

\begin{document}

\title{Modelling of directional data using Kent distributions}

\author{\name Parthan Kasarapu \email parthan.kasarapu@monash.edu\\
        \addr Faculty of Information Technology\\
             Monash University\\
             VIC 3800, Australia}

\editor{}

\maketitle

\begin{abstract}%
The modelling of data 
on a spherical surface requires the consideration of directional
probability distributions.
To model asymmetrically distributed data on a three-dimensional
sphere, Kent distributions are often used.
The moment estimates of the parameters are typically used 
in modelling tasks involving Kent distributions.
However, these 
lack a rigorous statistical treatment.
The focus of the paper is to introduce a Bayesian estimation of the
parameters of the Kent distribution 
which has not been carried out in the literature, partly because of its
complex mathematical form.
We employ the Bayesian information-theoretic paradigm of Minimum 
Message Length (MML) to bridge this gap and derive reliable estimators.
The inferred parameters are subsequently used in mixture modelling of 
Kent distributions. 
The problem of inferring the suitable number of mixture
components is also addressed using the MML criterion.
We demonstrate the superior performance of the derived MML-based parameter 
estimates against the traditional estimators.
We apply the MML principle to infer mixtures of Kent distributions
to model empirical data corresponding to protein conformations.
We demonstrate the effectiveness of Kent models to act as improved
descriptors of protein structural data as compared to commonly used von Mises-Fisher distributions.
\end{abstract}


\section{Introduction} \label{sec:introduction}
Directional statistics is a growing discipline with 
widespread applications in earth sciences, meteorology, 
physics, biology, and other areas. 
A sample of directional data corresponds to a collection of unit vectors. 
The modelling of directional data has been explored  
using several types of distributions 
described on surfaces of compact manifolds, such as spheres 
and tori \citep{fisher1953dispersion,fisher1993statistical,mardia-book}. 
The most popular amongst these distributions is the von Mises-Fisher (vMF) 
distribution \citep{watson1956construction}.
Its probability density function $f$ at any point $\boldx$ 
on a unit \emph{three-dimesional} 
sphere has the form:
\begin{equation*}
f(\boldx;\boldtheta) \propto \exp\{\kappa \mean\trans\boldx\}
\end{equation*}
where $\propto$ denotes proportionality,
$\boldtheta$ is the parameter vector comprising of
the unit mean vector $\mean$ 
and the \emph{concentration} parameter $\kappa \ge 0$.
The vMF distribution is analogous to a \emph{symmetric}
Gaussian distribution, wrapped around a unit sphere.
As such, it is useful for modelling directional data that is symmetrically
distributed with respect to a mean direction.
The modelling of asymmetrically distributed directional data, however,
requires distributions which generalize the vMF distribution.
A generalization of vMF is called the Fisher-Bingham distribution 
\citep{mardia1975statistics} which takes the form:
\begin{equation}
f(\boldx;\boldtheta) \propto \exp\{\kappa \mean\trans\boldx\ 
                        + \beta_2 (\major\trans\boldx)^2
                        + \beta_3 (\minor\trans\boldx)^2\} 
\label{eqn:fb_density}
\end{equation}
where the parameters $\mean,\major,\minor$ are unit vectors with $\major$ and $\minor$ being orthogonal
to each other, the parameters  
$\beta_2$ and $\beta_3$ are real values with $\beta_2 \ge \beta_3$.
As the distribution is characterized using an 8 real valued parameter vector $\boldtheta$
(2 for $\mean$, 3 for $\major$ and $\minor$, and 3 scalars $\kappa,\beta_2,\beta_3$), 
it is also referred to as the \fbgeneral~distribution. Notice that
compared to the vMF, the \fbgeneral~distribution 
has an exponential factor with additional quadratic terms.

Given more free parameters, the \fbgeneral~distribution is a better choice compared to the vMF
distribution in modelling real world three-dimensional directional data 
where symmetry cannot be assumed.
However, the use of the \fbgeneral~distribution in directional statistics 
poses difficulties owing to its complex mathematical form 
and also because of a lack of a natural understanding of its parameters
\citep{kent1982fisher}. 
In order to achieve a balance between the highly simplified vMF model
and the complex \fbgeneral~distribution,
\citet{kent1982fisher} suggested an alternative form that is relatively easy to work with
and whose parameters have natural interpretations.
This distribution, referred to as the \emph{Kent} distribution,
is obtained from Equation~\ref{eqn:fb_density} by assuming $\mean,\major,\minor$
form an orthogonal system of vectors and are subject to the constraint $\beta_2 = -\beta_3 = \beta$.
The probability density function is then given by
\begin{equation}
f(\boldx;\boldtheta) = c(\kappa,\beta)^{-1} \exp\{\kappa \mean^\mathsf{T}\boldx 
+ \beta[(\major^\mathsf{T}\boldx)^2 - (\minor^\mathsf{T}\boldx)^2]\}
\label{eqn:kent_pdf}
\end{equation}
where $\mean,\major,\minor$ are orthogonal unit vectors representing the 
\emph{mean}, \emph{major}, and \emph{minor} axes respectively;
$\kappa$, as before, measures the concentration, and $0\le\beta<\kappa/2$
describes the \emph{ovalness}.

The Kent distribution was proposed as a spherical analogue of
the \emph{general} Gaussian distribution and serves as a natural extension to the
vMF distribution.
The distribution has \emph{ellipse-like
contours} of constant probability density on the spherical surface. 
\citet{kent1982fisher} argued that by imposing the constraints $\beta < \kappa/2$
and $\mean,\major,\minor$ to be an orthogonal system, the distribution
would be unimodal and have a behaviour similar to the Gaussian distribution but on a spherical surface.

As the Kent distribution is characterized using a 5 real valued parameter vector $\boldtheta$
(3 for $\mean,\major,\minor$ because they are orthogonal and unit vectors, 2 for 
the scalar entities $\kappa,\beta$), it is popularly
referred to as the \fb~distribution.
We will denote the 5-parameter Fisher-Bingham distribution as 
$\fb(\mathbf{Q},\kappa,\beta)$,
where $\mathbf{Q}=(\mean,\major,\minor)$ is a $3\times 3$ orthogonal matrix.
The normalization constant $c(\kappa,\beta)$ of the distribution is derived as an infinite series
\begin{equation}
c(\kappa,\beta) = 2\pi \sum_{j=0}^{\infty} \frac{\Gamma(j+\frac{1}{2})}{\Gamma(j+1)} \beta^{2j} \left(\frac{2}{\kappa}\right)^{2j+\frac{1}{2}} I_{2j+\frac{1}{2}}(\kappa)
\label{eqn:norm_constant}
\end{equation}
that depends on the Gamma function $\Gamma$ and the modified Bessel function
$I_v$ of the first kind and order $v$ \citep{abramowitz1972handbook,kent1982fisher}. 

The importance of vMF and \fb~distributions in mixture modelling
tasks has been well established: vMF mixtures have been used in
large-scale text clustering \citep{Banerjee:generative-clustering,gopal2014mises}, 
clustering of protein dihedral angles 
\citep{dowe1996circular,mardia2007protein}, and gene expression 
analyses \citep{Banerjee:clustering-hypersphere}.
Mixtures of \fb~distributions
have been employed by \citet{peel2001fitting} to identify joint sets in rock masses, and by
\citet{hamelryck2006sampling} to sample random protein conformations.
The \fb~distribution has increasingly
found support in machine learning tasks in structural bioinformatics
\citep{kent2005using,boomsma2006graphical,hamelryck2009probabilistic}.

The analysis of data using \fb~distributions requires estimating the 
corresponding parameters. Due to the complex mathematical form of the
density function, these estimates are approximated. \citet{kent1982fisher}
derived the \emph{moment} estimates and suggested limiting case approximations.
However, the use of simplified approximations can have considerable effects
from a practical standpoint. To overcome this, we explore Bayesian
estimation using the minimum message length (MML) principle
as it results in reliable estimators as shown by the experiments in
Section~\ref{sec:exp_single_kent}.

The parameter inference of a statistical distribution is typically done by maximum likelihood (ML)
or Bayesian maximum \emph{a posteriori} probability (MAP) estimation. 
Bayesian inference using MML differs from the traditional approaches as follows:
(1) unlike ML, MML uses a prior over the parameters and considers their precision while encoding; 
(2) unlike MAP, MML estimators are invariant under 
non-linear transformations of the parameters \citep{oliver1994mml}.
The estimation of parameters using ML ignores the cost of stating the parameters, and MAP based
estimation uses the probability \textit{density} of parameters instead of their
probability measure. In contrast, the MML inference process takes into
account the optimal precision to which parameters should be stated 
and uses it to determine a corresponding probability value. 
The MML framework decomposes the inference problem into two parts: 
lossless encoding of the parameters, and encoding the data given those 
parameters. It then selects the parameters that result in the least 
\emph{overall} message length to explain the data. 
Thus, models with varying
parameters are evaluated based on their resultant total message lengths.

The MML principle has been used in the inference of several probability distributions 
\citep{WallaceBook}. In particular, the MML parameter estimates
of a three-dimensional vMF distribution were derived by 
\citet{vmf_mmlestimate}, wherein they
demonstrated that the MML estimates outperform
the traditional ML and MAP based ones.
For modelling higher dimensional directional data, \citet{multivariate_vmf}
demonstrated the reliable performance of MML-based vMF estimates
compared to other traditional estimates.
In this work, we derive the MML-based parameter estimates of a \fb~distribution 
and subsequently use them in the mixture modelling.
The MML estimates are shown to perform better than the traditionally used
moment and maximum likelihood estimates. Also, the invariance property
of MML estimates makes them reliable candidates when compared
to MAP estimates. 
We study the results of modelling the protein data using 
mixtures of vMF and \fb~distributions. 
Furthermore, we demonstrate that \fb~mixture models
serve as better candidate models when compared to vMF mixtures in modelling
protein directional data.

The paper is organized as follows: Section~\ref{sec:mml_framework} describes the
MML framework and highlights the key differences between the MML estimation
procedure and others. Section~\ref{subsec:fb5_parameterization} explains
the \fb~distribution and the associated geometrical construction.
Section~\ref{sec:existing_methods} describes the existing moment and maximum likelihood
parameter estimates of the \fb~distribution. Section~\ref{sec:map_estimation}
describes the MAP estimation procedure in the context of the \fb~distribution
and emphasizes its dependency on the manner the distribution is parameterized.
Section~\ref{sec:fb5_estimation} describes the MML-based estimation of the parameters
of the \fb~distribution. Section~\ref{sec:norm_constant_derivatives} outlines
the numerical implementation of methods to compute the normalization 
constant and the corresponding partial derivatives which are required as part of MML-based estimation.
Section~\ref{sec:fb5_mixture_modelling} describes mixture modelling using
\fb~distributions with emphasis on the search for the optimal number of
mixture components. Section~\ref{sec:exp_single_kent} presents the experimental
results of the various parameter estimation methods. Section~\ref{sec:mixture_experiments}
discusses the application of \fb~mixtures with respect to modelling protein
structural data. Section 11 concludes with a summary of the work.

\section{Minimum Message Length (MML) Inference} \label{sec:mml_framework}
In this section, we describe the model selection paradigm using the
Minimum Message Length criterion and proceed to give an overview
of MML-based parameter estimation for any distribution.

\subsection{Model selection using minimum message length criterion}
\citet{wallace68} developed the first practical criterion
for model selection based on information theory.
As per Bayes's theorem:
\[\Pr(\hypothesis\&\dataset) = \Pr(\hypothesis) \times \Pr(\dataset|\hypothesis) = \Pr(\dataset) \times \Pr(\hypothesis|\dataset)\]
where \dataset~denotes observed data, and \hypothesis~some
hypothesis about that data. Further, $\Pr(\hypothesis\&\dataset)$ is the joint probability
of data \dataset~and hypothesis \hypothesis, 
$\Pr(\hypothesis)$ and $\Pr(\dataset)$ are the prior probabilities of
hypothesis \hypothesis~and data \dataset~respectively, $\Pr(\hypothesis|\dataset)$
is the posterior probability, and $\Pr(\dataset|\hypothesis)$ is the
likelihood.  

As per \citet{shannon1948},
given an event $E$ with probability
$\Pr(E)$, the length of the optimal lossless code to represent that
event requires $I(E) = -\log_2 (\Pr(E))$ bits.  
Applying Shannon's insight to
Bayes's theorem, \citet{wallace68} got the following relationship 
between conditional probabilities in terms of optimal message lengths: 
\begin{equation*} 
I(\hypothesis\&\dataset) = I(\hypothesis) + I(\dataset|\hypothesis) = I(\dataset) + I(\hypothesis|\dataset) 
\end{equation*}
The above equation 
can be intrepreted as the \emph{total} cost to encode a
message comprising of the following two parts:
\begin{enumerate}
\item \emph{First part:} the hypothesis $\hypothesis$, which takes $I(\hypothesis)$ bits,
\item \emph{Second part:} the observed data $\dataset$ using knowledge of $\hypothesis$, which takes $I(\dataset|\hypothesis)$ bits.
\end{enumerate}
As a result, given two competing hypotheses \hypothesis~and $\hypothesis^\prime$, 
\begin{gather*}
\Delta I = I(\hypothesis\&\dataset) - I(\hypothesis^\prime\&\dataset) = I(\hypothesis|\dataset) - I(\hypothesis^\prime|\dataset)\quad\text{bits.}\\
\text{Hence,}\,\Pr(\hypothesis^\prime|\dataset) = 2^{\Delta I} \Pr(\hypothesis|\dataset) 
\end{gather*}
gives the log-odds posterior ratio between the two hypotheses.
The framework provides a rigorous means to objectively compare two
competing hypotheses. 
Clearly, the message length can vary depending on the complexity of \hypothesis~and
how well it can explain \dataset. 
A more complex $\hypothesis$ may explain $\dataset$
better but takes more bits to be stated itself.  The trade-off comes
from the fact that (hypothetically) transmitting the message requires the encoding of both
the hypothesis and the data given the hypothesis, that is, the model
complexity $I(\hypothesis)$ and the goodness of fit $I(\dataset|\hypothesis)$.

\subsection{MML-based parameter estimation}  \label{subsec:mml_parameter_estimation}
\citet{wallace-87} introduced a generalized framework to estimate a set of 
parameters $\boldsymbol{\Theta}$ given data \dataset. The method
requires a reasonable prior $h(\boldsymbol{\Theta})$ on the hypothesis and 
evaluating the
\textit{determinant} of the Fisher information matrix $|\fisher|$ of the 
\textit{expected} second-order partial derivatives of the negative 
log-likelihood function, $\mathcal{L}(D|\boldsymbol{\Theta})$. 
The parameter vector $\boldsymbol{\Theta}$ that minimizes 
the message length expression (given by Equation~\ref{eqn:two_part_msg})
is the MML estimate according to \citet{wallace-87}. 
\begin{equation}
I(\boldsymbol{\Theta},\dataset) = \underbrace{\frac{d}{2}\log q_d -\log\left(\frac{h(\boldsymbol{\Theta})}{\sqrt{|\mathcal{F}(\boldsymbol{\Theta})|}}\right)}_{\mathrm{I(\boldsymbol{\Theta})}} 
+ \underbrace{\mathcal{L}(\dataset|\boldsymbol{\Theta}) + \frac{d}{2}}_{\mathrm{I(\dataset|\boldsymbol{\Theta})}}
\label{eqn:two_part_msg}
\end{equation}
where $d$ is the number of free parameters in the model, and $q_d$ is the 
$d$-dimensional lattice quantization constant \citep{conwaySloane84}.
The total message length $I(\boldsymbol{\Theta},\dataset)$, therefore, comprises 
of two parts: (1)~the cost of encoding the parameters, $I(\boldsymbol{\Theta})$, and 
(2)~the cost of encoding the data given the parameters, $I(\dataset|\boldsymbol{\Theta})$.
A concise description of the MML method is presented in \citet{oliver1994mml}.

The key differences between ML, MAP, and MML estimation techniques are
as follows:
in ML estimation, the encoding cost of parameters is, in effect, considered constant,
and minimizing the message length corresponds to minimizing the negative
log-likelihood of the data (the second part).
In MAP based estimation, a probability \textit{density} 
rather than the probability is used.
It is self evident that continuous parameter values can
only be stated to some finite precision;
MML incorporates this in the framework
by determining the region of uncertainty in which the parameter is located.
The value of $V = \dfrac{q_d^{-d/2}}{\sqrt{|\mathcal{F}(\boldsymbol{\Theta})|}}$ gives a 
measure of the volume
of the region of uncertainty in which the parameter $\boldsymbol{\Theta}$ is centered.
This multiplied by the probability density $h(\boldsymbol{\Theta})$ gives the 
\emph{probability} of a particular $\boldsymbol{\Theta}$ as $\Pr(\boldsymbol{\Theta}) = h(\boldsymbol{\Theta})V$.
This probability is used to compute the message length associated with
encoding the continuous valued parameters (to a finite precision).

\section{The \fb~distribution and its parameterization}
\label{subsec:fb5_parameterization}
The \fb~distribution defined by Equation~\ref{eqn:kent_pdf} comprises
of three directional parameters and two scalar parameters. We describe the
following parameterization of the distribution that is intuitive
and relatively easy to comprehend.
Let $\mathbf{X}_1=(1~ 0~ 0)\trans, \mathbf{X}_2=(0~ 1~ 0)\trans,\mathbf{X}_3=(0~ 0~ 1)\trans$ 
be the unit vectors along the standard coordinate axes.
Let $\mathbf{R}$ be the rotation matrix that transforms the orientation axes
$\mean,\major,\minor$ supporting
a \fb~distribution to align with the standard coordinate axes. Then,
$\mathbf{R}\trans = (\mean,\major,\minor) = \mathbf{Q}$ based on the following
reasoning. 

\begin{figure}[ht]
\centering
\subfloat[$R_{\eta}$]{\includegraphics[width=0.3\textwidth]{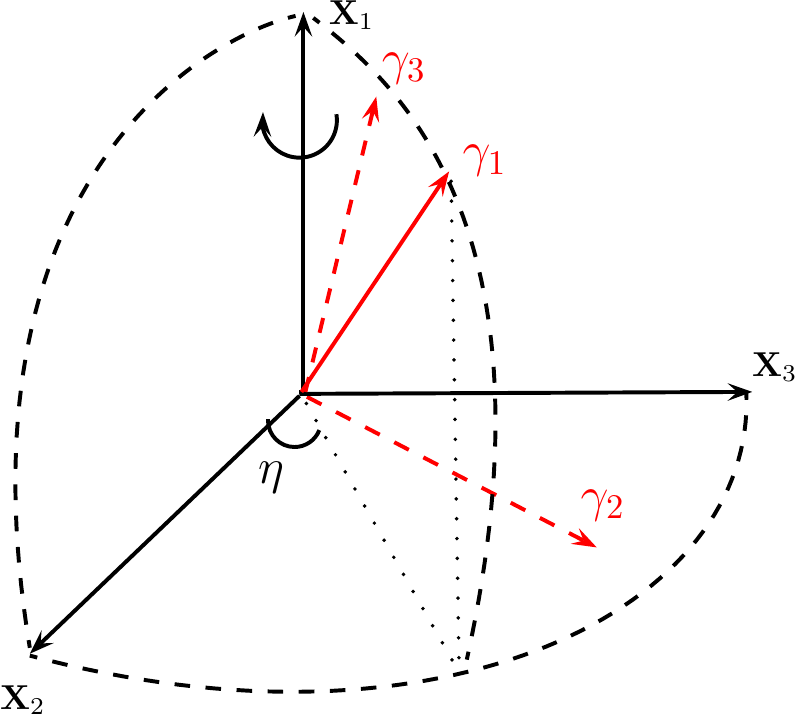}}\quad\quad
\subfloat[$R_{\alpha}$]{\includegraphics[width=0.3\textwidth]{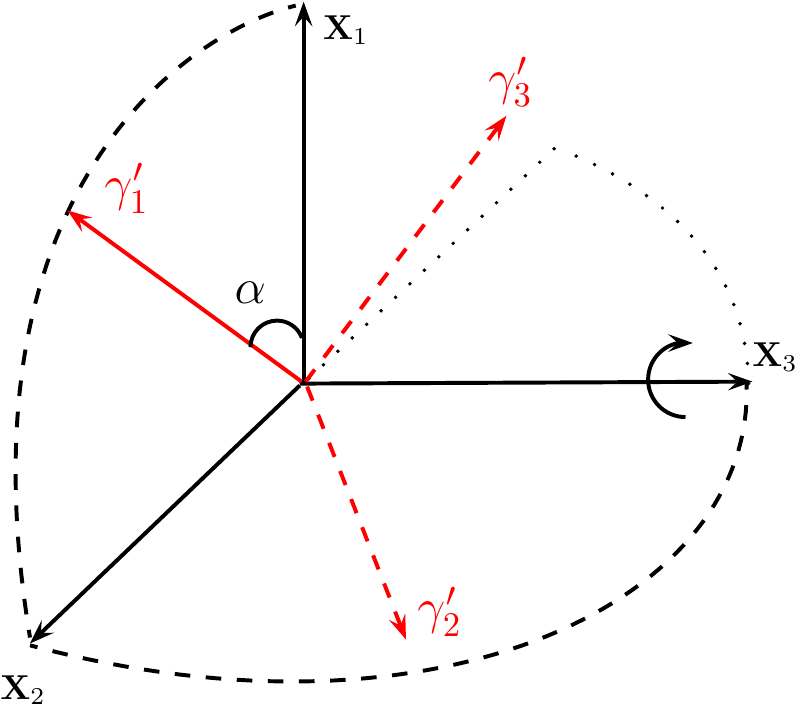}}\quad
\subfloat[$R_{\psi}$]{\includegraphics[width=0.3\textwidth]{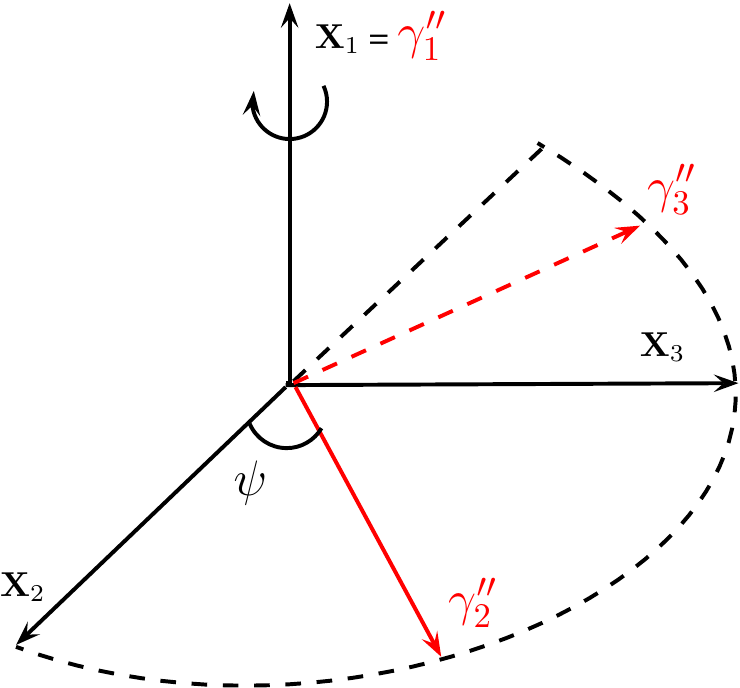}}
\caption{The series of rotations to orient $\mean, \major, \minor$ with the standard
         coordinate axes.
         The \textit{red dashed} lines indicate the axis that is not in the first octant.
         For example, in (b), $\major'$ is below the $\mathbf{X}_2\mathbf{X}_3$ plane whereas $\minor'$
         is above the $\mathbf{X}_2\mathbf{X}_3$ plane but behind the  $\mathbf{X}_1\mathbf{X}_3$ plane.
         } 
\label{fig:orientations}
\end{figure}

Let $\alpha\in[0,\pi]$ and $\eta\in[0,2\pi]$ be the co-latitude and 
longitude that determine the mean axis $\mean$ (shown in Figure~\ref{fig:orientations}a).
A clockwise rotation by an angle $\eta$ about $\mathbf{X}_1$ brings $\mean$ into the 
$\mathbf{X}_1\mathbf{X}_2$ plane. This operation transforms the axes to
$\mean',\major',\minor'$ respectively (Figure~\ref{fig:orientations}b).
A subsequent clockwise rotation by an angle $\alpha$
about $\mathbf{X}_3$ aligns $\mean'$ with $\mathbf{X}_1$.
This rotation brings the major and minor axes into the 
$\mathbf{X}_2\mathbf{X}_3$ plane (as orthogonality should be preserved). 
In this orientation (Figure~\ref{fig:orientations}c), let $\psi\in[0,\pi]$ be the angle between
the transformed axis $\major''$ and $\mathbf{X}_2$. A clockwise rotation
by $\psi$ about $\mathbf{X}_1$ aligns $\major''$ with $\mathbf{X}_2$
and $\minor''$ with $\mathbf{X}_3$.

If $\mathbf{R}_{\eta},\mathbf{R}_{\alpha},\mathbf{R}_{\psi}$ denote the 
respective rotation matrices given by 
\begin{gather*}
\mathbf{R}_{\eta} =  \begin{bmatrix}
                  1 & 0 & 0                   \\[0.3em]
                  0 & \cos\eta  & \sin\eta   \\[0.3em]
                  0 & -\sin\eta  & \cos\eta   
                \end{bmatrix},\,
\mathbf{R}_{\alpha} =  \begin{bmatrix}
                  \cos\alpha  & \sin\alpha & 0   \\[0.3em]
                  -\sin\alpha  & \cos\alpha & 0 \\[0.3em]
                  0 & 0 & 1
                \end{bmatrix},\,
\mathbf{R}_{\psi} =  \begin{bmatrix}
                  1 & 0 & 0                   \\[0.3em]
                  0 & \cos\psi  & \sin\psi   \\[0.3em]
                  0 & -\sin\psi  & \cos\psi   
                \end{bmatrix}
\end{gather*}
then the complete rotation matrix
that effects the transformation from $(\mean,\major,\minor)$ to 
$(\mathbf{X}_1,\mathbf{X}_2,\mathbf{X}_3)$ is given
by their product $\mathbf{R}=\mathbf{R}_{\psi}\mathbf{R}_{\alpha}\mathbf{R}_{\eta}$.
By construction, any $\mathbf{X}_i=\mathbf{R}\gammai, (i=1,2,3)$, and consequently,
$\mathbf{Q} =  \mathbf{R}^{\mathsf{T}}$.
Hence, the three orthogonal axes $\gammai$ of a \fb~distribution can effectively be
described using the three \emph{angular} parameters $\psi,\alpha,\eta$ as follows: 
\begin{gather}
\mean = (\cos\alpha,\, \sin\alpha \cos\eta,\, \sin\alpha \sin\eta)\trans \notag\\
\major = (-\cos\psi \sin\alpha,\,
          \cos\psi \cos\alpha \cos\eta - \sin\psi \sin\eta,\, 
          \cos\psi \cos\alpha \sin\eta + \sin\psi \cos\eta)\trans \notag\\
\minor = (\sin\psi \sin\alpha,\,
          -\sin\psi \cos\alpha \cos\eta - \cos\psi \sin\eta,\,
          -\sin\psi \cos\alpha \sin\eta + \cos\psi \cos\eta)\trans 
\label{eqn:net_rotation}
\end{gather}

\begin{figure}[ht]
\centering
\subfloat[eccentricity = 0.1]{\includegraphics[width=0.3\textwidth]{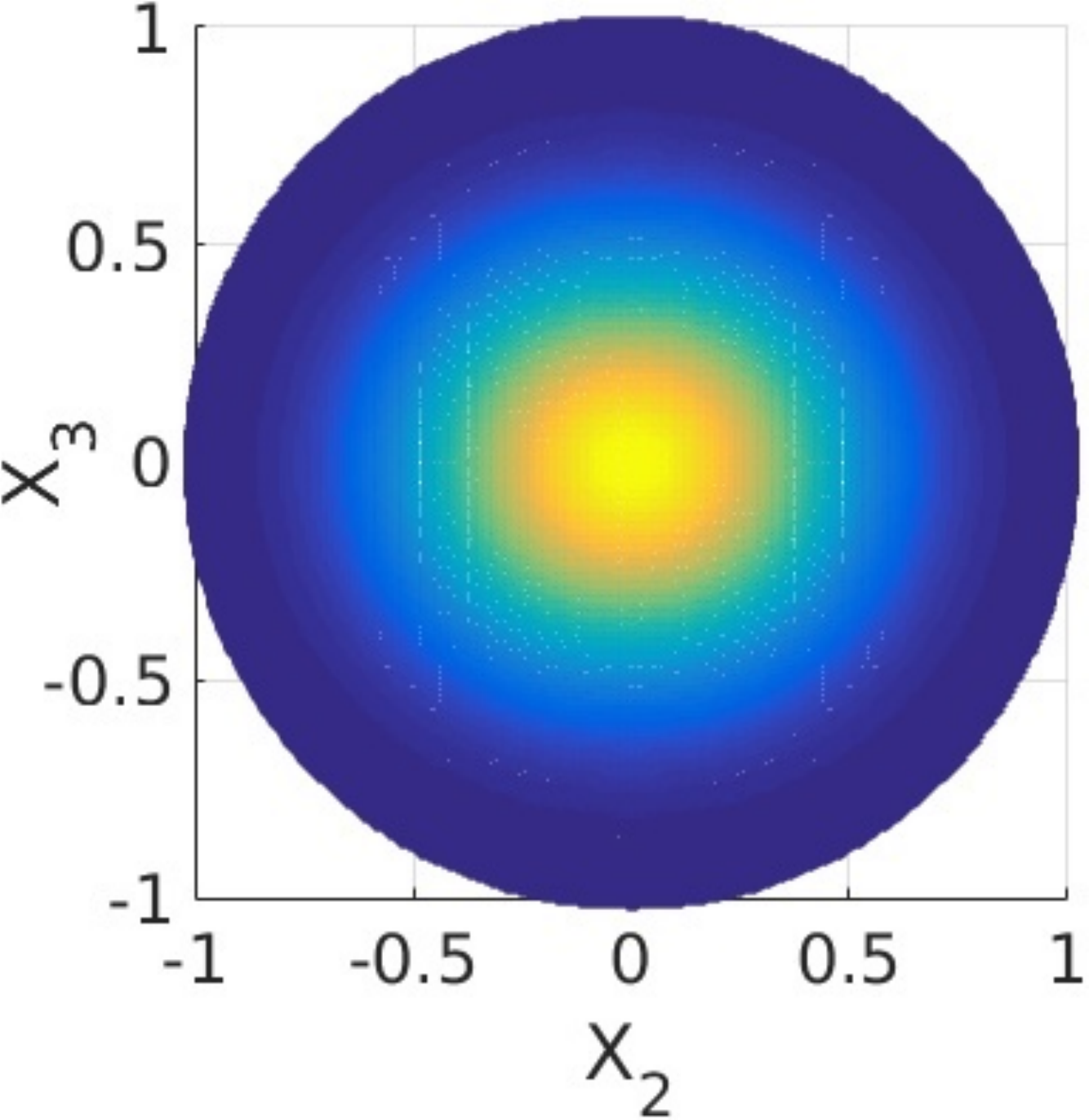}}\quad
\subfloat[eccentricity = 0.5]{\includegraphics[width=0.3\textwidth]{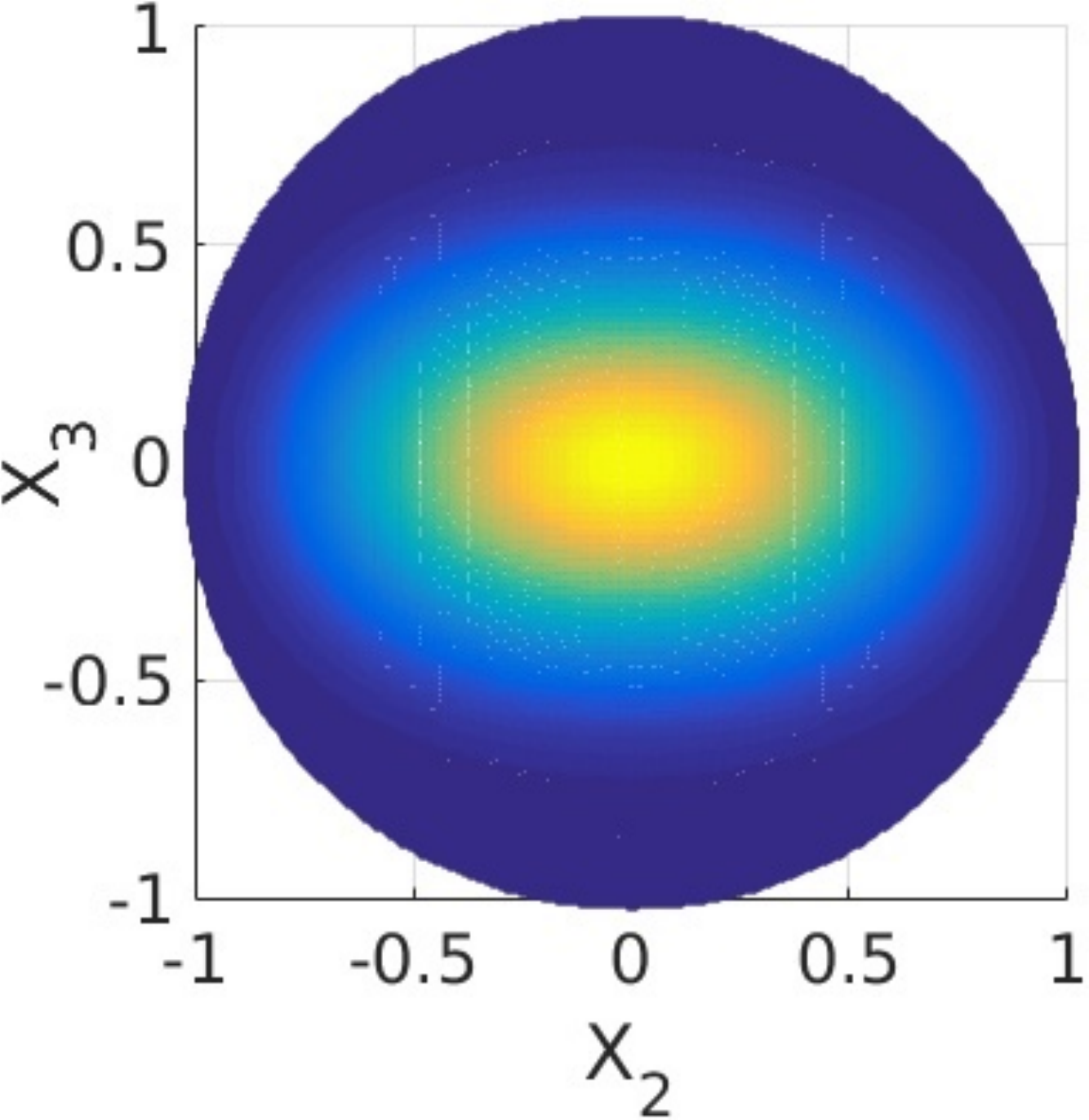}}\quad
\subfloat[eccentricity = 0.9]{\includegraphics[width=0.3\textwidth]{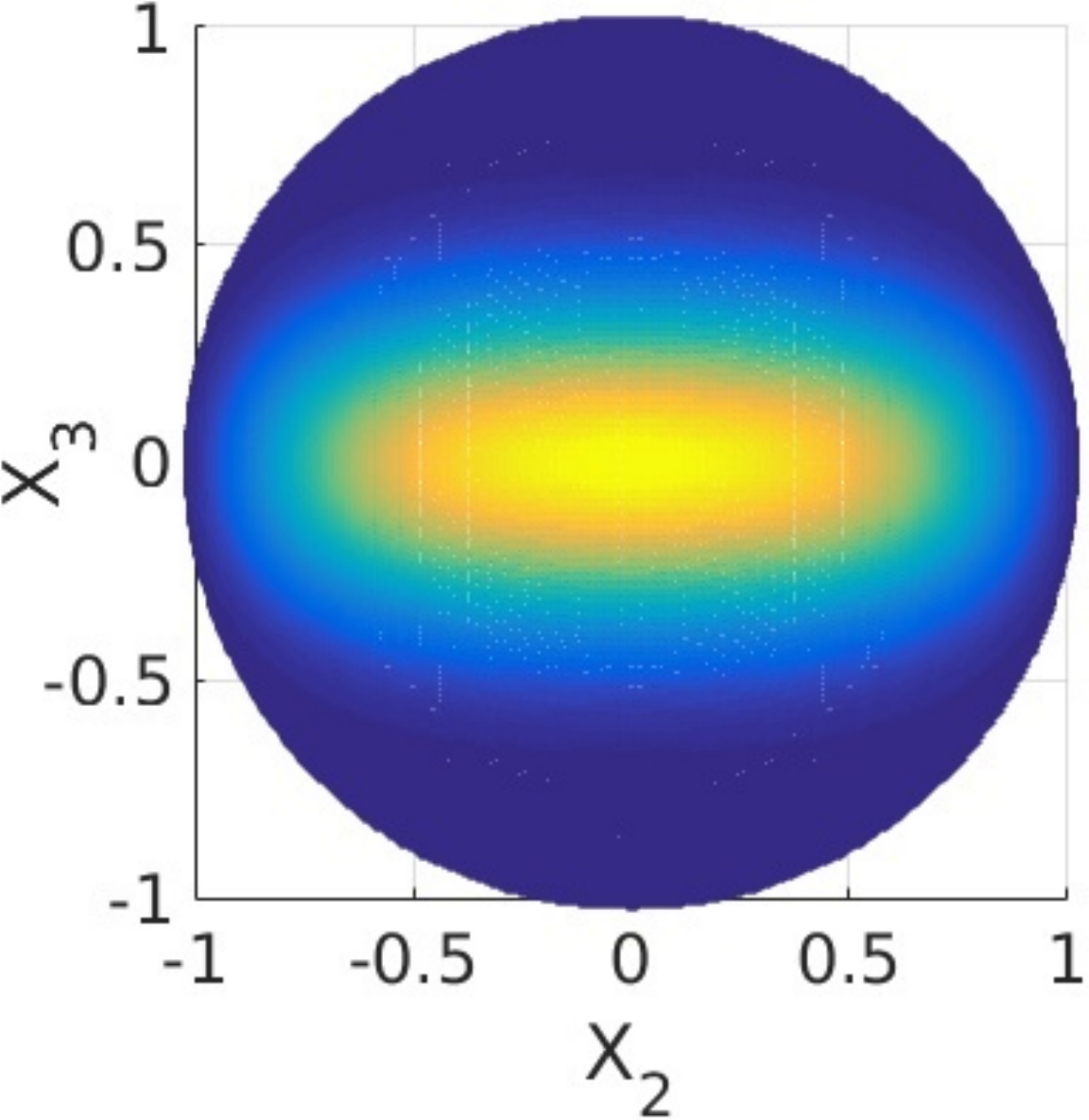}}
\caption{An example of a \fb~distribution with varying eccentricities for $\kappa=10$.}
\label{fig:varying_ecc_heatmap}
\end{figure}

The parameters $\kappa$ and $\beta$ are interpreted as scalars controlling the concentration and
ovalness of the distribution. Also, since the distribution has ellipse-shaped
contours on the spherical surface, it is easier to visualize the distribution and
relate $\kappa$ and $\beta$ terms using eccentricity. 
\citet{kent1982fisher} defined the \emph{eccentricity}\footnote{The definition of eccentricity in this context
differs from the traditional definition of eccentricity for a conic section such as a parabola, an ellipse, or a hyperbola
defined in the Euclidean plane.}
as $2\beta/\kappa$, which 
is constrained to be less than 1 (by definition), allowing correspondence between a specific Kent
distribution and its elliptical nature.
In order to better understand the interaction of $\kappa$ and eccentricity terms,
we provide examples in Figure~\ref{fig:varying_ecc_heatmap}. 

For a given $\kappa=10$, an eccentricity of 0.1 results in (almost) spherical contours 
(Figure~\ref{fig:varying_ecc_heatmap}a)
(reminiscent of a vMF distribution); an eccentricity of 0.5 results 
in contours which are moderately eccentric (Figure~\ref{fig:varying_ecc_heatmap}b);
an eccentricity of 0.9 further disperses the data along the major axis
(Figure~\ref{fig:varying_ecc_heatmap}c).

\section{Existing methods of parameter estimation of the \fb~distribution}
\label{sec:existing_methods}
The traditional methods of maximum
likelihood (ML) estimation or maximum a priori (MAP) based estimation 
require the optimization of negative log-likelihood
or the posterior density functions respectively. 
They, however, don't result in closed form solutions and 
present difficulties because of the complex form of the probability distribution.
Hence, the widely used method of estimating the parameters of a \fb~distribution
is done using moment estimation. \citet{kent1982fisher} 
formulated a procedure to obtain these estimates that may be subsequently
used as starting points to obtain the ML or MAP estimates.
\citet{kent1982fisher} derived the moment
estimates and suggested approximations based on these estimates.

\subsection{Moment estimation} \label{subsec:moment}
The moment estimates were proposed as an alternative to the maximum likelihood 
estimates. The approach adopted by \citet{kent1982fisher} is described here:
let data $\dataset=\{\boldx_1,\ldots,\boldx_N\}$ be a random sample from $\fb(\mathbf{Q},\kappa,\beta)$.
The \emph{sample mean} $\bar{\boldx}$ and \emph{sample dispersion $3\times 3$ matrix} 
$\mathbf{S}$ of the data are then
given as:
\begin{equation*}
\bar{\boldx} = \frac{1}{N} \sum_{i=1}^N \boldx_i
\quad\text{and}\quad
\mathbf{S} = \frac{1}{N} \sum_{i=1}^N \boldx_i \boldx_i\trans
\end{equation*}
Let $\widetilde{\kappa},\widetilde{\beta},\widetilde{\mathbf{Q}}=(\tildemean,\tildemajor,\tildeminor)$
be the respective moment estimates of $\kappa,\beta,$ and $\mathbf{Q}$.
Then the moment estimate $\tildemean$ of the unit mean vector is obtained by 
normalizing $\bar{\boldx}$.
The moment estimates $\tildemajor$ and $\tildeminor$ are obtained
by diagonalizing $\mathbf{S}$. 
The matrix $\widetilde{\mathbf{Q}}$ is obtained using the following two steps:
\begin{enumerate}
\item Choose an orthogonal matrix $\mathbf{H}$ to rotate $\bar{\boldx}$
to align with the $\mathbf{X}_1 =(1~ 0~ 0)\trans$ axis 
(based on the discussion in Section~\ref{subsec:fb5_parameterization},
$\mathbf{H} = \mathbf{R}_{\alpha}\mathbf{R}_{\eta}$, where $\alpha$ and $\eta$
are the co-latitude and longitude of $\bar{\boldx}$ respectively).
Let $\mathbf{B}=\mathbf{H}\trans\mathbf{S}\mathbf{H}$, so that $\mathbf{B}$
is the dispersion matrix in the transformed frame of reference.
\item The moment estimates of the major and minor axis correspond
to the respective directions of maximum and minimum variance of the data 
in this transformed reference frame.
If the angle between the direction of maximum variance and the 
$\mathbf{X}_2=(0~ 1~ 0)\trans$ axis is $\psi$, then a rotation 
defined by the orthogonal matrix $\mathbf{K}$ about $\mathbf{X}_1$ by $\psi$,
aligns the maximum and minimum variance directions with the 
$\mathbf{X}_2$ and $\mathbf{X}_3=(0~ 0~ 1)\trans$ axes respectively. To compute these directions, it is 
required to diagonalize 
$\mathbf{B}_L$, the lower $2\times 2$ submatrix of $\mathbf{B}$.
The eigenvalue decomposition of $\mathbf{B}_L$ gives the angle $\psi$
between the maximum variance direction and $\mathbf{X}_2$,
which can be subsequently used to determine $\mathbf{K}$.
If the $3\times 3$ dispersion matrix $\mathbf{B} = [b_{ij}], 1 \le i,j \le 3$,
the expression for $\psi$ is 
\begin{equation}
\tan2\psi=\frac{2 b_{23}}{b_{22}-b_{33}}
\quad\text{where}\quad 
\mathbf{B}_L =  \begin{bmatrix}
                  b_{22} & b_{23}       \\[0.3em]
                  b_{23} & b_{33}           
                \end{bmatrix}
\label{eqn:moment_psi_est}
\end{equation}
\end{enumerate}
The two rotations defined by the orthogonal transformations $\mathbf{H}$
followed by $\mathbf{K}$ transform the axes of a \fb~distribution to align
with the standard coordinate axes. In effect, 
the original data $\dataset$ is transformed to $\dataset'=\{\mathbf{y}_1,\ldots,\mathbf{y}_N\}$ 
such that $\dataset'$ corresponds to a random sample drawn from $\fb(\mathbf{I},\kappa,\beta)$, 
where $\mathbf{I}$ is the identity matrix.
Hence, an inverse transformation of the coordinate axes 
yields the moment estimates $\widetilde{\mathbf{Q}}$ of the axes of the
\fb~distribution.

Further, for $\mathbf{y}=(y_1,y_2,y_3)\trans$,
\citet{kent1982fisher} provided the moment expressions given below: 
\begin{equation}
\expect[y_1] = c_{\kappa}/c,\quad
\expect[y_2^2 - y_3^2] = c_{\beta}/c,\,\,
\text{where}\,\,c=c(\kappa,\beta), c_{\kappa}=\partial c/\partial\kappa,
c_{\beta}=\partial c/\partial\beta
\label{eqn:kent_standard_moments}
\end{equation}
For data $\dataset$, if $\|\bar{\boldx}\|$ is the magnitude of the sample mean 
$\bar{\boldx}$ and $l_1 > l_2$ are the eigenvalues of $\mathbf{B}_L$,
then \citet{kent1982fisher} defines the \emph{shape} and
\emph{size} and quantities as $r_1$ and $r_2$
respectively and are given as
\begin{equation}
r_1 = \expect[y_1] = \|\bar{\boldx}\|
\quad\text{and}\quad
r_2 = \expect[y_2^2 - y_3^2] = l_1 - l_2
\label{eqn:size_shape}
\end{equation}
Hence, solving these two simultaneous equations 
in conjunction with Equation~\ref{eqn:kent_standard_moments}
results in the moment estimates $\widetilde{\kappa}$ and $\widetilde{\beta}$.
As the expressions of the partial derivatives $c_{\kappa}$ and $c_{\beta}$
are difficult to work with, the following limiting case
approximations of $\widetilde{\kappa}$ and $\widetilde{\beta}$ are often used.
\begin{gather}
\widetilde{\kappa} \approx (2 - 2r_1 - r_2)^{-1} + (2 - 2r_1 + r_2)^{-1} \notag\\
\widetilde{\beta} \approx \frac{1}{2} \{(2 - 2r_1 - r_2)^{-1} - (2 - 2r_1 + r_2)^{-1}\} 
\label{eqn:moment_approx}
\end{gather}
These asymptotic approximations can also be used as starting points to
accurately determine $\widetilde{\kappa}$ and $\widetilde{\beta}$
using an optimization library.

\subsection{Maximum likelihood estimation} \label{subsec:mle}
To obtain the maximum likelihood estimates,
the negative log-likelihood function $\mathcal{L}(\dataset|\boldsymbol{\Theta})$ of the data $\dataset$,
given by Equation~\ref{eqn:negloglike_data}, needs to be minimized. It is to be
noted that $\mean,\major,\minor$ are expressed in terms of $\psi,\alpha,\eta$ 
(Equation~\ref{eqn:net_rotation}), so that 
$\boldsymbol{\Theta}=\{\psi,\alpha,\eta,\kappa,\beta\}$
is a vector of parameters.
\begin{equation}
\mathcal{L}(\dataset|\boldsymbol{\Theta}) = N\log c(\kappa,\beta) 
              - \kappa\,\mean\trans\sum_{i=1}^N\boldx_i
              - \beta\,\major\trans\left(\sum_{i=1}^N\boldx_i\boldxt_i\right)\major
              + \beta\,\minor\trans\left(\sum_{i=1}^N\boldx_i\boldxt_i\right)\minor
\label{eqn:negloglike_data}
\end{equation}
The maximum likelihood estimates are given as solutions to the equation
$\dfrac{\partial\mathcal{L}}{\partial\boldsymbol{\Theta}} = 0$. These
estimates are obtained through numerical optimization as the solution cannot be
written in an analytical form. The optimization routine 
often requires some initial values of the roots. These starting points
are taken to be the moment estimates that were discussed previously.

\section{Maximum a posteriori (MAP) based parameter estimation}
\label{sec:map_estimation}
The moment estimates of a \fb~distribution are typically used in a variety
of applications \citep{peel2001fitting,kent2005using,boomsma2006graphical,hamelryck2006sampling}.
In this section, we explore MAP based parameter estimation,
which we will later use in our discussion to compare the various estimators 
(see Section~\ref{sec:exp_single_kent}). 
The estimation procedure requires the maximization of the posterior
density given some observed data $\dataset$.
If $h(\boldtheta)$ is an appropriate prior density of the parameters
and $\Pr(\dataset|\boldtheta)$ is the likelihood of data given the parameters,
then the posterior density $\Pr(\boldtheta|\dataset)$ is given as 
\begin{equation*}
\Pr(\boldtheta|\dataset) \propto h(\boldtheta) \times \Pr(\dataset|\boldtheta)
\end{equation*}
For an independent and identically distributed sample $\dataset=\{\boldx_1,\ldots,\boldx_N\}$,
and a probability distribution $f(\boldx;\boldtheta)$, the likelihood term 
$\Pr(\dataset|\boldtheta) = \displaystyle\prod_{i=1}^N f(\boldx_i;\boldtheta)$.
The MAP estimator corresponds to the mode of the posterior distribution. The mode is, however, not the same
under varying parameterizations. As a result, the MAP estimate is \emph{not} invariant
under some non-linear transformation of the parameter space \citep{Murphy2012}.
This drawback is exemplified in the context of estimating the parameters of a
\fb~distribution.
A prior $h_{\boldtheta}$ is described  
on the parameter vector $\boldtheta$. It
is formulated based on the choice of priors for the individual
elements of the parameter vector.
We also consider its reparameterization in a transformed space
and demonstrate that the modes of the posterior in these alternative parameterizations
are not given by the same transformation of the parameter space.

\subsection{Prior density of the parameters} \label{subsec:prior_density}
The formulation of the prior density of the 5-parameter vector 
$\boldtheta$ is derived as a product of the priors
of the three angular parameters $\psi,\alpha,\eta$ and two scalar
parameters $\kappa,\beta$. Hence, the prior density of the complete 
set of parameters is given by 
$h_{\boldtheta}(\psi,\alpha,\eta,\kappa,\beta) = h_A(\psi,\alpha,\eta) \times h_S(\kappa,\beta)$.

\subsubsection{Prior density ($h_A$) on the angular parameters $\psi,\alpha,\eta$}
By construction (see Section~\ref{subsec:fb5_parameterization}), the pair
$\alpha,\eta$ uniquely defines the mean direction $\mean$ of a \fb~distribution.
The mean may be considered to be uniformly distributed on the spherical surface,
and hence, its prior density is $\dfrac{\sin\alpha}{4\pi}$. The angle $\psi$ which determines
the orientation of the major and minor axis in a plane perpendicular 
to $\mean$ is treated to be uniformly distributed on $[0,\pi]$. 
The joint prior of the angular parameters is, therefore, 
given by $h_A(\psi,\alpha,\eta) = \dfrac{\sin\alpha}{4\pi^2}$.

\subsubsection{Prior density ($h_S$) on the scale parameters $\kappa,\beta$}
The prior of the concentration parameter $\kappa$
corresponds to the one used by \citet{vmf_mmlestimate} in their
analysis of vMF distributions defined on the two-sphere and is given as:
$h(\kappa) = \dfrac{4\kappa^2}{\pi(1+\kappa^2)^2}$. 
For a given $\kappa$, as per the definition of a \fb~distribution, 
the parameter $\beta\in[0,\kappa/2)$. 
A uniform prior is considered for $\beta$ within this range, 
that is, the conditional density $h(\beta|\kappa) = 2/\kappa$. 
Therefore, the joint prior density of the scalar parameters
is $h_S(\kappa,\beta) = (2/\kappa)h(\kappa)$. 
The joint prior density $h_{\boldtheta}$ is, hence, given as:
\begin{equation}
h_{\boldtheta}(\psi,\alpha,\eta,\kappa,\beta) = \frac{2\kappa\sin\alpha}{\pi^3(1+\kappa^2)^2}
\label{eqn:prior3d}
\end{equation}

\subsection{Non-linear transformations of the parameter space}
The reason for considering another parameterization is to show that
MAP estimates are not invariant under non-linear transformations of the parameter space.
If $T(\boldtheta)=\boldtheta'$ denotes a transformation $T$ on the
parameter vector $\boldtheta$, then for invariance,
the parameter estimates in both the parameterizations should be
affected by the same transformation. The parameter estimate 
$\widehat{\boldtheta}'$ in the transformed space and the estimate
$\widehat{\boldtheta}$ should be related as
$T(\widehat{\boldtheta})=\widehat{\boldtheta}'$.
With the help of an example, we demonstrate that the invariance property
is not a characteristic of MAP-based estimation, thus,
making it an inconsistent estimator.

\subsubsection{An alternative parameterization involving $\beta$}
\label{subsubsec:transformation_beta}
An alternative parameterization is considered where the 
\emph{eccentricity} $e = 2\beta/\kappa$
(see Section~\ref{subsec:fb5_parameterization}) is used instead of $\beta$.
This is an example of a non-linear transformation of the parameter $\beta$.
The prior density $h_{\boldtheta'}$ (Equation~\ref{eqn:prior3dtrans}) 
of the modified parameter
vector $\boldtheta'=\{\psi,\alpha,\eta,\kappa,e\}$ is obtained by
dividing the prior density $h_{\boldtheta}$ 
by the Jacobian of the transformation given by $J = \partial e/\partial\beta  = 2/\kappa$.
The prior density $h_{\boldtheta'}$ (after reparameterization) is:
\begin{equation}
h_{\boldtheta'}(\psi,\alpha,\eta,\kappa,e) 
= \frac{h_{\boldtheta}(\psi,\alpha,\eta,\kappa,\beta)}{J}
= \frac{\kappa^2\sin\alpha}{\pi^3(1+\kappa^2)^2}
\label{eqn:prior3dtrans}
\end{equation}

\subsubsection{Alternative forms of the posterior distribution}
Based on the definitions of prior densities in varying parameter spaces,
one can estimate the parameters by maximizing the 
posterior density in the corresponding parameterization. 
The different expressions for the posterior density are summarized here.
\begin{align}
\text{Posterior}(\boldtheta|\dataset) &\propto h_{\boldtheta}(\psi,\alpha,\eta,\kappa,\beta)
\times \displaystyle\prod_{i=1}^N f(\boldx_i;\boldtheta) \notag\\
\text{Posterior}(\boldtheta'|\dataset) &\propto h_{\boldtheta'}(\psi,\alpha,\eta,\kappa,e) 
\times \displaystyle\prod_{i=1}^N f(\boldx_i;\boldtheta') 
\label{eqn:prior3d_posteriors}
\end{align}
The expression for $f(\boldx,\boldtheta')$ is obtained
by substituting $\beta=\kappa\,e / 2$ in 
the \fb~probability density function $f(\boldx,\boldtheta)$ 
(given by Equation~\ref{eqn:kent_pdf}). 
It should be noted that the value of likelihood expression is the same
across different parameterizations.

\subsection{An example demonstrating the effects of alternative parameterizations} \label{subsec:map_example}
An example of estimating parameters using the various posterior
distributions for a given dataset is shown here. A random sample of size $N=10$
is generated from a \fb~distribution \citep{kent2013new}.
The true parameters of the distribution
are $\{\psi,\alpha,\eta\} = \pi/2$ each, $\kappa = 10$, and $\beta=2.5$ (eccentricity = 0.5).
To obtain the MAP estimates, the objective functions corresponding to
the posterior density (Equation~\ref{eqn:prior3d_posteriors}) need to be maximized.
To solve for the parameter estimates,
the non-linear optimization library NLopt \citep{nlopt} in conjunction with derivative-free
optimization \citep{powell1994direct} is used. 
Maximizing the two versions of the posterior density results in the 
following MAP estimates of $\psi,\alpha,\eta$:
\begin{align*}
\widehat{\psi}=2.071,~\widehat{\alpha}=1.493,~\widehat{\eta}=1.522
& \quad\text{using}\,\,h_{\boldtheta}\\
\widehat{\psi}=2.071,~\widehat{\alpha}=1.493,~\widehat{\eta}=1.522
& \quad\text{using}\,\,h_{\boldtheta'} 
\end{align*}
It is observed that the MAP estimates of $\psi,\alpha,\eta$ are not
different from their counterparts obtained using the two variations of the posterior density.
However, the estimates $\widehat{\kappa}$
and $\widehat{\beta}$ under the parameterizations $\boldtheta$ and $\boldtheta'$
do not correspond to each other as illustrated in the results below.
\begin{gather*}
\widehat{\kappa}=17.023,~\widehat{\beta}=5.493 
\quad\text{using}\,\,h_{\boldtheta}\\
\widehat{\kappa}=20.549,~\widehat{e}=0.701 \implies \widehat{\beta} = \widehat{\kappa}\,\widehat{e}/2 = 7.199
\quad\text{using}\,\,h_{\boldtheta'} 
\end{gather*}
Ideally, the values of $\widehat{\kappa}$ and $\widehat{\beta}$ obtained through the use of
$h_{\boldtheta'}$ should be the same as that obtained when the posterior density is
maximized using $h_{\boldtheta}$ prior. Clearly, with MAP-based estimation,
the end results are different for the two cases.

\begin{figure}[ht]
\centering  
\subfloat[$h_{\boldtheta}$]{\includegraphics[width=0.33\textwidth]{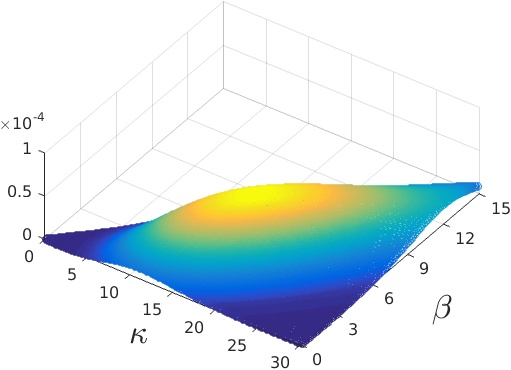}}
\subfloat[$h_{\boldtheta'}$]{\includegraphics[width=0.33\textwidth]{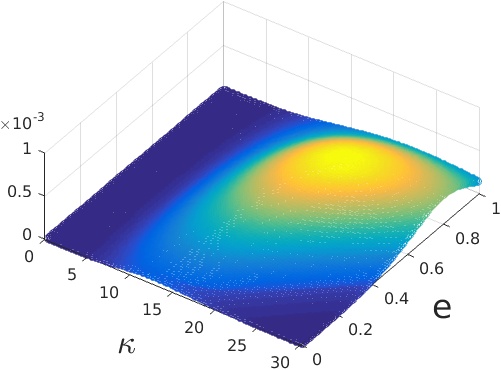}}
\subfloat[$h_{\boldtheta'}$]{\includegraphics[width=0.33\textwidth]{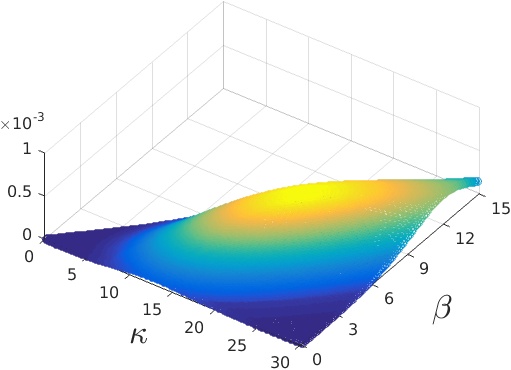}}
\caption{Heat maps depicting the modes (MAP estimate) of the posterior density 
as a function of $\kappa, \beta$ and  $\kappa, e$ parameterizations.
The Z-axis denotes 
the posterior density value in the respective parameterization.}
\label{fig:prior3d_heatmaps}
\end{figure}

The modes of the posterior in the $\kappa,~\beta$ and $\kappa,~e$ parameterizations 
are shown in Figure~\ref{fig:prior3d_heatmaps}(a) and (b) respectively. 
It is expected that the modes of the posterior shift as per the parameter space.
However, they should be invariant regardless of the transformation
affecting the two parameter spaces. 
It is observed that the mode in $\kappa,e$ space, when mapped back
to the $\kappa,\beta$ space, results in a posterior density as shown in
Figure~\ref{fig:prior3d_heatmaps}(c). This is different from the posterior
density shown in Figure~\ref{fig:prior3d_heatmaps}(a), as the modes are at
different locations. We emphasize that
the invariance property of parameter estimates is central to inductive inference. 
The example considered here shows that MAP estimation
of the parameters of a \fb~distribution
does not satisfy the invariance property, thus resulting in unreliable estimators.

The aforementioned eccentricity transform is a straightforward transformation
involving $\beta$. The remaining four parameters are left unchanged 
in this case.
Another parameterization involving all five parameters of the \fb~distribution
is outlined in Appendix~\ref{app:alternative_parameterization}.

\section{MML-based estimation of the parameters of the $\text{FB}_{\mathbf{5}}$ distribution} 
\label{sec:fb5_estimation}
We now shift our focus to deriving the MML-based parameter estimates
of a \fb~distribution which is among the main contributions of this work. 
As explained in Section~\ref{sec:mml_framework},
derivation of the MML estimates requires the formulation of the 
message length expression (Equation~\ref{eqn:two_part_msg}) for encoding
some observed data using the \fb~distribution.
The formulation requires the use of a suitable prior density on the
parameters (see Section~\ref{subsec:prior_density}). 
The prior for $\kappa$ is taken as $h(\kappa)$ 
(see Section~\ref{subsec:prior_density}), the prior of 
$\kappa$ for the vMF distribution
on the two-sphere \citep{vmf_mmlestimate}.
This results in the joint prior density 
$h_{\boldtheta}(\psi,\alpha,\eta,\kappa,\beta)$ (Equation~\ref{eqn:prior3d}).
The main bottleneck involved in the MML-based parameter estimation is,
however, the evaluation of the Fisher information matrix. 
As shown later, its computation involves the first and second order
moments corresponding to a \fb~distribution.  
The details are presented here.\\

\noindent\textbf{Notations}\label{subsec:notation}
Before we proceed with describing the 
approach based on MML inference, we define the following notations
which are used subsequently. We require the use of 
partial derivatives of the normalization constant $c(\kappa,\beta)$ 
given by Equation~\ref{eqn:norm_constant}. The following are the adopted notations 
to represent them.
\begin{gather*}
c(\kappa,\beta)  = c, \quad
c_{\kappa} = \partial c/\partial\kappa, \quad 
c_{\beta} = \partial c/\partial\beta \\
c_{\kappa\kappa} = \partial^2 c/\partial\kappa^2, \quad
c_{\beta\beta} = \partial^2 c/\partial\beta^2, \quad
c_{\kappa\beta} = \partial^2c/\partial\kappa\partial\beta
\end{gather*}

\subsection{Derivation of the moments of a general \fb~distribution}
\citet{kent1982fisher} provided the moment expressions in the case
of a \fb~distribution whose mean, major and minor axes are aligned
with the standard coordinate axes. In this setup,
consider a random vector $\mathbf{y}\sim\fb(\mathbf{I},\kappa,\beta)$, 
where $\mathbf{I}$ is the identity matrix. 
Then, \citet{kent1982fisher} provided the following moments: 
\begin{gather}
\mathbb{E}[\mathbf{y}] = 
\begin{pmatrix}
c_{\kappa}/c & 0 & 0
\end{pmatrix}\trans, \,\text{and}\, \notag\\
\mathbb{E}[\mathbf{y}\mathbf{y}\trans] = \boldsymbol{\Lambda} =
\begin{pmatrix}
\lambda_1 & 0 & 0 \\
0 & \lambda_2 & 0 \\
0 & 0 & \lambda_3
\end{pmatrix}, \text{where} \notag\\
\lambda_1 = \frac{c_{\kappa}}{c}, \lambda_2 = \frac{c-c_{\kappa\kappa}+c_{\beta}}{2c}, \lambda_3 = \frac{c-c_{\kappa\kappa}-c_{\beta}}{2c}
\label{eqn:expectation_standard}
\end{gather}
We derive here the moments in the case of a 
general \fb~distribution, that is, whose three mutually orthogonal axes
can be oriented in any fashion.
Let $\boldx\sim\fb(\mathbf{Q},\kappa,\beta)$,
a generic distribution whose axes are not aligned with the 
coordinate axes. 
Recall, from Section~\ref{subsec:fb5_parameterization}, that $\mathbf{Q}$ is the rotation matrix that aligns
the standard coordinate axes with the axes of a \fb~distribution.
Based on the parameterization of the \fb~distribution, 
we can deduce that 
$\forall\boldx,\exists\,\mathbf{y}\sim\fb(\mathbf{I},\kappa,\beta)$
such that $\boldx=\mathbf{Q}\mathbf{y}$, and hence, 
$\boldx\boldxt=\mathbf{Q}\mathbf{y}\mathbf{y}\trans\mathbf{Q}\trans$.
Using the results from Equation~\ref{eqn:expectation_standard}, we have
\begin{gather}
\mathbb{E}[\boldx] = \mathbf{Q}\,\mathbb{E}[\mathbf{y}] =
\begin{pmatrix}
\mean & \major & \minor
\end{pmatrix}
\begin{pmatrix}
c_{\kappa}/c & 0 & 0
\end{pmatrix}\trans 
= c_{\kappa}/c~\mean\notag\\
\text{and}\,\,\mathbb{E}[\boldx\boldxt] = \mathbf{Q}\,\mathbb{E}[\mathbf{y}\mathbf{y}\trans]\,\mathbf{Q}\trans
= \mathbf{Q}\boldsymbol{\Lambda}\mathbf{Q}\trans
\label{eqn:expectation_generic}
\end{gather}

\subsection{Computation of the Fisher information}
The computation of the \emph{determinant} of the Fisher information matrix
requires the evaluation of the second order partial derivatives
of the negative log-likelihood function with respect to the parameters of the distribution.
As per the density function (Equation~\ref{eqn:kent_pdf}),
the negative log-likelihood of a datum $\boldx$ is given by
\begin{equation}
\mathcal{L}(\boldx|\boldsymbol{\Theta}) = \log c(\kappa,\beta) - \kappa\mean\trans\boldx
              - \beta\,\major\trans\boldx\boldxt\major
              + \beta\,\minor\trans\boldx\boldxt\minor
\label{eqn:negloglike}
\end{equation}
For the \fb~distribution, the 5-parameter vector $\boldsymbol{\Theta}=\{\psi,\alpha,\eta,\kappa,\beta\}$.
Let $\fisherone$ denote the Fisher information for a \emph{single} observation.
The Fisher information matrix $\fisherone$ associated with the parameters of a \fb~distribution
is a $5\times5$ \emph{symmetric} matrix
whose $(i,j)^{\text{th}}$ element corresponding to parameters $\theta_i$,
$\theta_j \in \boldsymbol{\Theta}$ is
$\mathcal{F}_{\theta_i\theta_j} = 
\mathbb{E}\left[\dfrac{\partial^2\mathcal{L}}{\partial\theta_i\partial\theta_j}\right]$.
Further, as explained later, the determinant $|\fisherone|$ is decomposed as a product of $|\fisherij_A|$
and $|\fisherij_S|$, where $\fisherij_A$ is the Fisher matrix associated
with the angular parameters $\psi,\alpha,\eta$, and $\fisherij_S$ is the
Fisher matrix associated with the scale parameters $\kappa,\beta$.

\subsubsection{Fisher matrix ($\fisherij_{A}$) associated with $\psi,\alpha,\eta$}
\label{subsubsec:fisher_angle}
$\fisherij_A$ is a $3\times 3$ symmetric matrix whose elements are the expected values 
of the second order partial derivatives of $\mathcal{L}$ with respect to
$\theta_i,\theta_j\in\{\psi,\alpha,\eta\}$. Let the \emph{expectation} be given as
\begin{equation}
\mathbb{E}\left[\dfrac{\partial^2\mathcal{L}}{\partial\theta_i\partial\theta_j}\right] = 
-\kappa\,T(\mean) - \beta\,T(\major) + \beta\,T(\minor)
\label{eqn:fisher_ij}
\end{equation}
where the individual terms $T(\gammam),\,m\in\{1,2,3\}$ are comprised of 
the expectations of the corresponding partial differentials of $\gammam$.
They are computed using the following identities:
\begin{align}
T(\mean) &= 
\mathbb{E}\left[\frac{\partial^2 (\mean\trans\boldx)}{\partial\theta_i\partial\theta_j}\right] = 
\mathbb{E}\left[\boldx\right]\trans\frac{\partial^2 \mean}{\partial\theta_i\partial\theta_j}, \,\text{and}\notag\\
T(\gammam) &= 
\mathbb{E}\left[\frac{\partial^2 (\gammam\trans\boldx\boldxt\gammam)}{\partial\theta_i\partial\theta_j}\right] \quad(\text{for}\,\, m=2,3)\notag\\
&= 2 \left( 
\gammam\trans\mathbb{E}\left[\boldx\boldxt\right]\frac{\partial^2\gammam}{\partial\theta_i\partial\theta_j}
+ \left(\frac{\partial\gammam}{\partial\theta_i}\right)\trans\mathbb{E}\left[\boldx\boldxt\right]\left(\frac{\partial\gammam}{\partial\theta_j}\right)
\right)
\label{eqn:matrix_diff_identities}
\end{align}
The terms $T(\gammam)$ depend on the expressions for
the constituent first and second order partial differentials of $\gammam$, 
which we provide in Appendix~\ref{app:axes_differentials}.
Using Equations~\ref{eqn:expectation_generic}, \ref{eqn:fisher_ij} and \ref{eqn:matrix_diff_identities},
the elements of the Fisher information matrix $\fisherij_A$ are derived as follows:
\begin{align}
\fisherij_{\psi\psi} &= 4\,\beta\,c_{\beta}/c\,;\quad 
\fisherij_{\alpha\psi} = 0; \quad
\fisherij_{\eta\psi} = (\cos\alpha)\,4\,\beta\,c_{\beta}/c\notag\\
\fisherij_{\alpha\alpha} &= \kappa\,c_{\kappa}/c + 2\beta\left\{ (\lambda_1 - \lambda_3)\sin^2\psi - (\lambda_1 - \lambda_2)\cos^2\psi \right\} \notag\\
\fisherij_{\eta\alpha} &= \beta (1-3\lambda_1)\sin 2\psi\,\sin\alpha \notag\\
\fisherij_{\eta\eta} &= (\sin^2\alpha)\,\kappa\,c_{\kappa}/c \notag\\
+& 2\beta \left\{
                  \begin{array}{l}
                    \lambda_2 (\cos^2\psi\,\cos^2\alpha+\sin^2\psi) + (\lambda_2-\lambda_3)\cos^2\alpha\\
                    -\lambda_3 (\sin^2\psi\,\cos^2\alpha+\cos^2\psi) + \lambda_1\sin^2\alpha\cos 2\psi
                  \end{array}
          \right\} 
\label{eqn:fisher_angles}
\end{align}

\subsubsection{Fisher matrix ($\fisherij_S$) associated with $\kappa,\beta$}
$\fisherij_S$ is a $2\times 2$ symmetric matrix whose elements are the \emph{expectations}
of the second order partial derivatives of $\mathcal{L}$ with respect to
$\kappa$ and $\beta$. From Equation~\ref{eqn:negloglike}, we have
\begin{align}
\frac{\partial\mathcal{L}}{\partial\kappa} = \frac{c_\kappa}{c} - \mean\trans\boldx &\quad\text{and}\quad
\frac{\partial\mathcal{L}}{\partial\beta} = \frac{c_{\beta}}{c} -\major\trans\boldx\boldxt\major +\minor\trans\boldx\boldxt\minor \notag\\
\frac{\partial^2\mathcal{L}}{\partial\kappa^2} &= \frac{cc_{\kappa\kappa}-c_{\kappa}^2}{c^2} = \fisherij_{\kappa\kappa}\notag\\
\frac{\partial^2\mathcal{L}}{\partial\beta^2} &= \frac{cc_{\beta\beta}-c_{\beta}^2}{c^2} = \fisherij_{\beta\beta} ,\quad\text{and}\notag\\
\frac{\partial^2\mathcal{L}}{\partial\kappa\partial\beta} &= \frac{cc_{\kappa\beta}-c_{\kappa}c_{\beta}}{c^2} = \fisherij_{\kappa\beta}
\label{eqn:fisher_kb}
\end{align}

\subsubsection{Fisher matrix $\fisher$ associated with the 5-parameter vector $\boldtheta$}
It is to be noted that
for $\theta_i\in\{\kappa,\beta\}$ and $\theta_j\in\{\psi,\alpha,\eta\}$,
$T(\gammam)=0$ as $\dfrac{\partial\gammam}{\partial\theta_i} = 0$
($\gammam$ given by Equation~\ref{eqn:net_rotation} are independent of $\kappa,\beta$).
Consequently, $\fisherij_{\theta_i\theta_j}=0$. 
This allows for the computation of $|\fisherone|$ as the product of
$|\fisherij_A|$ and $|\fisherij_S|$. Thus,
\begin{equation*}
|\fisherone| =  \begin{vmatrix}
                  \fisherij_{\psi\psi} &  \fisherij_{\psi\alpha}  &  \fisherij_{\psi\eta}      & 0 & 0 \\
                  \fisherij_{\alpha\psi} &  \fisherij_{\alpha\alpha}  &  \fisherij_{\alpha\eta} & 0 & 0\\  
                  \fisherij_{\eta\psi} &  \fisherij_{\eta\alpha}  &  \fisherij_{\eta\eta}   & 0 & 0 \\
                  0 & 0 & 0 & \fisherij_{\kappa\kappa} & \fisherij_{\kappa\beta} \\
                  0 & 0 & 0 & \fisherij_{\beta\kappa} & \fisherij_{\beta\beta} 
                \end{vmatrix}
          = |\fisherij_A| |\fisherij_S|
\label{eqn:fisher}
\end{equation*}
Then, the Fisher information for some observed data $\dataset=\{\boldx_1,\ldots,\boldx_N\}$ 
is given by $|\fisher| = N^5 |\fisherone|$ (as each element in $|\fisherone|$ is multiplied by 
the sample size $N$).

\subsection{Message length formulation}
\label{subsec:mml_formulation}
The message length to encode some observed data $\dataset$
can now be formulated by substituting the prior density $h_{\boldtheta}$
(Equation~\ref{eqn:prior3d}), the Fisher information $|\fisher|$ 
and the negative log-likelihood of the data (Equation~\ref{eqn:negloglike_data})
in the message length expression (Equation~\ref{eqn:two_part_msg}).
The MML parameter estimates are the ones that minimize the entire
message length. As there is no analytical form of the MML estimates,
the solution is obtained, as for the maximum likelihood and MAP case,
by using the NLopt\footnote{\url{http://ab-initio.mit.edu/nlopt}} optimization library \citep{nlopt} .
At each stage of the optimization routine, the Fisher information
needs to be calculated. However, this involves the computation of 
complex entities such as the normalization constant $c(\kappa,\beta)$ and its partial derivatives.
The computation of these intricate mathematical forms using numerical methods
is discussed in Section~\ref{sec:norm_constant_derivatives}.

\section{Computation of the normalization constant and the associated derivatives} 
\label{sec:norm_constant_derivatives}
The computation of the negative log-likelihood function and the message length
is hindered because of the presence of the normalization constant
and its associated derivatives. \citet{kent1982fisher} provided an
asymptotic formula for $c(\kappa,\beta)$ as $2\pi\exp(\kappa) [(\kappa^2-4\beta^2)]^{-1/2}$.
However, this approximation is valid for large $\kappa$ and when
$2\beta/\kappa$ is sufficiently small. In this section, we describe the
methods that can be employed to efficiently compute these complex functions
without making any assumptions.

\subsection{Computing $\boldsymbol{\log c(\kappa,\beta)}$ and the logarithm of 
the derivatives: $\boldsymbol{c_{\kappa} = \partial c/\partial\kappa}$
and $\boldsymbol{c_{\kappa\kappa} = \partial^2c/\partial\kappa^2}$}
\label{subsec:kappa_der}
The expressions of $c,c_{\kappa},c_{\kappa\kappa}$ are related.
to each other. 
These are explained by defining the quantity
$S^{(m)}_1$, a logarithm sum where $m\in\{0,1,2\}$, $p=2j+\dfrac{1}{2}$,
$\delta_1=2\pi\sqrt{\dfrac{2}{\kappa}}$, and $e = \dfrac{2\beta}{\kappa} < 1$ (by definition).
\begin{equation}
S^{(m)}_1 = \log\delta_1 + \log\sum_{j=0}^\infty\underbrace{\frac{\Gamma(j+\frac{1}{2})}{\Gamma(j+1)} e^{2j} I_{p+m}(\kappa)}_{f_j} 
\label{eqn:series1}
\end{equation}
\noindent\emph{Computation of the series $S^{(m)}_1$:}
We first establish that $f_{j+1}<f_j \,\forall j \ge 0$ and 
show that $S^{(m)}_1$ converges to a finite sum as $j \to \infty$.
Consider the logarithm of the ratio of consecutive terms $f_j$ and $f_{j+1}$
in $S^{(m)}_1$. 
\begin{align}
\log\frac{f_{j+1}}{f_j} = \log\frac{j+\frac{1}{2}}{j+1} + 2\log e + 
\log\frac{I_{p+m+2}(\kappa)}{I_{p+m}(\kappa)}
\label{eqn:ratio_series1}
\end{align}
For $p,v>0$, $I_{p+v} < I_{p}$, and the ratio $\frac{I_{p+v}}{I_p} \to 0$
for large $v$ \citep{amos1974computation}.
Further, $e < 1$ implies the above equation is the sum of negative terms. Hence,
$\log\frac{f_{j+1}}{f_j} < 0$, which means $f_{j+1}<f_j$. Also, 
\begin{equation*}
\lim_{j\to\infty} \log\frac{f_{j+1}}{f_j} = 0 + 2\log e + \lim_{j\to\infty}
\log\frac{I_{2j+\frac{1}{2}+2}(\kappa)}{I_{2j+\frac{1}{2}}(\kappa)} = -\infty
\end{equation*}
Hence, as $\displaystyle\lim_{j\to\infty} \dfrac{f_{j+1}}{f_j} = 0$, $S^{(m)}_1$ is a convergent series.

For practical implementation of the sum, 
we express $S^{(m)}_1$ as the modified summation, 
\begin{equation}
S^{(m)}_1 = \log\delta_1 + \log f_0 + \log \sum_{j=0}^\infty t_j
\label{eqn:series1_modified}
\end{equation}
where each $f_j$ is divided by the 
\emph{maximum} term $f_0$. For each $j > 0, \log f_{j}$ is calculated using
the previous term $\log f_{j-1}$ (Equation~\ref{eqn:ratio_series1}).
The new term $t_j = f_j/f_0$ is then 
computed\footnote{Because of the nature of Bessel functions,
$\log f_j$ can get very large and can result in overflow when
calculating the exponent $\exp(\log f_j)$. However, dividing by $f_0$
results in $f_j/f_0 < 1$.
}
as  $\exp(\log f_j - \log f_0)$ (computing the difference with the maximum value 
and then taking the exponent ensures numerical stability).
The summation is terminated when the ratio $\dfrac{t_j}{\sum_{k=1}^j t_k} < \epsilon$
(a small threshold $\sim10^{-6}$).

\begin{itemize}
\item Let $S(c)=\log c(\kappa,\beta)$:
Substituting $m=0$ in Equation~\ref{eqn:series1}
gives the logarithm of the normalization constant (given in Equation~\ref{eqn:norm_constant}).
Hence, $S(c)=S^{(0)}_1$. 

\item Let the $j^{th}$ term dependent on $\kappa$ in Equation~\ref{eqn:norm_constant}
be represented as
$g_j(\kappa) = I_{p}/\kappa^p$, where 
$I_p$ implicitly refers to $I_p(\kappa)$. 
We use the relationship between the Bessel functions $I_{p},I_{p-1}$, and the derivative $I'_p$
in Equation~\ref{eqn:bessel_identity} \citep{abramowitz1972handbook},
to derive the expressions for the first and second derivatives of $g_j(\kappa)$
(Equation~\ref{eqn:gk_derivatives}).
\begin{gather}
\kappa I'_p = \kappa I_{p-1} - p I_p \label{eqn:bessel_identity}\\
g_j'(\kappa) = \frac{I_{p+1}}{\kappa^p} \quad\text{and}\quad g_j''(\kappa) = \frac{I_{p+2}}{\kappa^p} + \frac{1}{\kappa} . \frac{I_{p+1}}{\kappa^{p}} \label{eqn:gk_derivatives}
\end{gather}
Let $S(c_\kappa) = \log c_{\kappa}$: 
Because of the similar forms of $g_j(\kappa)$ and $g_j'(\kappa)$,
the expression for $S(c_\kappa)$ will be 
similar to $S(c)$ with a change in \emph{order} of the Bessel functions 
from $m=0$ in Equation~\ref{eqn:series1} to $m=1$. Hence, $S(c_{\kappa})=S^{(1)}_1$ and
an expression akin to Equation~\ref{eqn:series1_modified} can be derived for $S(c_\kappa)$.

\item Let $S(c_{\kappa\kappa}) = \log c_{\kappa\kappa}$:
Substituting $m=2$ in Equation~\ref{eqn:series1} gives the 
logarithm sum $S^{(2)}_1$ corresponding to the series with terms $\dfrac{I_{p+2}}{\kappa^p}$.
Based on the nature of $g_j''(\kappa)$ (Equation~\ref{eqn:gk_derivatives}), 
and noting that $S(c_{\kappa}) > S^{(2)}_1$ (as $I_{p+1} > I_{p+2}\,\forall\,p\ge0$),
we can formulate $S(c_{\kappa\kappa})$ as given below. 
\begin{equation*}
S(c_{\kappa\kappa}) = S(c_{\kappa}) + \log \left(\exp(S^{(2)}_1-S(c_{\kappa})) + \frac{1}{\kappa}\right)
\end{equation*}
\end{itemize}

\subsection{The logarithm of the
derivatives: $\boldsymbol{c_{\beta} = \partial c/\partial\beta}$,
$\boldsymbol{c_{\kappa\beta} = \partial^2c/\partial\kappa\partial\beta}$, and
$\boldsymbol{c_{\beta\beta} = \partial^2c/\partial\beta^2}$}
The expressions of $c_{\beta}$ and $c_{\kappa\beta}$ are related
and are explained using 
the logarithm sum $S^{(n)}_2$ where $n\in\{0,1\}$, 
$\delta_2=\dfrac{4\pi}{\beta}\sqrt{\dfrac{2}{\kappa}}$, 
$p=2j+\frac{1}{2}$, and $e=\dfrac{2\beta}{\kappa}$.
\begin{equation}
S^{(n)}_2 = \log\delta_2 + \log\sum_{j=1}^\infty\underbrace{\frac{\Gamma(j+\frac{1}{2})}{\Gamma(j)} e^{2j} I_{p+n}(\kappa)}_{f_j}
\label{eqn:series2}
\end{equation}
We note that $S^{(n)}_2$ is a convergent series (proof is based on the same reasoning
as in Section~\ref{subsec:kappa_der}).

Let the $j^{th}$ term dependent on $\beta,\kappa$ in Equation~\ref{eqn:norm_constant}
be represented as
$g_j(\beta,\kappa) = \beta^{2j}\dfrac{I_{p}}{\kappa^p}$. Its partial derivatives are
given below.
These derivatives are the terms in the series $S^{(n)}_2$ (after
factoring out the common elements as $\delta_2$).
\begin{equation*}
\frac{\partial g_j}{\partial\beta} = 2j \beta^{2j-1}\frac{I_{p}}{\kappa^p} 
\quad\text{and}\quad
\frac{\partial^2 g_j}{\partial\kappa\partial\beta} = 2j \beta^{2j-1}\frac{I_{p+1}}{\kappa^p}
\end{equation*} 

\begin{itemize}
\item Let $S(c_\beta)=\log c_\beta$: this is obtained by
substituting $n=0$ in Equation~\ref{eqn:series2}.
Hence, $S(c_\beta)=S^{(0)}_2$. 

\item Similarly, $S(c_{\kappa\beta})=\log c_{\kappa\beta}=S^{(1)}_2$.

\item The expression to compute $S(c_{\beta\beta})=\log c_{\beta\beta}$ is given by
\begin{equation*}
S(c_{\beta\beta}) = \log\left(\frac{\delta_2}{\beta}\right) + \log\sum_{j=1}^\infty\underbrace{\frac{\Gamma(j+\frac{1}{2})}{\Gamma(j)} (2j-1) e^{2j} I_{p}(\kappa)}_{f_j}
\label{eqn:series3}
\end{equation*}
\end{itemize}
The practical implementation of $S_2^{(n)}$ and $S(c_{\beta\beta})$
is similar to that of $S^{(m)}_1$ 
given by Equation~\ref{eqn:series1_modified}.
However, in these cases, the expressions of $f_j$ and consequently $t_j$, are modified
accordingly.
Also, the series begin from $j=1$, and hence,
the maximum terms will correspond to $f_1$.

\section{Mixture modelling of \fb~distributions}
\label{sec:fb5_mixture_modelling}
In this section, we provide an overview of the mixture modelling apparatus
in the context of modelling directional data using \fb~distributions.
The probability distribution of a mixture $\fancym$ is of the form:
\begin{equation*}
f(\boldx;\boldphi) = \sum_{j=1}^K w_j f_j(\boldx;\boldtheta_j)
\end{equation*}
where $K$ is the number of component \fb~distributions, $w_j \ge 0$ is the 
component weight such that $\sum_{j=1}^K w_j = 1$, and $\boldtheta_j$ denotes the 5-parameter vector
of the $j^{\text{th}}$ \fb~distribution. The parameters of the mixture 
are collectively given by
$\boldphi = \{w_1,\cdots,w_K,\boldtheta_1,\cdots,\boldtheta_K\}$.

\subsection{Estimating the mixture parameters}
\label{subsec:em_ml}
For a mixture with $K$ number of components,
the traditional method of estimating the mixture
parameters is done by minimizing the negative log-likelihood function of the data given by
\begin{equation}
\mathcal{L}(\dataset|\boldphi) = -\displaystyle\sum_{i=1}^N \log\displaystyle\sum_{j=1}^K w_j f_j(\boldx_i;\boldtheta_j)
\label{eqn:negloglike_mixture}
\end{equation}
where $\dataset = \{\boldx_1,\ldots,\boldx_N\}$ is the observed data of size $N$.
The maximum likelihood estimation procedure, in this case, involves
an \emph{expectation-maximization} (EM) algorithm 
\citep{dempster1977maximum,krishnan1997algorithm} which is decomposed into the following
steps:
\begin{itemize}
\item\emph{Expectation (E-step):} The membership of each datum $\boldx_i\in\dataset$
in a mixture component $j \in \{1,K\}$ is updated as:
\begin{equation*}
  r_{ij} = \frac{w_j f(\boldx_i;\boldtheta_j)}{\sum_{k=1}^K w_k f(\boldx_i;\boldtheta_k)}, 
  \quad\text{and}\quad 
  n_j = \sum_{i=1}^N r_{ij} 
\end{equation*}
where the table of memberships $r_{ij}$ is termed the \emph{responsibility matrix}
and $n_j$ is the effective membership of the $j^{\text{th}}$ component.

\item\emph{Maximization (M-step):} The parameters of each component are updated by
their respective maximum likelihood estimates. These are obtained
by minimizing $\mathcal{L}(\dataset|\boldphi)$.
Differentiating Equation~\ref{eqn:negloglike_mixture}
with respect to $\boldtheta_j$ leads to the following modified form
\begin{align}
\mathcal{L}(\dataset|\boldtheta_j)
      &= n_j \log c(\kappa_j,\beta_j) 
      - \kappa_j\,\meanj\trans\sum_{i=1}^N r_{ij} \boldx_i \notag\\
      &- \beta_j\,\majorj\trans\left(\sum_{i=1}^N r_{ij} \boldx_i\boldxt_i\right)\majorj
      + \beta_j\,\minorj\trans\left(\sum_{i=1}^N r_{ij} \boldx_i\boldxt_i\right)\minorj
\label{eqn:negloglike_data_mixcomp}
\end{align}
where $\boldtheta_j = \{\psi_j,\alpha_j,\eta_j,\kappa_j,\beta_j\}$ and
$\meanj,\majorj,\minorj$ are functions of $\psi_j,\alpha_j,\eta_j$ 
(Equation~\ref{eqn:net_rotation}).
The above equation resembles the negative log-likelihood function 
due to a single \fb~component (Equation~\ref{eqn:negloglike_data}) after accounting for 
the partial memberships of data within that component. 
Minimizing Equation~\ref{eqn:negloglike_data_mixcomp} yields the maximum likelihood
estimate of $\boldtheta_j$. The component weights are updated as $w_j = n_j/N$.
\end{itemize}

\subsection{Estimating the mixture parameters using the MML framework}
\label{subsec:em_mml}
The seminal work on minimum message length inference of mixture
models was carried out by \citet{wallace68}. 
As per the MML framework (Section~\ref{sec:mml_framework}), the estimation
of parameters of a mixture distribution requires the encoding of the 
parameters and the data given those parameters. The resultant total
message length expression needs to be minimized to obtain the MML estimates.
The formulation of a mixture modelling problem using MML framework
can be decomposed into: 
\begin{enumerate}
\item \emph{First part:} encoding the mixture parameters $\boldphi$, namely, 
number of components $K$, mixture weights $\mathbf{w} = \{w_1,\ldots,w_K\}$, 
and the component parameters $\boldtheta_j$. 
\item \emph{Second part:} encoding the data $\dataset$
given the parameters $\boldphi$.
\end{enumerate}
The schemes for encoding $K$ and $\mathbf{w}$ are generic \citep{WallaceBook} 
and are summarized in \citet{multivariate_vmf}. However, encoding
the component parameters requires the evaluation of the corresponding Fisher
information. Using the appropriate encoding schemes, the general form of
the total message length expression provided by \citet{wallace-87} is:
\begin{equation}
I(\boldphi,\dataset) = \underbrace{I(K) + I(\mathbf{w}) + 
                       \sum_{j=1}^K I(\boldtheta_j)}_{\text{first part: } I(\boldphi)}
+ \underbrace{\mathcal{L}(\dataset|\boldphi) + \text{constant}}_{\text{second part:} I(\dataset|\boldphi)}
\label{eqn:mixture_msglen}
\end{equation}
where $I(K)$ and $I(\mathbf{w})$ are the message lengths to encode
$K$ and $\mathbf{w}$ respectively. The cumulative Fisher information 
of mixtures for the components' parameters
is given by the summation $\displaystyle\sum_{j=1}^K I(\boldtheta_j)$, where 
$I(\boldtheta_j) = -\log\dfrac{h(\boldtheta_j)}{\sqrt{|\mathcal{F}(\boldtheta_j)|}}$
is the message length to encode the parameters of the $j^{\text{th}}$
component (Section~\ref{subsec:mml_parameter_estimation}).
The second part of the message $I(\dataset|\boldphi)$ is a measure of the 
goodness of fit to the data and corresponds to the negative log-likelihood 
(Equation~\ref{eqn:negloglike_mixture}).

To obtain the MML estimates, an EM algorithm is employed to minimize the two-part
message length $I(\boldphi,\dataset)$. In the E-step, the memberships
of the data are updated, while in the M-step, the component parameters are
updated using their respective MML estimates (Section~\ref{subsec:mml_formulation}). 
The EM algorithm is continued
until there is no change in message length, that is, when the algorithm 
converges to a local minimum.

\subsection{Determining the optimal number of mixture components}
\label{subsec:optimal_components}
The parameters of a mixture can be estimated once the number of mixture
components are known. A mixture modelling problem also needs to address
the issue of selection of optimal number of components. The EM algorithms
that are used in the estimation of mixture parameters are carried out
with a fixed number of components. As the number of
mixture components $K$ increases, the negative log-likelihood 
(Equation~\ref{eqn:negloglike_mixture}) decreases and consequently
results in an improvement to the quality of fit to the data. However, increasing 
$K$ results in the mixtures becoming overly complex. Thus, a reliable tradeoff 
in terms of balancing the model complexity
and the quality of fit should be achieved. 
There are two aspects concerning the determination of suitable
number of mixture components: 
\begin{enumerate}
\item a scoring function to evaluate a given mixture
\item a search strategy to infer such a mixture
\end{enumerate}

\noindent \emph{Scoring function:} 
There have been numerous scoring functions proposed in the literature
that aim to evaluate a given mixture model. 
A review of these methods is presented in \citet{mclachlan2000finite}.
The common motivation is to
balance the model complexity and the quality of fit.
The scoring functions which quantify the model complexity based 
on the number of components are Akaike Information Criterion (AIC) \citep{aic}, 
Bayesian Information Criterion (BIC) \citep{bic,rissanen1978modeling},
and Integrated Completed Likelihood (ICL) criterion \citep{icl}.
It is to be noted that AIC and BIC are shown to be approximations of the 
general MML framework \citep{figueiredo2002unsupervised}.
The information-theoretic criteria that account for not just the
number of components but also the components' parameters 
are \emph{ICOMP} \citep{icomp}, 
\emph{Laplace empirical criterion} (LEC) \citep{roberts}, and
\emph{approximated MML criterion} \citep{oliver1996unsupervised,figueiredo2002unsupervised}.
Further, these criteria are derived using a MML interpretation.
However, as detailed in \citet{multivariate_vmf},
these criteria are oversimplified versions of the generic MML framework
and are incomplete in objectively addressing the tradeoff associated
with selecting a suitable mixture model.\\

\noindent \emph{Search strategy:} 
To determine the optimal number of mixture components using the 
aforementioned criteria 
\citep{aic,bic,oliver1996unsupervised,roberts,icl},
mixtures are inferred for varying number
of components using the EM algorithm, and the mixture that has the least score is treated
as the optimal one. 
As the EM only guarantees convergence to a local optimum,
a few trials are conducted with different starting points in an effort
to minimize the possibility of getting trapped in a local optimum
\citep{krishnan1997algorithm,mclachlan2000finite}.
In order to rectify the issues arising from the use of EM method
which plays a central role in identifying the right mixture model,
methods based on iteratively splitting and merging constituent mixture 
components have been proposed so as to enable the intermediate
mixtures to escape from local optima. The notable amongst these
are \emph{split-merge} based EM (SMEM) method proposed by \citet{ueda2000smem}
and component-deletion based unsupervised learning approach 
proposed by \citet{figueiredo2002unsupervised}. 

Given a mixture with $K$ components, the SMEM method 
selects the top three candidates, merges two of them, and splits the other into
two, thus, leaving the effective number of components unchanged. 
Further, the potential candidates are chosen depending on the improvement
to the complete data log-likelihood function (used to formulate the ICL criterion).
In contrast, the method of \citet{figueiredo2002unsupervised} 
starts off by assuming a large number of components and iteratively
eliminates those that are deemed redundant as per their
objective function (a simplified MML-like formulation).
\citet{figueiredo2002unsupervised} demonstrated their competitive edge 
against the contemporary BIC \citep{bic}, LEC \citep{roberts},
and ICL criterion \citep{icl}.

The SMEM algorithm does not facilitate an increase or decrease in the mixture size.
In contrast, the method of \citet{figueiredo2002unsupervised} 
progressively reduces the mixture size, and hence, has no provision for
recovering a component if it is deleted by chance. Also, the assumptions
made in formulating their MML-like scoring function lack
the rigor to objectively weigh the mixture model complexity against
the quality of data fit, as explained in \citet{multivariate_vmf}.
In order to address the limitations resulting from approximating the scoring functions 
and the search strategies, 
more recently, \citet{multivariate_vmf} proposed a search heuristic
in conjunction with a comprehensive MML formulation (with no approximations) 
to infer a suitable mixture model. 
This was demonstrated in the context of inference of mixtures of multivariate
Gaussian and vMF distributions. 
In our previous work \citep{multivariate_vmf},
we have established that the proposed approach outperforms
the widely used method of \citet{figueiredo2002unsupervised}.

\subsection{The optimal number of components of a \fb-component mixture}
\label{subsec:search_method}
We briefly review the search method of \citet{multivariate_vmf}
here that extended the MML-based Snob program \citep{wallace68,wallace1986improved} 
for unsupervised learning.
The method is adapted to the present scenario of mixture modelling of 
\fb~distributions. The general idea is to perform a series of perturbations 
(split, delete, and merge operations) to a current sub-optimal mixture
to obtain an improved mixture with a lower message length. The method begins by
assuming a one-component mixture. The mixture is split into two children
which are locally optimized. If the resultant mixture has a lower message length,
the current mixture is updated. If, at any stage, a mixture has $K$
components, each component is separately split into two, deleted, and merged with
an appropriate component. The split operation results in a $(K+1)$-component
mixture, while the delete and merge operations result in $(K-1)$-component mixtures.
Each of the intermediate mixtures are optimized using an EM algorithm
(Section~\ref{subsec:em_mml}).
The perturbation corresponding to a component
that results in the greatest reduction in message length is considered.
This heuristic exhaustively considers all possible operations 
giving the $K$-component mixture the best chance to
escape from a sub-optimal state. The method terminates when none of the
perturbations result in improved mixtures. Each of these operations
are explained below in the context of \fb~distributions.

\subsubsection{Splitting a component}
\label{subsubsec:splitting}
The split operation is critical as it leads to mixtures with greater
number of components. It is not desirable to have overly complex mixtures
unless required. While splitting a (parent) component, the initial means of the
two children should be reasonably apart so that they 
form distinct components and the $K$-component mixture
has the best chance to move from a sub-optimal state to a
more optimal state (if one exists). After the inital means are chosen,
an EM is carried out just on the two child components until they are
stabilized, leaving the remaining $(K-1)$-components unchanged.
After optimizing the children, they are then integrated with the original
$K-1$ components and an EM is subsequently performed on the $K+1$
components to reach an optimal state. If the new $(K+1)$-component mixture 
results in a lower total message length, that implies the perturbation
of the $K$-component mixture resulted in an improved mixture. \\

\noindent\emph{Selection of initial means of the two child components:}
In the case of Gaussian distributions, \citet{multivariate_vmf}
chose the initial means 
such that they are one standard deviation away on either side of the
component mean along the direction of maximum variance. 
In the present work, for directional distributions (vMF and \fb) defined on the 
three-dimensional spherical surface, we provide an analogous form.
As described in Section~\ref{subsec:moment}, the procedure for moment estimation
of the major and minor axes of a \fb~distribution involves the eigenvalue
decomposition of the matrix $\mathbf{B}_L$, 
the submatrix derived from the dispersion matrix $\mathbf{B}$. 
If $l_1$ and $l_2$ are the eigenvalues of $\mathbf{B}_L$ 
(Equation~\ref{eqn:moment_psi_est}), then $l_1,l_2$ are roots of the
characteristic equation:
\begin{equation*}
l^2 - (b_{22}+b_{33})l + b_{22}b_{33}-b_{23}^2 = 0
\quad\text{so that}\quad
l_1 + l_2 = b_{22}+b_{33}
\end{equation*}
According to Equation~\ref{eqn:size_shape}, we have $l_1 - l_2 = r_2$,
and hence, $l_1 = (b_{22}+b_{33}+r_2)/2$.
The maximum variance is along the direction of major axis and is equal to 
the eigenvalue $l_1$.
Hence, one standard deviation would correspond to $\sqrt{l_1}$.
It is to be noted that these calculations are done in the 
$\mathbf{X}_2\mathbf{X}_3$ plane which contains the major and minor axes
(that is, after the mean of the parent, as part of moment estimation,
is aligned with $\mathbf{X}_1$).
However, it is now required to map this point 
back onto the \emph{unit} sphere.

Consider Figure~\ref{fig:setup_spherical}(a) where $\major''$ and $\minor''$ are the major
and minor axes in the $\mathbf{X}_2\mathbf{X}_3$ plane respectively.
The mean axis $\mean''$ of the parent component
being split is aligned with $\mathbf{X}_1$.
The segment OP is of length $\sqrt{l_1}$ corresponding to unit
standard deviation along $\major''$. Let $M_1$ be the mean of one of the children.
Then, for $M_1$ such that $M_1P$ is perpendicular to the 
$\mathbf{X}_2\mathbf{X}_3$ plane, we have $M_1P = \sqrt{1-l_1}$
(as $OM_1 = 1$ is the radius of the sphere). If $\theta\in[0,180^{\circ}]$ measures
the co-latitude of the mean $M_1$ as shown, we have 
$\theta = \arccos\sqrt{1-l_1}$. The mean $M_2$ (not shown in the figure) 
of the second child component
lies in the plane containing $OM_1P$ such that the angle between 
$OM_2$ and $OX_1$ is $\theta$. The two means are then transformed in order
to conform with the axes of the parent \fb~component.
With these as starting points for the EM algorithm, the two child components
are locally optimized. The children along with the untouched $(K-1)$-components
serve as a starting point for estimating the parameters of the $(K+1)$-component
mixture using the EM algorithm. 

\subsubsection{Deletion of a component}
While deleting a component, its memberships are adjusted by
proportionally distributing among the remaining $K-1$ components. With this
new starting point, the parameters of the $(K-1)$-component mixture
are estimated using an EM algorithm.

\subsubsection{Merging two components}
\label{subsubsec:merging}
The choice of merging a pair of components is determined by their closeness.
To identify the closest component, \citet{multivariate_vmf} compute the 
Kullback-Leibler (KL) divergence \citep{kullback1951information} of the component 
in consideration with the remaining
$K-1$ components in the mixture. The chosen pair is then merged 
to form a single component whose initial weight and memberships are
given by the sum of the individual components' weights and memberships respectively.
This acts as the starting point for the EM algorithm to estimate the
parameters of the merged $(K-1)$-component mixture. \\

\noindent\emph{Kullback-Leibler divergence of \fb~distributions:}
The analytical form of the KL divergence between two \fb~distributions is derived below.
The KL divergence between two probability distributions $f_a$ and $f_b$ is defined as 
\begin{equation*} 
D_{KL}(f_a||f_b)
= \expect_a\left[\log\frac{f_a(\boldx)}{f_b(\boldx)} \right] 
\end{equation*}
where $\expect_a[.]$ is the expectation of the quantity $[.]$ using $f_a$.

Let $f_a(\boldx) = \fb(\kappa_a,\beta_a,\mathbf{Q}_a)$ 
and $f_b(\boldx) = \fb(\kappa_b,\beta_b,\mathbf{Q}_b)$ 
be two distributions such that $\mathbf{Q}_a=(\meana,\majora,\minora)$
and $\mathbf{Q}_b=(\meanb,\majorb,\minorb)$.
Let $c_a$ and $c_b$ be the respective normalization constants. Then,
\begin{align}
\expect_a\left[\log\frac{f_a(\boldx)}{f_b(\boldx)}\right] &= 
\log\frac{c_b}{c_a}
+ (\kappa_a\meana\trans - \kappa_b\meanb\trans)\,\expect_a[\boldx] \notag\\
&+ \beta_a \,\majora\trans \,\expect_a[\boldx\boldxt]\, \majora
- \beta_b \,\majorb\trans \,\expect_a[\boldx\boldxt]\, \majorb \notag\\
&- \beta_a \,\minora\trans \,\expect_a[\boldx\boldxt]\, \minora
+ \beta_b \,\minorb\trans \,\expect_a[\boldx\boldxt]\, \minorb 
\label{eqn:kldiv}
\end{align}
gives the analytical form of the KL divergence of two \fb~distributions. 
The expressions for $\expect_a[\boldx]$ and $\expect_a[\boldx\boldxt]$ are
derived in Equation~\ref{eqn:expectation_generic}.

\begin{figure}[h]
\centering
\subfloat[]{\includegraphics[width=0.48\textwidth]{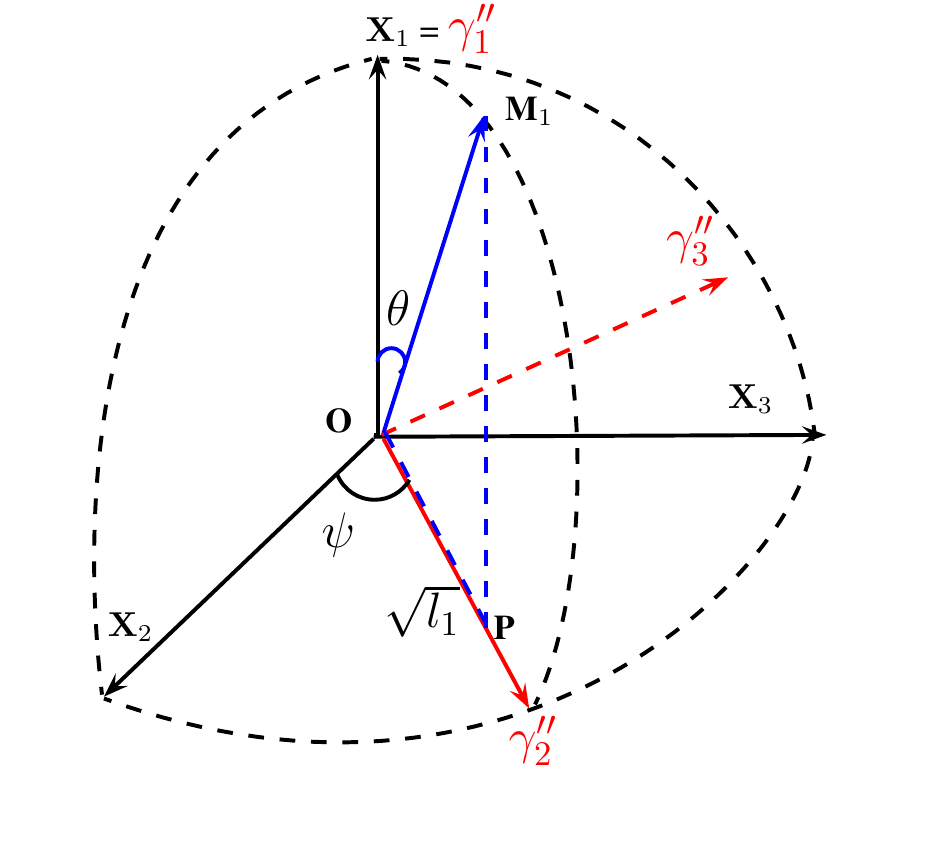}}\quad
\subfloat[]{\includegraphics[width=0.45\textwidth]{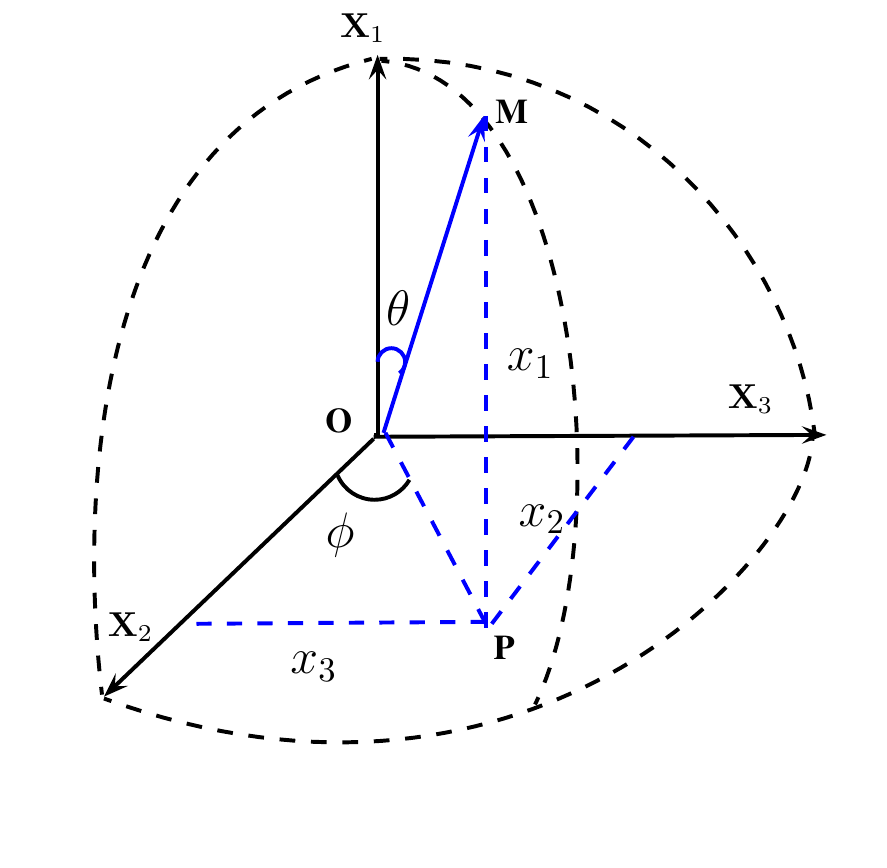}}
\caption{
         (a) Selection of initial means of children during splitting a parent 
             component (see Section~\ref{subsubsec:splitting}).
         (b) Transformation between spherical and Cartesian coordinates.
        }
\label{fig:setup_spherical}
\end{figure}
Through these perturbations, the search method aims to find an optimal state by leveraging
information about the sub-optimal state. In doing so, we are cautiously splitting,
deleting, or merging potential candidate components. The heuristic attempts
to find improved mixtures without compromising the optimality of the intermediate
solution.

\subsection{An illustrative example of the search procedure}
\label{subsec:mix_example}
The mechanics of the inference of a suitable \fb~mixture model has been explained
previously in Section~\ref{subsec:search_method}. For further details of the search method, 
we refer the reader to \citet{multivariate_vmf}. 
To better illustrate the search process, this subsection presents
a detailed example.

\begin{figure}[ht]
\centering
\subfloat[]{\includegraphics[width=0.55\textwidth]{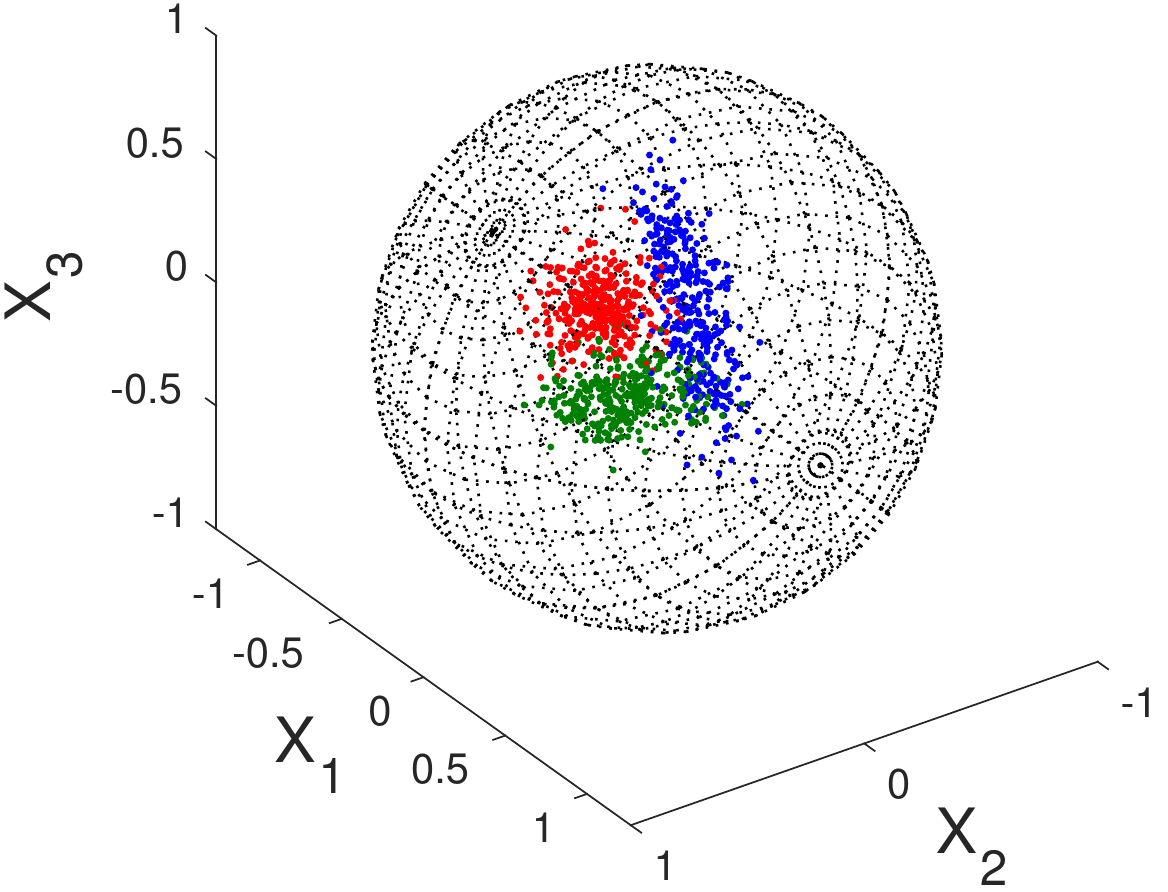}}\quad
\subfloat[]{\includegraphics[width=0.40\textwidth]{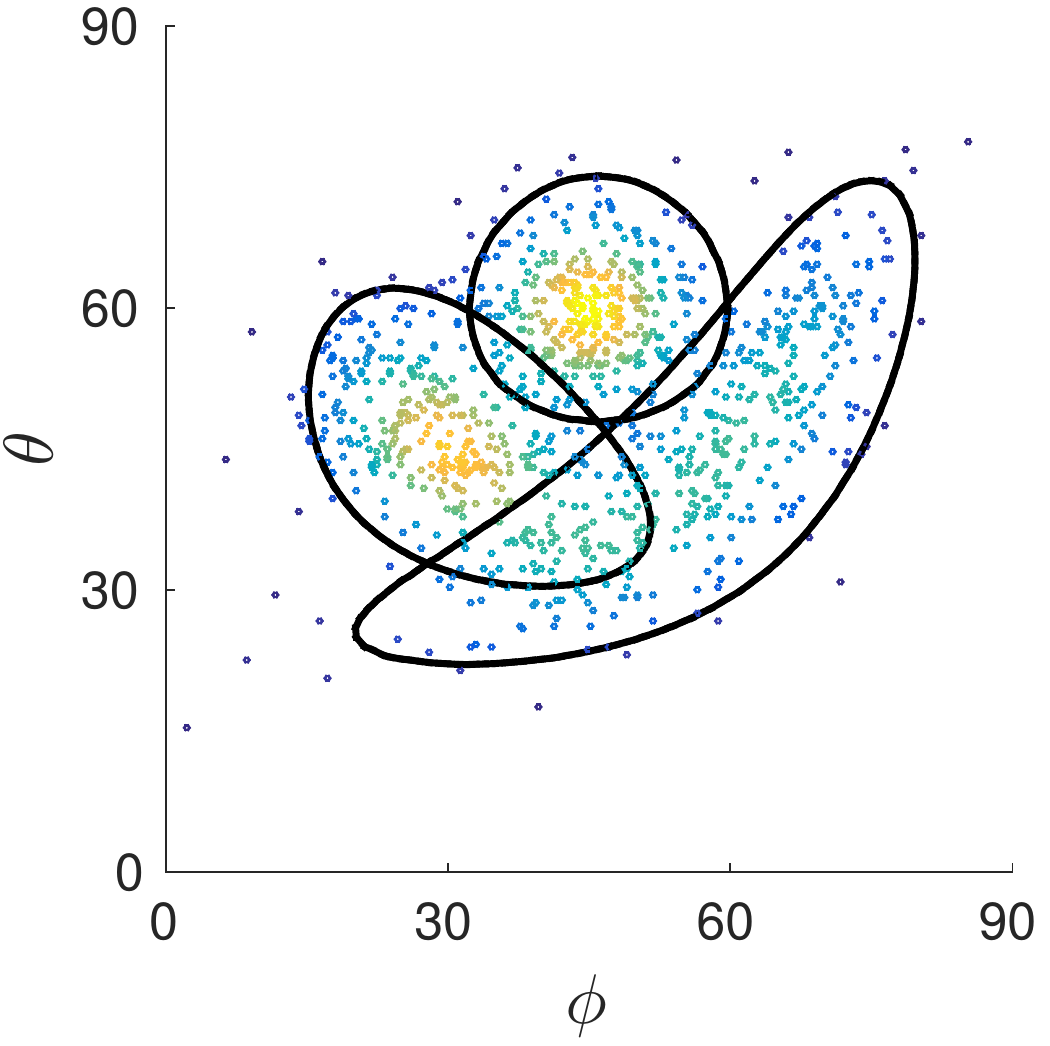}}
\caption{Original mixture consisting of three components with equal weights and $\kappa=100$.
(a) individual components with varying eccentricities: $e = 0.1$ (red), $e = 0.5$ (green), and $e = 0.9$ (blue)
(b) simulated data plotted in degrees in the $\theta\phi$ space (contours encompass 90\% of the data),
} 
\label{fig:original_mix}
\end{figure}

Consider a mixture with three \fb~components (Figure~\ref{fig:original_mix})
that have equal mixing proportions, the same concentration parameter $\kappa=100$
and different eccentricities.
The red component has eccentricity $e=0.1$ and the angular parameters 
defining its axes are $(\psi,\alpha,\eta)=(0,60^{\circ},45^{\circ})$.
The green component has $e=0.5$ and $(\psi,\alpha,\eta)=(150^{\circ},45^{\circ},30^{\circ})$.
The blue component has $e=0.9$ and $(\psi,\alpha,\eta)=(30^{\circ},45^{\circ},60^{\circ})$.
The parameters are chosen such that the components are close to each other.
A sample of size $N=1000$ was generated from the mixture using the method
of \citet{kent2013new}. 
The mixture density is shown as 
a heat map in Figure~\ref{fig:original_mix}(b). 
For ease of visualization, the density is represented in $\theta\phi$ space, 
where $\theta$ is the co-latitude and $\phi$ is the longitude
(Figure~\ref{fig:setup_spherical}(b)).
The Cartesian coordinates of each datum $\boldx=(x_1,x_2,x_3)$ in the sampled data 
are transformed into the spherical coordinates defined by unit radius, 
co-latitude, and longitude. The transformation\footnote{It is to be noted that 
transforming data generated from a \fb~distribution
(that has elliptical contours on the spherical surface)
and representing in the $\theta\phi$ space produces shapes that do not
have any decipherable pattern as can be seen through this example.}
is effected by:
\begin{equation*}
x_1 = \cos\theta, \quad x_2 = \sin\theta\,\cos\phi, \quad x_3 = \sin\theta\,\sin\phi
\end{equation*}

\subsubsection{The seach method explained}
\label{subsubsec:search_method_explained}
\begin{figure}[ht]
\centering
\subfloat[$\fancym_1$ ($I=19364$ bits)]{\includegraphics[width=0.33\textwidth]{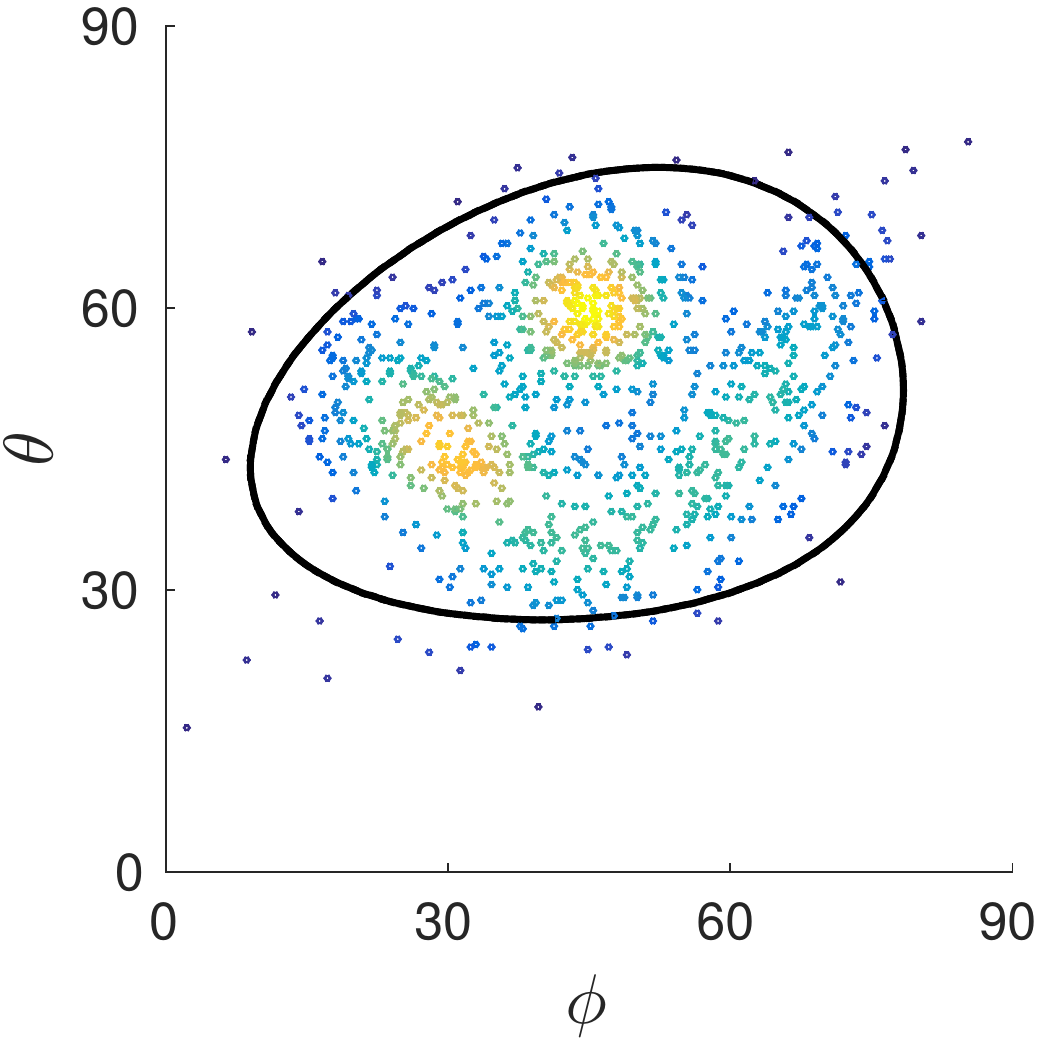}}
\subfloat[Initialization of means]{\includegraphics[width=0.33\textwidth]{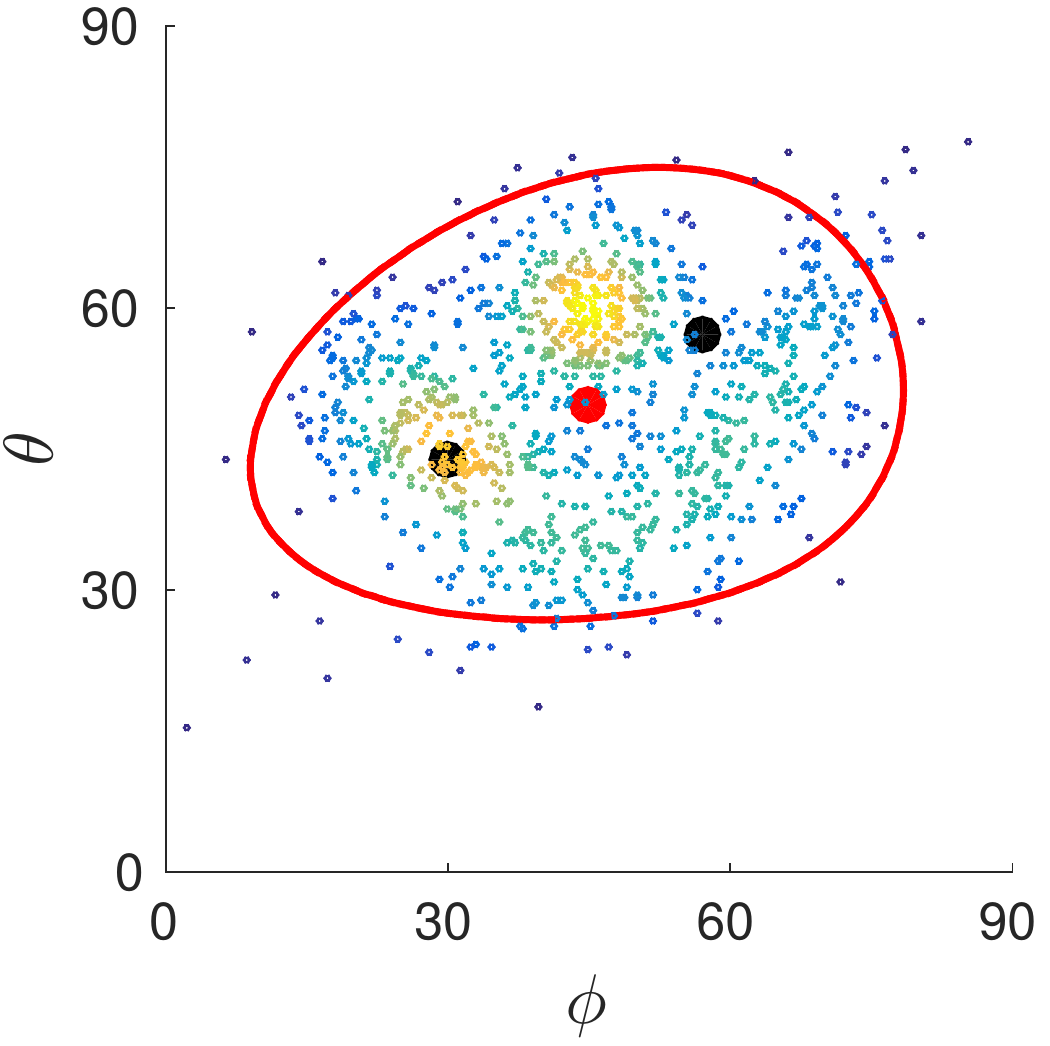}}
\subfloat[$\fancym_2$ ($I=19319$ bits)]{\includegraphics[width=0.33\textwidth]{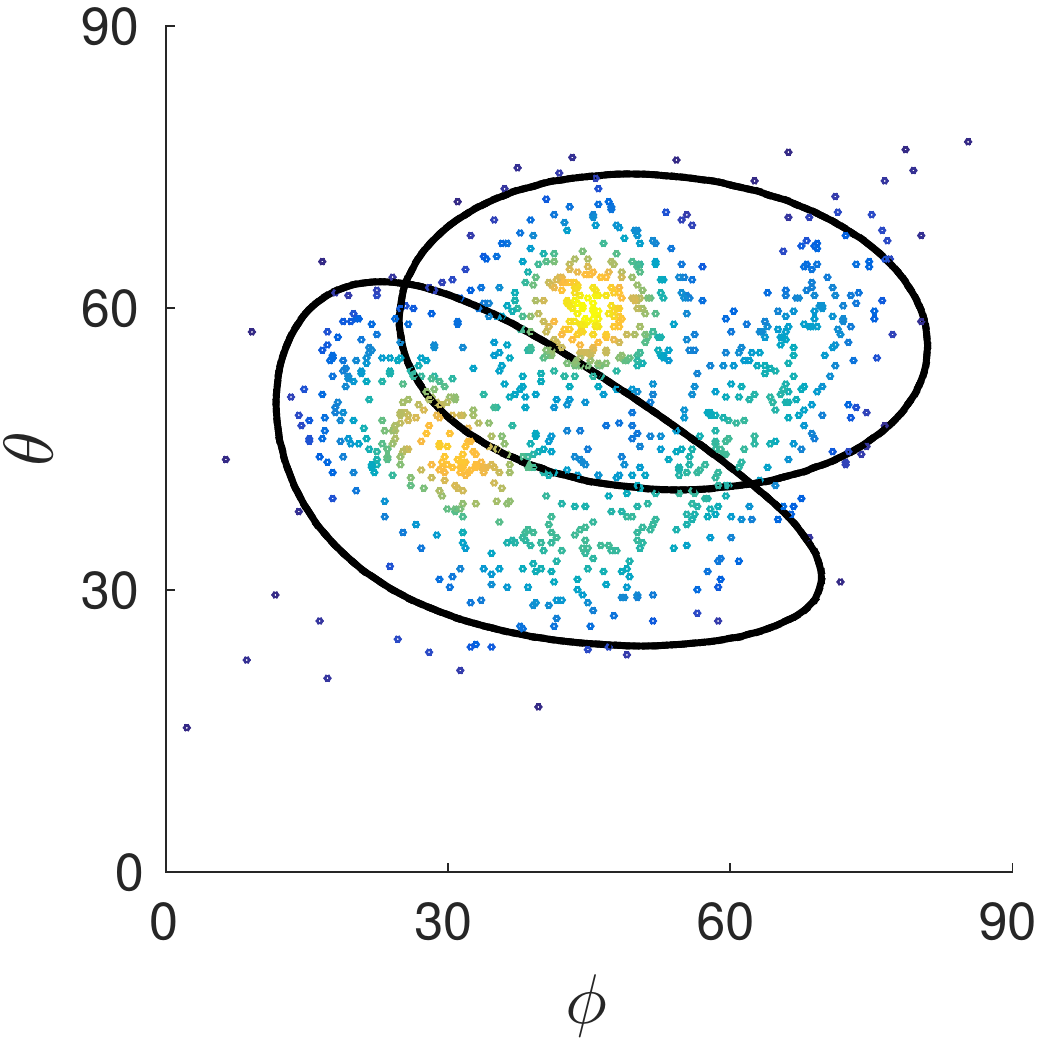}}
\caption{Iteration 1 (a) one-component mixture, 
(b) Red colour denotes the parent component being split, the red dot indicates the mean
of the parent and the black dots (on either side) indicate the initial means of the children,
(c) improved mixture.
\label{fig:mix_iter1}
} 
\end{figure}

The search begins by inferring a one-component mixture $\fancym_1$ (Figure~\ref{fig:mix_iter1}a).
It has an associated message length of $I=19364$ bits.
Before splitting the component, the means of the children are initialized
as shown in Figure~\ref{fig:mix_iter1}(b). These means are determined as explained in
Section~\ref{subsubsec:splitting}. The children are optimized
using the EM algorithm to generate the two-component mixture $\fancym_2$
(Figure~\ref{fig:mix_iter1}c). $\fancym_2$ has a message length of $I=19319$ bits,
and hence improves $\fancym_1$ by 45 bits.

\begin{figure}[ht]
\centering
\subfloat[Splitting $P_1$ ($I=19319$)]{\includegraphics[width=0.33\textwidth]{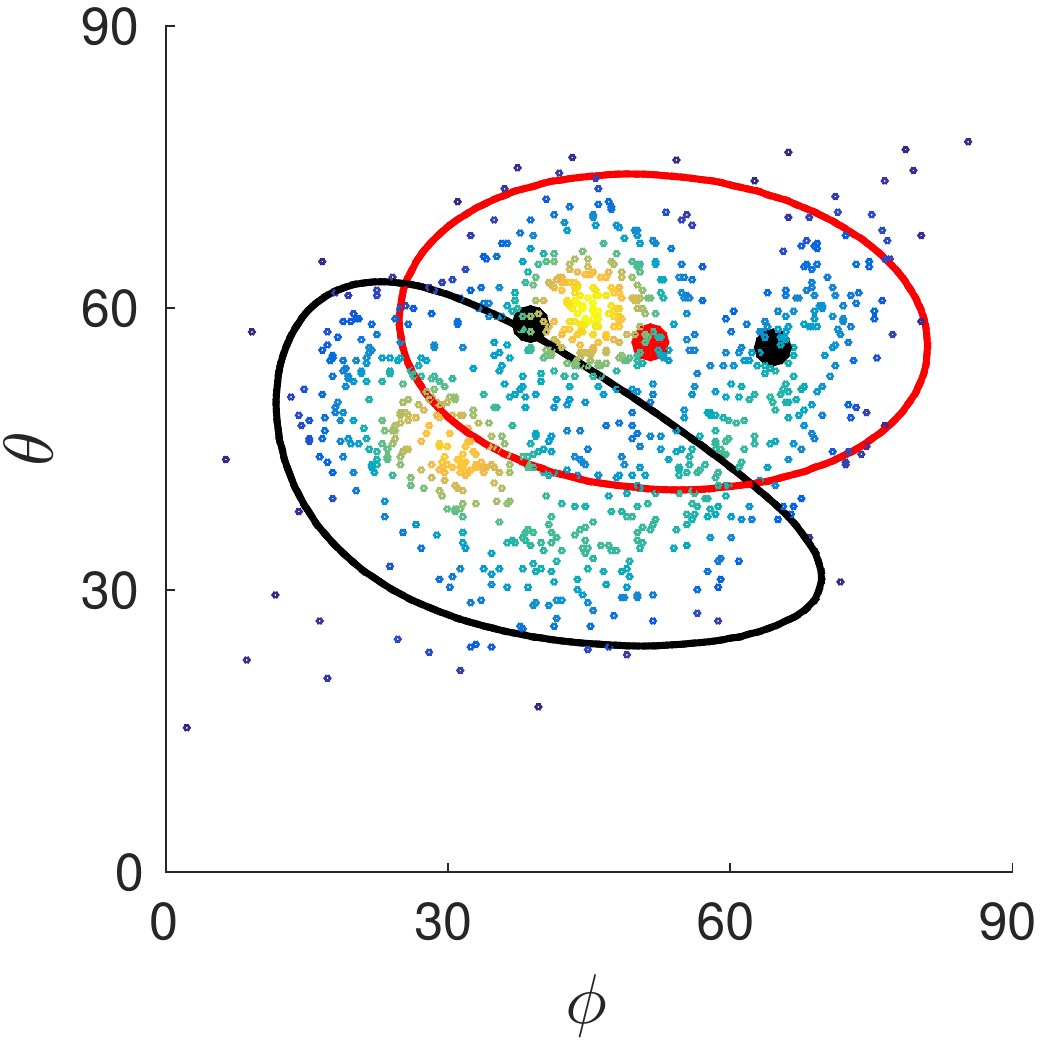}}
\subfloat[Optimized children ($I=19259$)]{\includegraphics[width=0.33\textwidth]{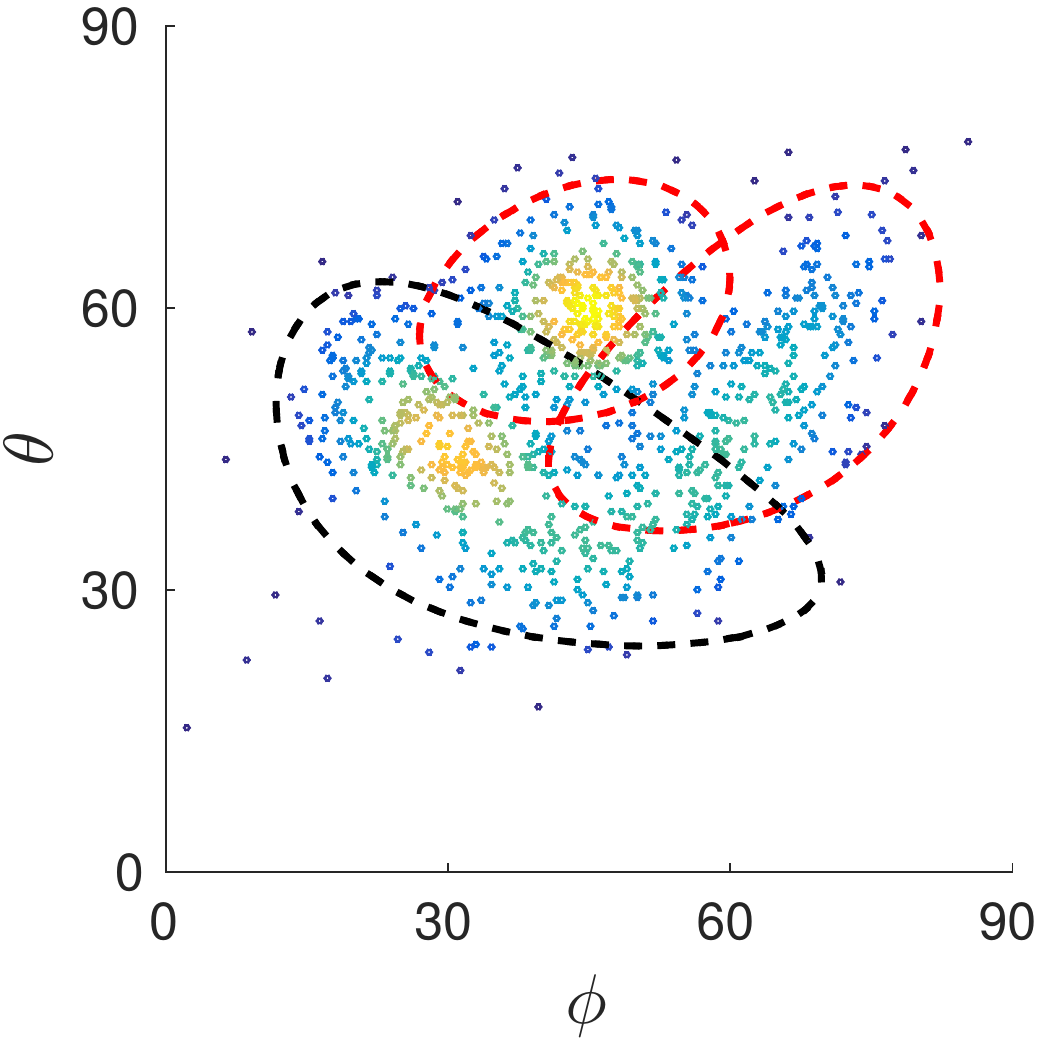}}
\subfloat[$\fancym_3:$ post-EM ($I=19143$)]{\includegraphics[width=0.33\textwidth]{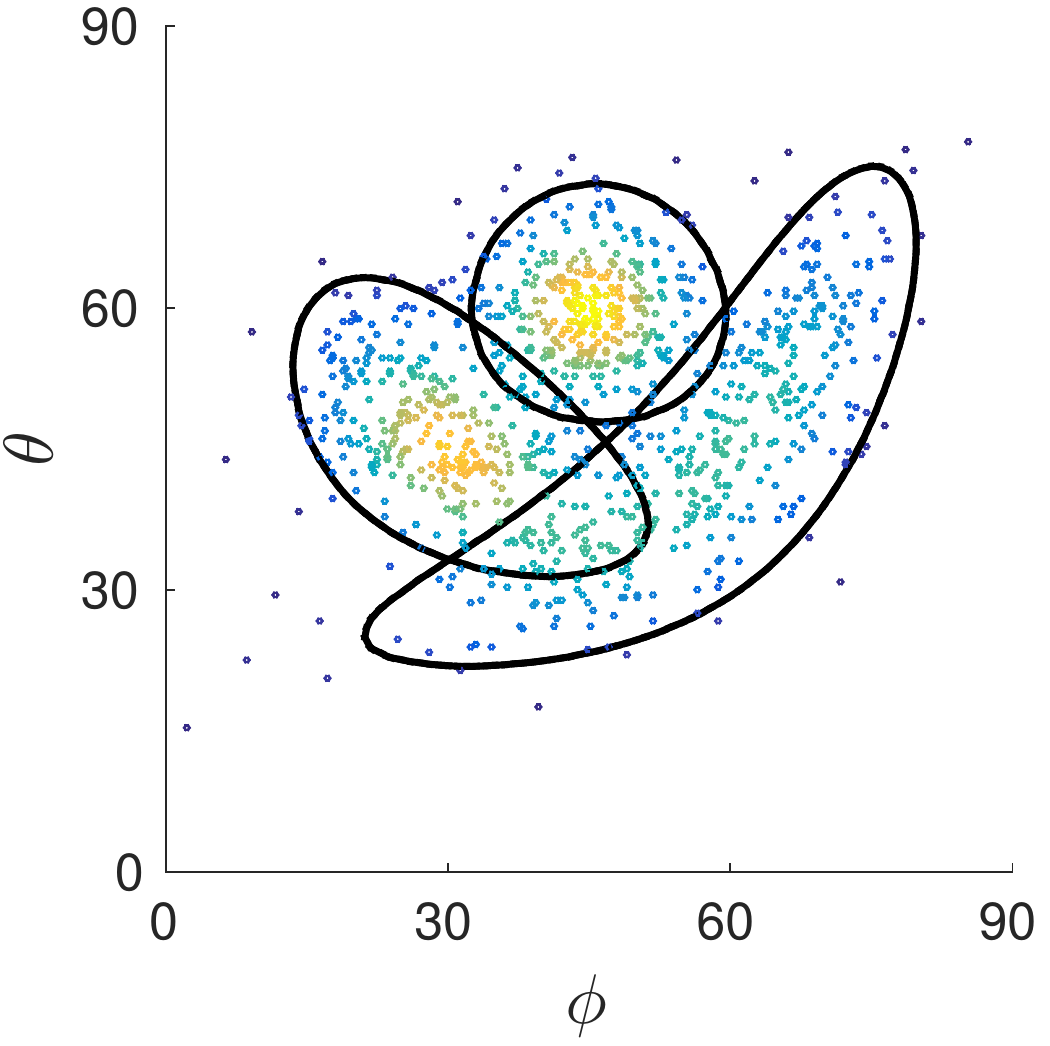}}\\
\subfloat[Splitting $P_2$ ($I=19319$)]{\includegraphics[width=0.33\textwidth]{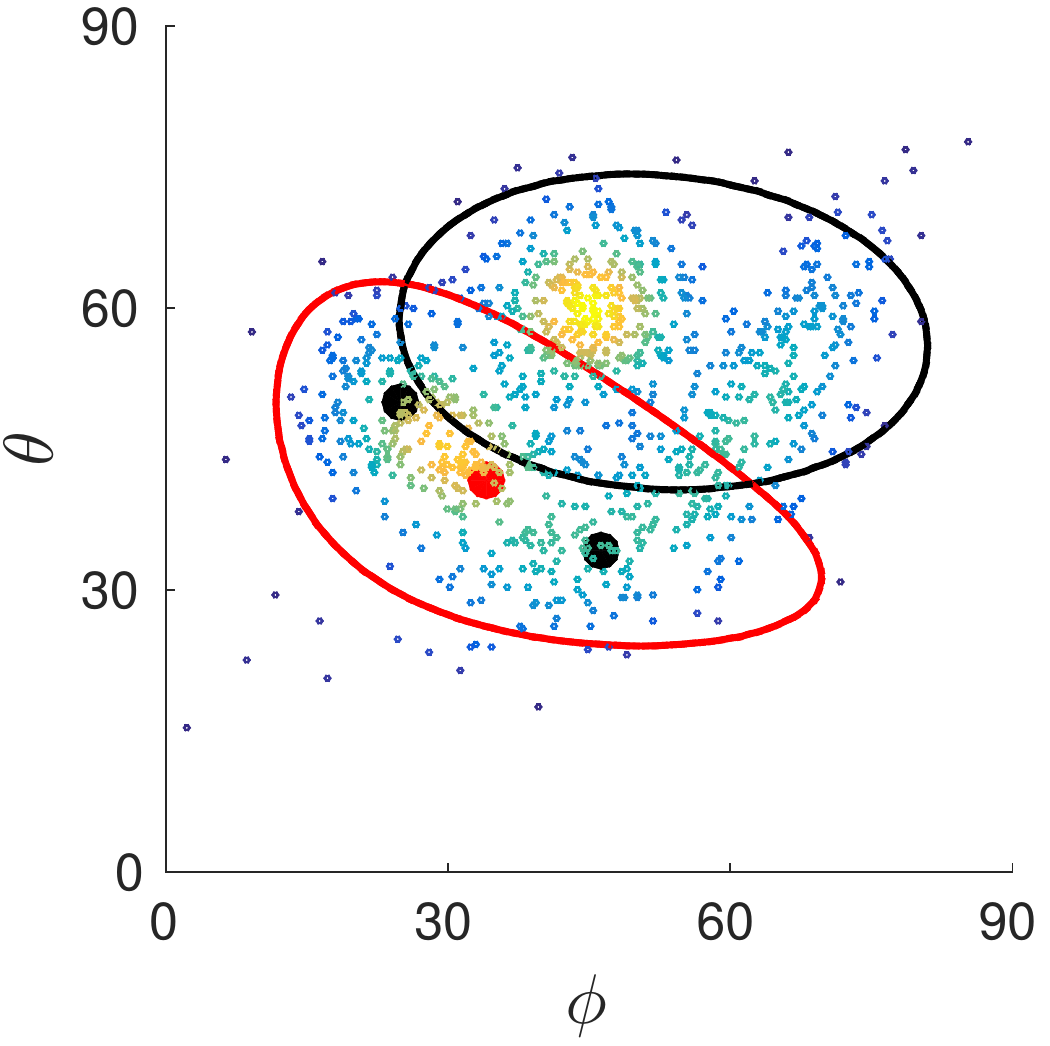}}
\subfloat[Optimized children ($I=19333$)]{\includegraphics[width=0.33\textwidth]{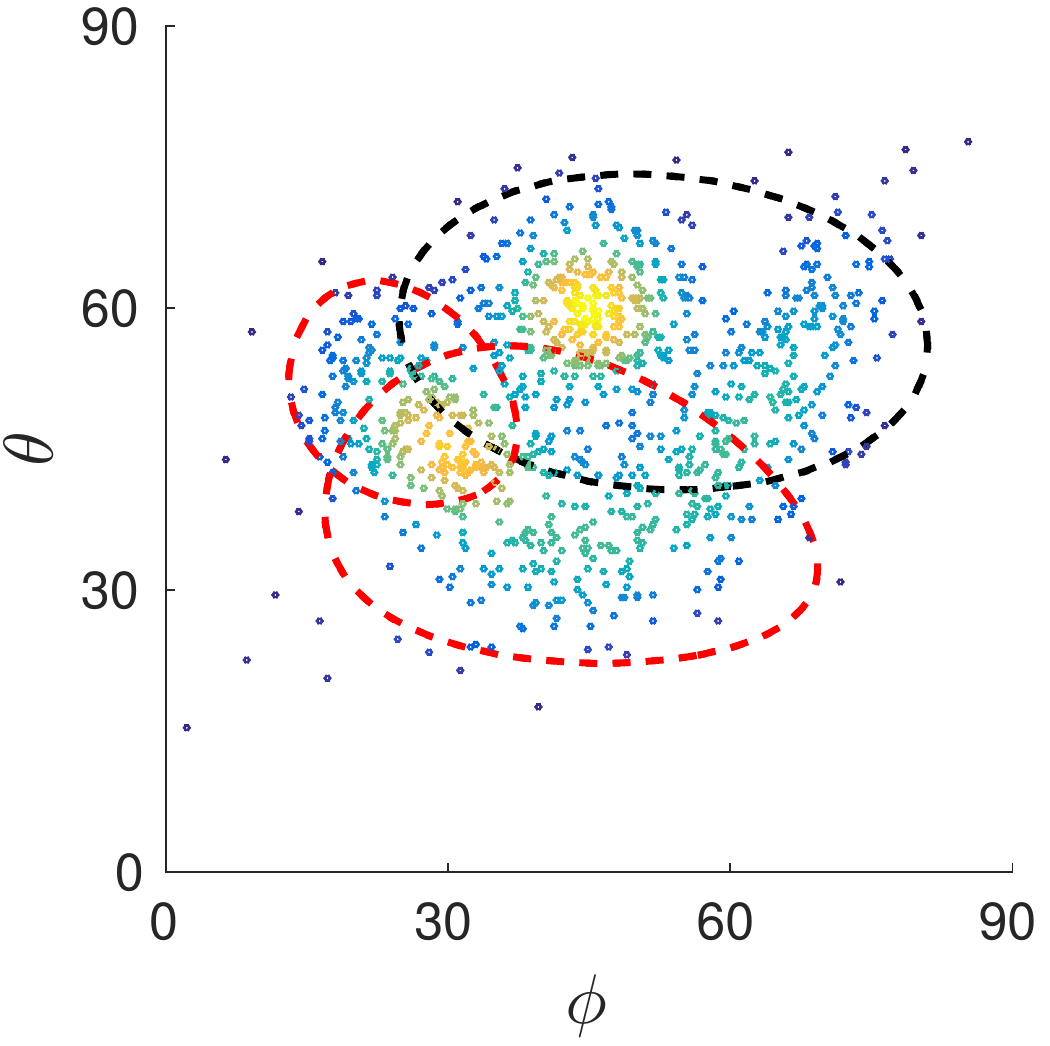}}
\subfloat[$\fancym_3:$ post-EM ($I=19143$)]{\includegraphics[width=0.33\textwidth]{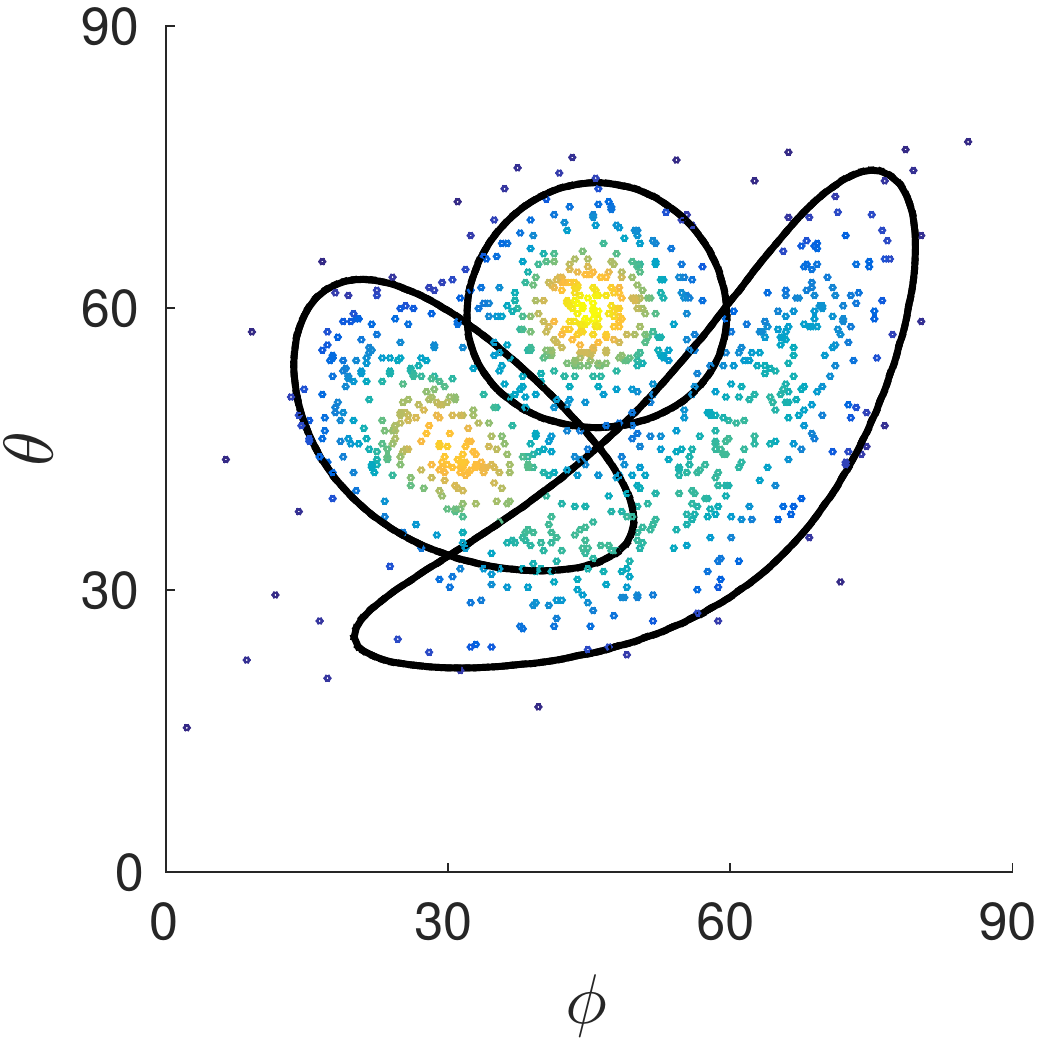}}
\caption{Iteration 2 -- \emph{Split} operations.
(a)-(c) splitting the first component $P_1$ in $\fancym_2$, and
(d)-(f) splitting the second component $P_2$ in $\fancym_2$.
The red dashed lines in (b),(e) represent the optimized children (prior to integration),
and black dashed lines in (b),(e) represent the unchanged components.
}
\label{fig:mix_iter2_split}
\end{figure}

\begin{figure}[ht]
\centering
\subfloat[Deleting $P_1$ ($I=19319$)]{\includegraphics[width=0.33\textwidth]{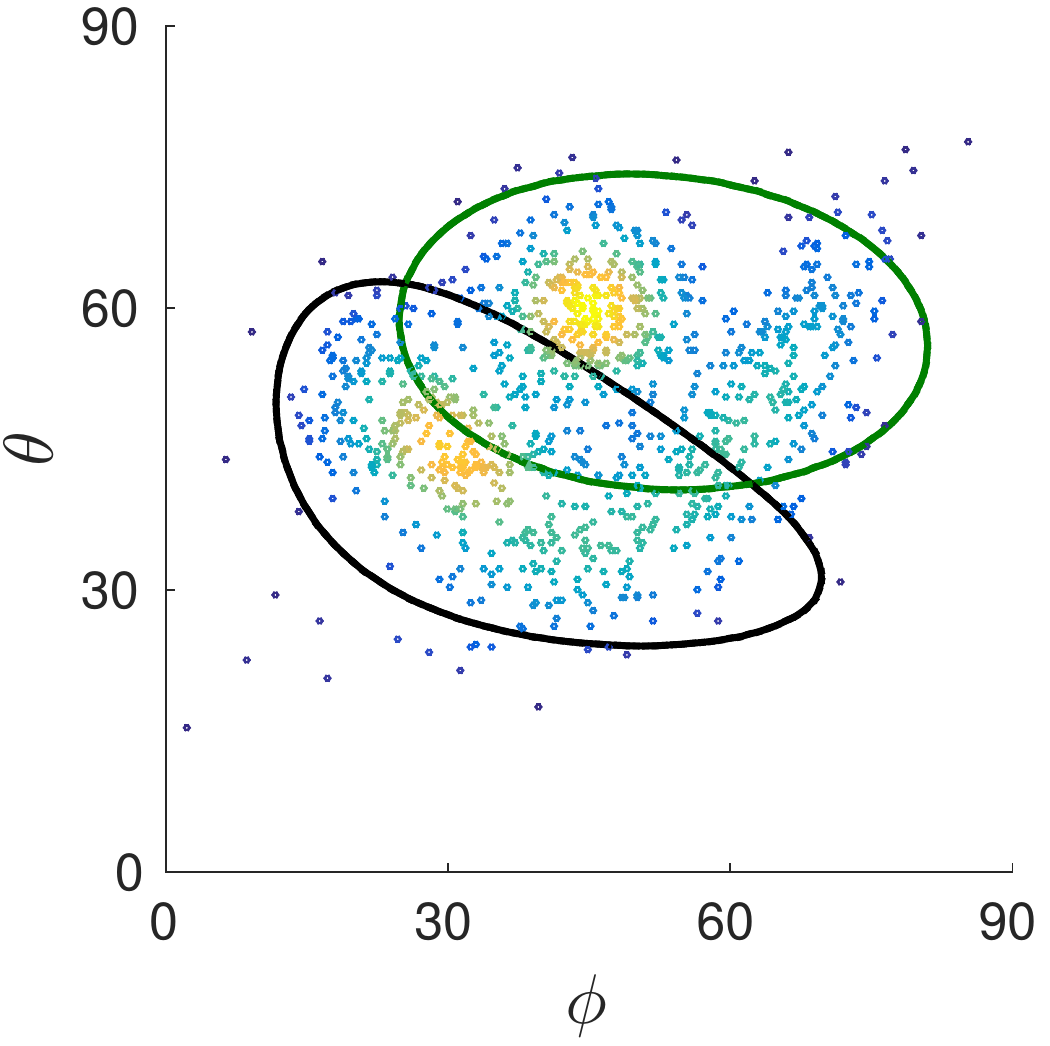}}
\subfloat[Before optimizing ($I=23019$)]{\includegraphics[width=0.33\textwidth]{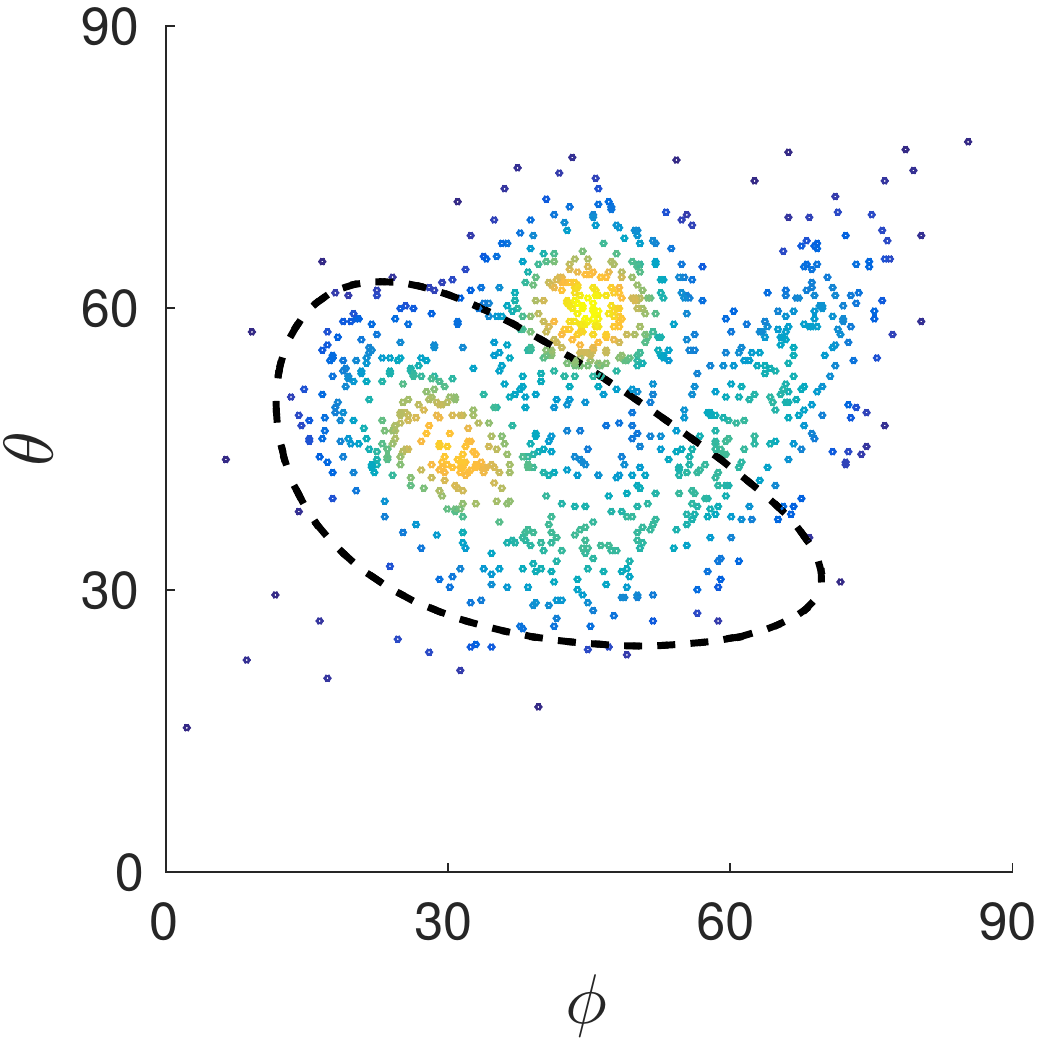}}
\subfloat[post-EM ($I=19364$)]{\includegraphics[width=0.33\textwidth]{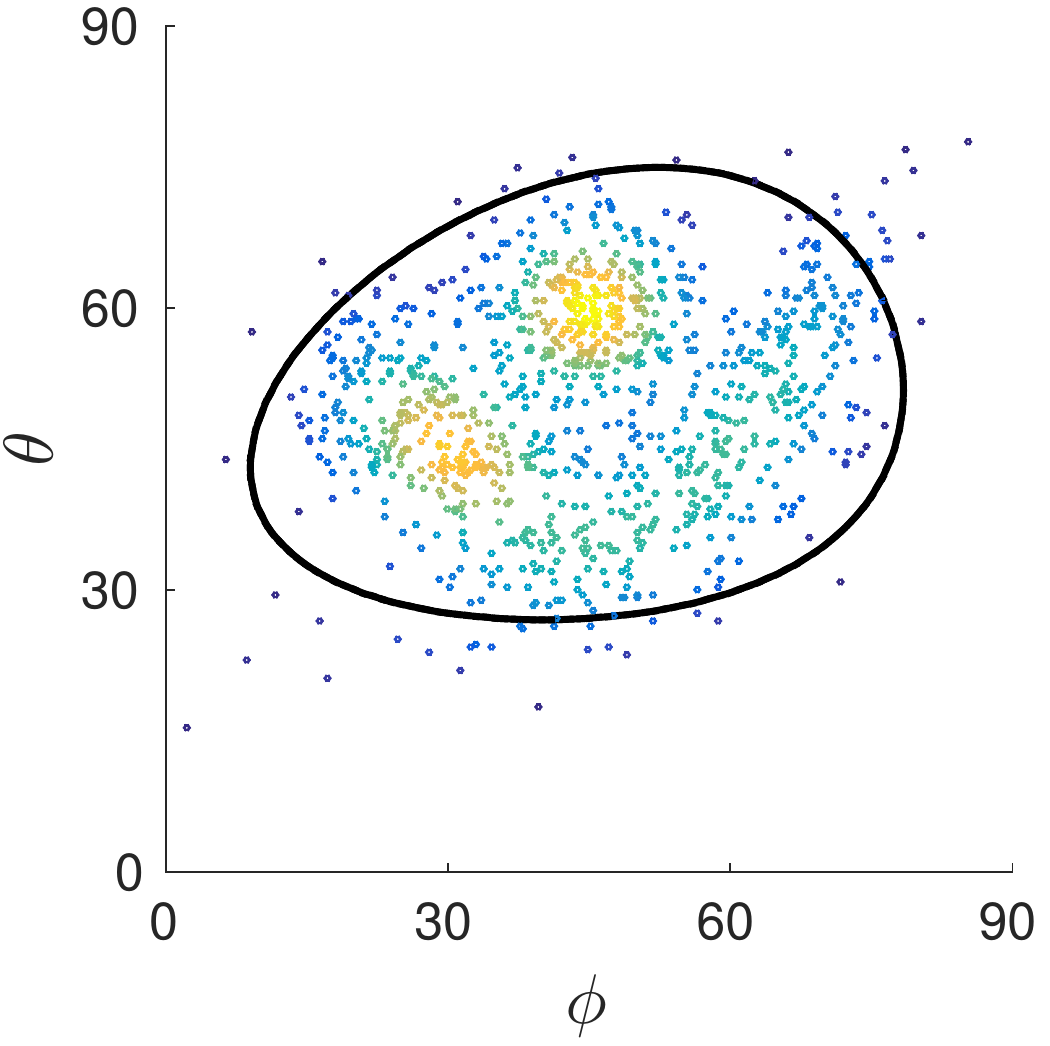}}\\
\subfloat[Deleting $P_2$ ($I=19319$)]{\includegraphics[width=0.33\textwidth]{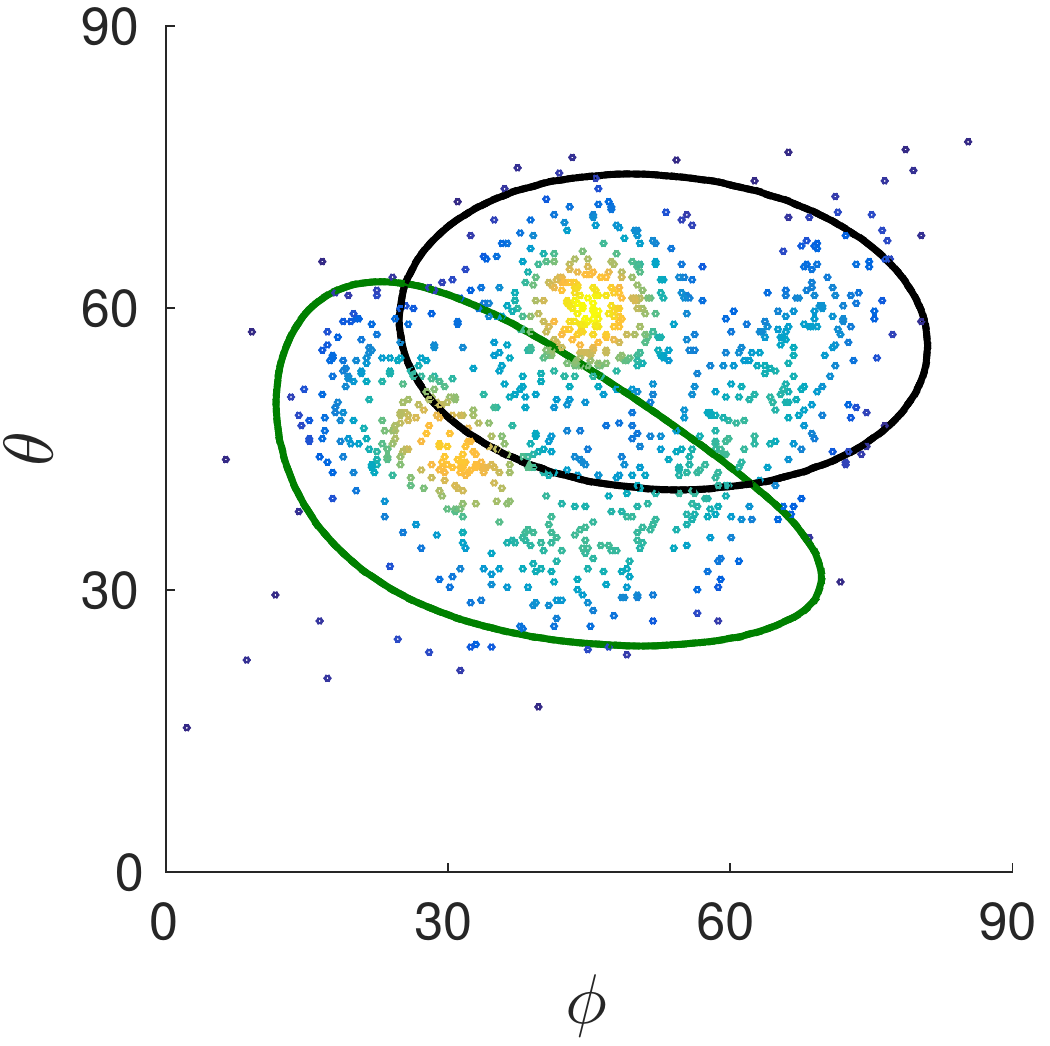}}
\subfloat[Before optimizing ($I=20416$)]{\includegraphics[width=0.33\textwidth]{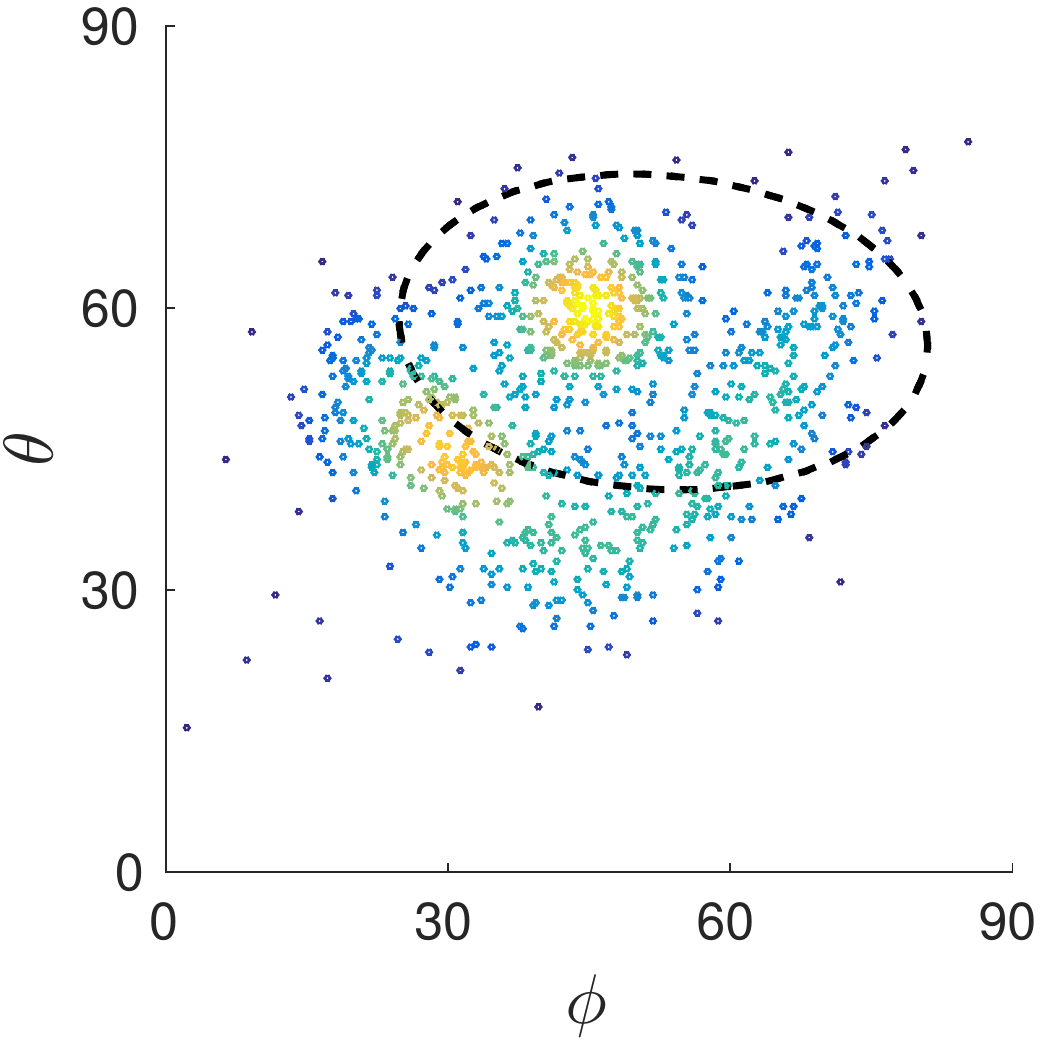}}
\subfloat[post-EM ($I=19364$)]{\includegraphics[width=0.33\textwidth]{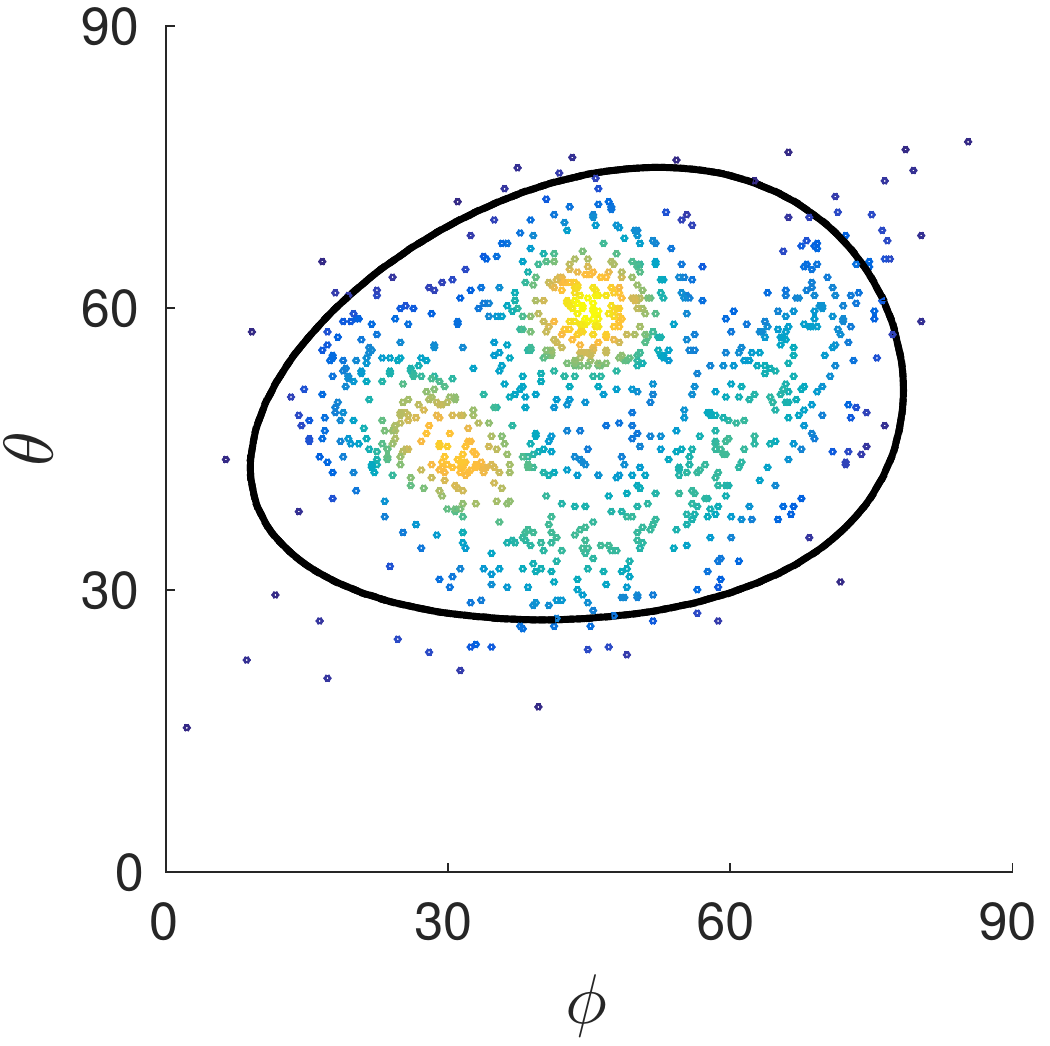}}\\
\subfloat[Merging $P_1$ and $P_2$ ($I=19319$)]{\includegraphics[width=0.33\textwidth]{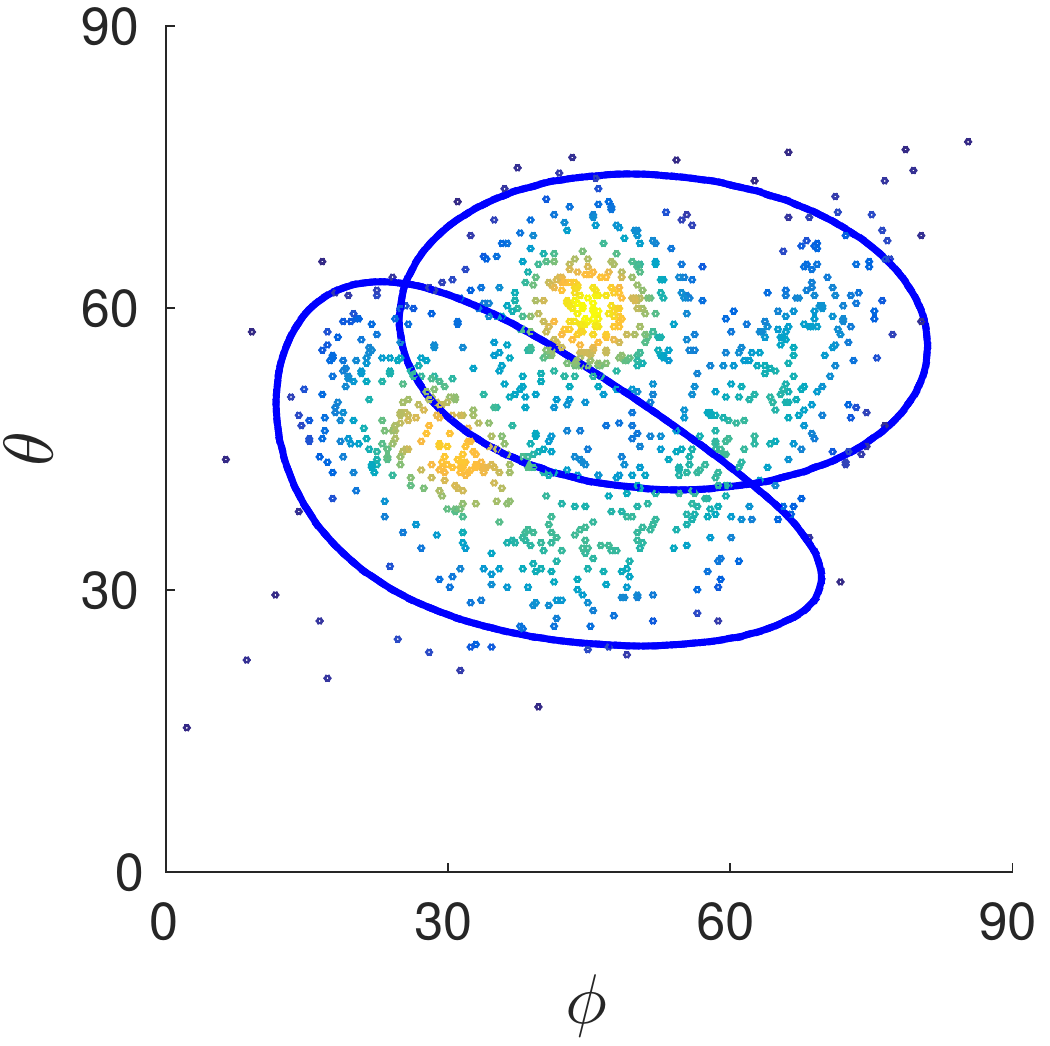}}
\subfloat[Before optimizing ($I=19364$)]{\includegraphics[width=0.33\textwidth]{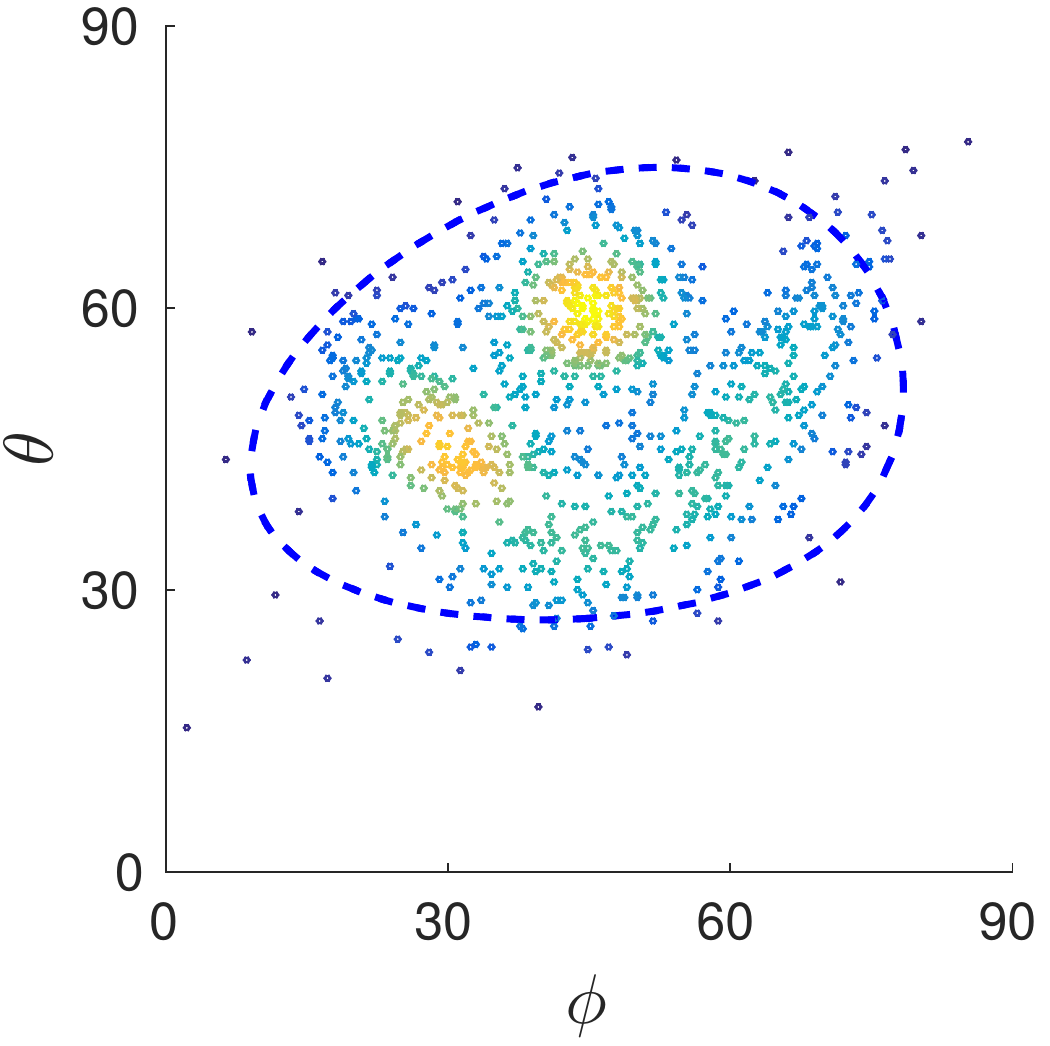}}
\subfloat[post-EM ($I=19364$)]{\includegraphics[width=0.33\textwidth]{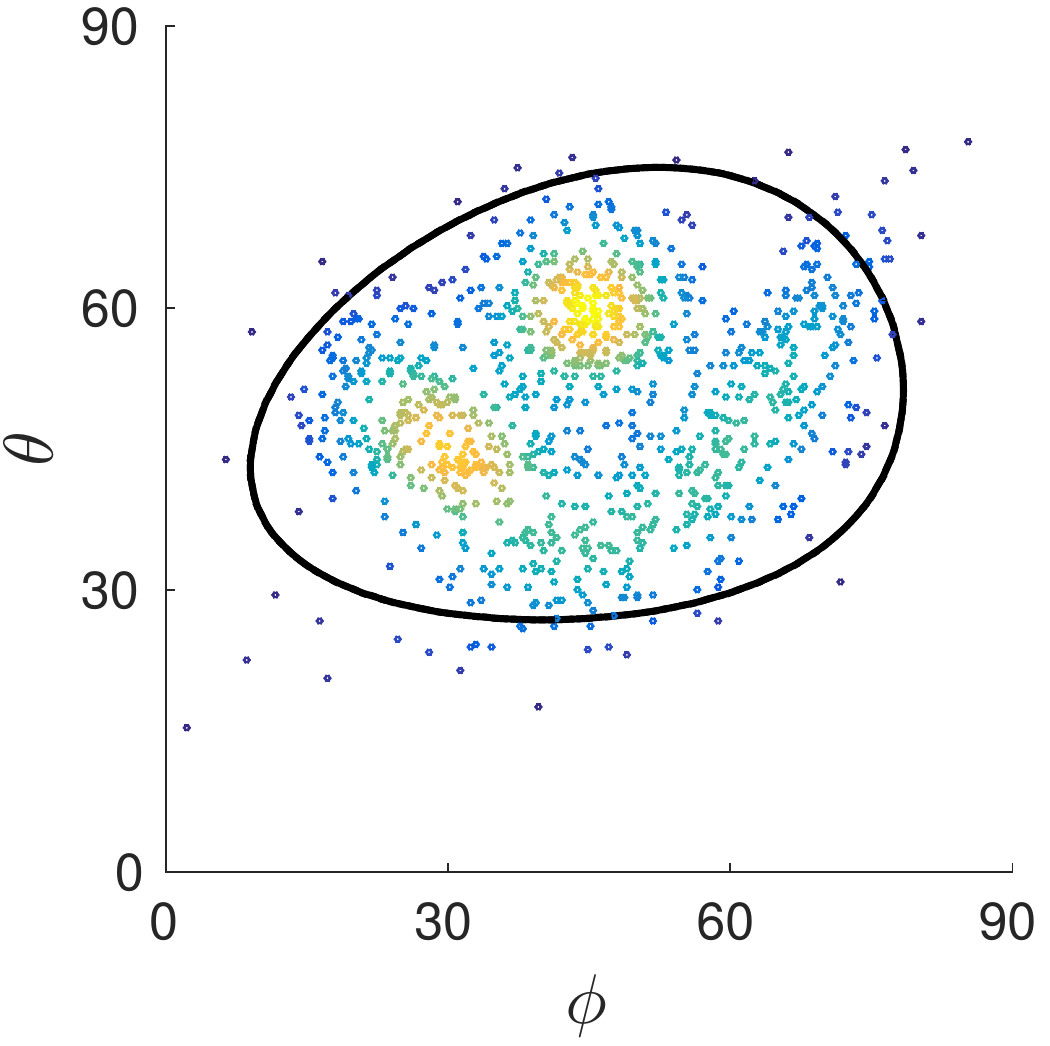}}
\caption{Iteration 2 -- \emph{Deletions} and \emph{Merging}
(green colour represents the component being deleted and blue denotes the pair being merged).
}
\label{fig:mix_iter2_delete_merge}
\end{figure}

In the second iteration, each of the two components in $\fancym_2$
are split, deleted, and merged. Figure~\ref{fig:mix_iter2_split}(a)-(c)
illustrates the splitting of component $P_1$. After integrating the 
optimized children and subsequently optimizing the resulting
3-component mixture using an EM algorithm, an improved mixture $\fancym_3$
is obtained. Figure~\ref{fig:mix_iter2_split}(d)-(f) illustrates
the splitting of component $P_2$.
In this case, splitting $P_2$ results in the same 
3-component mixture $\fancym_3$. It is to be noted that while splitting $P_1$
and $P_2$ produce different intermediate states,
as shown in Figure~\ref{fig:mix_iter2_split}(b) and (e),
the EM converges to the same optimal state in these cases.
Figure~\ref{fig:mix_iter2_delete_merge}(a)-(f) illustrate the
deletion of $P_1$ and $P_2$.
While their deletions also have different intermediate starting points,
as shown in Figure~\ref{fig:mix_iter2_delete_merge}(b) and (e), 
the EM algorithm results in the same sub-optimal state 
(same as $\fancym_1$). As this one-component mixture has a greater
message length than that of $\fancym_2$, the deletion operations
do not result in improved mixtures. 
The merging of $P_1$ and $P_2$ components, as shown in Figure~\ref{fig:mix_iter2_delete_merge}(g)-(i), 
also does not improve on $\fancym_2$.
Hence, after the second iteration, it is observed that amongst all perturbations,
the splitting of $P_1$ or $P_2$ results in an improved mixture $\fancym_3$.

\begin{figure}[ht]
\centering
\subfloat[Initialize means of children]{\includegraphics[width=0.33\textwidth]{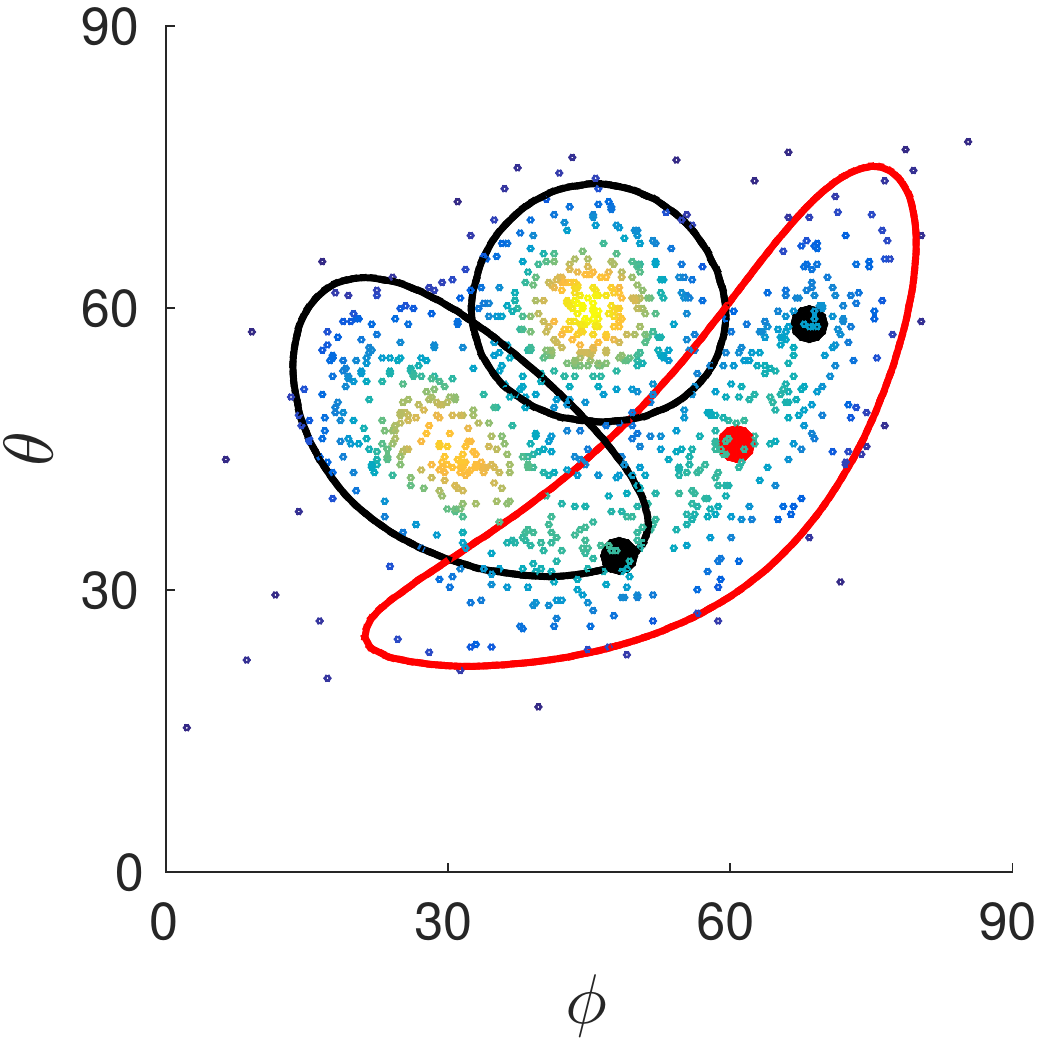}}
\subfloat[Optimized children ($I=19186$)]{\includegraphics[width=0.33\textwidth]{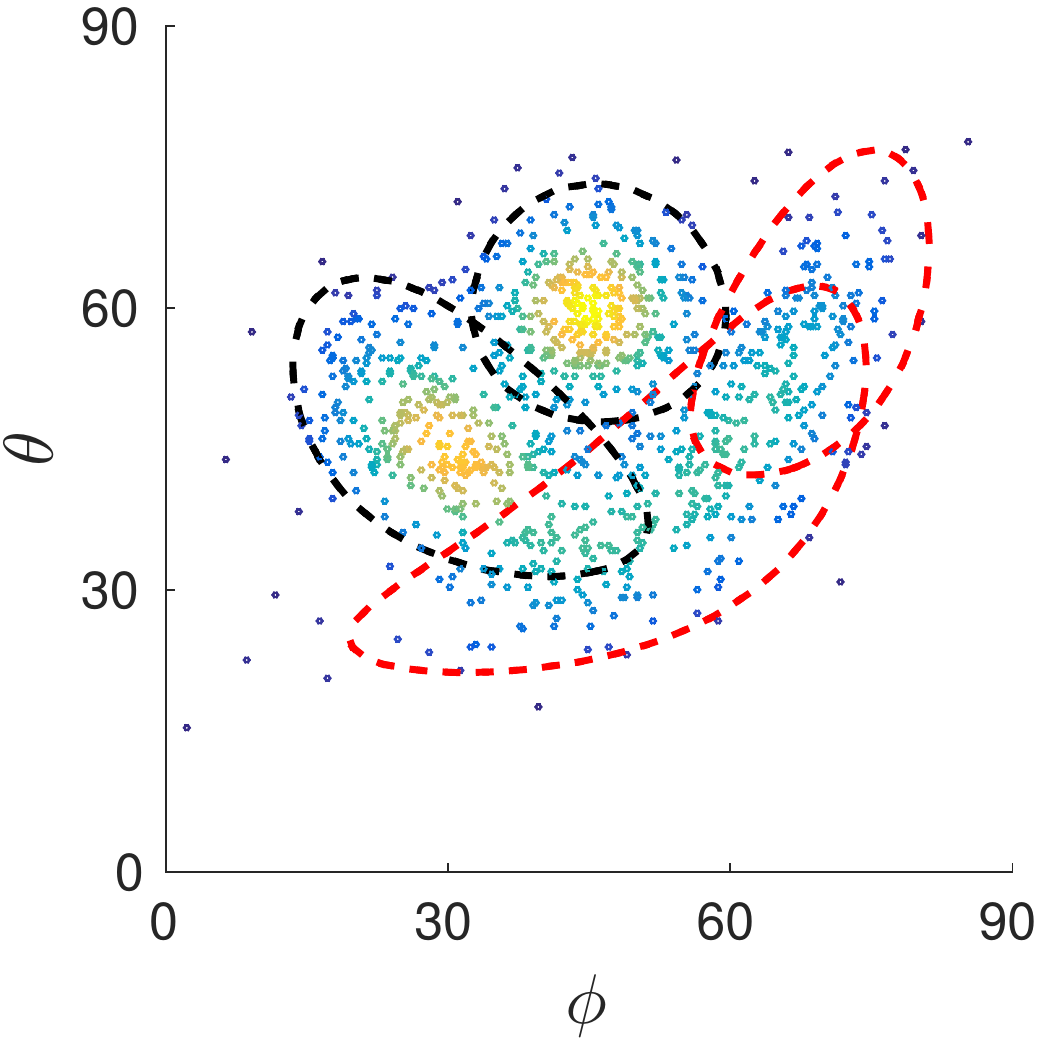}}
\subfloat[$\fancym_4:$ post-EM ($I=19174$)]{\includegraphics[width=0.33\textwidth]{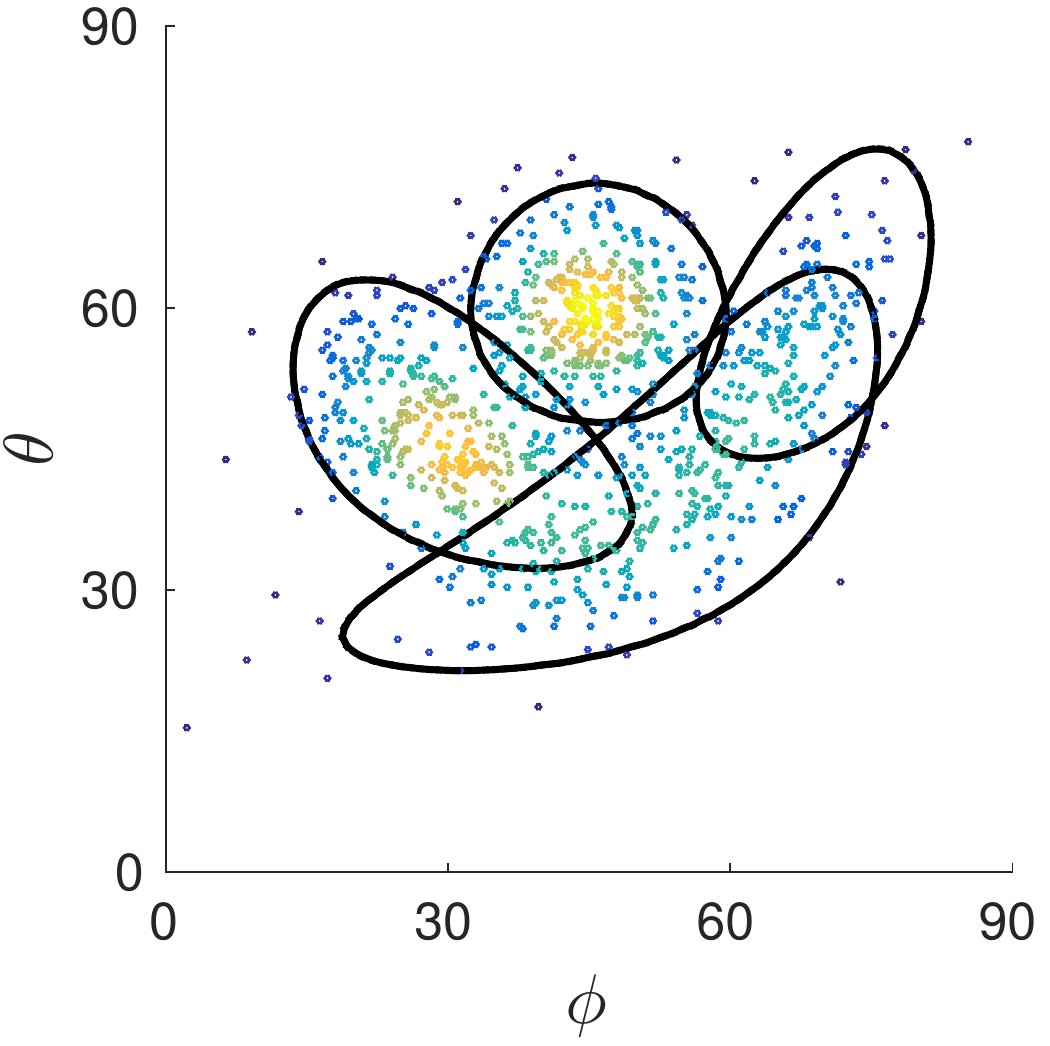}}\\
\subfloat[]{\includegraphics[width=0.33\textwidth]{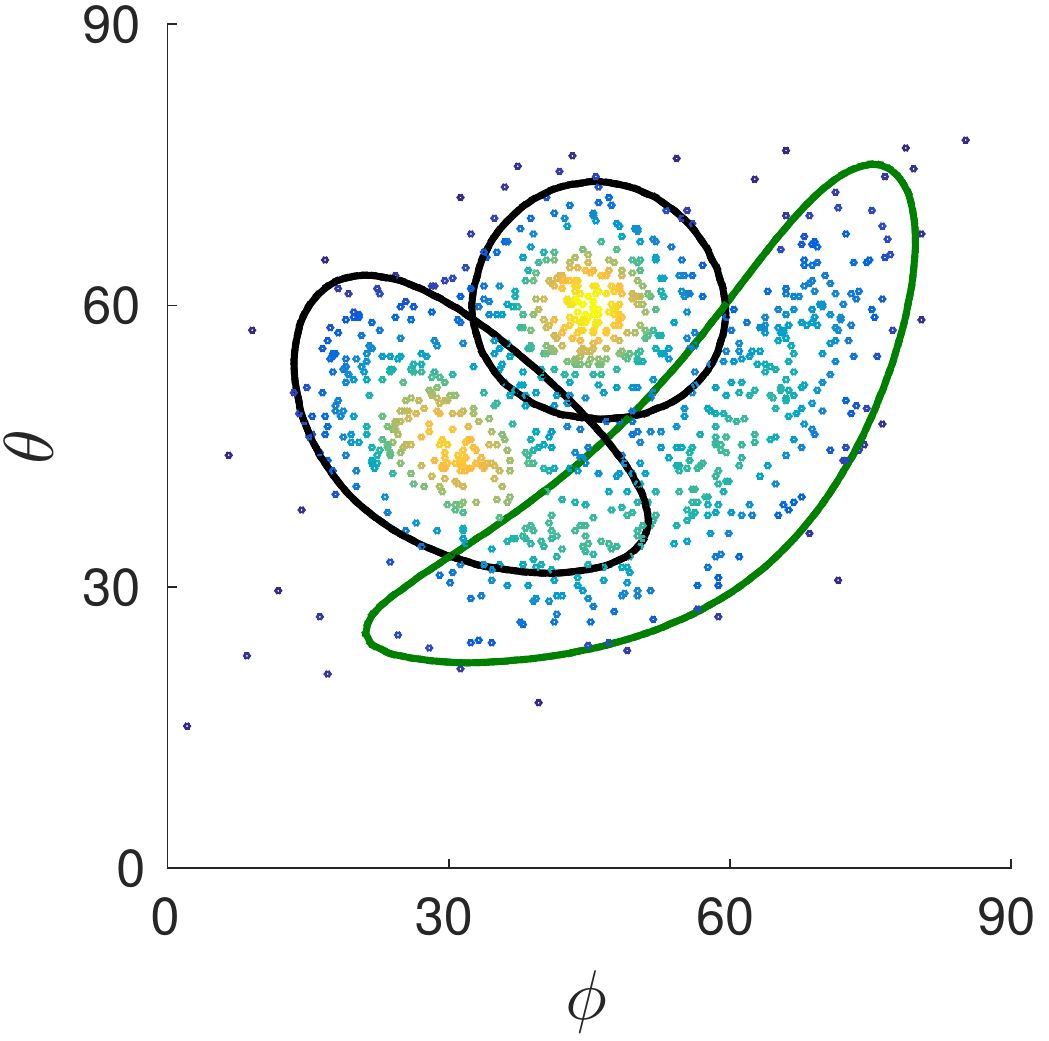}}
\subfloat[Before optimizing ($I=20741$)]{\includegraphics[width=0.33\textwidth]{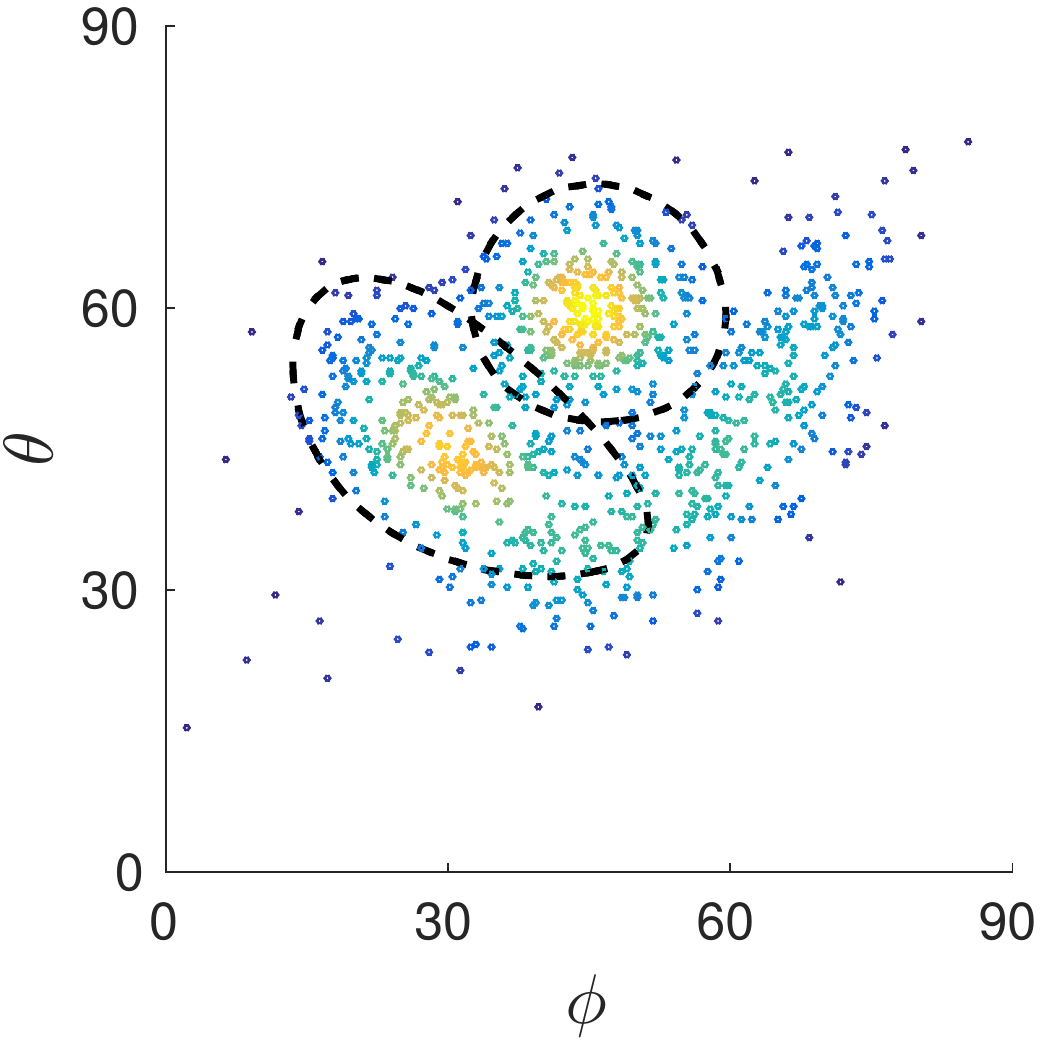}}
\subfloat[post-EM ($I=19320$)]{\includegraphics[width=0.33\textwidth]{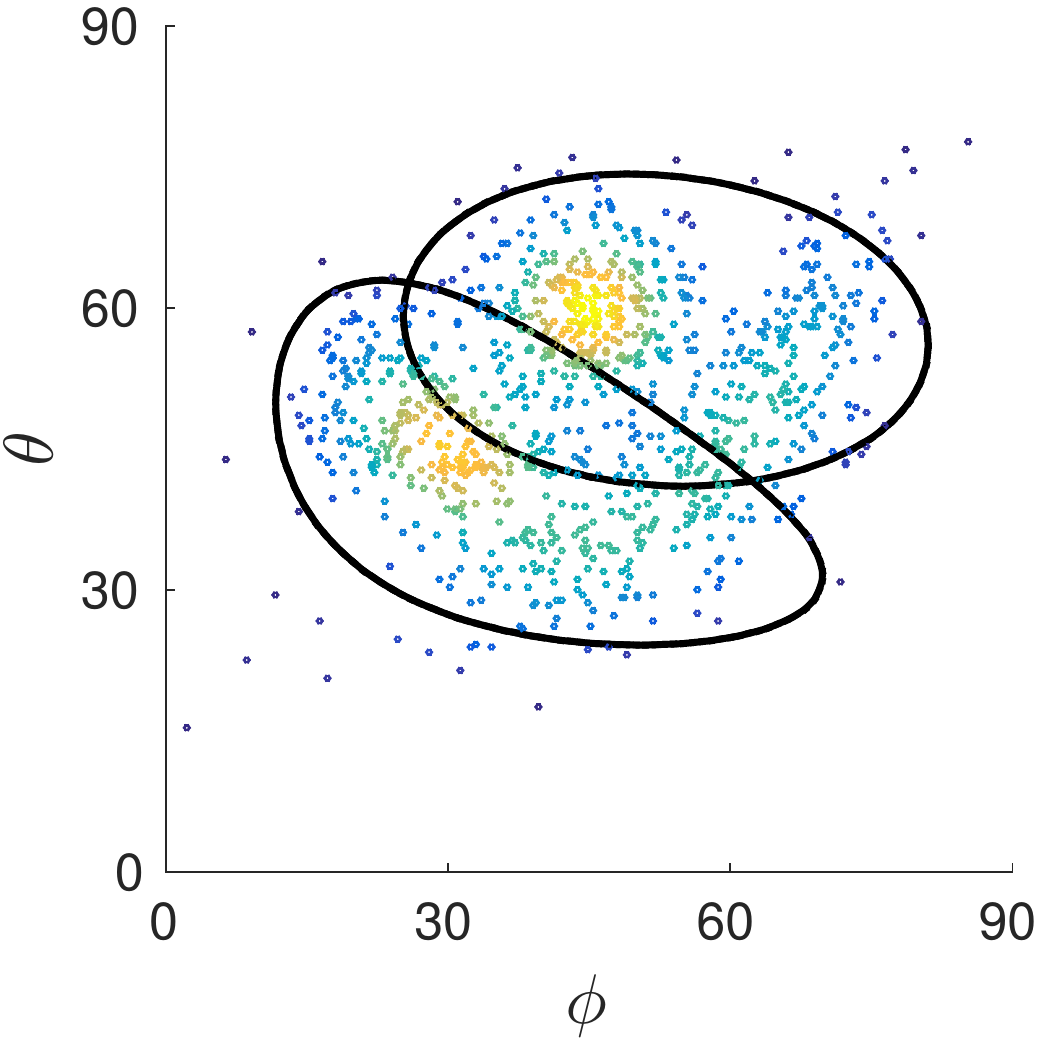}}\\
\subfloat[]{\includegraphics[width=0.33\textwidth]{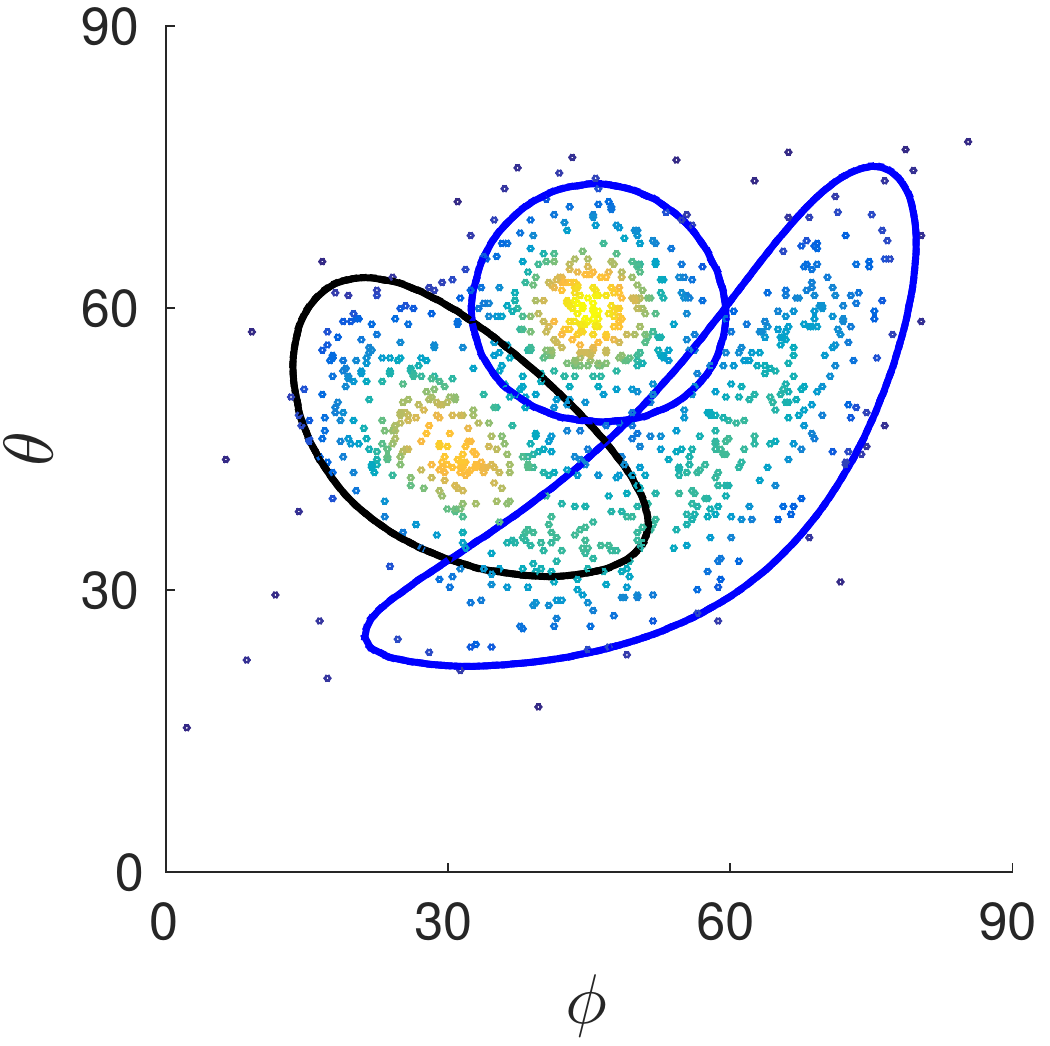}}
\subfloat[Before optimizing ($I=19371$)]{\includegraphics[width=0.33\textwidth]{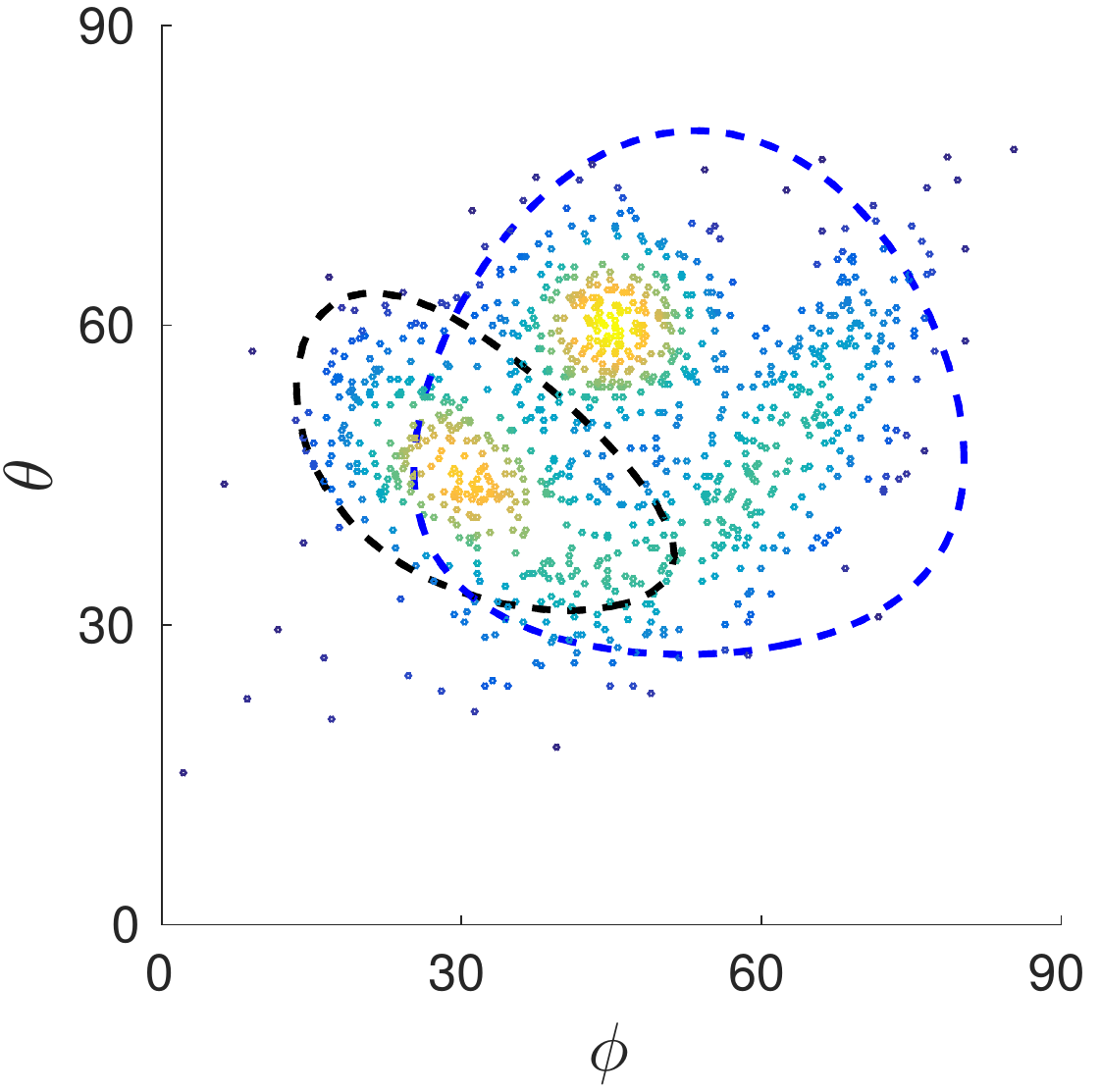}}
\subfloat[post-EM ($I=19312$)]{\includegraphics[width=0.33\textwidth]{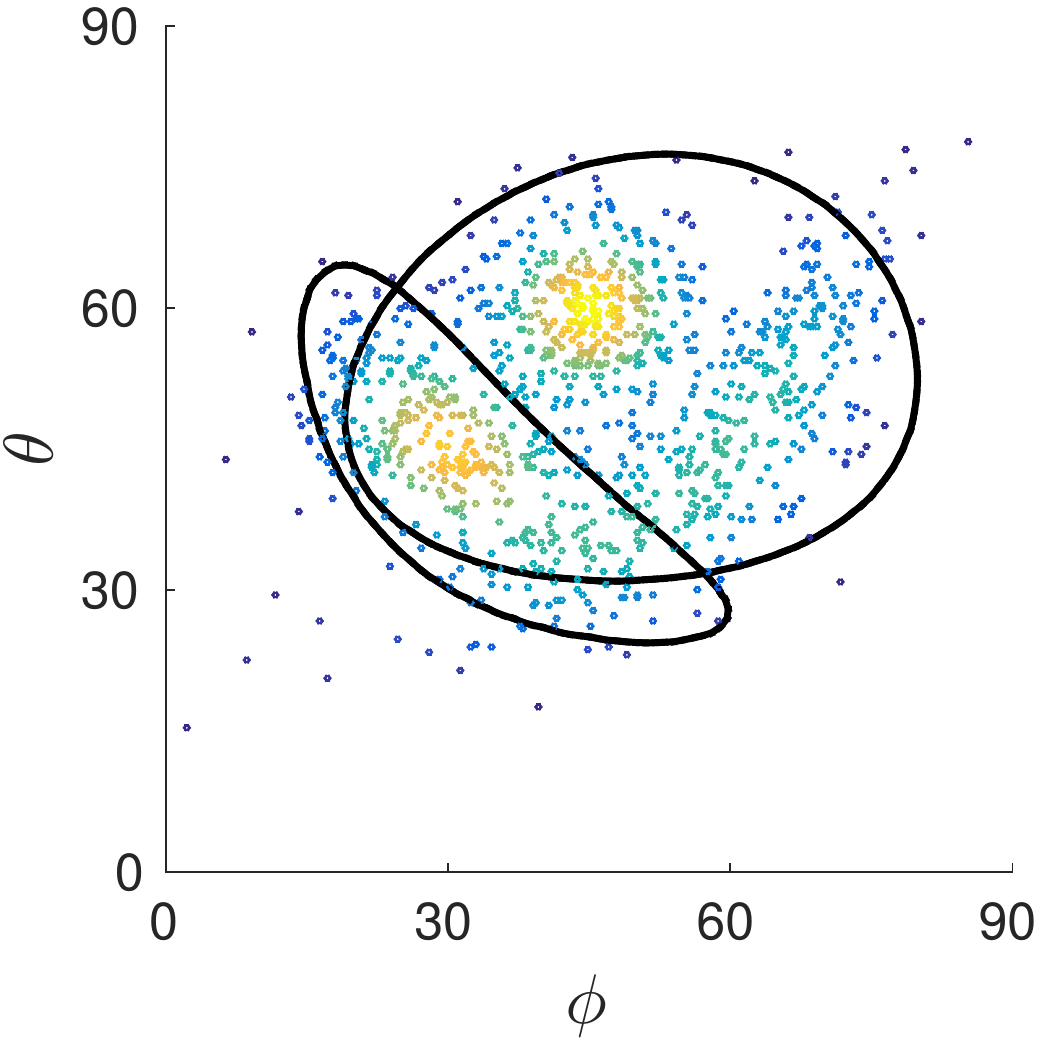}}
\caption{Iteration 3 -- perturbations of $P_1$ 
(a)-(c) splitting,
(d)-(f) deletion, 
(g)-(i) merging
} 
\label{fig:mix_iter3_c1}
\end{figure}

In the third iteration, all perturbations are carried out exhaustively.
Figure~\ref{fig:mix_iter3_c1} depicts the splitting, deletion, and merging 
of one of the three components ($P_1$) in $\fancym_3$.
During splitting, observe the initial selection of means of the child components.
The procedure outlined in Section~\ref{subsubsec:splitting} faithfully
separates the two children and results in a mixture with a greater number
of components. However, in this case, the optimized mixture $\fancym_4$ (Figure~\ref{fig:mix_iter3_c1}c)
does not improve the message length. 
Similarly, the deletion of $P_1$ does not lead to an improved mixture (Figure~\ref{fig:mix_iter3_c1}f).
While merging $P_1$, KL divergence is used to determine an appropriate
candidate that is closest. Accordingly, the pair is selected (Figure~\ref{fig:mix_iter3_c1}g)
which also does not result in an improved mixture (Figure~\ref{fig:mix_iter3_c1}i).
The other two components in $\fancym_3$ are also perturbed similarly. However,
the operations do not result in an improvement 
(the series of steps and the resulting mixtures are included in 
Appendix~\ref{app:search_continued}).

\subsubsection{Variation of the two-part message length}
\begin{figure}[htb]
\centering
\includegraphics[width=0.7\textwidth]{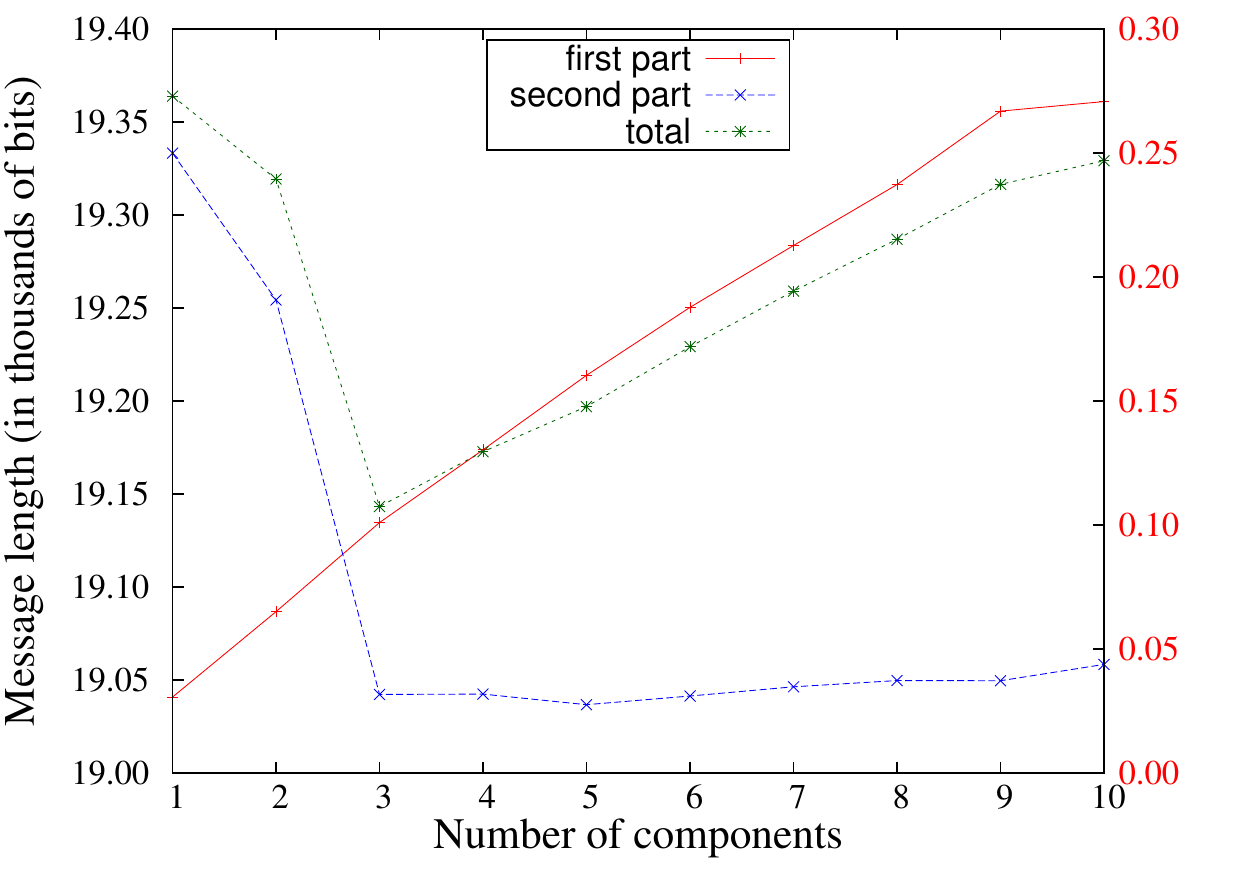}
\caption{Variation of the individual parts of the total message length 
         with increasing number of components (note that the two Y-axes have 
         different scales: the first part of the message follows the right side Y-axis;
         while the second part of the message and total message lengths
         follow the left side Y-axis)
        }
\label{fig:individual_msglens}
\end{figure}

Let us now explain the evolution of the mixture model in terms of
the two-part message length (Equation~\ref{eqn:mixture_msglen}).
While increasing the number of mixture components leads to increased
mixture complexity, the fit to the data improves.
The first part of the message corresponds to the overhead related to encoding the mixture
parameters (number of components, weights, and constituent components' parameters).
The second part mainly corresponds to the negative log-likelihood
of the data using a given mixture model. In the previous example,
the search method infers three components and terminates thereafter.
The message lengths corresponding to the optimal mixtures during
the associated search process are plotted in Figure~\ref{fig:individual_msglens}.
It is observed that, until $K=3$, the total message length (green curve) decreases.
We wanted to examine the variation of the message length beyond
the inferred number of components. For this, starting from $K=4$ until $K=10$,
we estimated the mixture parameters using the EM algorithm (Section~\ref{subsec:em_mml})
for each value of $K > 3$. The results indicate that the
total message length steadily increases beyond $K=3$.
The reason is that although the negative log-likelihood of the data decreases 
(with increasing $K$), the second part of the message (blue curve) only changes marginally,
while the first part continually increases. Thus, as mixtures become
overly complex, there is a greater cost associated with encoding their parameters.
This affects the total message length as the minimal gain in negative log-likelihood is
overshadowed by the increase in the first part of the message.
Hence, this example demonstrates the effectiveness of the search method in the
context of \fb~distributions. Furthermore, it also demonstrates the ability of the MML
criterion to balance the tradeoff between the model complexity and the quality
of data fit.

\section{Experimental analyses of the various parameter estimates} 
\label{sec:exp_single_kent}
For a given \fb~distribution characterized by concentration $\kappa$ and eccentricity $e$,
a random sample of size $N$ is generated using the method proposed by \citet{kent2013new}.
We set the true distribution to have $\{\psi,\alpha,\eta\} = \pi/2$
each. The scale parameters $\kappa$ and $e$ are varied
to obtain different \fb~distributions and corresponding random samples.
The parameters are estimated using the sampled data and the different
estimation methods. The procedure is repeated
1000 times for each combination of $N,\kappa,$ and $e$.

\subsection{Methods of comparison}
\label{subsec:comparison_methods}
We conduct a comparison between the moment, maximum likelihood (ML), MAP, and MML-based
estimates. The results include the two versions of MAP
estimates resulting from the two forms of the posterior distributions 
(Equation~\ref{eqn:prior3d_posteriors}): \emph{MAP1} corresponds to the posterior
with parameterization $\kappa, \beta$, and \emph{MAP2} corresponds to the posterior
with parameterization $\kappa, e$. The two versions are considered so as to
show that MAP estimates are inconsistent and are
dependent on the parameterization used.

The MML estimates are obtained by minimizing the message length expression. Naturally,
the estimates due to other methods do not result in lower message lengths.
Similarly, if we use the negative log-likelihood as the comparison criterion,
the maximum likelihood estimates have a lower value compared to the other estimates.
As each estimation technique optimizes a different objective function,
it is required to have a metric that impartially evaluates the different estimates. 
The mean squared error of the estimates and Kullback-Leibler (KL) divergence 
\citep{kullback1951information} are therefore used to compare the various estimates.
The estimates are also compared using statistical hypothesis testing.

\subsubsection{Mean squared error of the estimates}
For a parameter vector $\boldtheta$, and its estimate $\widehat{\boldtheta}$,
the mean squared error (MSE) is given by $\expect[(\widehat{\boldtheta}-\boldtheta)^2]$.
Further, the MSE can be decomposed into bias and variance terms, as given below
\citep{lebanon2010bias,taboga2012lectures}.
\begin{equation*}
\expect[(\widehat{\boldtheta}-\boldtheta)^2] = \text{Bias}^2(\widehat{\boldtheta}) + \text{trace}(\text{Var}(\widehat{\boldtheta}))
\end{equation*}
where $\text{Bias}^2(\widehat{\boldtheta}) = \|\expect[\widehat{\boldtheta}]-\boldtheta\|^2$
and $\text{Var}{(\widehat{\boldtheta})}$ is the covariance matrix of the estimator.
Ideally, it is expected that the estimates result in low MSE values
and it depends on the bias and variance of the parameter estimates.
An estimate that results in lower values of MSE is usually preferred
over the other estimates.

\subsubsection{Kullback-Leibler (KL) divergence of the estimated distribution}
The KL divergence is a similarity measure that is used to determine the
``distance" between the true distribution and the distribution thats uses the 
estimated parameters. An estimate
that results in lower KL divergence is considered a better estimate.
The analytical form of the KL divergence between two
\fb~distributions is derived in Section~\ref{subsubsec:merging}. 
We report the percentage of times (out of 1000 random simulations)
that the KL divergence of a particular estimator is lower than that of others.
An estimator \emph{wins} when its associated KL divergence 
is less than that of the other estimates.

When the KL divergence of different estimates is compared, because of two different
versions of MAP estimation, we present two separate frequency plots.
The KL divergence of the moment, ML, and MML estimates is contrasted with
KL divergence of MAP1 or MAP2 estimates.

\subsubsection{Statistical hypothesis testing}
The likelihood ratio test is typically used to determine the 
suitability of modelling data $\dataset$ using a simpler or a nested model, corresponding 
to one with fewer free parameters (null hypothesis $\hypothesis_0$) against a more general model
(alternate hypothesis $\hypothesis_A$).
The likelihood ratio $\lambda$ is used to determine the preference of a null
hypothesis $\hypothesis_0$ over an alternate hypothesis $\hypothesis_A$,
and is as follows:
\begin{equation*}
\lambda = \frac{\displaystyle\max_{\hypothesis_0}\,\,\text{likelihood}(\dataset|\hypothesis_0)}{\displaystyle\max_{\hypothesis_A}\,\,\text{likelihood}(\dataset|\hypothesis_A)}
\end{equation*}
The test statistic resulting from the use of $\lambda$ is related to the negative logarithm
of the likelihood ratio and is given by $\Lambda=-2\log\lambda$. The distribution of the
statistic $\Lambda$ is asymptotically approximated as a $\chi^2$ distribution with degrees 
of freedom equal to the difference in the 
number of free parameters between the alternate and the null hypothesis \citep{wilks1938}.
If $\lambda$ is sufficiently small, it would lead to a rejection of the
null hypothesis. Conversely, if $\Lambda$ exceeds some confidence threshold,
$\hypothesis_0$ is rejected.

In the current analysis of the various parameter estimates,
we compare the likelihood ratio resulting from the use of a particular
estimate $\widehat{\boldtheta}$ (that is, moment, MAP or MML-based) against 
a general \fb~distribution. It is equivalent to testing the null hypothesis 
$\hypothesis_0: \boldtheta = \widehat{\boldtheta}$ (explicit parameters)
against the alternate hypothesis $\hypothesis_A: \boldtheta \neq \widehat{\boldtheta}$ 
(with 5 free parameters). Assuming a statistical significance of the test as $1\%$,
$\hypothesis_0$ is rejected
when $\Lambda > \tau$, where $\tau = 13.086$ corresponds to the $99^{\text{th}}$ percentile
of a $\chi^2$ distribution with 5 degrees of freedom. 
Alternatively, the test statistic can be used to evaluate the p-value, which
if less than $1\%$ (significance of the test) amounts to rejection of $\hypothesis_0$.

For the various parameter estimates compared here, it is expected that
at especially large sample sizes, the estimates are close to the maximum likelihood estimate
as determined by the corresponding test statistic. In other words,
the empirically determined test statistic is expected to be lower than the critical value $\tau$,
which implies it has a corresponding p-value greater than 0.01.

\subsection{Empirical analysis}
The estimates are analyzed here in two controlled cases: (1) fixing sample size $N$
with varying $\kappa$ and $e$, and (2) varying $N$ while fixing $\kappa$ and $e$.

\subsubsection{Fixed sample size, varying concentration $\kappa$ and eccentricity $e$}
\begin{figure}[ht]
\centering
\subfloat[Bias-squared]{\includegraphics[width=0.5\textwidth]{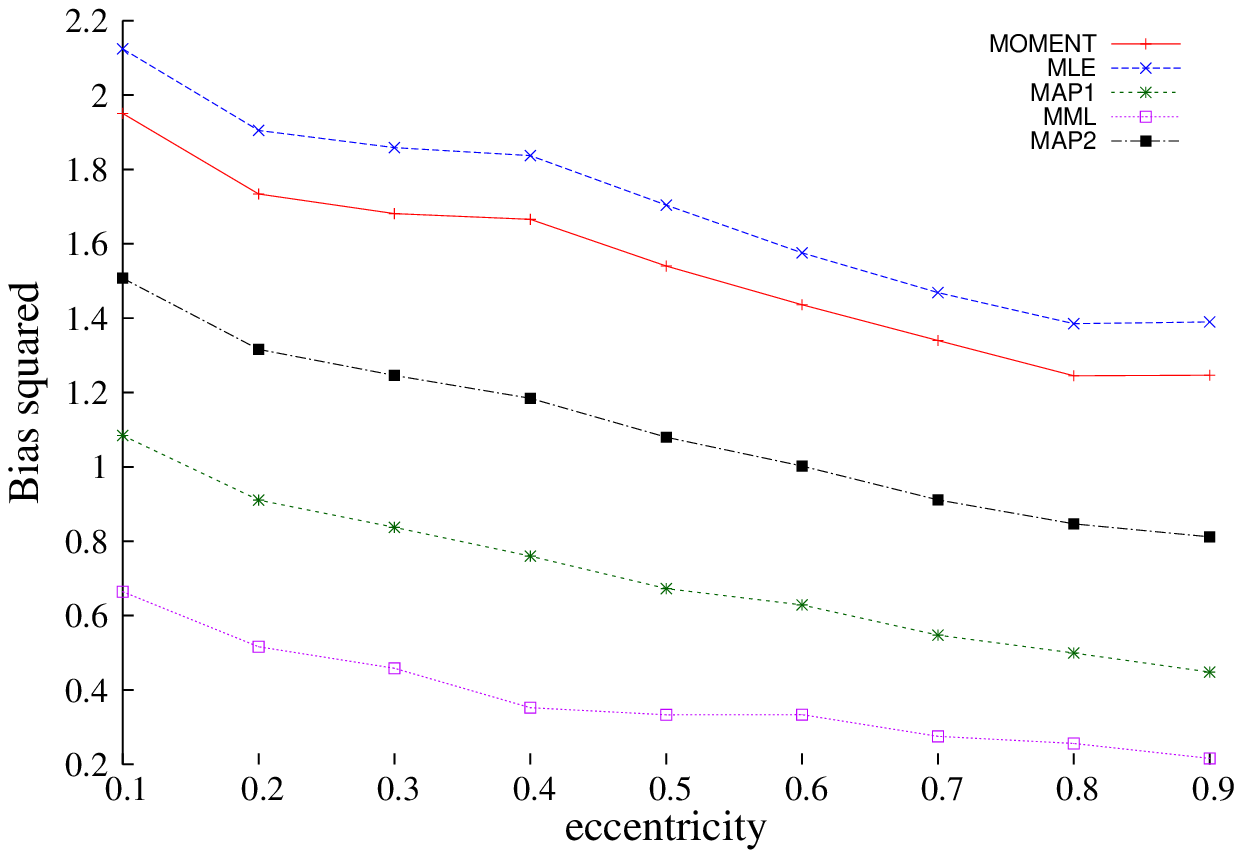}}
\subfloat[Mean squared error]{\includegraphics[width=0.5\textwidth]{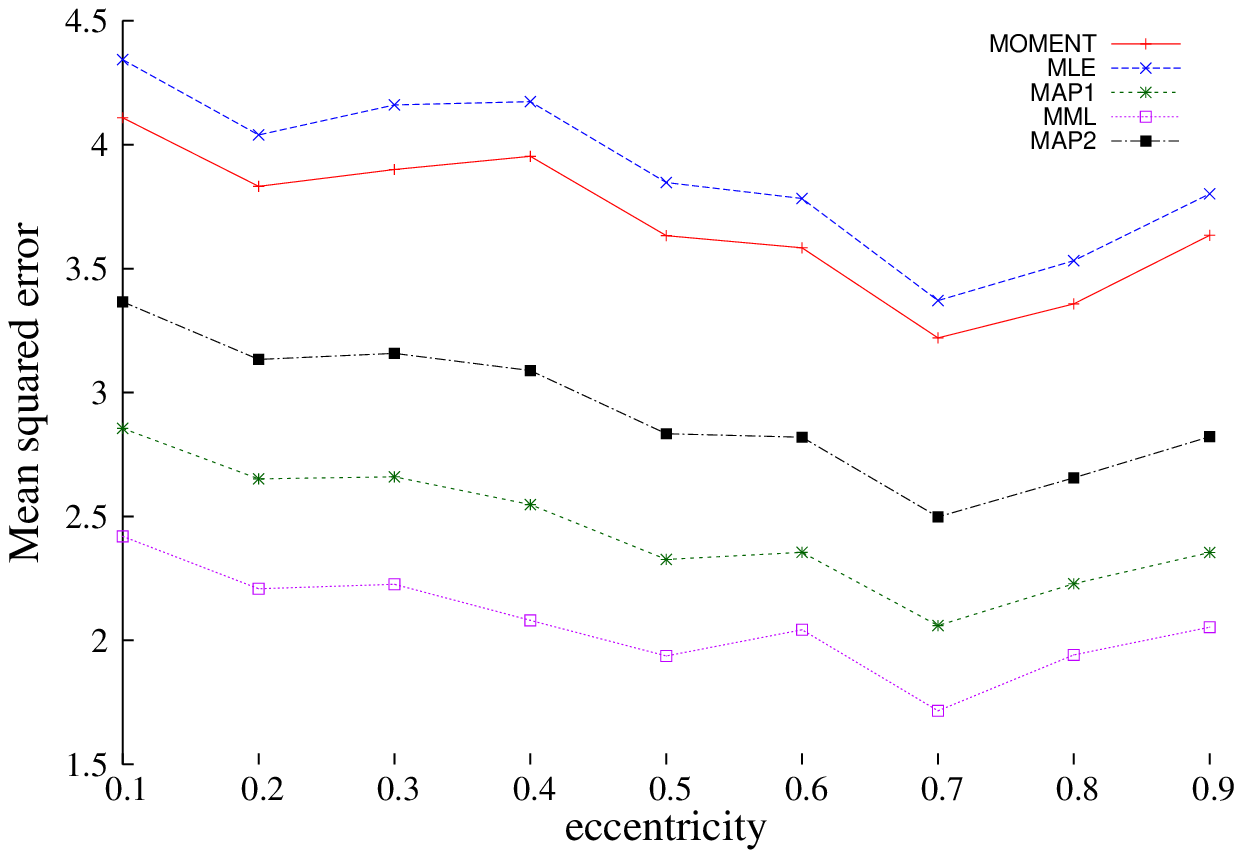}}\\
\subfloat[KL divergence (MAP version 1)]{\includegraphics[width=0.5\textwidth]{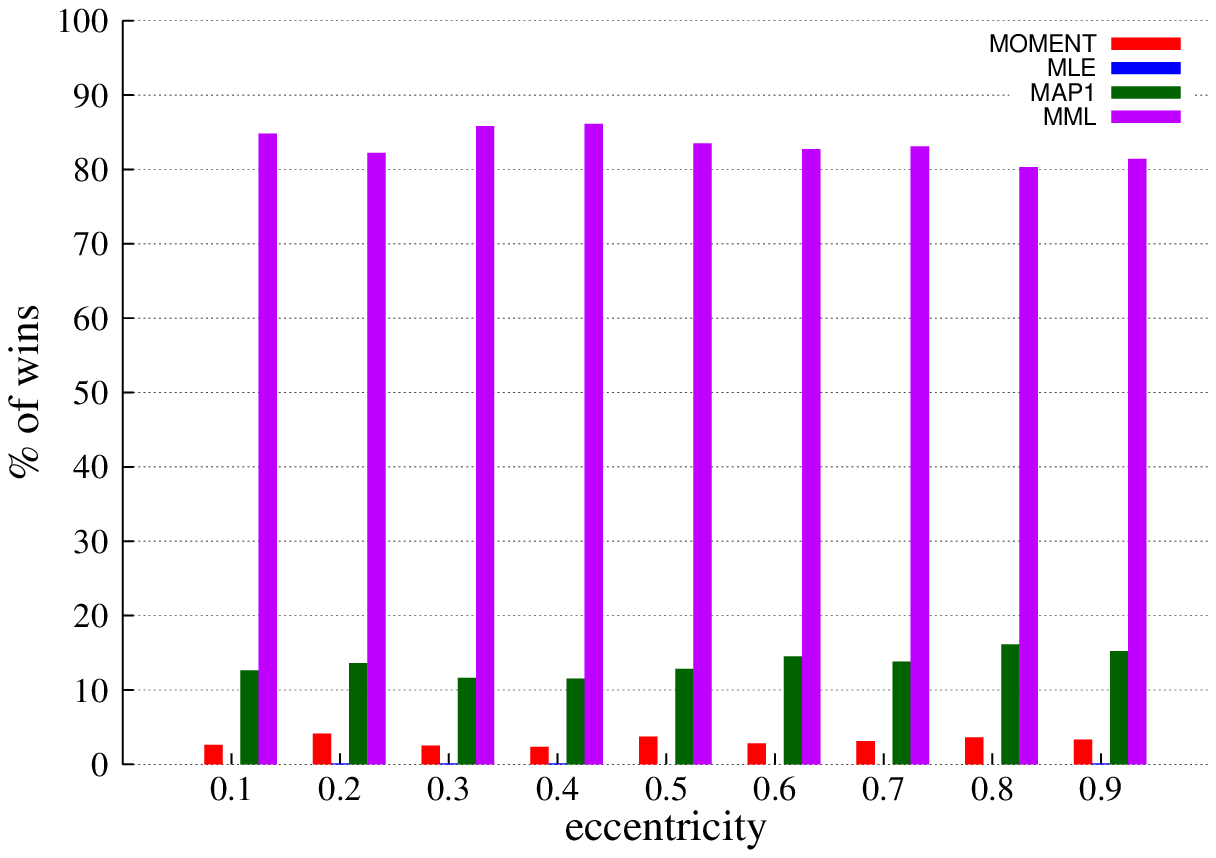}}
\subfloat[KL divergence (MAP version 2)]{\includegraphics[width=0.5\textwidth]{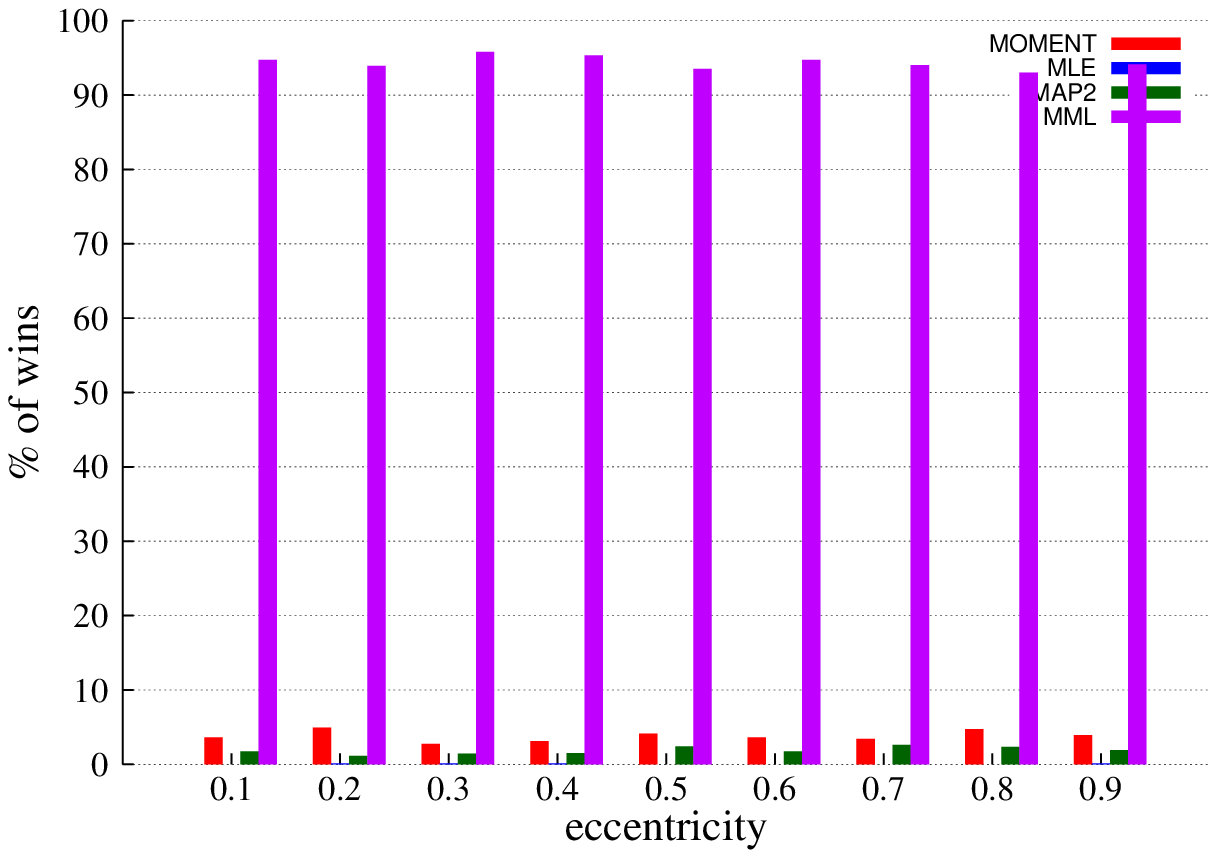}}\\
\subfloat[Variation of test statistics]{\includegraphics[width=0.5\textwidth]{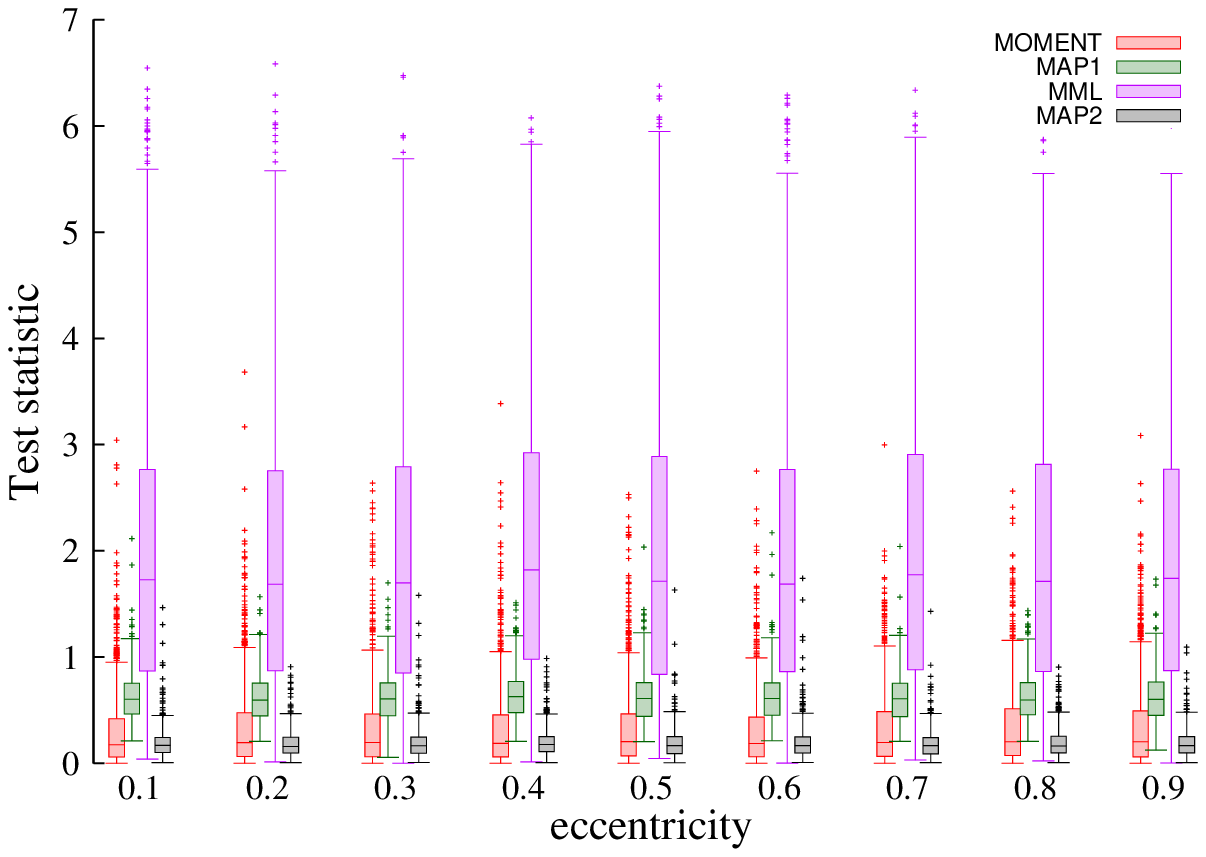}}
\subfloat[Variation of p-values]{\includegraphics[width=0.5\textwidth]{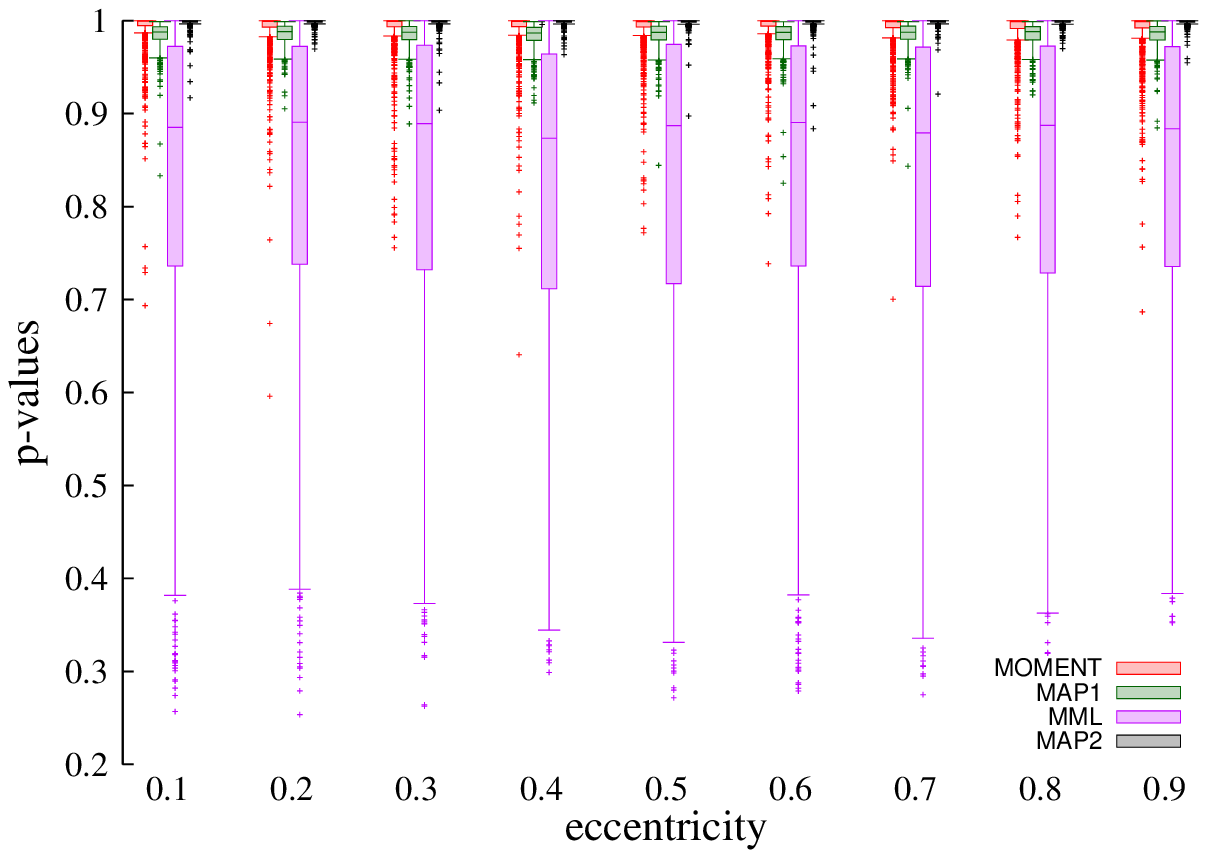}}
\caption{$N=10, \kappa=1$.} 
\label{fig:n10_p3_k1}
\end{figure}

\begin{figure}[ht]
\centering
\subfloat[Bias-squared]{\includegraphics[width=0.5\textwidth]{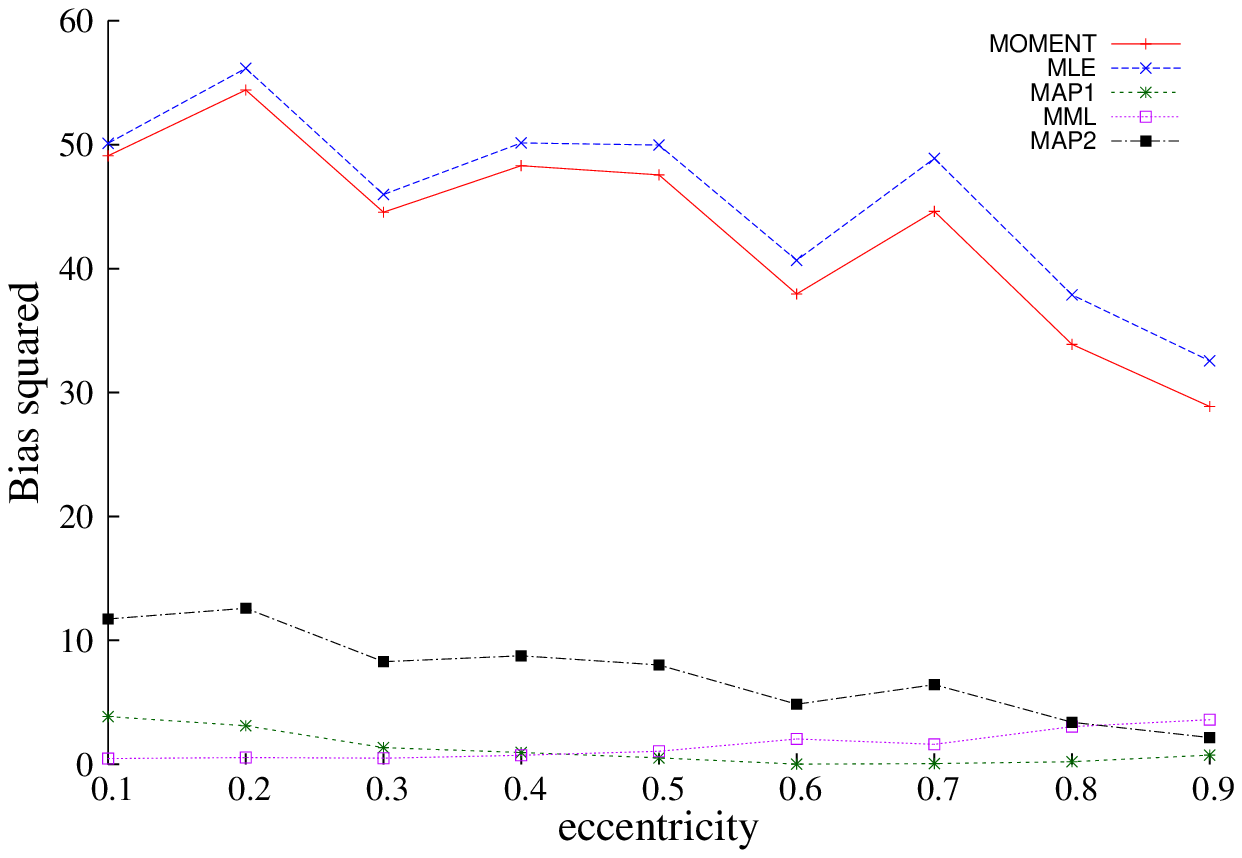}}
\subfloat[Mean squared error]{\includegraphics[width=0.5\textwidth]{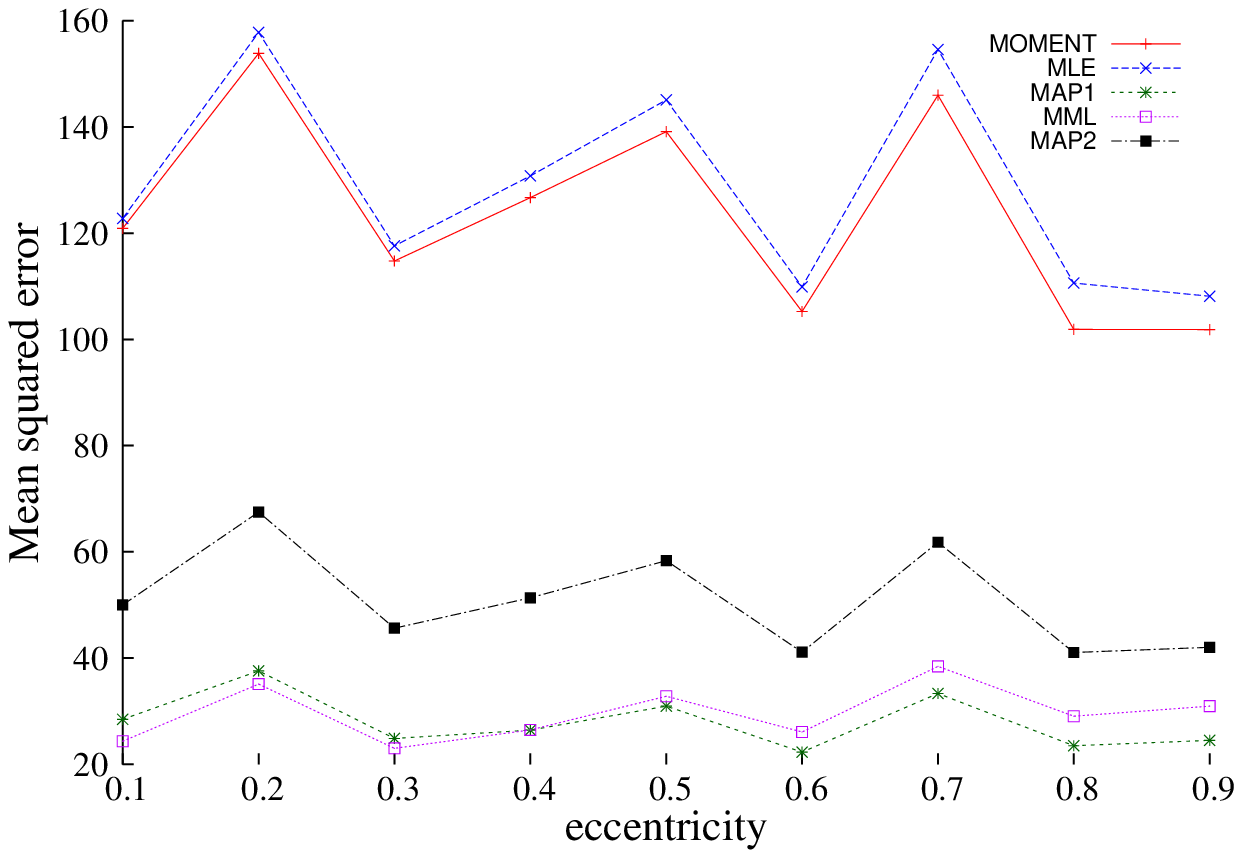}}\\
\subfloat[KL divergence (MAP version 1)]{\includegraphics[width=0.5\textwidth]{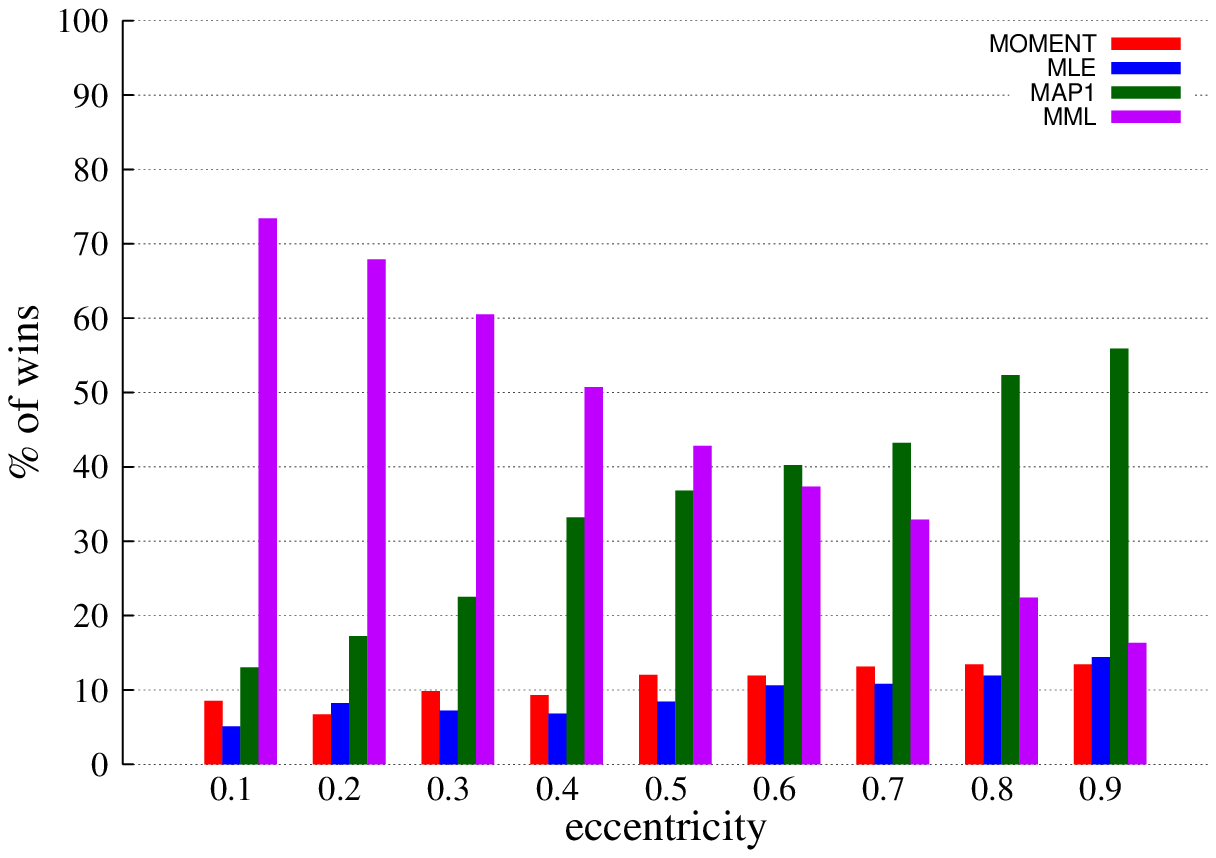}}
\subfloat[KL divergence (MAP version 2)]{\includegraphics[width=0.5\textwidth]{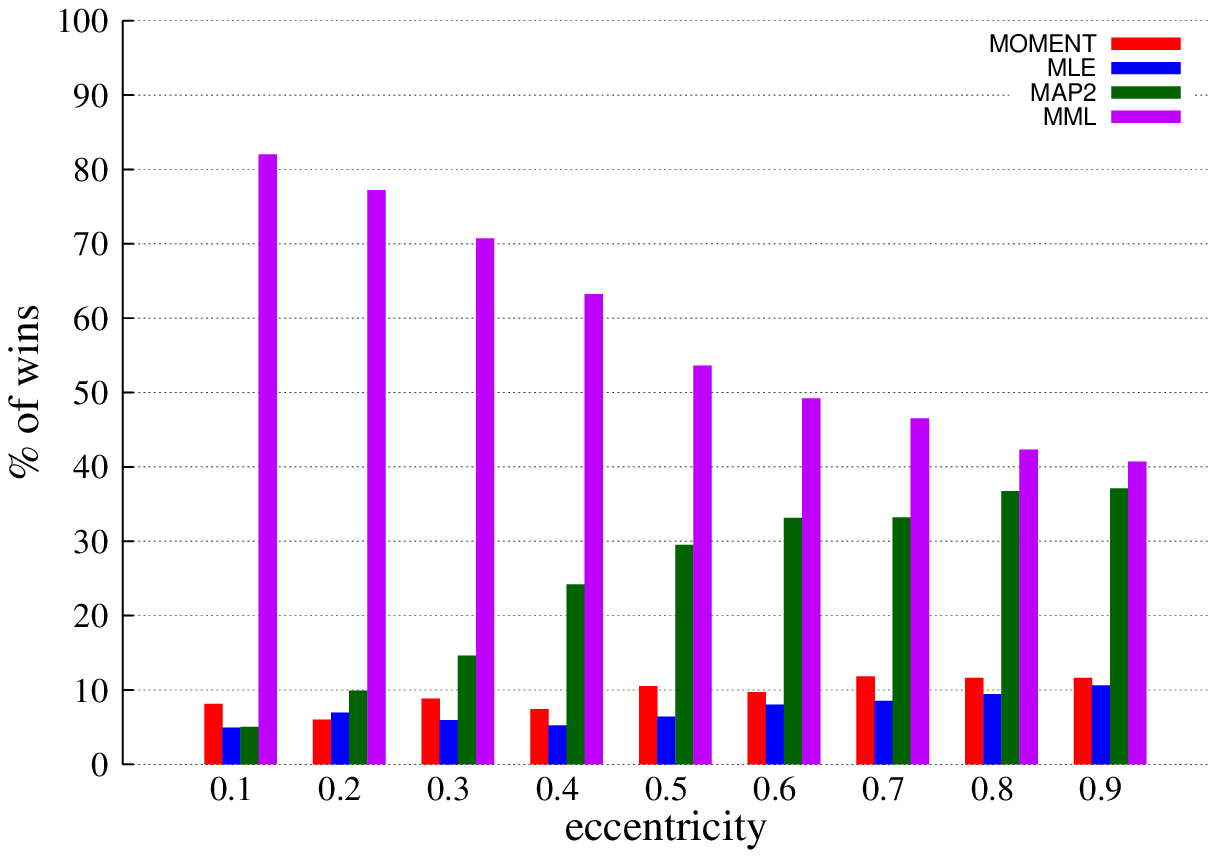}}\\
\subfloat[Variation of test statistics]{\includegraphics[width=0.5\textwidth]{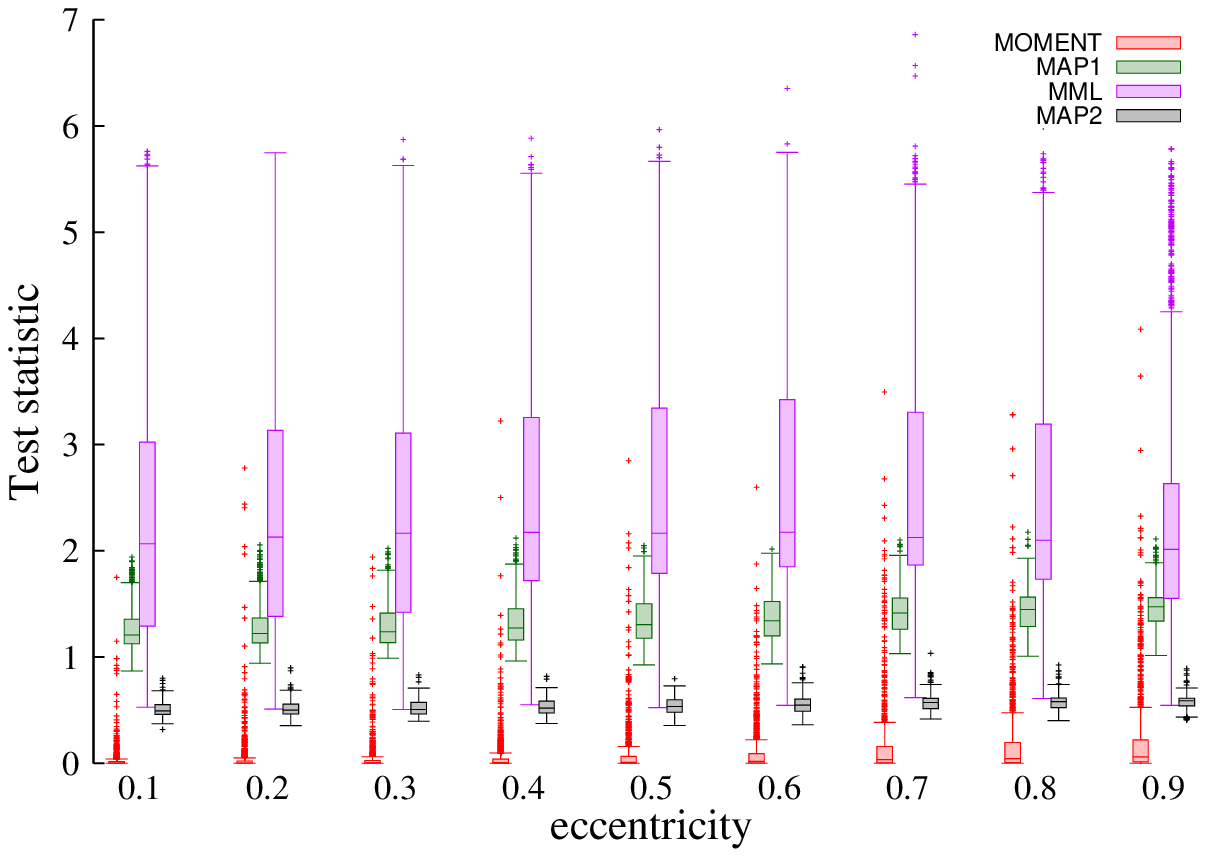}}
\subfloat[Variation of p-values]{\includegraphics[width=0.5\textwidth]{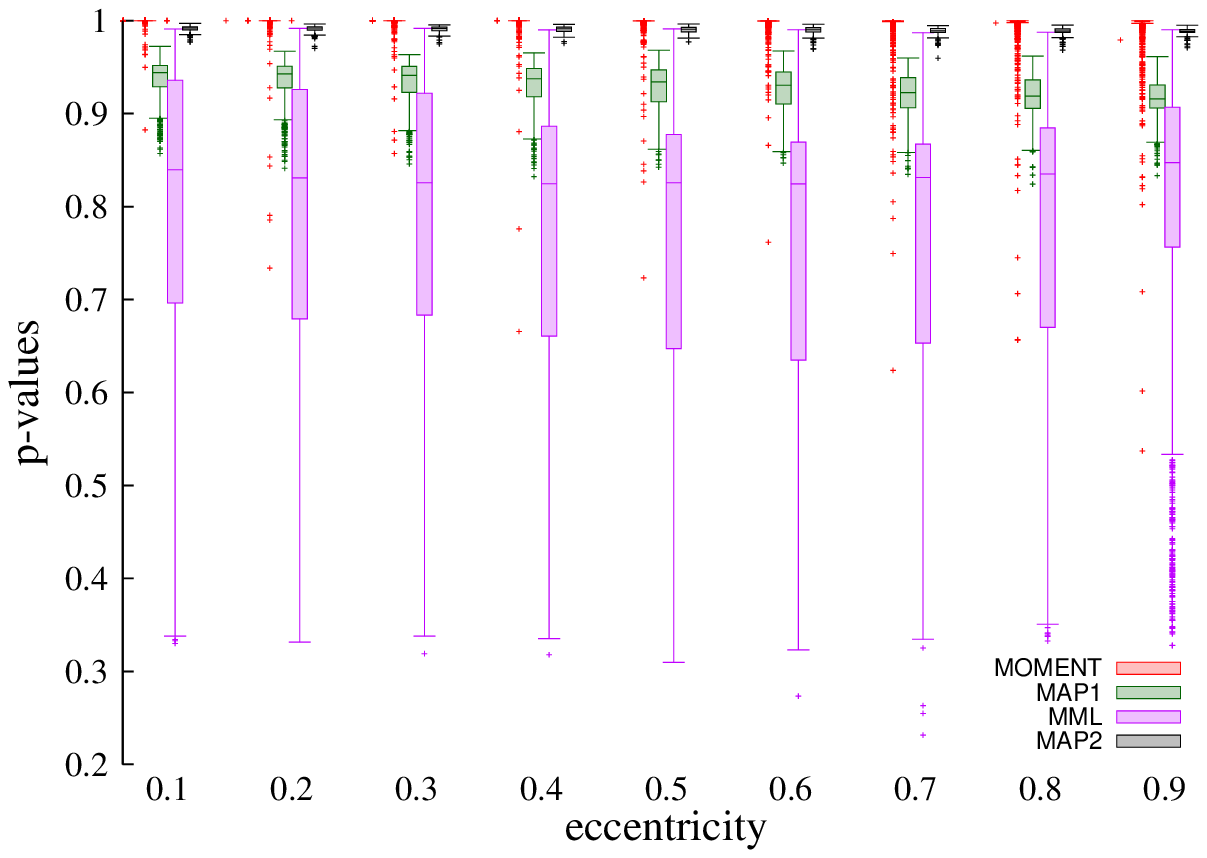}}
\caption{$N=10, \kappa=10$.} 
\label{fig:n10_p3_k10}
\end{figure}

The results are presented when a random sample of size $N=10$ is generated
from the \fb~distribution for a 
$\kappa$ that is increased by an order of magnitude starting from 1 to 100.
The behaviour of the estimates is analyzed below. 
\begin{itemize}
\item $\kappa = 1$:
The performance of the various estimates using the comparison methodologies
(Section~\ref{subsec:comparison_methods}) is illustrated in Figure~\ref{fig:n10_p3_k1}.
It is observed that the bias and MSE of moment and ML estimates is greater than that of
MAP and MML-based estimates. The two versions of the MAP estimates
also have a greater bias and MSE as compared to the MML estimates 
shown in Figure~\ref{fig:n10_p3_k1}(a) and (b).

It is also observed that the MML-based estimates result in lower KL divergence 
more than 80\% of time as compared to other estimates
when MAP1 is used (see Figure~\ref{fig:n10_p3_k1}c).
With MAP2, the frequency of wins for the MML-based estimates increases to more 
than 90\% (see Figure~\ref{fig:n10_p3_k1}d). This suggests that transforming
the parameter space greatly impacts the MAP-based estimates.
The ML estimates win less than 5\% of the time. This is in agreement with the
relatively greater MSE observed for the ML estimates.

The boxplots shown in Figure~\ref{fig:n10_p3_k1}(e) and (f) show the variation
of the test statistics and the corresponding p-values. There is a greater 
variation for the MML-based estimates. However, across all values of eccentricity,
the test statistic $\Lambda$ is less than the threshold $\tau = 13.086$ and the smallest 
p-value is greater than 0.01. This is true across all estimation methods, thus,
suggesting that the null hypothesis of modelling data using a particular estimate
(moment, MAP or MML) is accepted at the 1\% significance level. 

\item $\kappa = 10$:
The comparison results are presented in Figure~\ref{fig:n10_p3_k10}. Similar to the
previous case ($\kappa = 1$), the moment and ML estimates have greater bias and MSE. 
It is interesting to note that MAP2 has greater bias 
and MSE compared to MML estimates (Figure~\ref{fig:n10_p3_k10}(a) and (b) respectively). 
However, MAP1 estimates are in close competition with the MML. The bias and MSE
are lower for MML estimates until $e \leq 0.5$ and greater compared to MAP1
estimates for $e > 0.5$.

The number of times KL divergence is lower for the MML estimates
decreases with increasing eccentricity (for both versions of MAP estimates). 
For $e \leq 0.5$, the percentage of wins for the MML estimates is
greater than all other estimates. However, for $e > 0.5$, MAP1 wins 
majority of the time (Figure~\ref{fig:n10_p3_k10}c). In the case of comparison with MAP2, 
the percentage of wins
of MML estimates continuously decreases. However, the number of wins of MML estimates 
is always in the majority (Figure~\ref{fig:n10_p3_k10}d).

The observations are in contrast to what was observed in the case of $\kappa=1$ 
where MML estimates emerged as consistently better estimates. 
In terms of statistical hypothesis testing, the null hypotheses corresponding to
modelling using moment, ML, MAP or MML estimates are accepted at the 1\% significance level. 

\item $\kappa = 100$:
The comparison results in this case follow the same pattern 
as that of $\kappa=10$
(not illustrated here as they are similar to Figure~\ref{fig:n10_p3_k10}).
\end{itemize}
When $\kappa=10$ and $100$, the MAP1 estimates perform competitively compared to the
MML estimates (with respect to bias and MSE). 
Further, the proportion of times MAP1 estimates win with respect to KL divergence 
progressively increases as the eccentricity increases.
In general, similar results are observed for $\kappa>10$.
However, as discussed previously, MAP-based estimation is subjective to the
parameterization of the distribution as shown by the stark contrast between
MAP1 and MAP2 estimates by the two parameterizations even 
though they are both reasonable.
The moment, ML, and MML estimates, on the other hand, are not affected by
parameterization. Amongst these, MML-based estimates outperform with
respect to all objective metrics as described here.

\subsubsection{Varying sample size $N$, fixed concentration $\kappa$ and eccentricity $e$}
We have also explored the behaviour of different estimates with increasing 
sample size from $N=10$ to $N=50$. For space reasons, we only include
the results for $\kappa=10$. 
The results are discussed for three specific eccentricity values,
ranging from low eccentricity ($e=0.1$), to moderate
eccentric ($e=0.5$) to high ($e=0.9$).
\begin{itemize}
\item $e=0.1$:
The comparison results are presented in Figure~\ref{fig:p3_k10_e1}
which clearly shows how,
across all estimators, the bias and MSE decrease as $N$ increases.
This is expected: as more data becomes available, the accuracy of
estimation increases. Figure~\ref{fig:p3_k10_e1}(a) and (b) illustrate that 
the bias and MSE
are prominent for moment and ML estimators. The bias of MML estimates
is close to zero and convincingly lower than both versions of MAP 
estimates, especially when $N<25$.
The MSE of MML estimates is smaller but close to that of MAP1 estimate.

The proportion of wins of MML estimates with respect to KL-divergence
is the highest with values of at least 70\% and 80\% when compared with 
MAP1 and MAP2, respectively (see Figure~\ref{fig:p3_k10_e1}c,d). Also, hypothesis testing indicate that
the respective estimates constituting the null hypothesis are
accepted at the 1\% significance level, as observed from the
boxplots of test statistics and p-values in Figure~\ref{fig:p3_k10_e1}(e) and (f).
\begin{figure}[ht]
\centering
\subfloat[Bias-squared]{\includegraphics[width=0.5\textwidth]{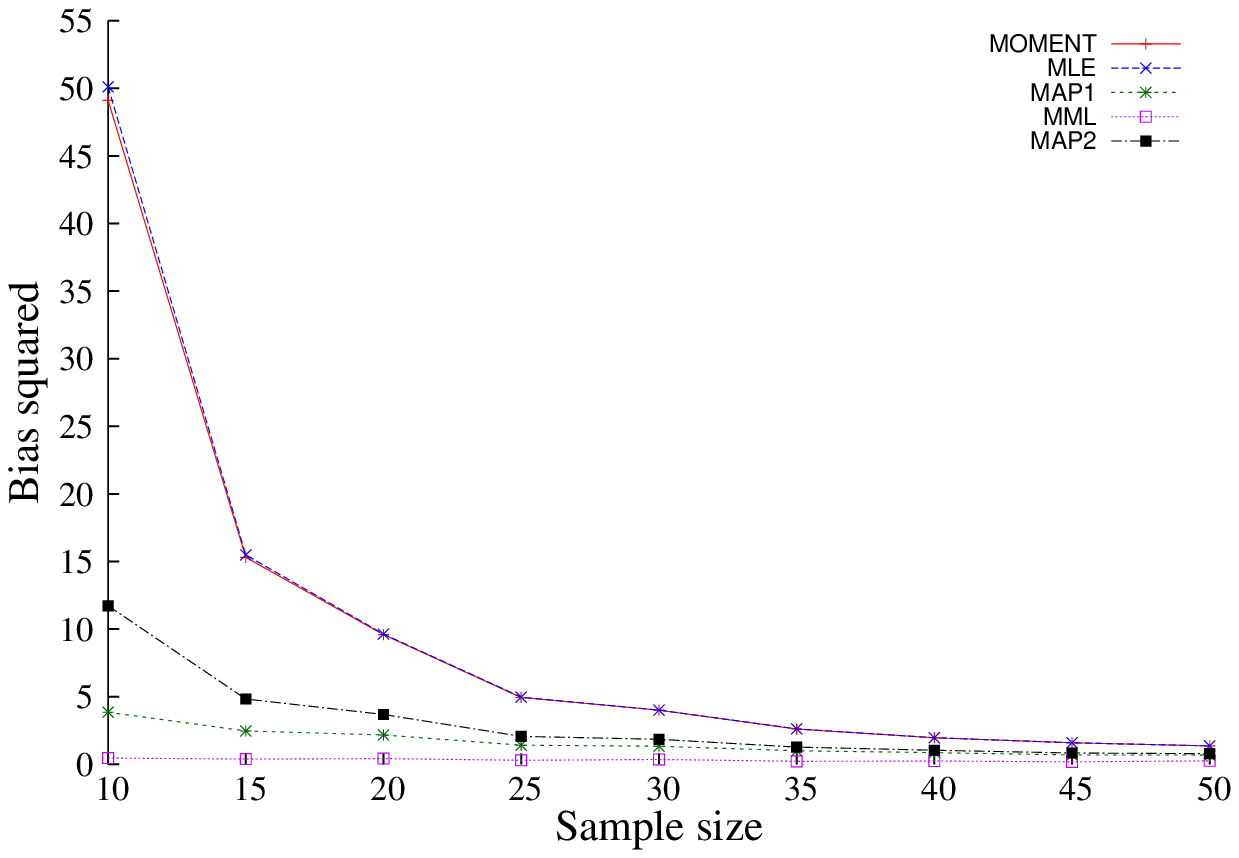}}
\subfloat[Mean squared error]{\includegraphics[width=0.5\textwidth]{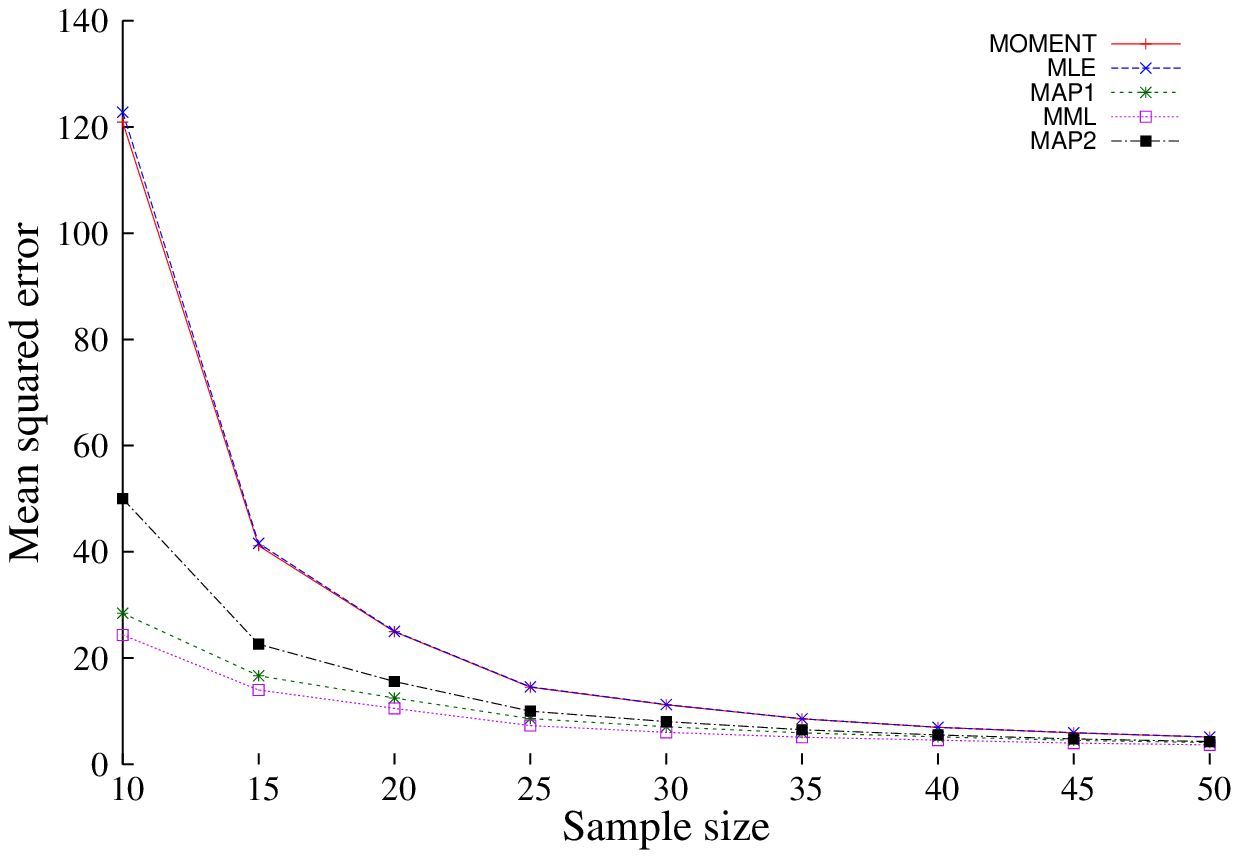}}\\
\subfloat[KL divergence (MAP version 1)]{\includegraphics[width=0.5\textwidth]{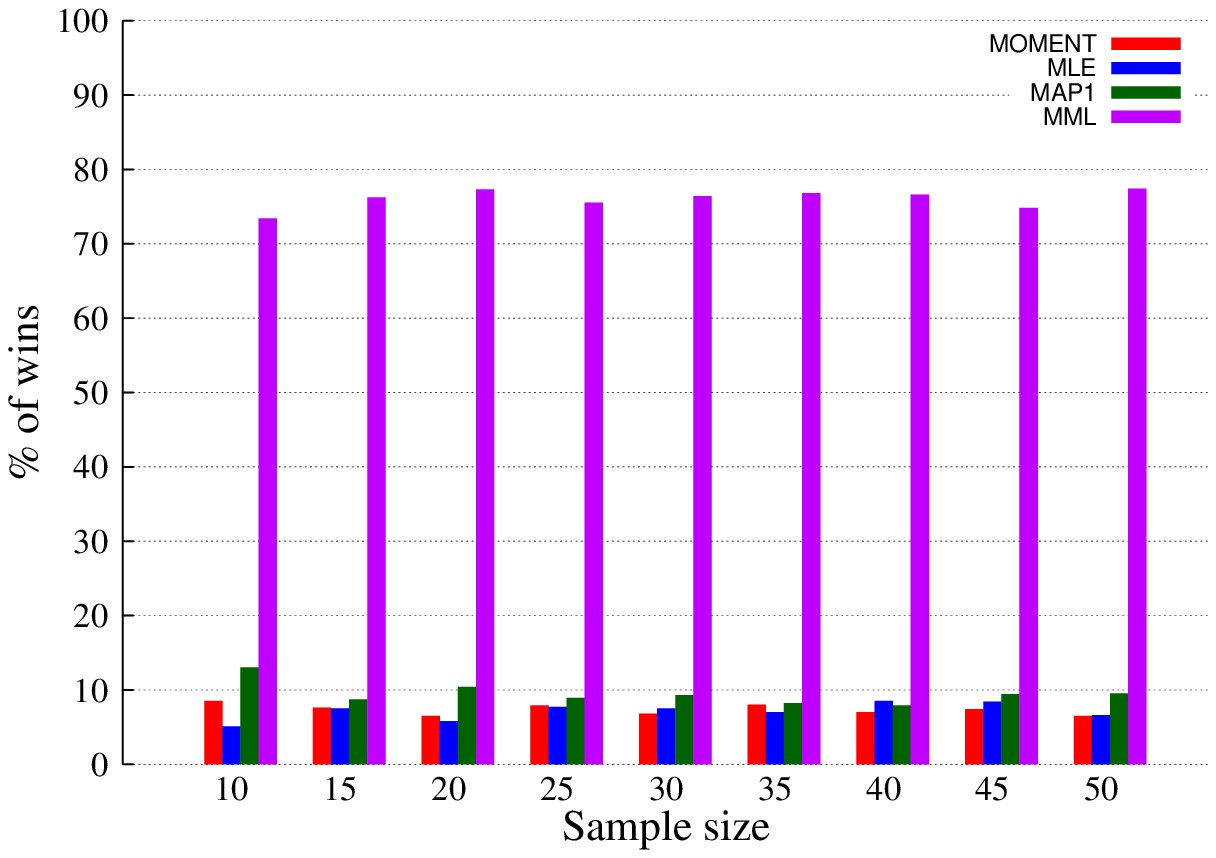}}
\subfloat[KL divergence (MAP version 2)]{\includegraphics[width=0.5\textwidth]{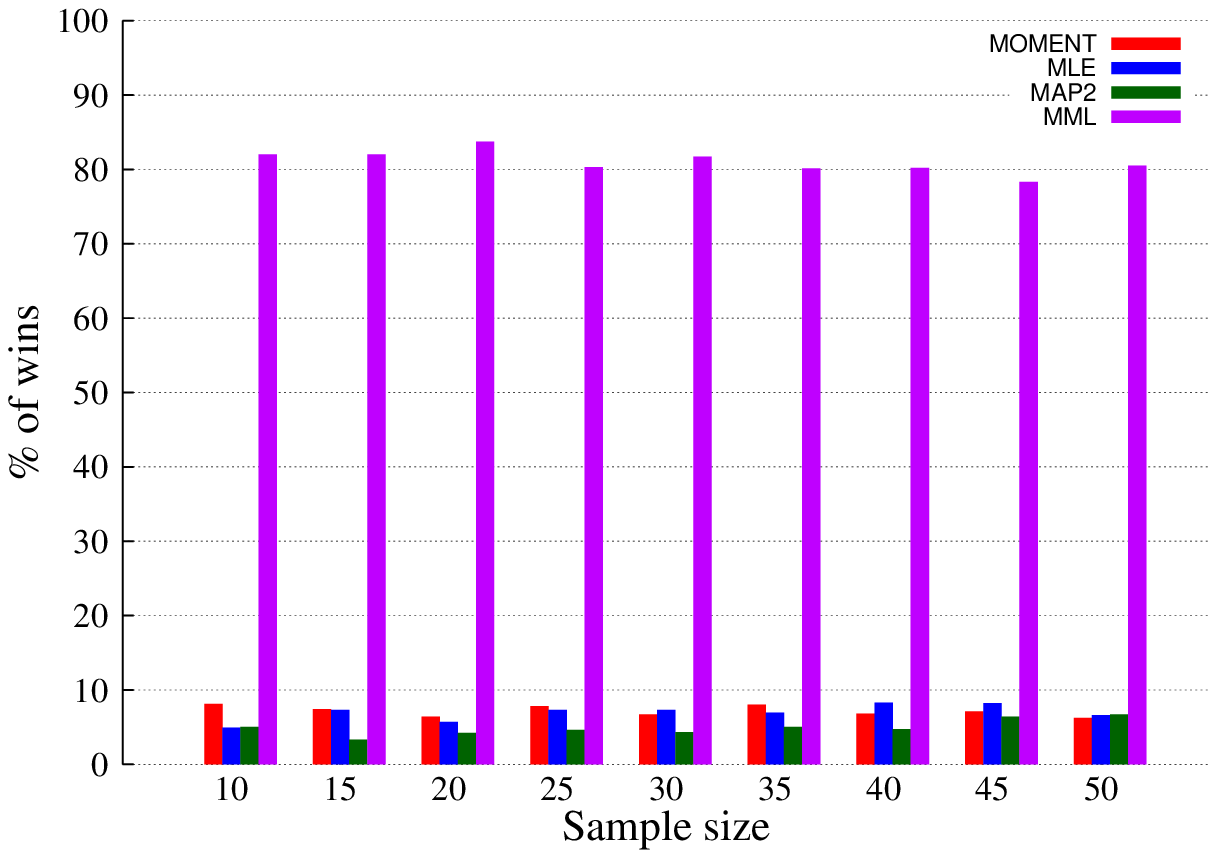}}\\
\subfloat[Variation of test statistics]{\includegraphics[width=0.5\textwidth]{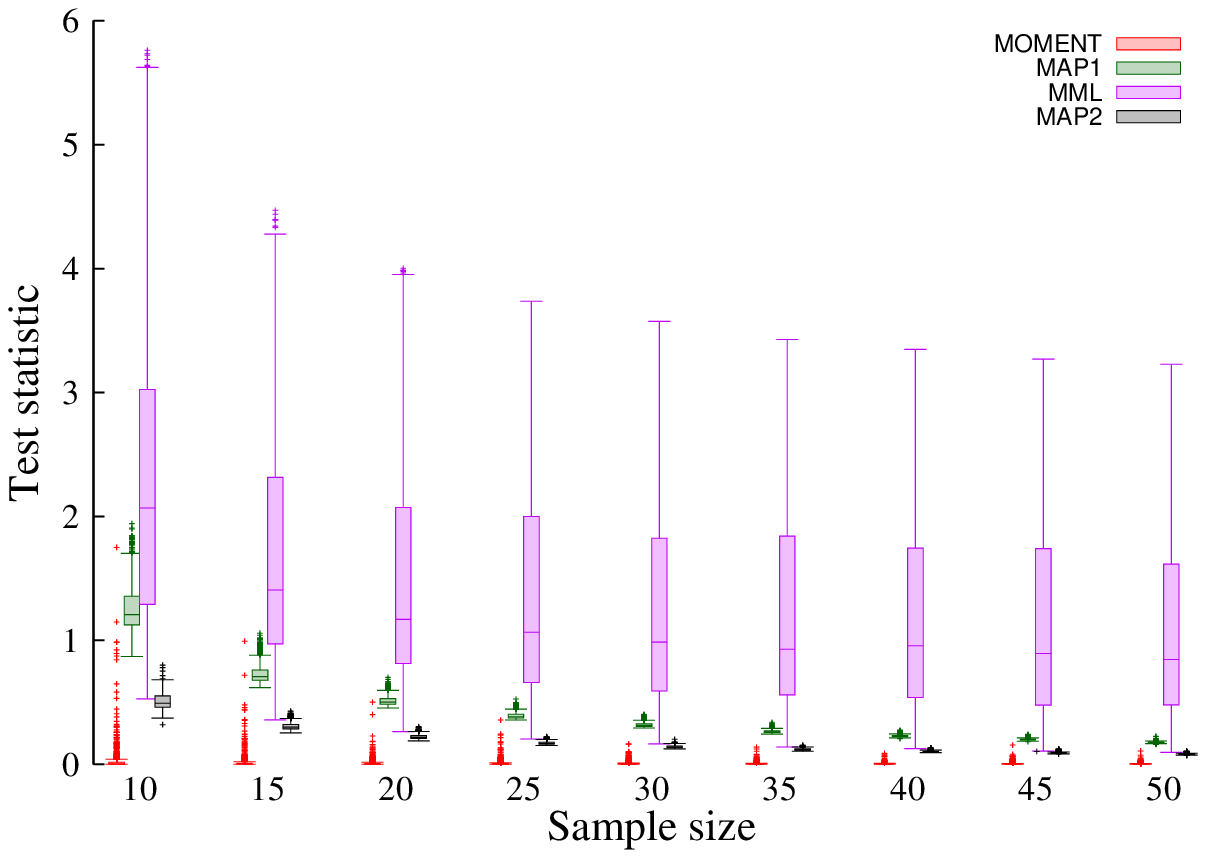}}
\subfloat[Variation of p-values]{\includegraphics[width=0.5\textwidth]{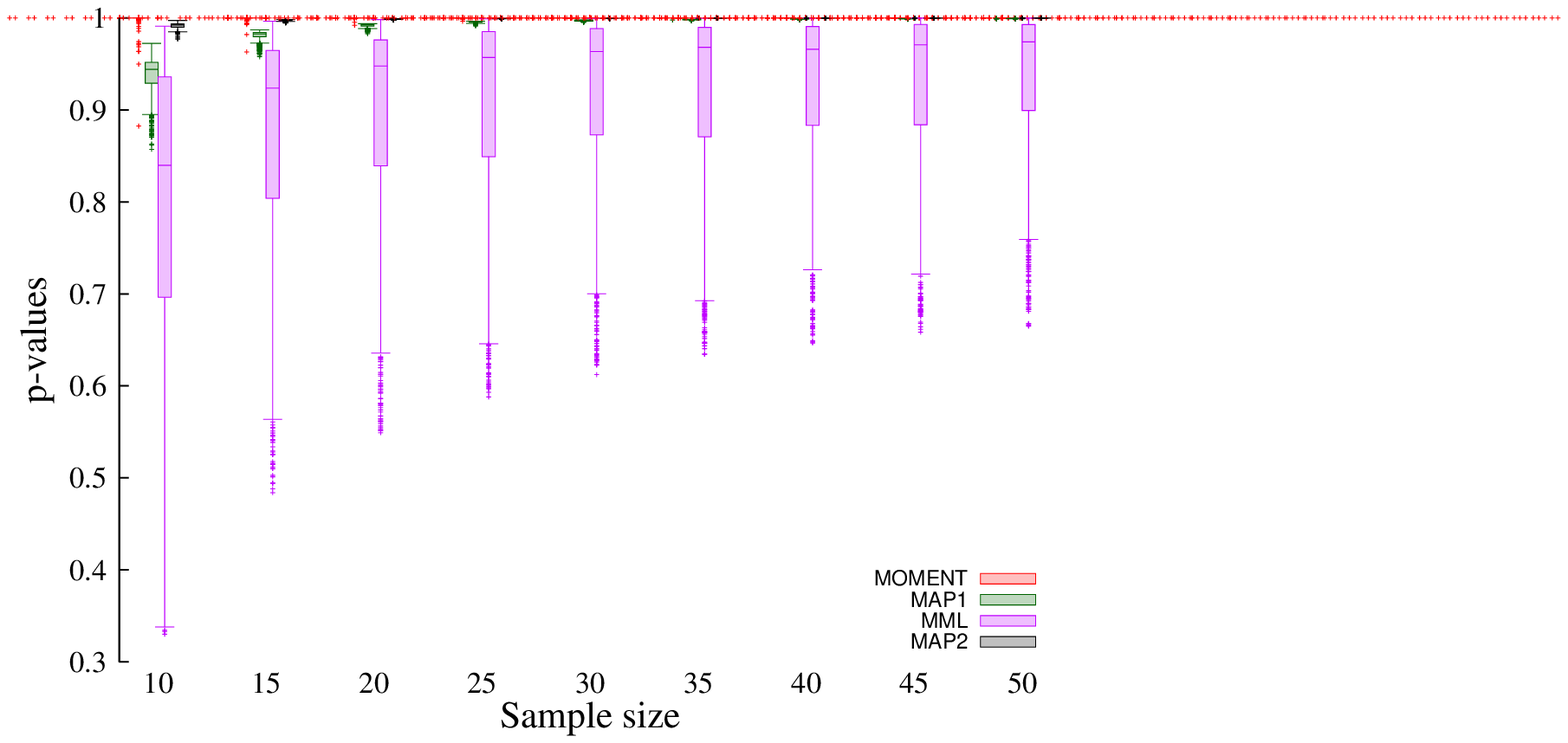}}
\caption{$\kappa=10$, eccentricity = 0.1} 
\label{fig:p3_k10_e1}
\end{figure}

\item $e=0.5$:
The comparison results are presented in Figure~\ref{fig:p3_k10_e5}.
Similar to the previous case, the bias and MSE of moment and ML 
estimates are considerable high compared to those of the MAP and MML estimates. 
Also, MAP1 estimates have greater bias and MSE as compared to MAP2 estimates.
In this case, the bias and MSE of MAP2 and MML are close to zero.

The proportion of wins of MML estimates with respect to KL divergence
is higher with about 40\% and 50\% when compared against MAP1 and MAP2
estimates, respectively. The proportion of wins are, however, lower
compared to the previous case when $e=0.1$ as shown in 
Figure~\ref{fig:p3_k10_e5}(c) and (d).
\begin{figure}[htb]
\centering
\subfloat[Bias-squared]{\includegraphics[width=0.5\textwidth]{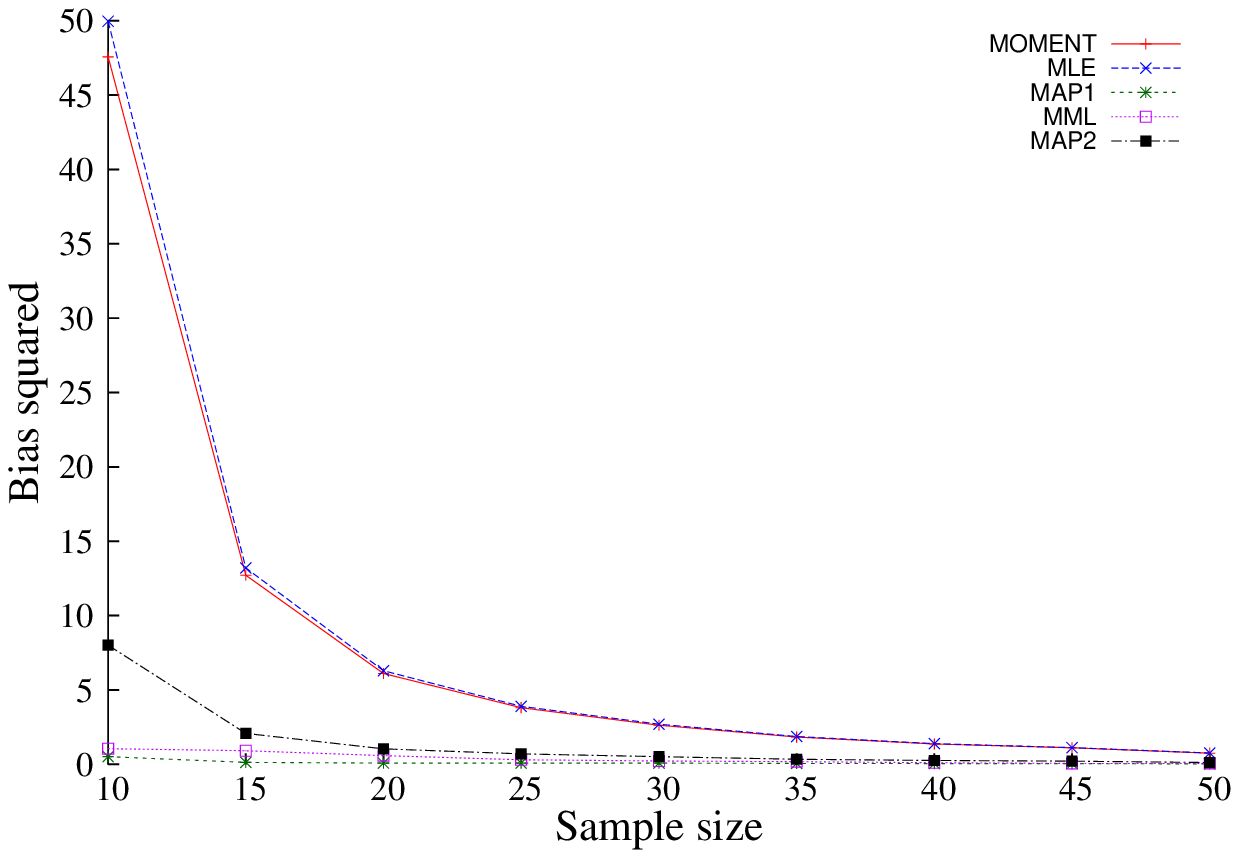}}
\subfloat[Mean squared error]{\includegraphics[width=0.5\textwidth]{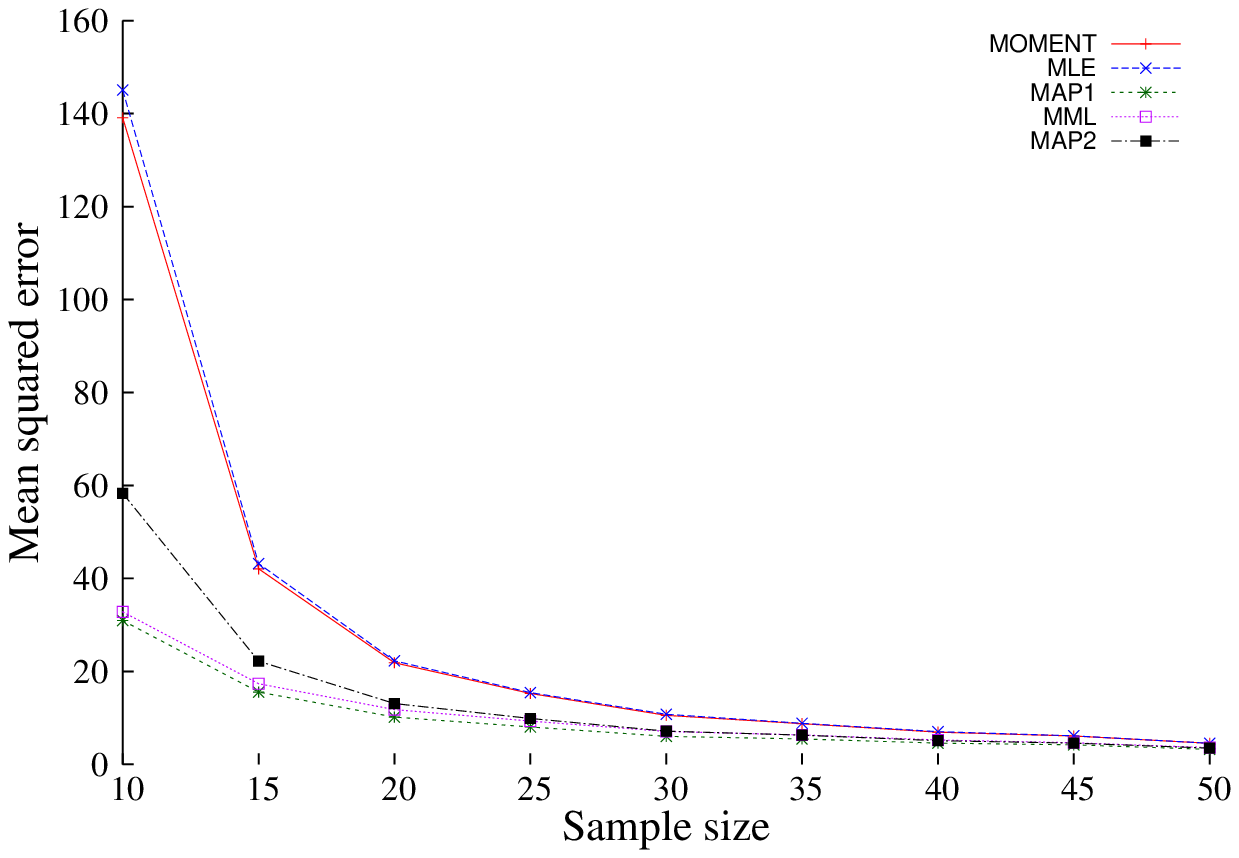}}\\
\subfloat[KL divergence (MAP version 1)]{\includegraphics[width=0.5\textwidth]{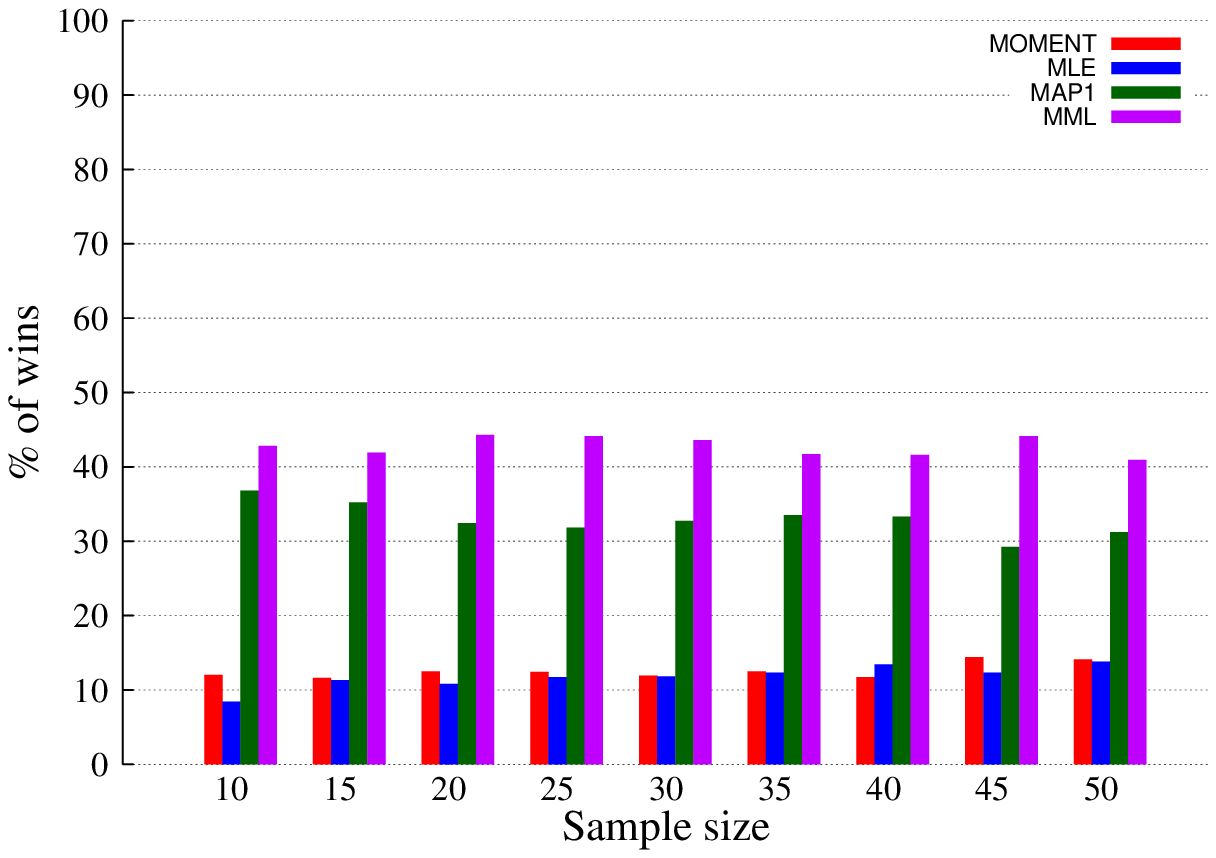}}
\subfloat[KL divergence (MAP version 2)]{\includegraphics[width=0.5\textwidth]{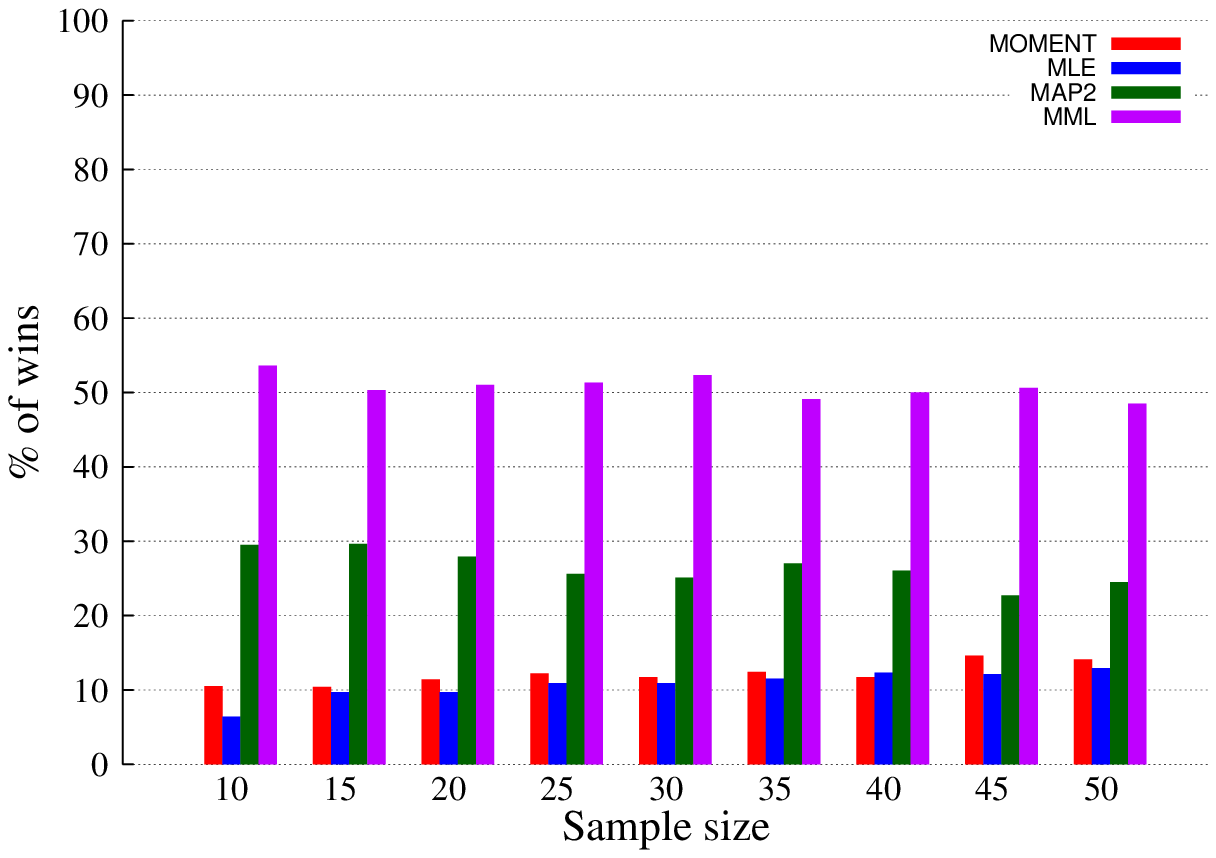}}\\
\subfloat[Variation of test statistics]{\includegraphics[width=0.5\textwidth]{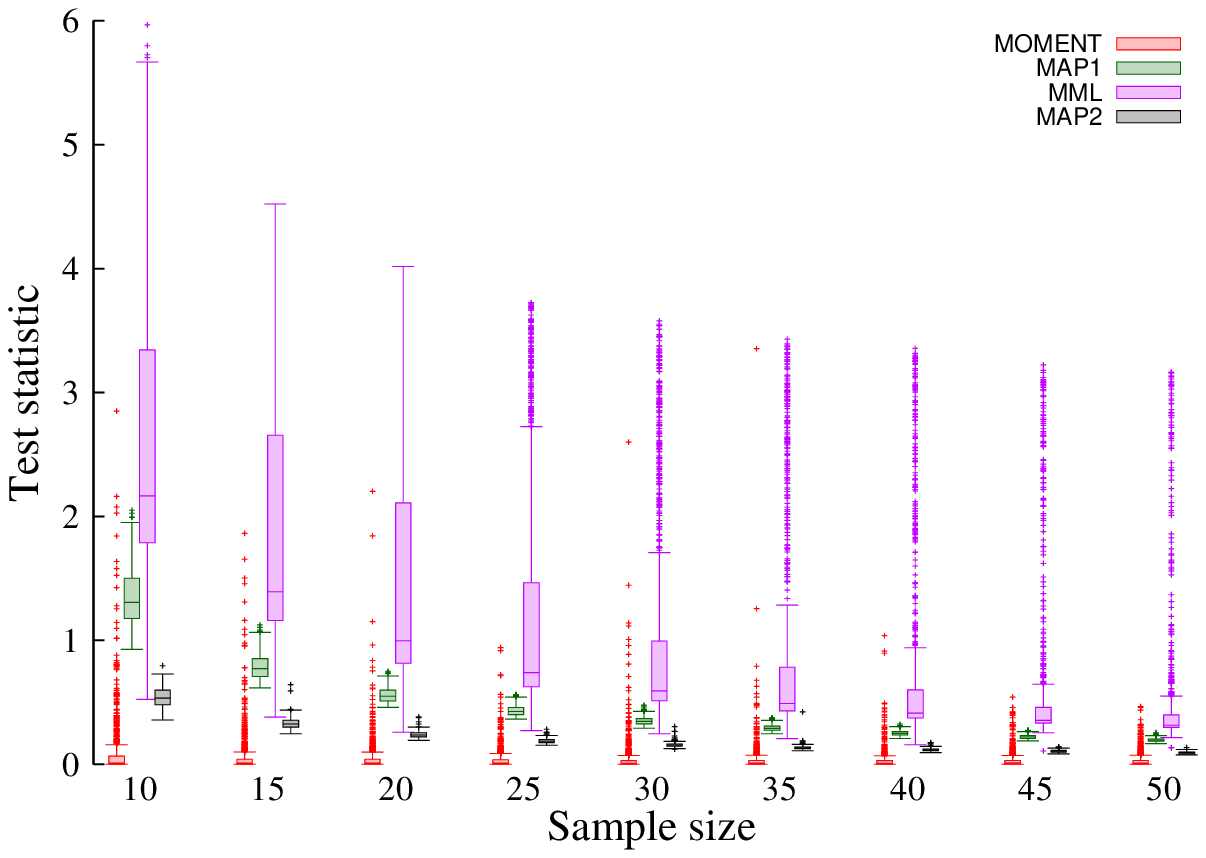}}
\subfloat[Variation of p-values]{\includegraphics[width=0.5\textwidth]{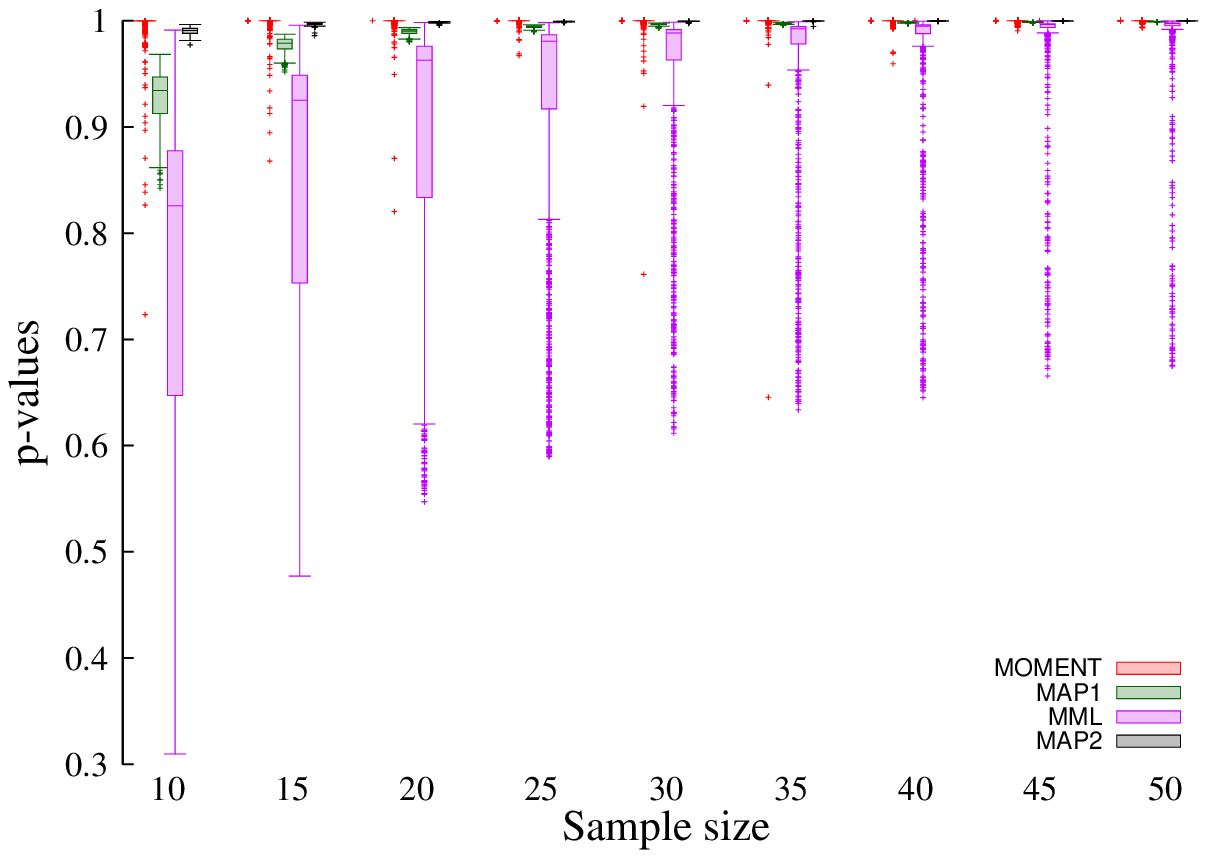}}
\caption{$\kappa=10$, eccentricity = 0.5} 
\label{fig:p3_k10_e5}
\end{figure}

\item $e=0.9$:
The comparison results are presented in Figure~\ref{fig:p3_k10_e9}.
In this case, again, the bias and MSE of moment and ML estimates
are greater compared to others. For $N<20$, the bias of MML estimates is greater
when compared to those of MAP1 and MAP2 (Figure~\ref{fig:p3_k10_e9}a).
Further, the MSE of MML estimates is greater than that of MAP1
and lower than that of MAP2. As the MSE combines the bias and
variance terms, there is a tradeoff that leads to this result.
When $N>25$, there is almost no difference in the bias and MSE due to MAP
and MML estimates.

Also, the proportion of wins of KL divergence for MAP1 is greater than all others
(Figure~\ref{fig:p3_k10_e9}c). This corresponds to the proportion
of wins as illustrated through Figure~\ref{fig:n10_p3_k10}(c), similar to 
the $N=10, \kappa=10, e = 0.9$ case. However, when compared with MAP2,
the MML estimates have greater proportion of wins $\sim40\%$
(Figure~\ref{fig:p3_k10_e9}d).
\begin{figure}[ht]
\centering
\subfloat[Bias-squared]{\includegraphics[width=0.5\textwidth]{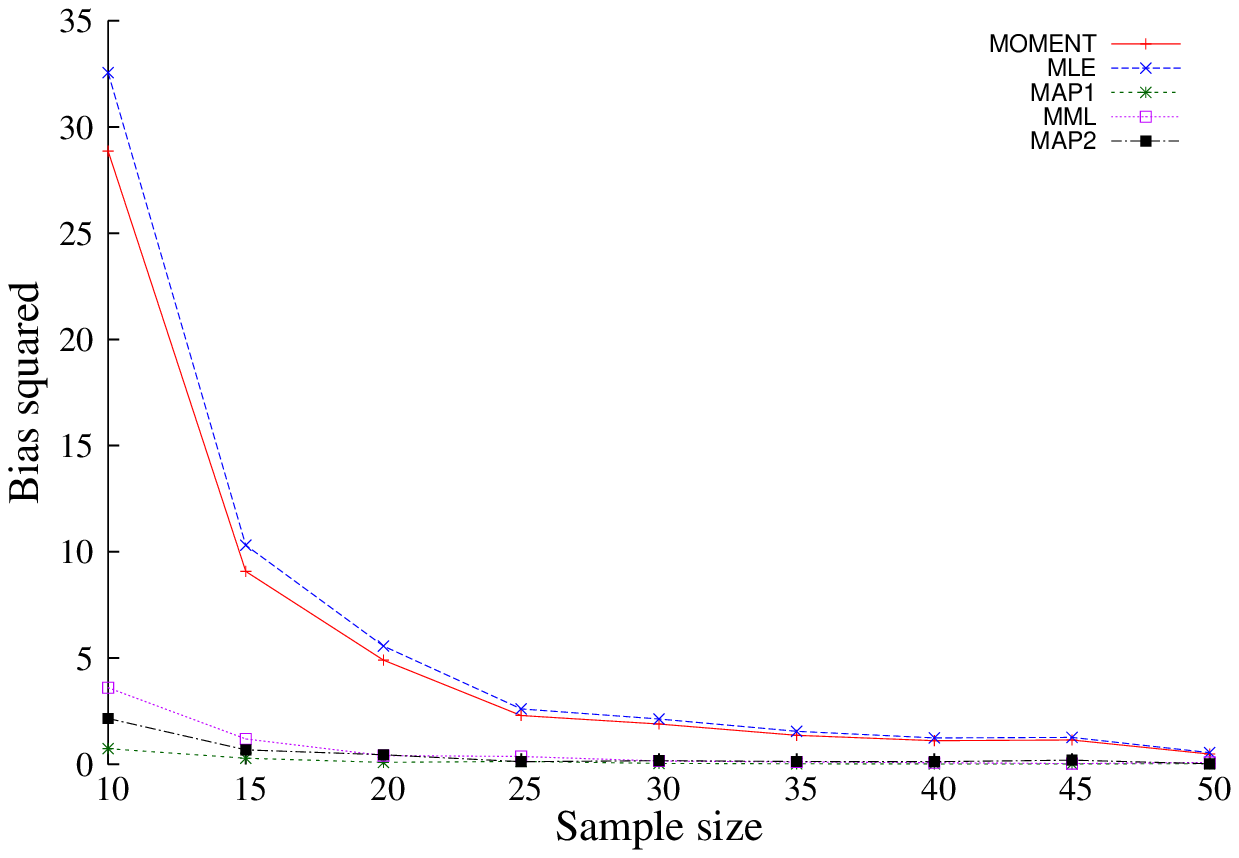}}
\subfloat[Mean squared error]{\includegraphics[width=0.5\textwidth]{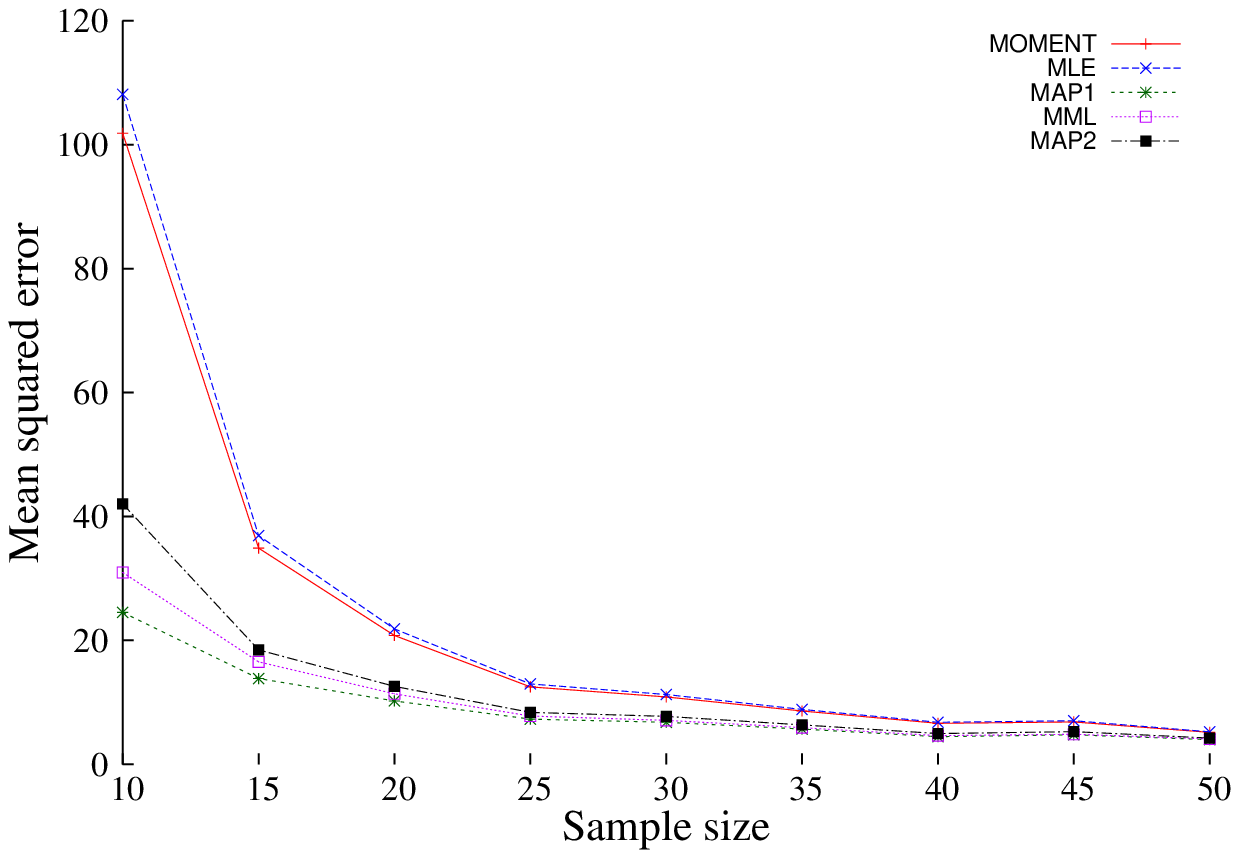}}\\
\subfloat[KL divergence (MAP version 1)]{\includegraphics[width=0.5\textwidth]{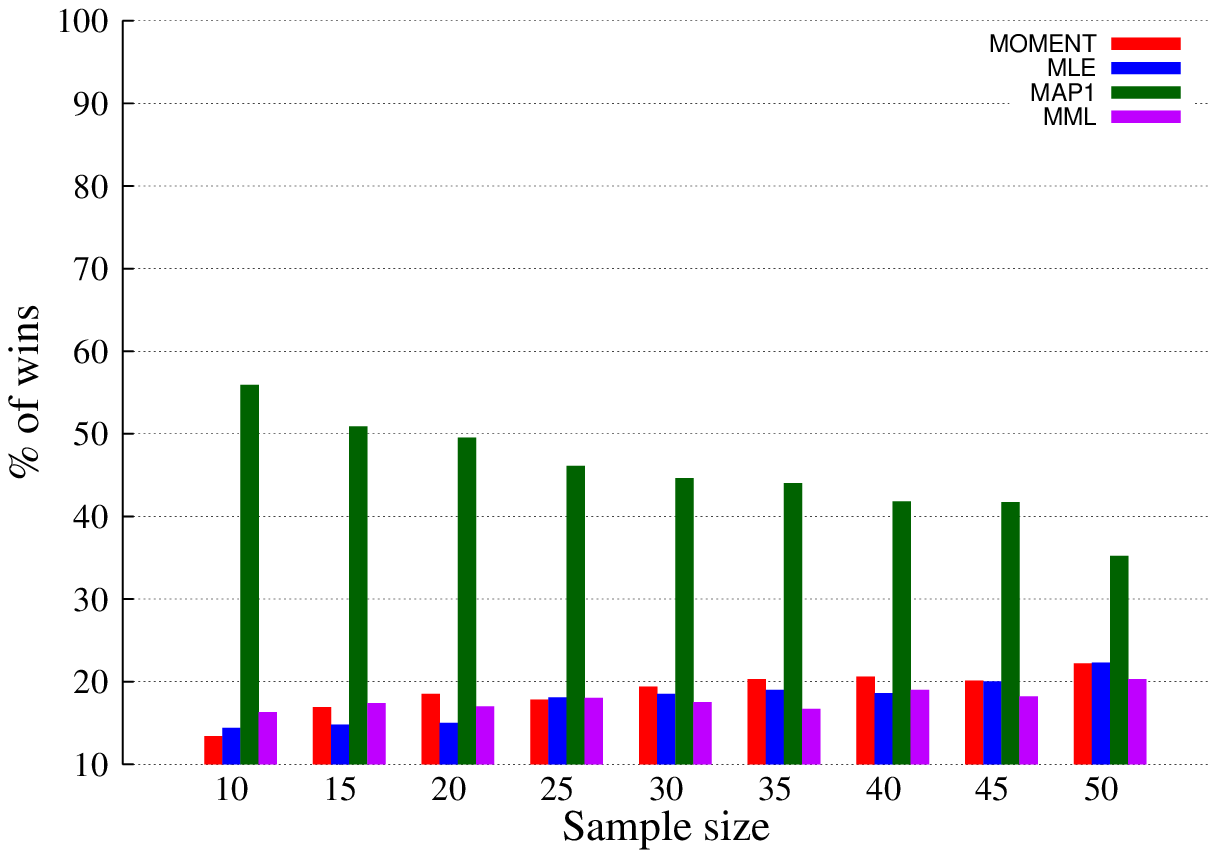}}
\subfloat[KL divergence (MAP version 2)]{\includegraphics[width=0.5\textwidth]{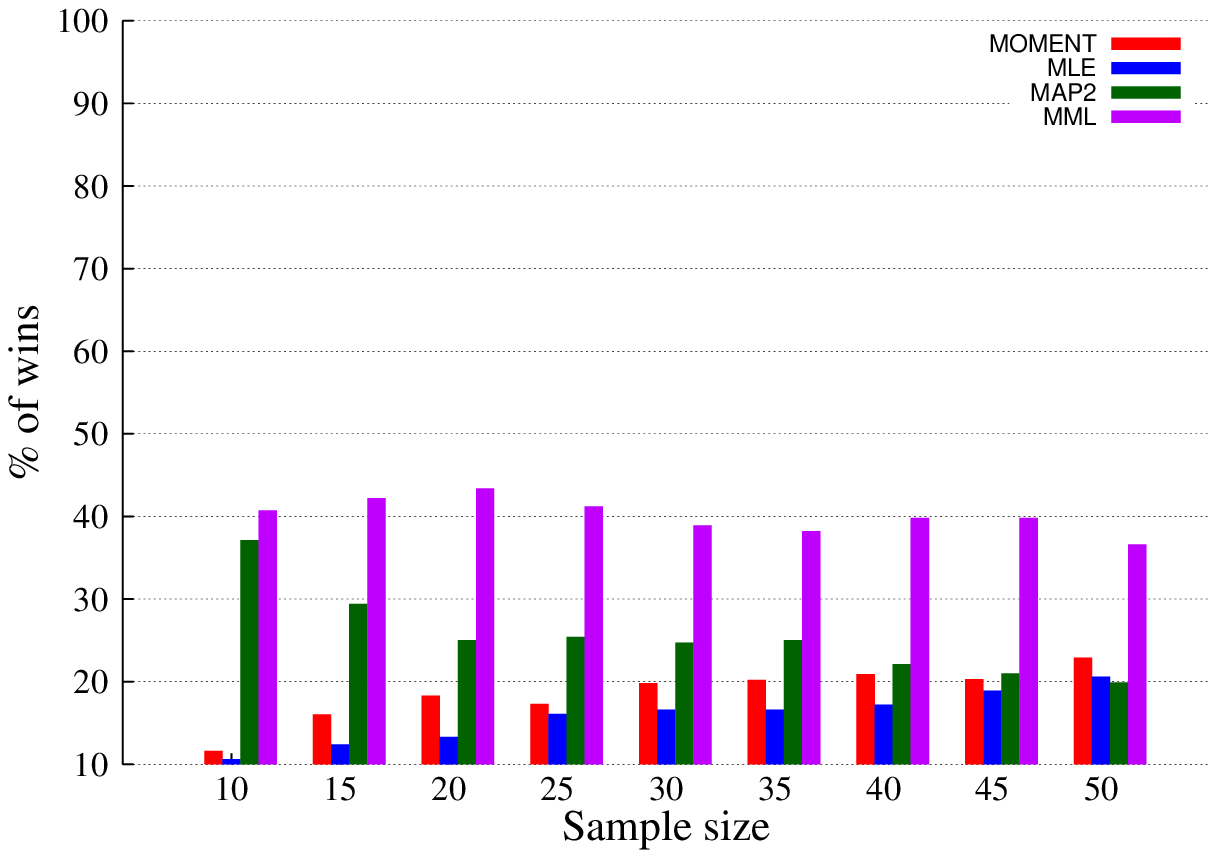}}\\
\subfloat[Variation of test statistics]{\includegraphics[width=0.5\textwidth]{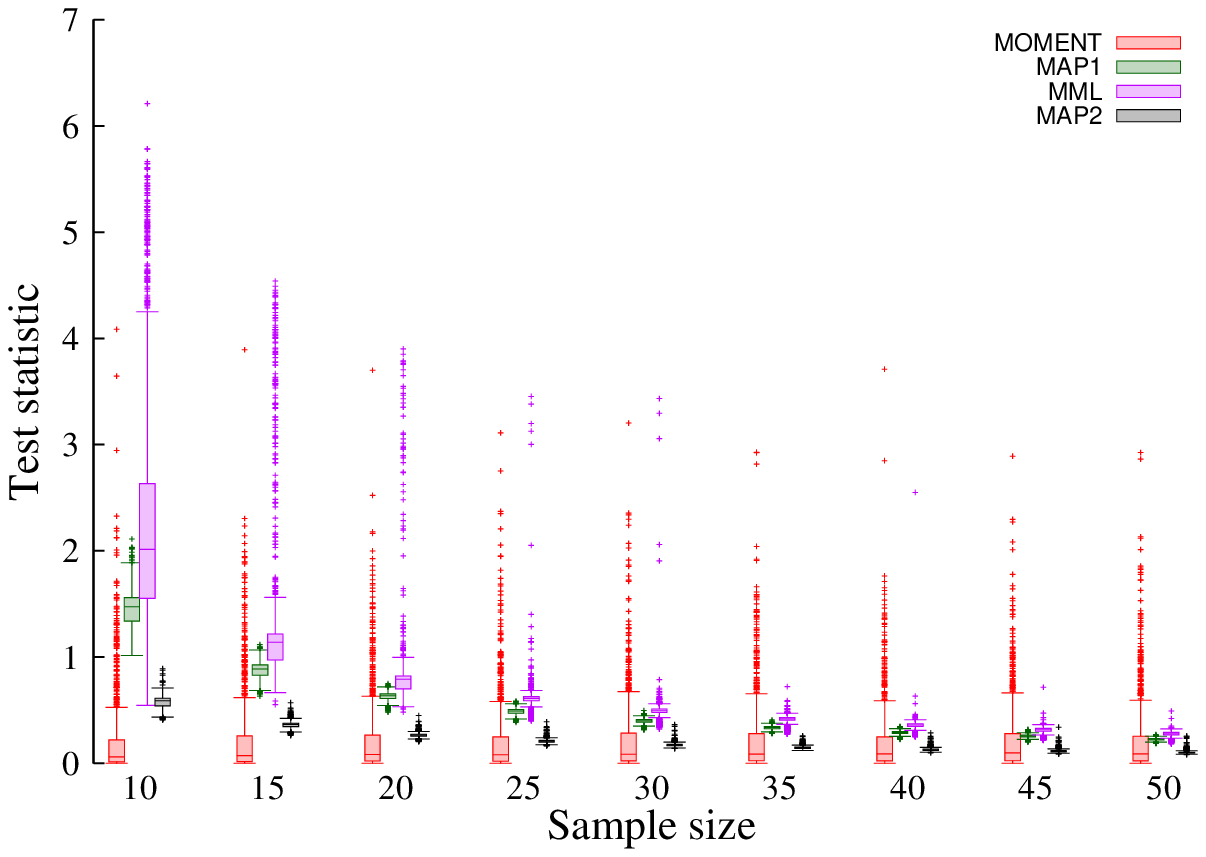}}
\subfloat[Variation of p-values]{\includegraphics[width=0.5\textwidth]{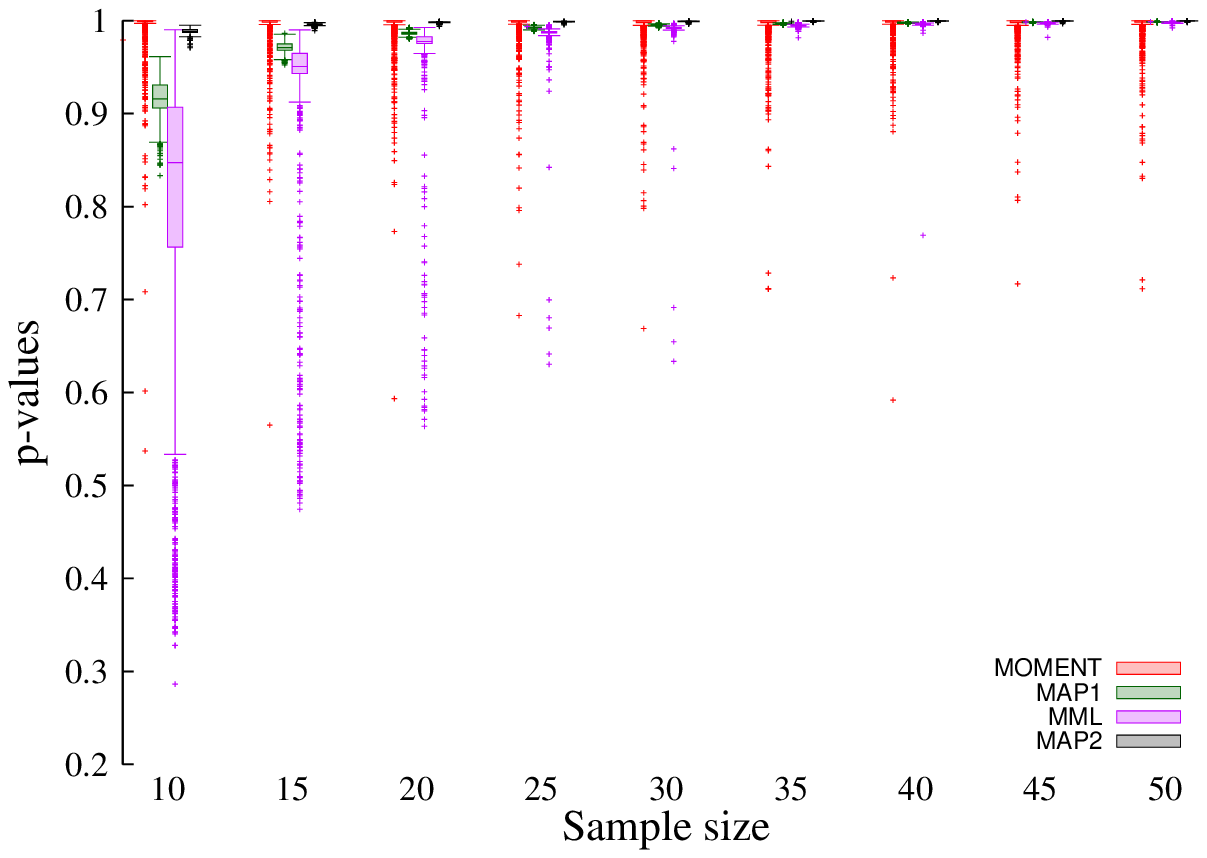}}
\caption{$\kappa=10$, eccentricity = 0.9} 
\label{fig:p3_k10_e9}
\end{figure}
\end{itemize}

The traditional ML estimators 
are known to have considerable bias, especially at lower sample sizes 
\citep{dryden1998statistical,dorebias}. 
The ML estimates of $\kappa$ in the case of a vMF distribution
are known to be biased \citep{schou1978estimation,best1981bias,cordeiro1999theory}.
Similarly, the ML estimates of a Bingham distribution, which is a special case of a \fb~distribution,
are also shown to be biased and corrections have been proposed 
\citep{cordeiro1994bias,kume2007derivatives,dorebias}. 

The MML-based estimates have been shown to be effective in reducing bias
in the case of a vMF distribution \citep{multivariate_vmf}. For an \fb~distribution,
we empirically demonstrated that, in comparison to the moment and ML estimates, the MML-based
estimates have lower bias and MSE. Further, when compared to MAP
estimates, MML estimates are competitive, particularly considering that MAP estimates are dependent 
on the parameterization. 
As a result, MAP estimates are inconsistent and should therefore be avoided. 
In contrast, MML-based estimates are invariant to alternative parameterizations
\citep{oliver1994mml,WallaceBook}.
In this regard, we discuss another parameterization in Appendix~\ref{app:alternative_parameterization}
involving all parameters of an \fb~distribution to further strengthen our case.

\section{Experiments involving \fb~mixtures}
\label{sec:mixture_experiments}
To demonstrate the applicability of \fb~mixtures,
we consider the problem of mixture modelling of directional data
arising out of protein three-dimensional conformations.
A protein chain consists of a sequence of amino acids (residues).
Each residue has a central carbon atom $C_{\alpha}$.
If $C_{\alpha}^i$ and $C_{\alpha}^{i+1}$ denote the carbon atoms
at positions $i$ and $i+1$ in the protein chain, then
the distance between these successive atoms is highly
constrained to be 3.8\AA~because of the
chemical interactions between the constituent atoms. Thus,
$C_{\alpha}^{i+1}$ atom lies on a sphere of radius $\sim3.8$\AA~whose 
centre is $C_{\alpha}^i$. The direction vector from $C_{\alpha}^i$ to
$C_{\alpha}^{i+1}$ is considered a point in the data set.
Given the Cartesian coordinates
of a $C_{\alpha}$ atom, its co-latitude ($\theta$) and longitude ($\phi$)
are determined with respect to the previous $C_{\alpha}$ atom
in a consistent manner \citep{multivariate_vmf}.
The set of all $(\theta,\phi)$ pairs form the directional data corresponding
to a given set of protein structures.
The protein data set considered is the publicly available ASTRAL SCOP-40 
(version 1.75) database \citep{murzin1995scop}. Out of the entire dataset,
the ``$\beta$ class'' proteins comprising of 1802 structures is a case in point.
The empirical distribution consists of 251,346 $(\theta,\phi)$ pairs
and we infer mixtures on this directional data using the search method
described in Section~\ref{subsec:search_method}.

\subsection{Evolution of vMF and \fb~mixtures}
Mixtures of vMF distributions were previously explored by \citet{multivariate_vmf}.
This entailed estimating the vMF concentration parameter $\kappa$ using MML.
They use Taylor series approximations (Newton's and Halley's root-finding methods)
in the computation of the MML estimates of $\kappa$. Both these root-finding approaches
are truncated after two iterations in order to do a fair comparison with the other 
contemporary $\kappa$ approximations that were 
discussed in that work (see Equations 8 and 9 in \citet{multivariate_vmf}).
The obtained vMF $\kappa$ estimates were used as part of the mixture modelling apparatus. 
As a result, the search method employed for determining the optimal number of
vMF components was terminated prematurely. 
Furthermore, the search heuristic employed in \citet{multivariate_vmf}
does a random selection of initial means of the child components
while splitting a parent component. 
In contrast to these, the vMF $\kappa$ estimates used in the current work
correspond to the converged values (without truncating prematurely). 
Further, as per the search method described in this work, while
splitting a parent component,
the initial means of the children are chosen such that they are reasonably apart 
which gives them the best chance to form two distinct sub-components in order to
escape a local optimum (explained in Section~\ref{subsubsec:splitting}).

The search method infers a 37-component vMF mixture and terminates
after 49 iterations involving split, delete, and merge operations. 
When modelled using \fb~distributions, the search method infers
23 components and terminates after 33 iterations.
In each of these iterations, for every intermediate
$K$-component mixture, each constituent component is split, deleted, and merged
(with an appropriate component) to generate improved mixtures.
The method terminates when these perturbations do not result in an improvement.

\begin{figure}[ht]
\centering
\subfloat[vMF mixture]{\includegraphics[width=0.5\textwidth]{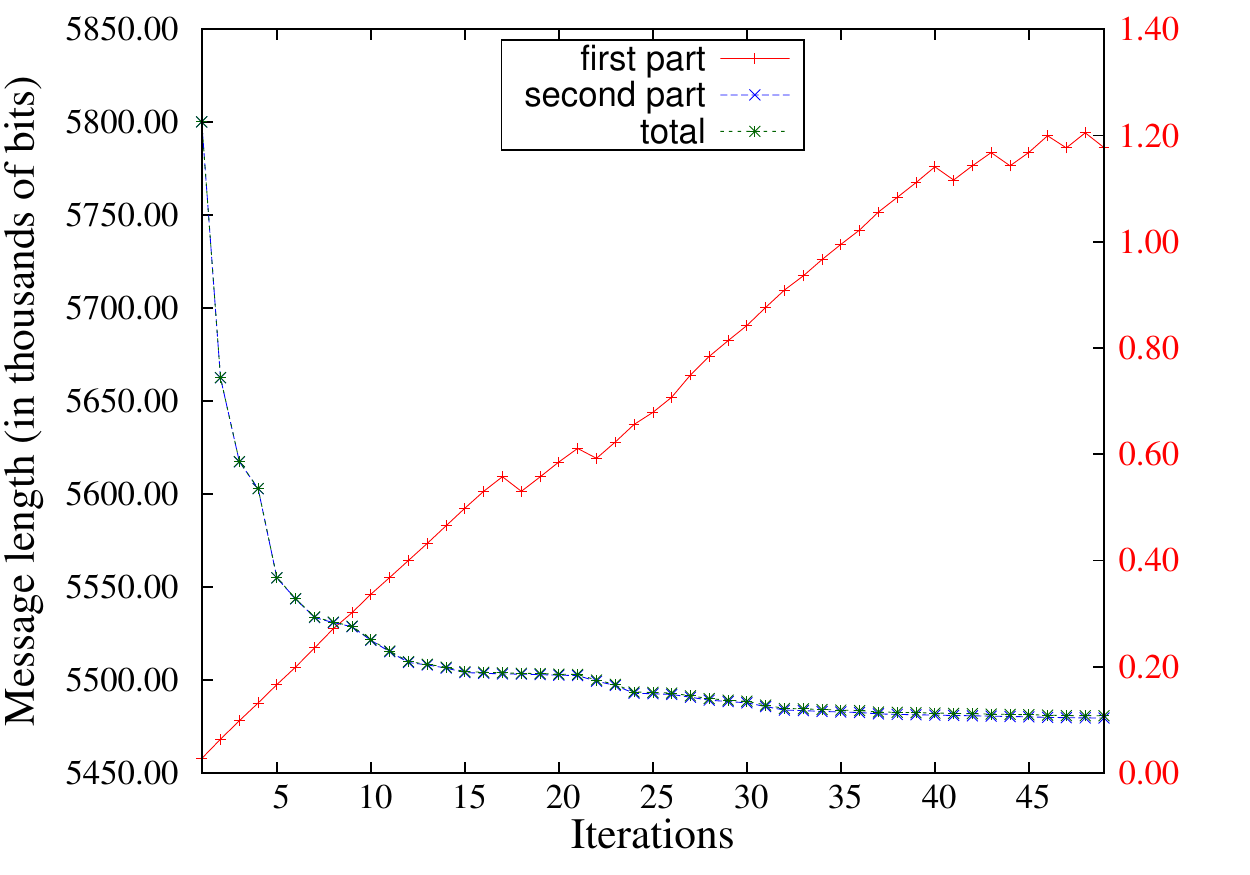}}
\subfloat[\fb~mixture]{\includegraphics[width=0.5\textwidth]{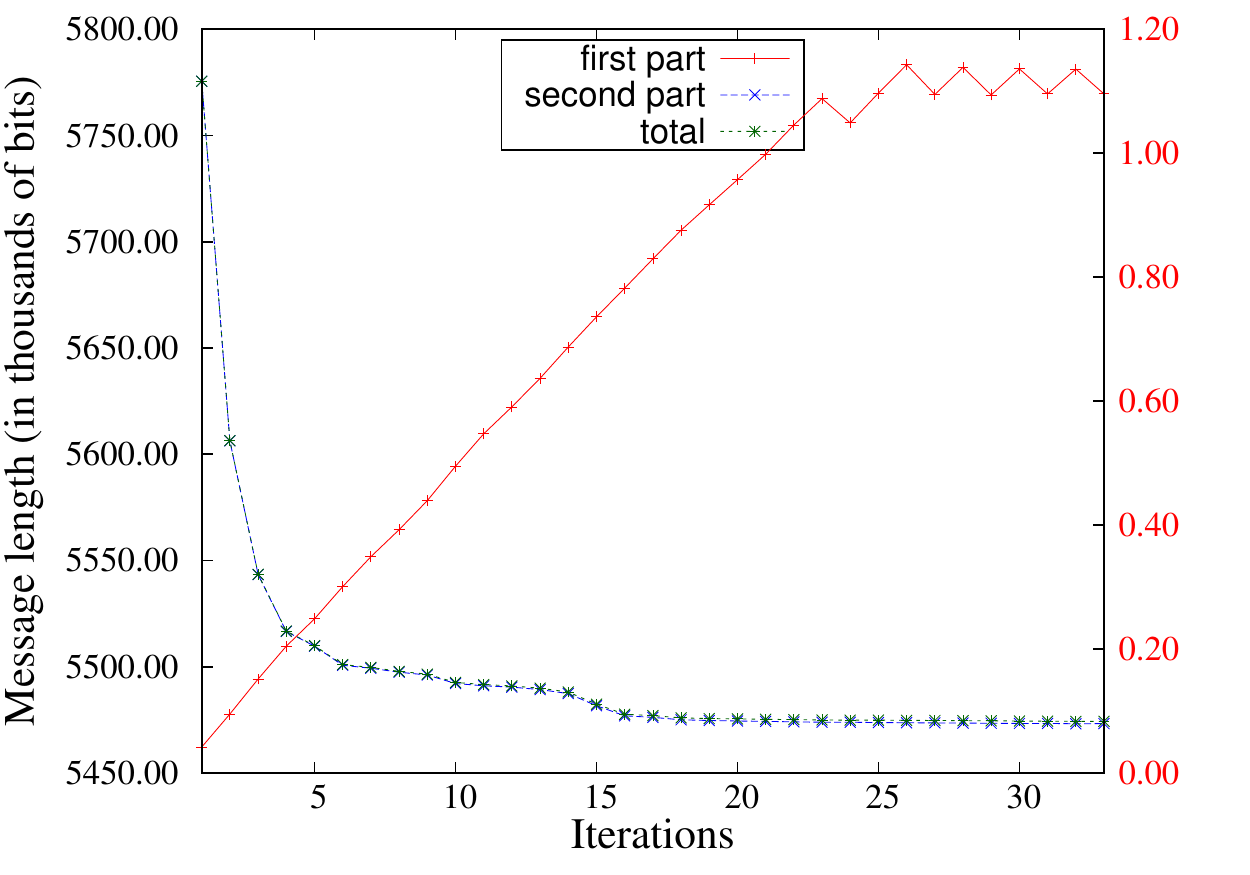}}
\caption{Evolution of mixtures inferred by the search method.
         Note there are two Y-axes in both (a) and (b) with 
         different scales: the first part of the message follows the right side Y-axis (red);
         while the second part and total message lengths
         follow the left side Y-axis (black).
} 
\label{fig:protein_mix_evolution}
\end{figure}

In the case of vMF mixture, the search method begins with a
one-component mixture, continuously favours splits over
delete and merge operations until a 17-component mixture
is inferred. This corresponds to the steady increase in the first
part of the message length as observed by the red curve in
Figure~\ref{fig:protein_mix_evolution}(a) until the $17^{\text{th}}$
iteration. Thereafter, a series of deletions and splits result in an 
intermediate sub-optimal 19-component mixture at the end of the $23^{\text{rd}}$ 
iteration. This is characterized by the
step-like behaviour of the red curve between the $17^{\text{th}}$ 
and $24^{\text{th}}$ iteration. The first part of the message
is dependent on the number of components
(model complexity) and an increase in number of mixture components
leads to an increase in the encoding cost of the parameters.
From the $24^{\text{th}}$ iteration, the method continues to split 
the constituent components until a 36-component mixture is inferred
after 40 iterations. 
This is reflected in the continuous rise of the red curve in
Figure~\ref{fig:protein_mix_evolution}(a) between $25^{\text{th}}$
and $40^{\text{th}}$ iterations.
Thereafter, through a series of perturbations, 
the final resultant mixture has 37 components at the end of 49 iterations,
characterized by a step-like behaviour towards the end between
$40^{\text{th}}$ and $49^{\text{th}}$ iterations.

In the case of \fb~mixture, the search method infers a 
23-component mixture at the end of 23 iterations by continuous
splitting. This corresponds to the steady increase in the first part
of the message length denoted by the red curve in 
Figure~\ref{fig:protein_mix_evolution}(b).
From here on, after a series of perturbations,
the final mixture stabilizes at the end of $33^{\text{rd}}$ iteration
thereby resulting in a 23-component mixture.
This is characterized by the step-like behaviour corresponding to
intermediate reduction and increase in the number of mixture
components between the $24^{\text{th}}$ and $33^{\text{rd}}$ iterations.

In both cases, the second part of the message length continues to
decrease with an increase in the number of mixture components. 
An initial sharp decrease is observed in both mixture types.
The search method terminates when the increase in first part 
dominates the reduction in the second part leading to an increase in
total message length.

\subsection{Comparison of vMF and \fb~mixture models}
\label{subsec:vmf_vs_kent_mixtures}
\begin{figure}[htb]
\centering
\subfloat[vMF MML mixture (37 components)]{\includegraphics[width=0.725\textwidth]{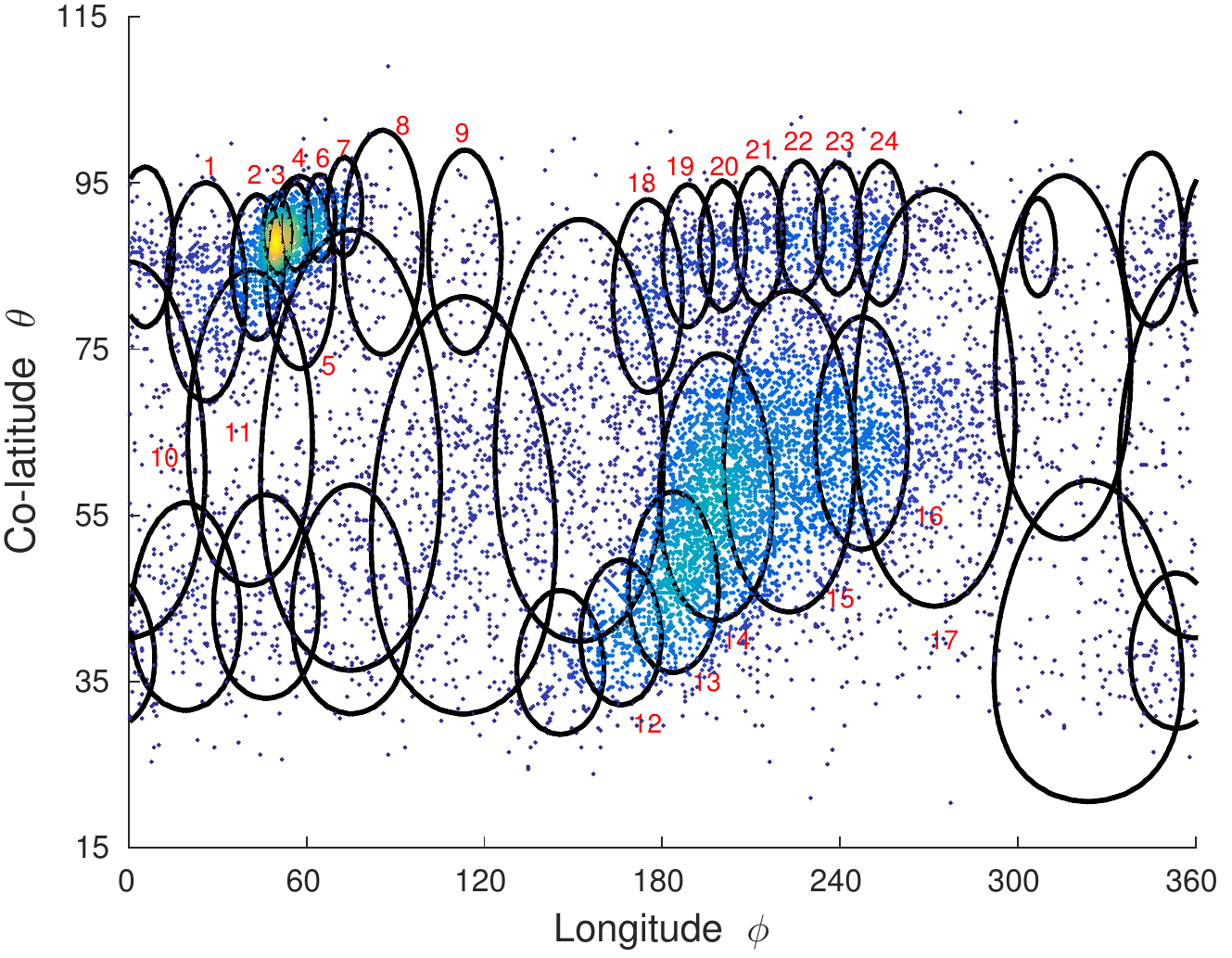}}\\
\subfloat[\fb~MML mixture (23 components)]{\includegraphics[width=0.725\textwidth]{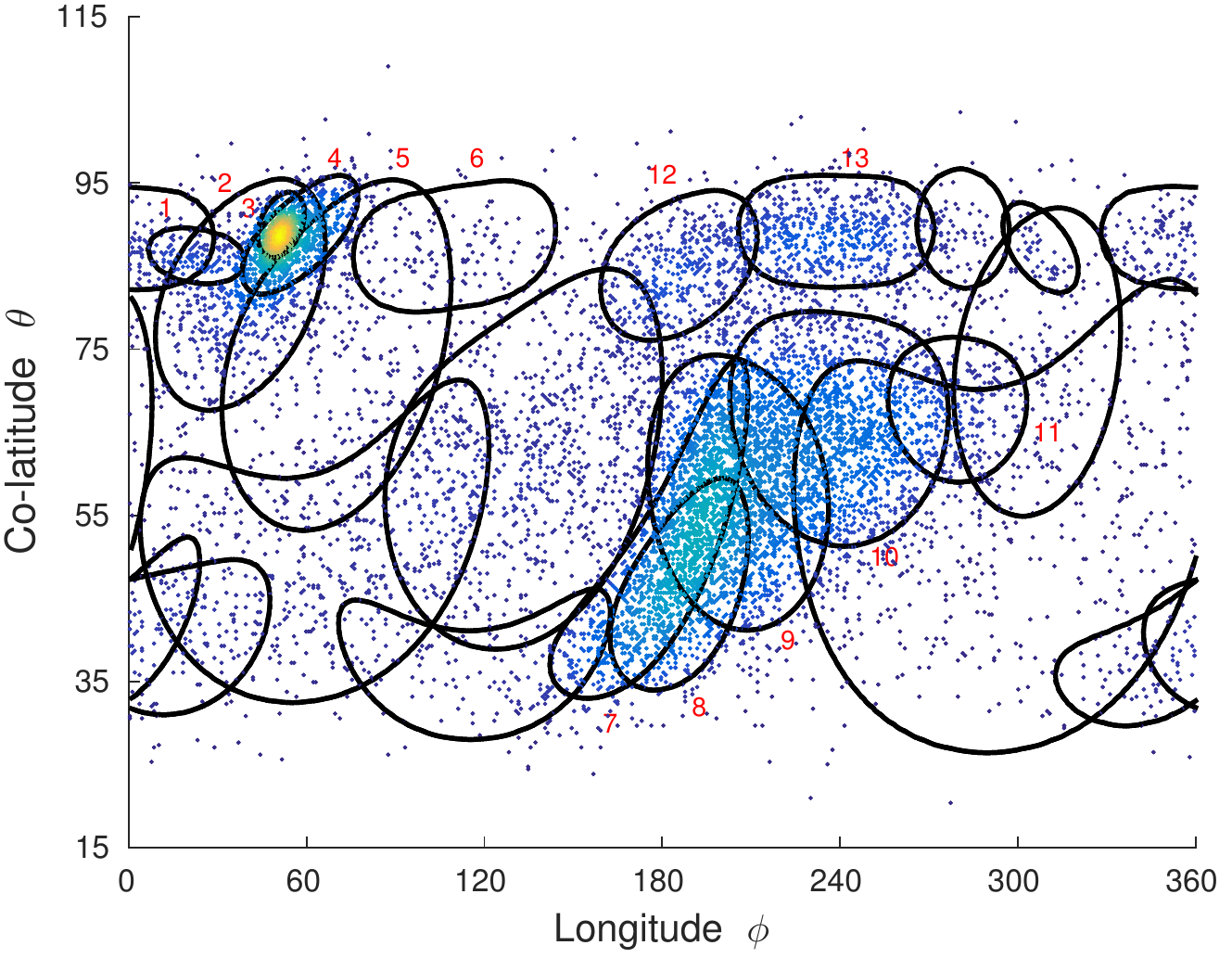}}
\caption{Mixtures inferred on the $\beta$-class proteins ($\theta$ and $\phi$ are in degrees).} 
\label{fig:vmf_kent_mml_mix}
\end{figure}

The resulting vMF and \fb~mixtures are shown in Figure~\ref{fig:vmf_kent_mml_mix}.
In order for effective visualization of the individual mixture components,
the illustration includes the contours of the components such that
they encompass 80\% of the probability corresponding to each component. 
The data plotted is a random sample of $10000~(\theta,\phi)$ pairs drawn from the empirical
distribution of $\beta$ class of proteins. 
The regions in Figure~\ref{fig:vmf_kent_mml_mix} are coloured based on the empirical distribution
(heat map). There are two distinguishable regions of the distribution of
$\theta$ and $\phi$ values. At $(\theta,\phi)\sim(90^{\circ},60^{\circ})$, there
is a concentrated mass which corresponds to the \emph{helical} region in a typical protein.
The area characterized by $\theta\in(40^{\circ},80^{\circ}),~\phi\in(180^{\circ},240^{\circ})$
roughly corresponds to the \emph{strand} region in a typical protein.

The search method inferred a 37-component vMF mixture and a 23-component \fb~mixture.
It is observed that the number of components used to model the entire collection of 251,346 $(\theta,\phi)$-pairs
using a \fb~mixture model is fewer compared to a vMF mixture.
This is expected as a vMF distribution is a specific case of a \fb~distribution
and, hence, a vMF mixture requires more number of components to model data that is
asymmetrically distributed. 
In Figure~\ref{fig:vmf_kent_mml_mix}(a), the vMF mixture components 1-11 
are used to model the helical region (approximately), whereas
in Figure~\ref{fig:vmf_kent_mml_mix}(b), the same region is modelled using
\fb~mixture components 1-6.
Similarly, the strand region in the proteins is modelled by 
components 12-17 in the vMF case, whereas, it is modelled 
by components 1-11 using \fb~mixture. Further, components 18-24 in the vMF mixture
and components 12, 13 in the \fb~mixture model the same region.
The other regions in the protein directional data space follow the same
modelling pattern, that is, with fewer \fb~components.
These observations reflect the better explanatory power of \fb~mixtures
compared to vMF mixtures.

Compared to a singleton vMF distribution, the encoding cost of the parameters of
a \fb~distribution would be greater as it is a complex model with more number
of parameters. As shown in Table~\ref{tab:vmf_kent_mml_mix}, 
the encoding cost of the parameters of the inferred 
23-component \fb~mixture is 1095 bits. 
A vMF mixture with the same number of components has a first part equal to 749 bits
(a difference of 346 bits).
However, the second part of the message (the fit to the data) is lower
for the \fb~mixture (a difference of $\sim17,000$ bits).
Hence, the gain in the second part outweighs the greater cost of encoding
the more complex \fb~mixture.
Thus, the total message length is lower for the \fb~mixture and serves
as a better model to explain the data.
If the 23-component vMF mixture is compared with the 37-component vMF mixture
inferred by our search method, the vMF mixture with 23 components 
has smaller first part and greater second part
(Table~\ref{tab:vmf_kent_mml_mix}). The 37-component mixture
has a first part equal to 1177 bits compared to 749 bits in the 23-component case
(a difference of 428 bits). However, there is a gain of 11,000 bits in the second
part, thus, resulting in a lower total message length in the 37-component case.
Through this analysis, it is shown how the tradeoff of choosing a complex
model and the quality of fit is addressed using the MML framework. 
\begin{table}[htb]
  \centering
  \caption{Message lengths corresponding to mixtures inferred on the protein directional data.}
  \begin{tabular}{|c|c|c|c|c|}
  \hline
   Mixture  & Number of   &         First part        & Second part         &  Total message length                \\ \cline{4-5}
    model   & components  &  (thousands of bits)  &\multicolumn{2}{c|}{(millions of bits)} \\ \hline
    vMF     &   23        &   0.749               & 5.490         & 5.491                  \\
    vMF     &   37        &   1.177               & 5.479         & 5.481                  \\ 
    \fb     &   23        &   1.095               & 5.473         & \textbf{5.474}         \\ 
    \hline
  \end{tabular}
  \label{tab:vmf_kent_mml_mix}
\end{table}

It is also interesting to note the shape of the contours generated by both vMF
and \fb~mixtures. A vMF distribution caters to symmetrically distributed data 
and has circular contours of constant probability on a spherical surface. 
Hence, in the $\theta\phi$ space, we see regular oval-shaped contours as 
shown in Figure~\ref{fig:vmf_kent_mml_mix}(a). In contrast, a \fb~distribution 
has ellipse-like contours on a spherical surface (Figure~\ref{fig:varying_ecc_heatmap}). 
Thus, when projected onto the $\theta\phi$ space, it results in a myriad of contour shapes
(Figure~\ref{fig:vmf_kent_mml_mix}b)
depending on the parameters defining a \fb~distribution.

\noindent\emph{Compressibility of protein structures:}
The better explanatory power of \fb~mixtures over vMF mixtures leads to enhanced 
data compression (demonstrated through Table~\ref{tab:vmf_kent_mml_mix}), 
and hence, serve as efficient descriptors to model directional data.
In the context of proteins,
previous null model descriptors are based on the
uniform distribution on the sphere \citep{konagurthu-sst}, and due to
vMF mixtures \citep{multivariate_vmf}. The null model descriptions
provide a baseline for encoding protein coordinate data
in varied structure modelling tasks 
\citep{konagurthu-sst,konagurthu2013statistical,collier2014new}.
In this regard, the use of \fb~mixture offers a better alternative
as opposed to encoding using uniform distribution or vMF mixtures.

The message length expressions to encode the orientation angles using 
uniform, vMF, and \fb~null models are given by Equation~\ref{eqn:null_models}
\citep{konagurthu-sst,multivariate_vmf},
where $\boldx$ corresponds to a unit vector described by $(\theta,\phi)$
on the surface of the sphere, $\epsilon$ is the precision\footnote{Protein
coordinate data is measured to an accuracy of $\epsilon=0.001$\AA.}
to which each coordinate is measured, and $r$ denotes
the distance between successive $C_{\alpha}$ atoms.
In Equation~\ref{eqn:null_models}, for the vMF mixture, $K=37$ and
the null model corresponding to the \fb~mixture has $K=23$ components.
\begin{align}
\text{Uniform Null} &= -\log_2\left(\frac{\epsilon^2}{4\pi r^2}\right) = \log_2(4\pi) - 2 \log_2\left(\frac{\epsilon}{r}\right)\quad{\text{bits.}}\notag\\ 
\text{vMF~\&~\fb~Null} &= -\log_2 \left(\sum_{j=1}^K w_j f_j(\mathbf{x};\Theta_j)\right)- 2 \log_2\left(\frac{\epsilon}{r}\right)\quad{\text{bits.}} 
\label{eqn:null_models}
\end{align}

The inferred mixture models are then used to encode the entire protein data.
After accounting for the distances between the successive atoms, the total message
lengths obtained are given in Table~\ref{tab:null_models}.
The uniform distribution is clearly not an appropriate descriptor and this can be
reasoned from the empirical distribution which has multiple modes
(Figure~\ref{fig:vmf_kent_mml_mix}).
The inferred vMF mixture has better explanatory power over the uniform distribution
as it has a corresponding saving of 446,000 bits over 251,346 data points (residues).
This translates to an enhanced compression of 1.778 bits per residue (on average).
The inferred \fb~mixture, however, encodes the same amount of data with a saving of
7,000 bits against the vMF mixture (an average of 0.026 bits extra compression per residue).
The results following the application
of \fb~mixtures to modelling protein directional data demonstrate that
that they supersede the vMF mixture models (Table~\ref{tab:null_models}). The ability of \fb~distributions
to model asymmetrical data leads to improved encoding of the protein data.
Hence, they serve as natural successors to the vMF null model descriptors. 
\begin{table}[htb]
  \centering
  \caption{Comparison of the null model encoding lengths based on uniform distribution,
vMF mixture (37 components), and \fb~mixture (23 components).}
  \begin{tabular}{|c|c|c|}
  \hline
  \multirow{2}{*}{Null model}      & Total message length & Bits per \\ 
                                   & (millions of bits)   & residue \\ \hline
  Uniform                          & 6.895                & 27.434 \\
  vMF mixture                      & 6.449                & 25.656 \\
  \fb~mixture                      & \textbf{6.442}       & \textbf{25.630} \\
  \hline
  \end{tabular}
  \label{tab:null_models}
\end{table}

\subsection{Comparison of MML criterion with other information-theoretic criteria}
\label{subsec:criteria_comparison}
The MML criterion is used in computing the score associated with a mixture
model by separately encoding the parameters (first part) and the data given those
parameters (second part). This yields the total message length (Equation~\ref{eqn:mixture_msglen})
which is used to find improved mixtures during the search process.
In addition to the MML criterion, as discussed in Section~\ref{subsec:optimal_components},
the traditional information-theoretic criteria used are AIC \citep{aic} 
and BIC \citep{bic,rissanen1978modeling}.
These two criteria introduce constant term penalties depending on the 
number of free parameters in the mixture model.
If $p$ denotes the number of a model's free parameters\footnote{The number
of free parameters in a \fb~mixture with $K$ components is $p=5K + (K-1) = 6K-1$.}, 
$\mathcal{L}(\dataset|\boldsymbol{\Phi})$ is the minimized negative log-likelihood of data 
given the parameters $\boldphi$, and $N$ the sample size, then AIC and BIC are given by
\begin{equation*}
  \text{AIC}(p) = p + \mathcal{L}(\dataset|\boldsymbol{\Phi}) 
  \quad\text{and}\quad
  \text{BIC}(p) = \frac{p}{2}\log N + \mathcal{L}(\dataset|\boldsymbol{\Phi})
\end{equation*}

Mixture modelling of some observed data based on these criteria can be
done as follows:
\begin{itemize}
\item \emph{Exhaustive search:} 
The search heuristic (Section~\ref{subsec:search_method}) to determine the optimal mixture 
can be used alongside any objective function and not necessarily the MML criterion.
The series of perturbations are carried out as described and the improvement
to mixtures is determined based on the criterion in use.
\item \emph{Traditional search:} 
As discussed in Section~\ref{subsec:optimal_components}, the traditional
search method using AIC/BIC involves estimating the mixture parameters
using the EM algorithm (Section~\ref{subsec:em_ml}) for varying number
of components $K$ and choosing the one which results in minimum criterion value.
\end{itemize}
It is to be noted that with the MML criterion, the EM algorithm in Section~\ref{subsec:em_mml}
is used to obtain the MML estimates of the mixture parameters. However, with AIC and BIC,
the EM algorithm in Section~\ref{subsec:em_ml} results in the maximum likelihood (ML) estimates
for a given $K$. For \fb~mixtures, the ML estimates are often approximated 
by the moment estimates which are
used in the M-step of the EM algorithm \citep{peel2001fitting,kent2005using,hamelryck2006sampling}.
We compare the results for mixtures obtained using the ML estimates
and their approximations against those obtained using MML-based estimates.

The results pertaining to the \emph{exhaustive} search method are shown in
Table~\ref{tab:proposed_search_aic_bic}. 
It is observed that when search is based on AIC, the mixtures resulting
due to moment and ML estimation have 37 and 34 components respectively. 
The moment and the ML mixtures have the same AIC values in this case.
With BIC, the mixture resulting from ML estimation has the
lower BIC value. In this case, the moment and the ML mixtures
have 23 and 24 components respectively. This number resembles the one 
obtained by the exhaustive search but MML-based parameter estimation.
The ML mixture has the lowest BIC score.
\begin{table}[htb]
  \centering
  \caption{\fb~mixtures inferred by employing the \emph{exhaustive search} method
and changing the evaluation criteria and methods to estimate mixture parameters.}
  \begin{tabular}{|c|c|c|c|c|c|c|}
  \hline
  \multirow{4}{*}{Criterion}  &         \multicolumn{3}{c|}{Moment mixtures}                      &           \multicolumn{3}{c|}{Maximum likelihood mixtures}         \\ \cline{2-7} 
                              & \multirow{3}{*}{$K$}  &     Criterion        &    Message           &  \multirow{3}{*}{$K$}   &     Criterion        &    Message      \\ 
                              &                     &       value          &    length            &                       &       value          &    length           \\ 
                              &                     & ($\times10^5$ bits)  & ($\times10^6$ bits)  &                       & ($\times10^5$ bits)  & ($\times10^6$ bits) \\ \hline
                        AIC   &        37           &     \textbf{2.313}   &    5.474             &           34          &    \textbf{2.313}    &    5.474            \\
                        BIC   &        23           &     4.647            &    5.475             &           24          &    \textbf{4.645}    &    5.474            \\
  \hline
  \end{tabular}
  \label{tab:proposed_search_aic_bic}
\end{table}

\begin{figure}[htb]
  \begin{minipage}{0.48\textwidth}
      \begin{figure}[H]
      \centering
      \includegraphics[width=\textwidth]{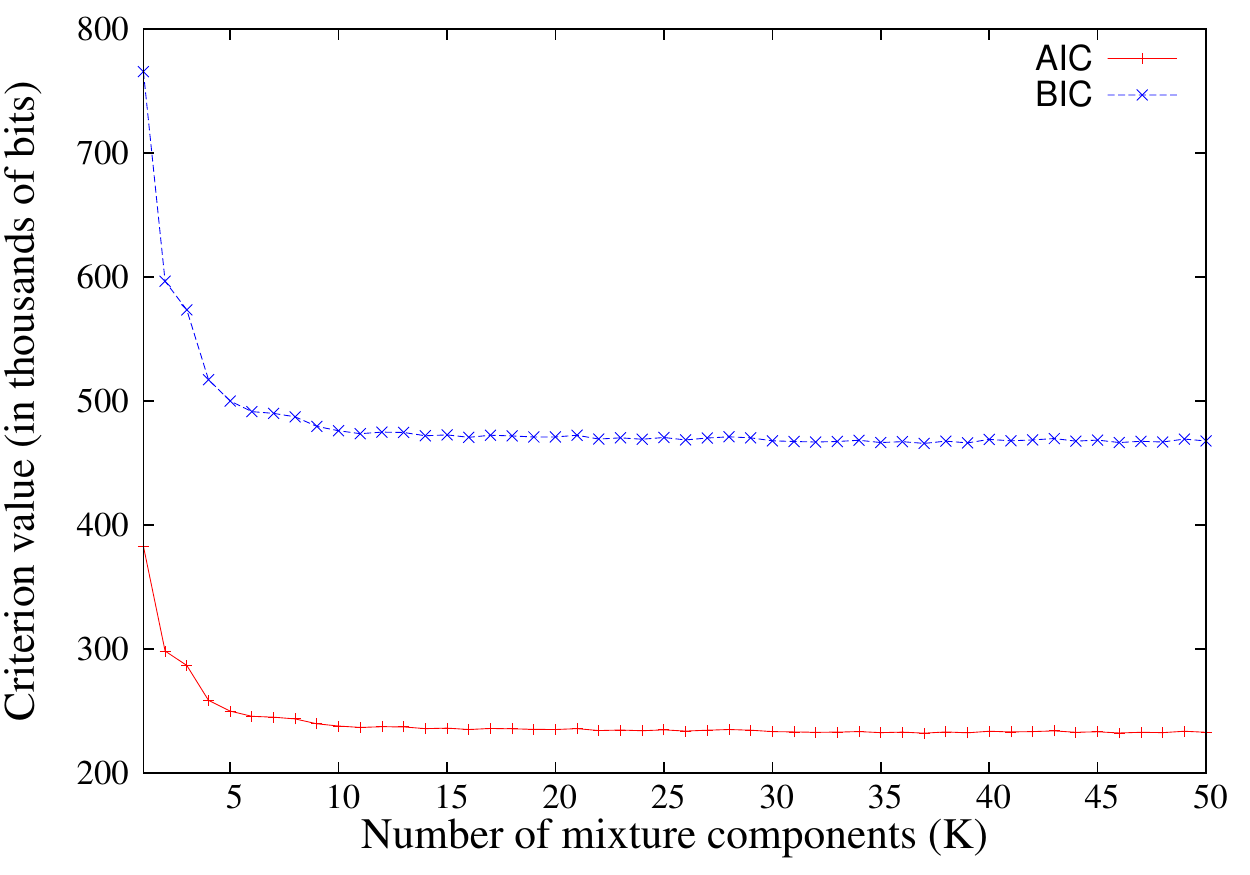}
      \caption{The criteria computed for maximum likelihood mixtures
               (moment mixtures have the same behaviour and are hence not shown) }
      \label{fig:mle_aic_bic}
      \end{figure}
  \end{minipage}
  \quad
  \begin{minipage}{0.48\textwidth}
      \begin{figure}[H]
      \centering
      \includegraphics[width=\textwidth]{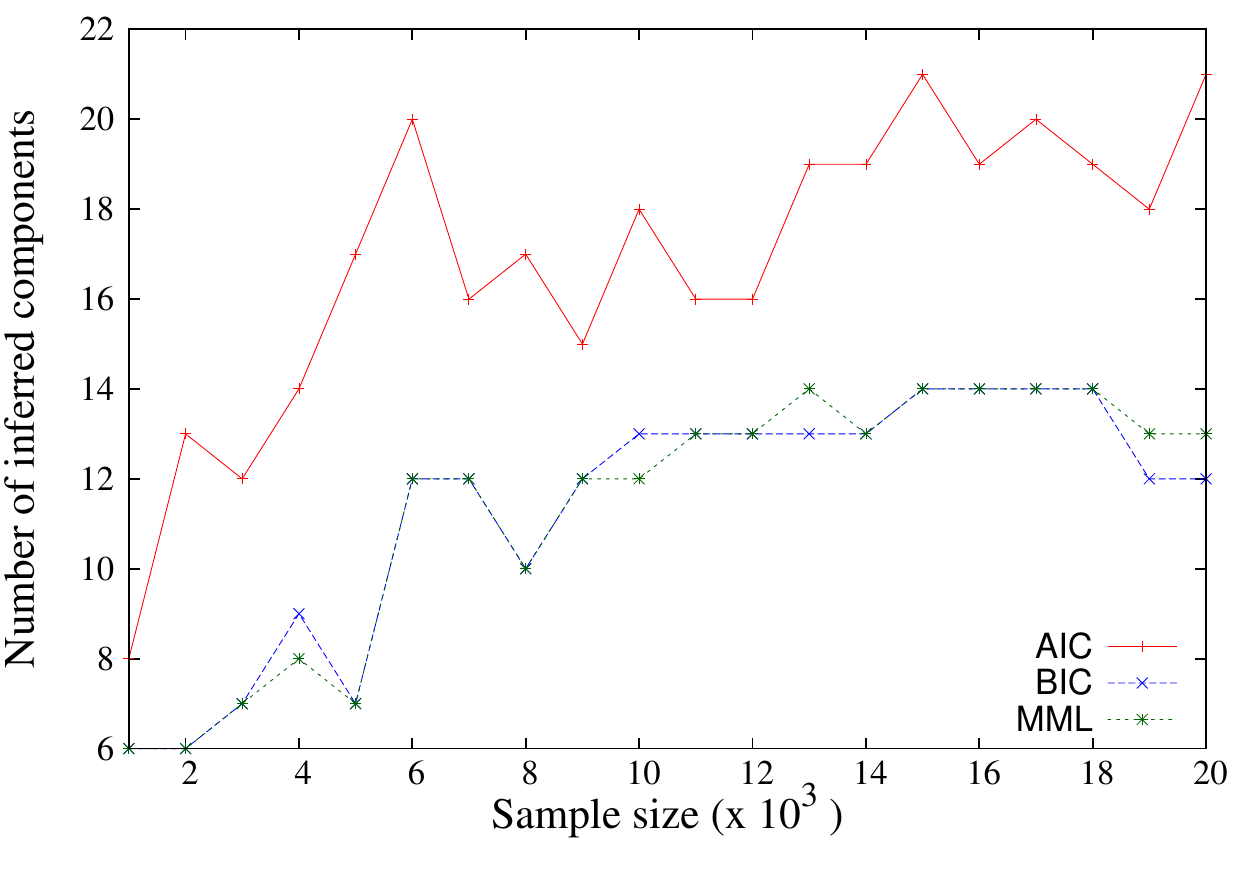}
      \caption{Variation of the number of inferred components using the 
search method based on exhaustive perturbations.} 
      \label{fig:inferred_components_varying_n}
      \end{figure}
  \end{minipage}
\end{figure}

The results pertaining to the traditional search method are shown in
Figure~\ref{fig:mle_aic_bic}. As the number of components $K$ is increased,
it is expected that the AIC and BIC scores decrease until some minimum
is reached and then increase thereafter. The value of $K$ at which this behaviour happens
is treated to be the optimal mixture that models the data.
It is observed that initially, both criteria decrease and after $K=30$,
the values do not change dramatically.
By increasing $K$, the linear increase in penalty factors and the associated
increase in log-likelihood are of the same magnitude, and hence, the 
difference in criteria is not apparent. 
Thus, using the traditional search, it is difficult to decide
on an appropriate number of mixture components.

The trend observed in Figure~\ref{fig:mle_aic_bic} is the same for mixtures 
obtained using both moment and ML estimates. The expressions for AIC and BIC do not help
in distinguishing the moment and ML mixtures because for different types of estimates
and a given $K$, the penalty terms are the same.
Also, the log-likelihood is approximately the same because for huge amounts
of data, as is the case here, all the estimates converge to the same value.

\begin{figure}[htb]
\centering
\subfloat[]{\includegraphics[width=0.5\textwidth]{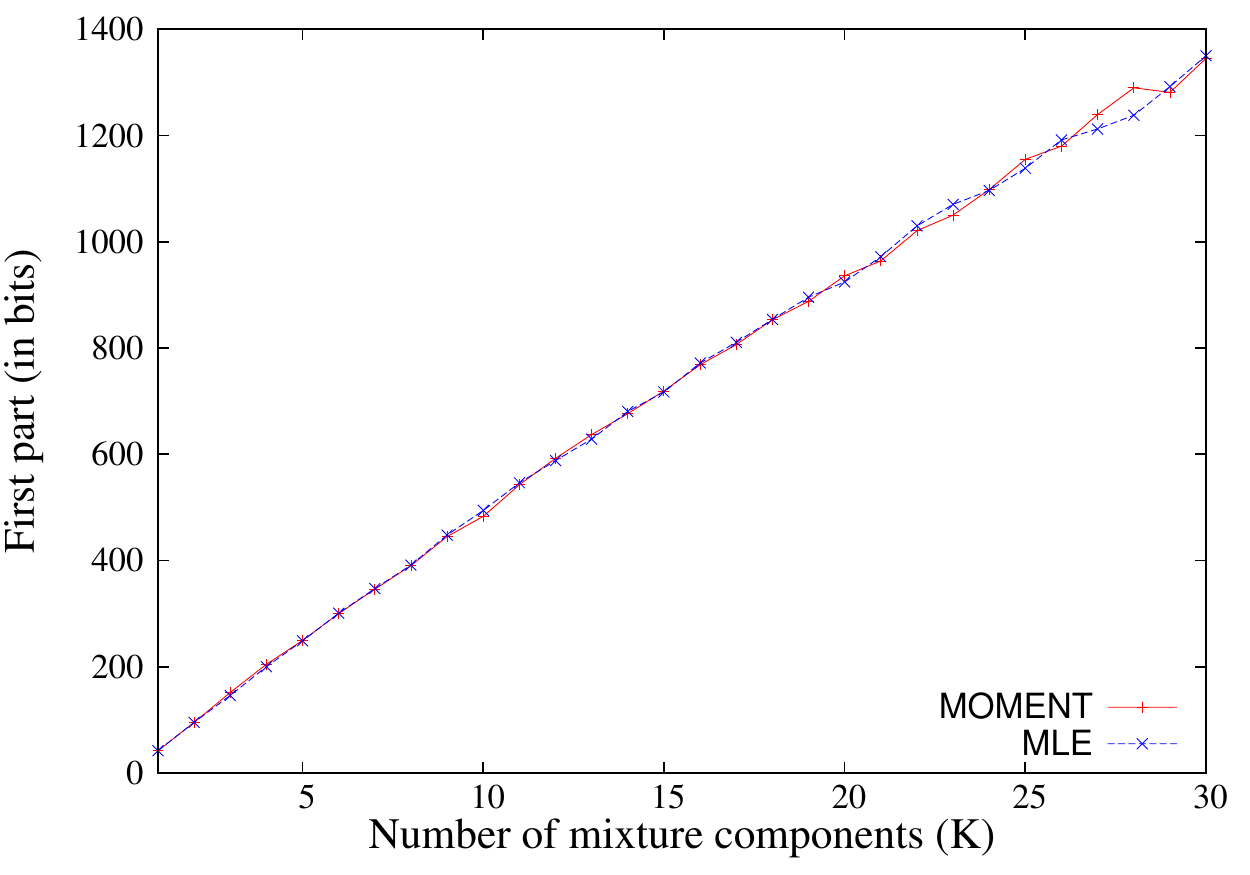}}
\subfloat[]{\includegraphics[width=0.5\textwidth]{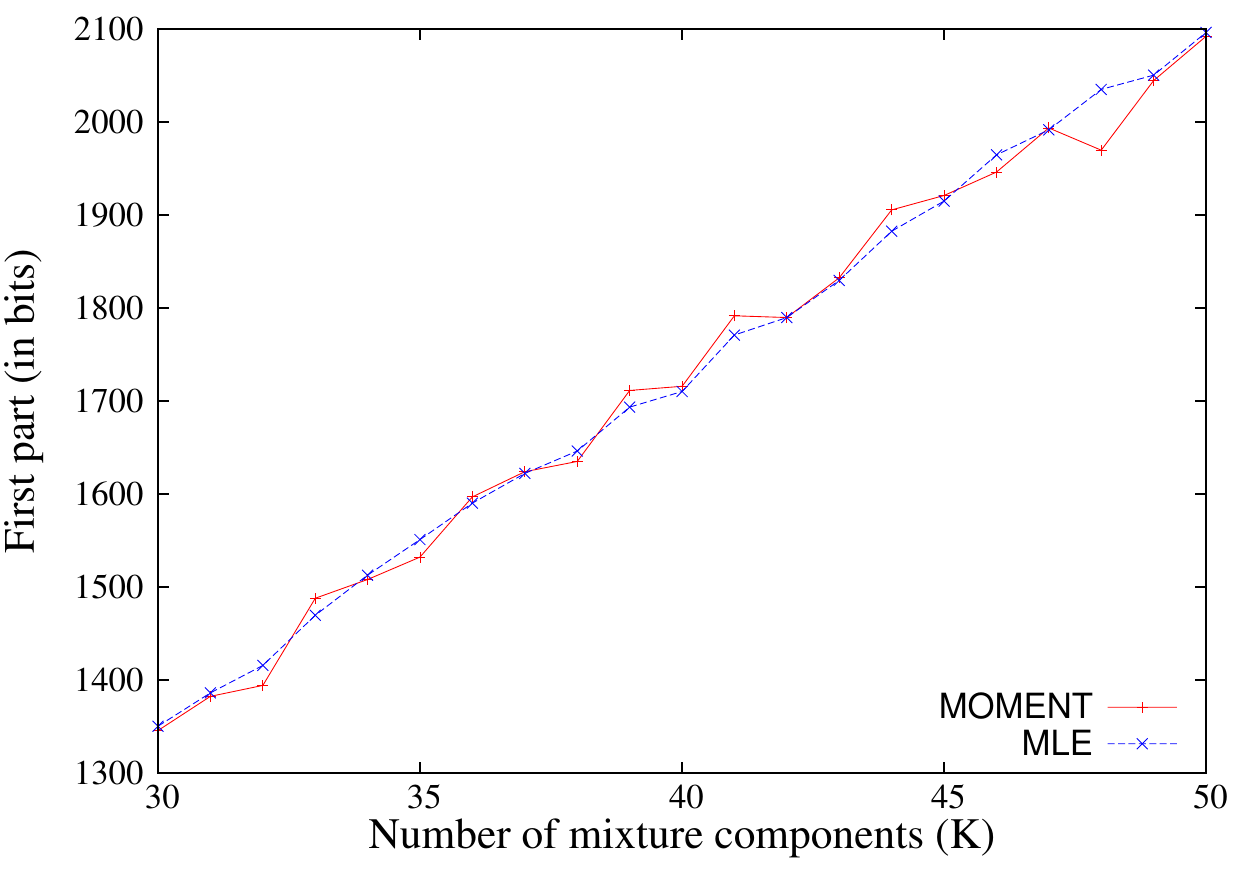}}
\caption{First part message length corresponding to mixtures evaluated using AIC.
The results for BIC display the same pattern and are hence not shown.
(the range of $K\in[1,50]$ is split into two sub-figures (a) and (b)
in order to highlight the differences in the message lengths).} 
\label{fig:aic_first_part}
\end{figure}

In contrast, if we compute the first part message lengths
corresponding to the moment and ML mixtures, for a given $K$, the 
differences in their encoding lengths become apparent.
The variation in the first part message lengths
for the moment and ML mixtures resulting from the traditional search
are shown in Figure~\ref{fig:aic_first_part}. It is observed
that until $K=30$, the first part message lengths of moment and ML mixtures
are close to each other. In Figure~\ref{fig:aic_first_part}(b), when $K>30$,
there are minute differences between encoding lengths of mixture
parameters obtained using moment and ML estimates.
Thus, unlike AIC/BIC, the MML criterion is able to distinguish mixtures
with equal number of components.
The first part corresponds to the model complexity and is dependent
on not just the number of components $K$ but also on the components'
parameters themselves according to the MML framework. 

The above discussion is aimed at projecting the limitations of the traditional
search method and also the use of AIC and BIC as evaluation criteria. We find 
in MML an objective way to assess mixtures and in conjunction with the search method
offers a better alternative to determine reliable mixture models.
We further illustrate the behaviour of the search method with smaller
amount of data. The previous discussion pertains to the entire empirical data set
containing $N=251,346~(\theta,\phi)$ pairs. In the current context,
from the empirical distribution,
we randomly sample varying amounts of data ranging from $N=1000$ to $N=20,000$.
The experiment is conducted by fixing the search method (exhaustive) but changing the
evaluation criteria to infer suitable \fb~mixtures.
It is observed that mixtures based on AIC have greater number
of components as compared to BIC and MML (see Figure~\ref{fig:inferred_components_varying_n}). 
The mixtures corresponding to BIC and MML have the same number of components
in most of the experimental trials.
These results are in agreement with what was observed on the complete protein data 
(Table~\ref{tab:proposed_search_aic_bic}), where AIC resulted in greater number of components.

\section{Conclusion}
We derived the parameter estimates of a \fb~distribution defined on a 
three-dimensional unit sphere based on the
Bayesian information-theoretic minimum message length criterion.
The derived estimators have lower bias and mean squared error compared 
to the traditionally used moment and maximum likelihood estimators.
The MML-based estimates are also invariant to transformations of the 
parameter space unlike the MAP estimates. Hence, the MML-based estimates are
improvements over the traditionally used estimators. Further, we have
designed the mixture modelling apparatus to be used in conjunction with
\fb~mixtures and demonstrated their applicability in modelling
real-world directional data resulting from protein spatial orientations.
The results obtained from modelling using \fb~mixtures 
is contrasted with commonly used vMF mixtures.
The \fb~mixture models supersede the vMF models in describing 
protein data, and serve as improved null model descriptors
that are important to modelling tasks in structural biology.


\appendix
\section{Prior density of $\boldtheta$ governed by the $\boldsymbol{\kappa}$ prior for the 2D vMF}
\label{app:alternative_parameterization}
In the MML estimation of parameters of a vMF distribution on a circle,
\citet{wallace1994estimation} use $h(\kappa) = \dfrac{\kappa}{(1+\kappa^2)^{3/2}}$.
We discuss this prior additionally as this leads to an invertible transformation
of all five parameters (described below) in the context of a \fb~distribution.
As in Section~\ref{subsec:prior_density}, $h_A(\psi,\alpha,\eta) = \dfrac{\sin\alpha}{4\pi^2}$
and $h(\beta|\kappa) = 2/\kappa$. Hence, the joint prior density $h_{\boldtheta}$ is formulated 
as shown below. Further, using the eccentricity transform described in 
Section~\ref{subsubsec:transformation_beta}, the joint prior density $h_{\boldtheta'}$
in $\boldtheta'$ parameterization is given below.
\begin{equation*}
h_{\boldtheta}(\psi,\alpha,\eta,\kappa,\beta) 
= \frac{\sin\alpha}{2\pi^2(1+\kappa^2)^{3/2}} 
\quad\text{and}\quad
h_{\boldtheta'}(\psi,\alpha,\eta,\kappa,e) 
= \frac{\kappa\sin\alpha}{4\pi^2(1+\kappa^2)^{3/2}} 
\end{equation*}

\subsection{An alternative parameterization of the parameter vector $\boldtheta$}
In addition to the eccentricity transform, 
we study a transformation proposed by \citet{rosenblatt1952remarks}, that transforms
a given continuous $k$-variate probability distribution into the uniform distribution
on the $k$-dimensional hypercube. Such a transformation applied on the 
prior density of the \fb~parameter vector $\boldtheta$ results in
the prior transforming to a uniform distribution. Hence, estimation in this
transformed parameter space is equivalent to the corresponding 
maximum likelihood estimation.
For the 5-parameter vector $\boldtheta=\{\psi,\alpha,\eta,\kappa,\beta\}$, 
the \citet{rosenblatt1952remarks} 
transformation to $\boldtheta''=\{z_1,z_2,z_3,z_4,z_5\}$ is given by
\begin{align*}
z_1 &= \Pr(X_1 \le \psi) = F_1(\psi)\\
z_2 &= \Pr(X_2 \le \alpha | X_1 = \psi) = F_2(\alpha|\psi)\\
z_3 &= \Pr(X_3 \le \eta | X_2 = \alpha, X_1 = \psi) = F_3(\eta|\alpha,\psi)\\
z_4 &= \Pr(X_4 \le \kappa | X_3 = \eta, X_2 = \alpha, X_1 = \psi) = F_4(\kappa|\eta,\alpha,\psi)\\
z_5 &= \Pr(X_5 \le \beta | X_4 = \kappa, X_3 = \eta, X_2 = \alpha, X_1=\psi) = F_5(\beta|\kappa,\eta,\alpha,\psi)
\end{align*}
This transformation results in $0\le z_i \le 1, i = 1,\ldots,5$. 
Further, \citet{rosenblatt1952remarks} argues that each $z_i$ is uniformly 
and independently distributed on $[0,1]$, so that the prior density 
in this transformed parameter space is $h_{\boldtheta''}(z_1,z_2,z_3,z_4,z_5) = 1$. In order
to achieve such a transformation, we need to express $z_i$ in terms of 
the original parameters. As per the definitions of the prior on $\boldtheta$
(Section~\ref{subsec:prior_density}), the following relationships are derived:
\begin{align}
z_1 &= \psi/\pi \implies \psi = \pi z_1 \notag\\
z_2 &= (1-\cos\alpha)/2 \implies \alpha = \arccos(1-2z_2) \notag\\
z_3 &= \eta/(2\pi) \implies \eta = 2\pi z_3
\label{eqn:rosenblatt_angle}
\end{align}
Based on the independence assumption in the formulation of priors of angular
and scale parameters (Section~\ref{subsec:prior_density}), 
$z_4 = F_4(\kappa|\eta,\alpha,\psi) = F_4(\kappa)$.
Similarly, $F_5(\beta|\kappa,\eta,\alpha,\psi) = F_5(\beta|\kappa)$.
Hence, the invertible transformations corresponding to
$\kappa$ and $\beta$ are as follows:
\begin{align}
z_4 &= \int_{0}^{\kappa} h(\kappa) d\kappa
= \int_{0}^{\kappa} \frac{\kappa}{(1+\kappa^2)^{3/2}} d\kappa = 1 - \cos(\arctan\kappa) 
\implies \kappa = \tan(\arccos(1-z_4)) \notag\\
z_5 &= F_5(\beta|\kappa) = 2\beta/\kappa \implies \beta = \kappa\,z_5/2 
\label{eqn:rosenblatt_scale}
\end{align}
With the 3D version of vMF $\kappa$ prior (see Section~\ref{subsec:prior_density}),
$z_4$ evaluates to $\dfrac{2}{\pi}\left(\arctan\kappa - \dfrac{\kappa}{1+\kappa^2} \right)$.
This version of $z_4$ is not invertible as it does not allow us to express $\kappa$ as a closed form
expression in $z_4$. Hence, the \citet{rosenblatt1952remarks} transformation is 
discussed only in the context when 2D vMF $\kappa$ prior is considered, as it
is possible to find an inverse transformation.

\subsection{The example demonstrating the effects of alternative parameterizations} 
The above discussed prior and its variants are used in the MAP-based
parameter estimation of the data from the example discussed in Section~\ref{subsec:map_example}.
The resulting estimates of $\psi,\alpha,\eta$ are given below using:
\begin{align*}
h_{\boldtheta}&: 
\widehat{\psi}=2.070,~\widehat{\alpha}=1.493,~\widehat{\eta}=1.522\\
h_{\boldtheta'}&: 
\widehat{\psi}=2.070,~\widehat{\alpha}=1.493,~\widehat{\eta}=1.522\\
h_{\boldtheta''}&:
\widehat{z}_1=0.659,~\widehat{z}_2=0.461,~\widehat{z}_3=0.242
\end{align*}
As observed, $\widehat{\psi},\widehat{\alpha}$, and $\widehat{\eta}$ are the same when posteriors
corresponding to $h_{\boldtheta}$ and $h_{\boldtheta'}$ are used.
In the case of $h_{\boldtheta''}$, the mapping of $\widehat{z}_1,\widehat{z}_2,\widehat{z}_3$
back to $\widehat{\psi},\widehat{\alpha},\widehat{\eta}$ (Equation~\ref{eqn:rosenblatt_angle}), 
results in the same estimates as that of $h_{\boldtheta}$ and $h_{\boldtheta'}$.
Hence, the MAP estimates of $\psi,\alpha,\eta$ are the same across the different versions.
The estimates of $\kappa$ and $\beta$ are, however, not the same under the various transformations.
They are as follows depending on the parameterization:
\begin{align*}
h_{\boldtheta}&: 
\widehat{\kappa}=16.975,~\widehat{\beta}=5.467 \\
h_{\boldtheta'}&: 
\widehat{\kappa}=20.547,~\widehat{e}=0.701 \implies \widehat{\beta} = \widehat{\kappa}\,\widehat{e}/2 = 7.205\\
h_{\boldtheta''}&:
\widehat{z}_4=0.964,~\widehat{z}_5=0.779 \implies \widehat{\kappa} = 28.065,~\widehat{\beta} = 10.925 \quad\text{(as per Equation~\ref{eqn:rosenblatt_scale})}
\end{align*}
The estimated value of $\kappa$ using $h_{\boldtheta}$ is 16.975 whereas it is
20.547 using $h_{\boldtheta'}$. The value of $\widehat{e}$ corresponds to a 
$\widehat{\beta} = 7.205$. 
Similarly, the value of $\widehat{\kappa}$ and $\widehat{\beta}$ corresponding to
$\widehat{z}_4=0.964$ and $\widehat{z}_5=0.0.779$ are 28.065 and 10.925 respectively.
Clearly, the value of the parameter estimates depend on the parameterization.
However, it is required that the estimates obtained in different parameterizations
should be the same irrespective of the space in which the parameters are defined. 
However, through this example, it is observed that for the various parameterizations, the value
of MAP estimates differ.

\begin{figure}[ht]
\centering
\subfloat[$h_{\boldtheta}$]{\includegraphics[width=0.33\textwidth]{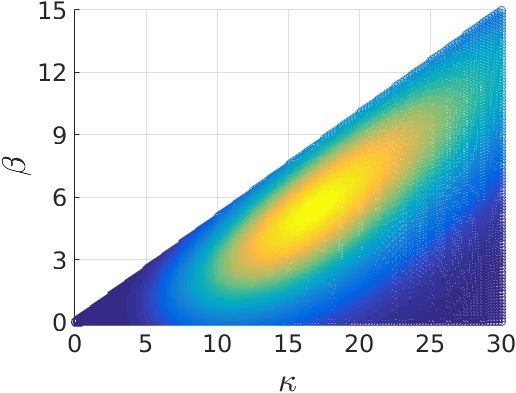}}
\subfloat[$h_{\boldtheta'}$]{\includegraphics[width=0.33\textwidth]{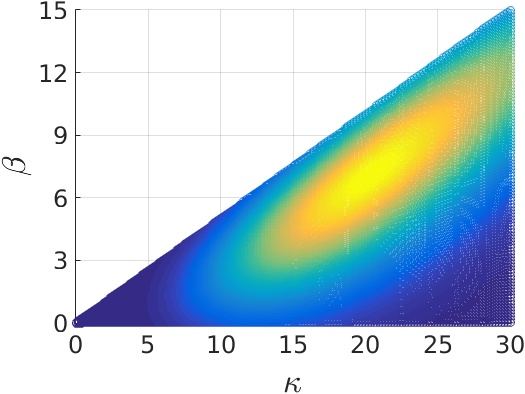}}
\subfloat[$h_{\boldtheta''}$]{\includegraphics[width=0.33\textwidth]{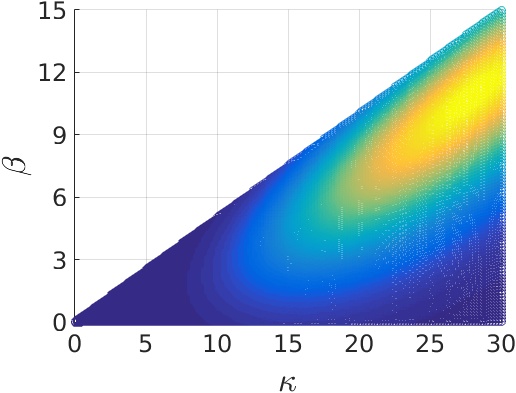}} \\
\subfloat[$h_{\boldtheta'}$]{\includegraphics[width=0.325\textwidth]{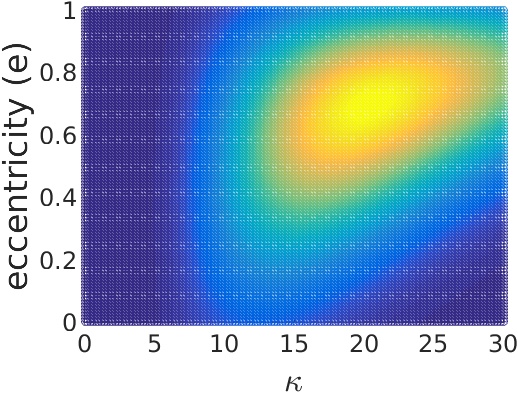}}\quad
\subfloat[$h_{\boldtheta''}$]{\includegraphics[width=0.32\textwidth]{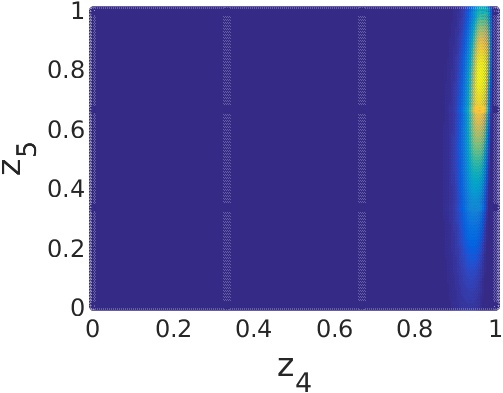}}
\caption{Heat maps depicting the modes (MAP estimates) of the posterior density 
resulting from different parameterizations.
}
\label{fig:prior2d_heatmaps}
\end{figure}

The variation of the posterior density under various transformations of the parameter space
are shown in Figure~\ref{fig:prior2d_heatmaps}. These are plotted as a function of $\kappa,\beta$ 
(in case of $h_{\boldtheta}$), $\kappa,e$ (in case of $h_{\boldtheta'}$),
and $z_4,z_5$ (in case of $h_{\boldtheta''}$), each
reflecting the space in which the posterior is defined. It is observed that the modes
of the respective posterior distributions occur at different positions and they
are not equivalent to each other. 
The posterior density plots in Figure~\ref{fig:prior2d_heatmaps}(b) and (c) correspond to those in
Figure~\ref{fig:prior2d_heatmaps}(d) and (e) respectively.
Ideally, (the modes in) Figure~\ref{fig:prior2d_heatmaps}(a)-(c) should be the same.
However, as demonstrated, that is not the case.
Thus, maximizing the posterior density
does not yield consistent estimates as observed through this example.

\section{The partial derivatives of $\mean,\major,\minor$ with respect to $\psi,\alpha,\eta$}
\label{app:axes_differentials}
The first and second order partial derivatives of the axes 
are required for the evaluation of the elements of the Fisher
information matrix (see Section~\ref{subsubsec:fisher_angle}).
The expressions for $\mean,\major,\minor$ as a function of $\psi,\alpha,\eta$
are given by Equation~\ref{eqn:net_rotation}.
\begin{itemize}
\item Derivatives of $\mean$:
\begin{gather*}
  \frac{\partial\mean}{\partial\alpha} =
  \begin{pmatrix}
  \cos\alpha  \\
  \sin\alpha \cos\eta \\
  \sin\alpha \sin\eta
  \end{pmatrix}\,,\quad
  \frac{\partial\mean}{\partial\eta} =
  \begin{pmatrix}
  0  \\
  -\sin\alpha \sin\eta \\
  \sin\alpha \cos\eta
  \end{pmatrix}\\
  \frac{\partial^2\mean}{\partial\alpha^2} = -\,\mean\,,\quad
  \frac{\partial^2\mean}{\partial\eta^2} = 
  \begin{pmatrix}
  0  \\
  -\sin\alpha \cos\eta \\
  -\sin\alpha \sin\eta
  \end{pmatrix}\,,\quad
  \frac{\partial^2\mean}{\partial\eta\partial\alpha} = 
  \begin{pmatrix}
  0  \\
  -\cos\alpha \sin\eta \\
  \cos\alpha \cos\eta
  \end{pmatrix}
\end{gather*}
The partial derivatives of $\mean$ involving the parameter $\psi$ are zero vectors.

\item Derivatives of $\major$:
\begin{gather*}
  \frac{\partial\major}{\partial\alpha} = -\cos\psi\,\mean\,,\quad
  \frac{\partial\major}{\partial\eta} = 
  \begin{pmatrix}
  0  \\
  -\cos\psi \cos\alpha \sin\eta - \sin\psi \cos\eta\\
  \cos\psi \cos\alpha \cos\eta - \sin\psi \sin\eta
  \end{pmatrix}\,,\quad
  \frac{\partial\major}{\partial\psi} = \minor\\
  \frac{\partial^2\major}{\partial\alpha^2} = -\cos\psi \frac{\partial\mean}{\partial\alpha}\,,\quad
  \frac{\partial^2\major}{\partial\eta^2} = 
  \begin{pmatrix}
  0  \\
  -\cos\psi \cos\alpha \cos\eta + \sin\psi \sin\eta\\
  -\cos\psi \cos\alpha \sin\eta - \sin\psi \cos\eta
  \end{pmatrix}\,,\quad
  \frac{\partial^2\major}{\partial\psi^2} = -\,\major\\
  \frac{\partial^2\major}{\partial\eta\partial\alpha} = -\cos\psi \frac{\partial\mean}{\partial\eta}\,,\quad
  \frac{\partial^2\major}{\partial\psi\partial\alpha} = \sin\psi\,\mean\,,\quad
  \frac{\partial^2\major}{\partial\psi\partial\eta} = \frac{\partial\minor}{\partial\eta}
\end{gather*}

\item Derivatives of $\minor$:
\begin{gather*}
  \frac{\partial\minor}{\partial\alpha} = \sin\psi\,\mean\,,\quad
  \frac{\partial\minor}{\partial\eta} = 
  \begin{pmatrix}
  0  \\
  \sin\psi \cos\alpha \sin\eta - \cos\psi \cos\eta\\
  -\sin\psi \cos\alpha \cos\eta - \cos\psi \sin\eta
  \end{pmatrix}\,,\quad
  \frac{\partial\minor}{\partial\psi} = -\,\major\\
  \frac{\partial^2\minor}{\partial\alpha^2} = -\sin\psi \frac{\partial\mean}{\partial\alpha}\,,\quad
  \frac{\partial^2\minor}{\partial\eta^2} = 
  \begin{pmatrix}
  0  \\
  \sin\psi \cos\alpha \cos\eta + \cos\psi \sin\eta\\
  \sin\psi \cos\alpha \sin\eta - \cos\psi \cos\eta
  \end{pmatrix}\,,\quad
  \frac{\partial^2\minor}{\partial\psi^2} = -\,\minor\\
  \frac{\partial^2\minor}{\partial\eta\partial\alpha} = \sin\psi \frac{\partial\mean}{\partial\eta}\,,\quad
  \frac{\partial^2\minor}{\partial\psi\partial\alpha} = \cos\psi\,\mean\,,\quad
  \frac{\partial^2\minor}{\partial\psi\partial\eta} = -\,\frac{\partial\major}{\partial\eta}
\end{gather*}
\end{itemize}

\section{The search process continued}
\label{app:search_continued}
\begin{figure}[ht]
\centering
\subfloat[Initialize means of children]{\includegraphics[width=0.33\textwidth]{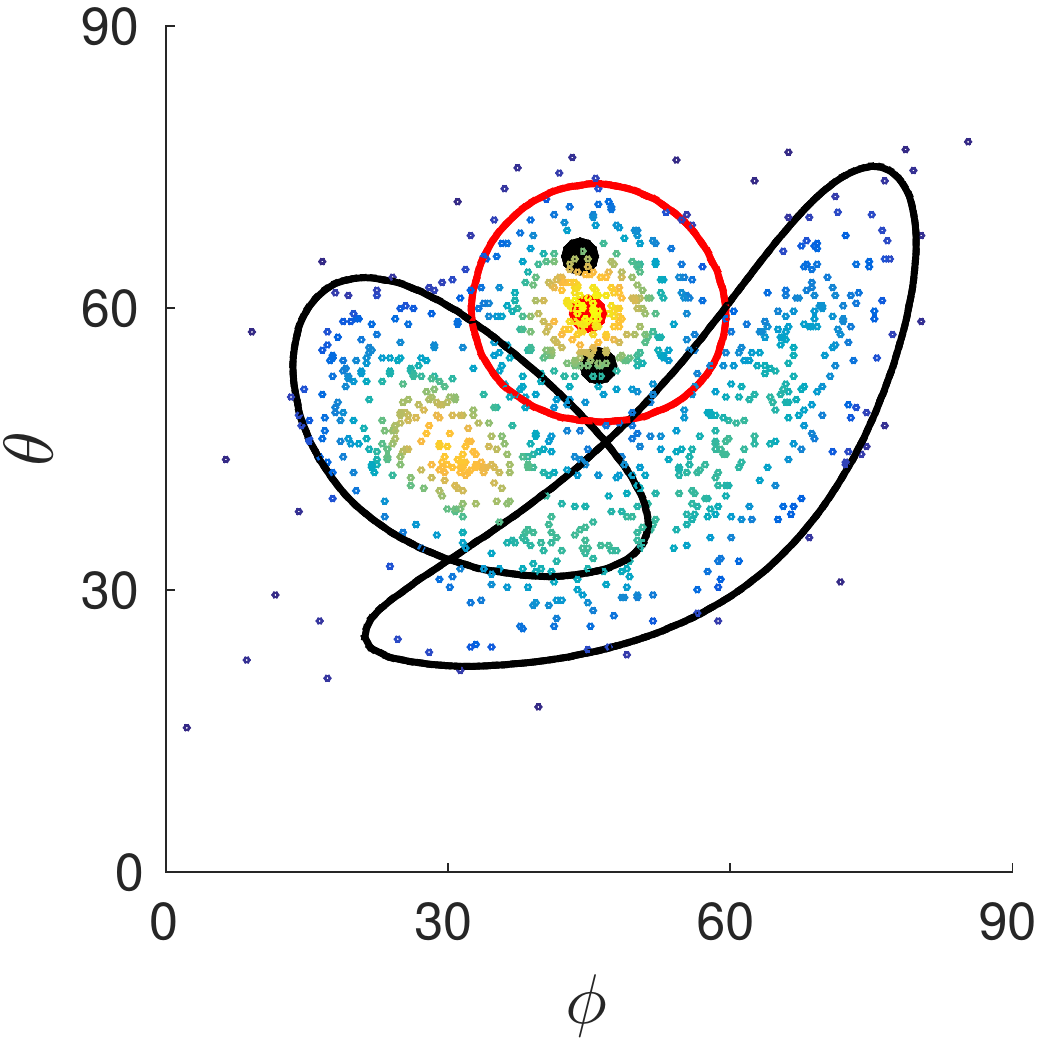}}
\subfloat[Optimized children ($I=19172$)]{\includegraphics[width=0.33\textwidth]{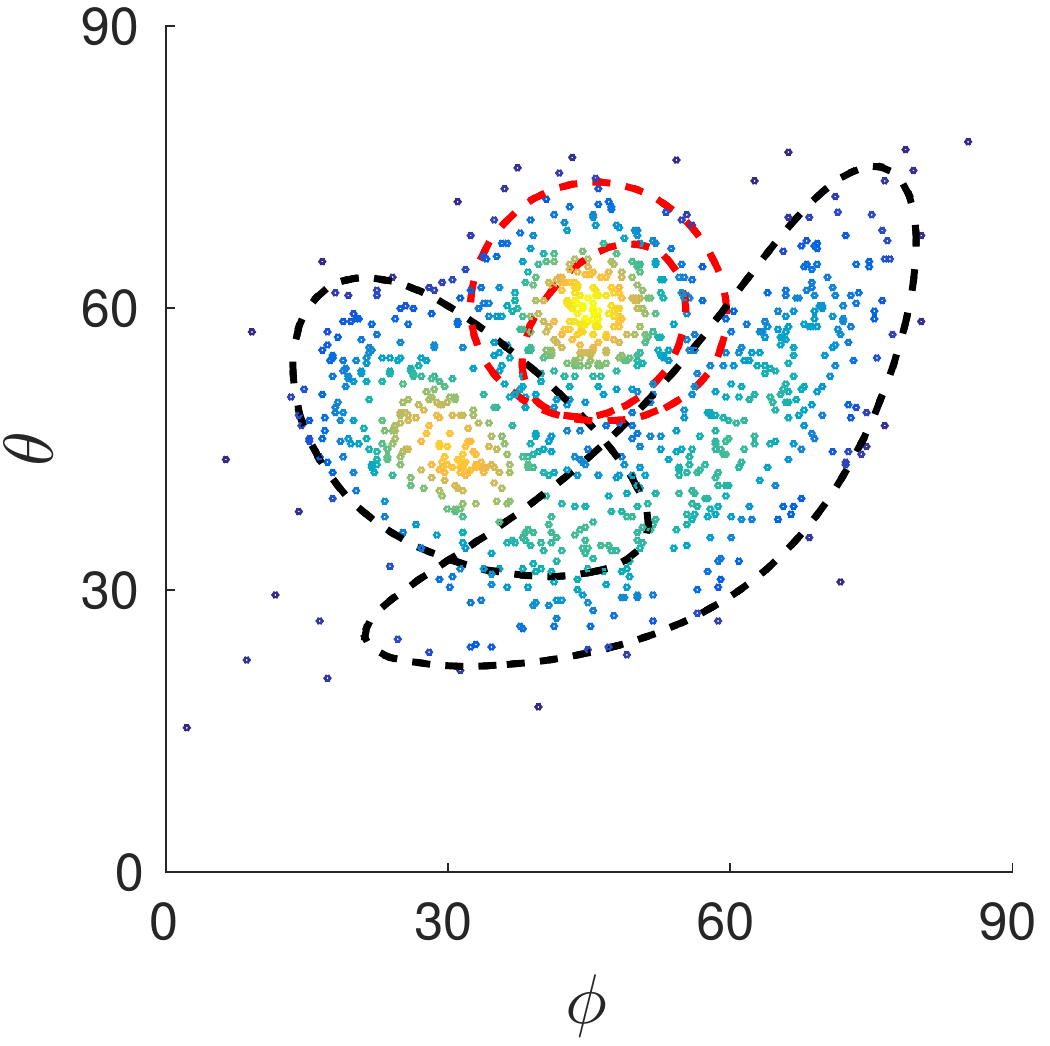}}
\subfloat[post-EM ($I=19170$)]{\includegraphics[width=0.33\textwidth]{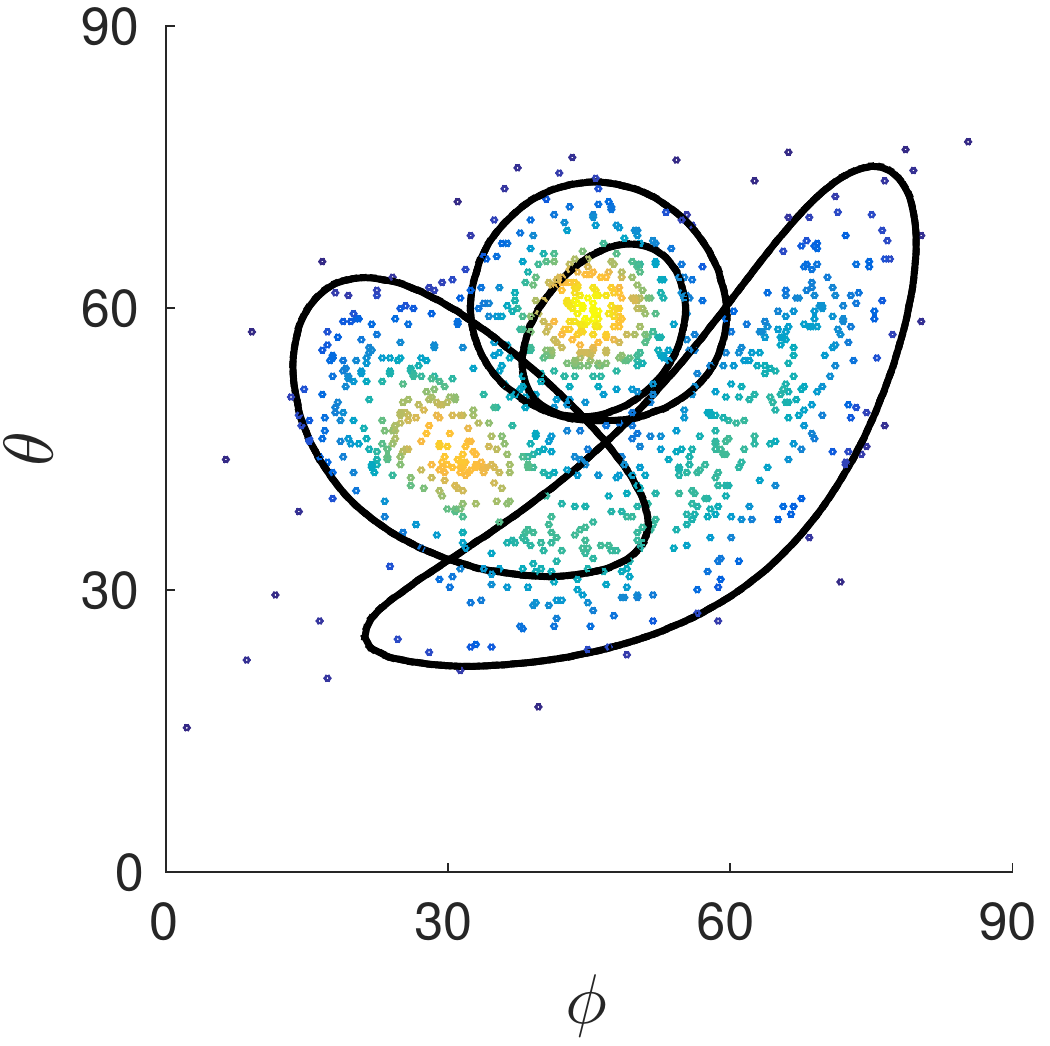}}\\
\subfloat[]{\includegraphics[width=0.33\textwidth]{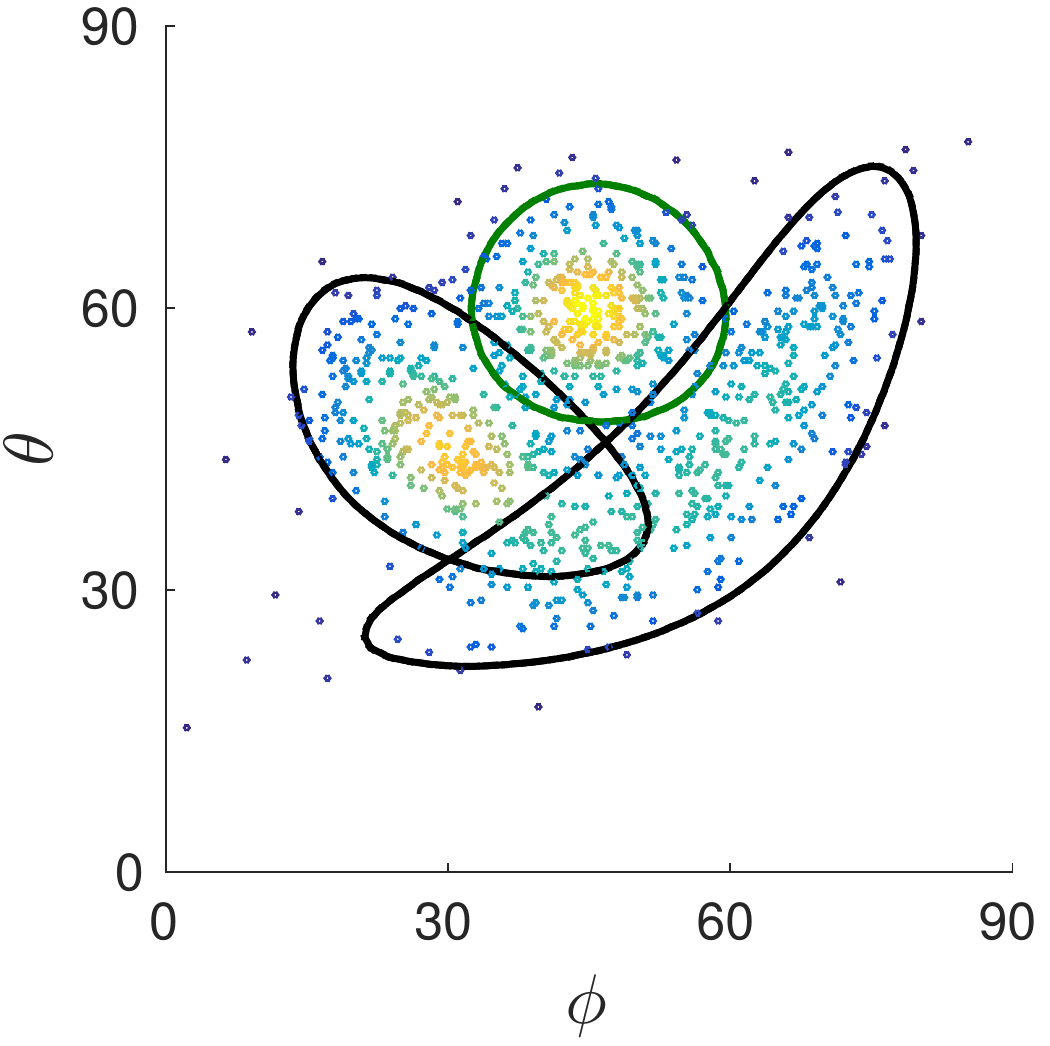}}
\subfloat[Before optimizing ($I=21730$)]{\includegraphics[width=0.33\textwidth]{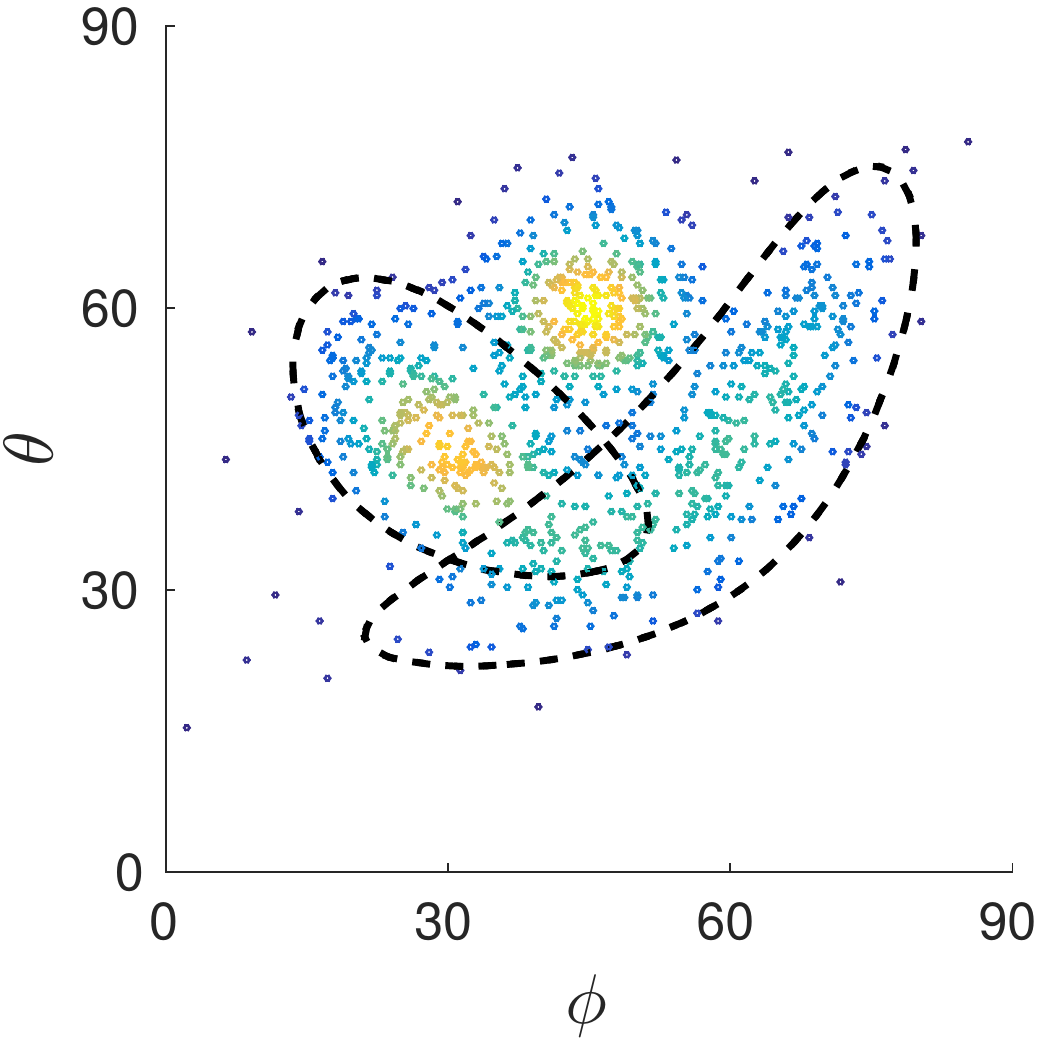}}
\subfloat[post-EM ($I=19237$)]{\includegraphics[width=0.33\textwidth]{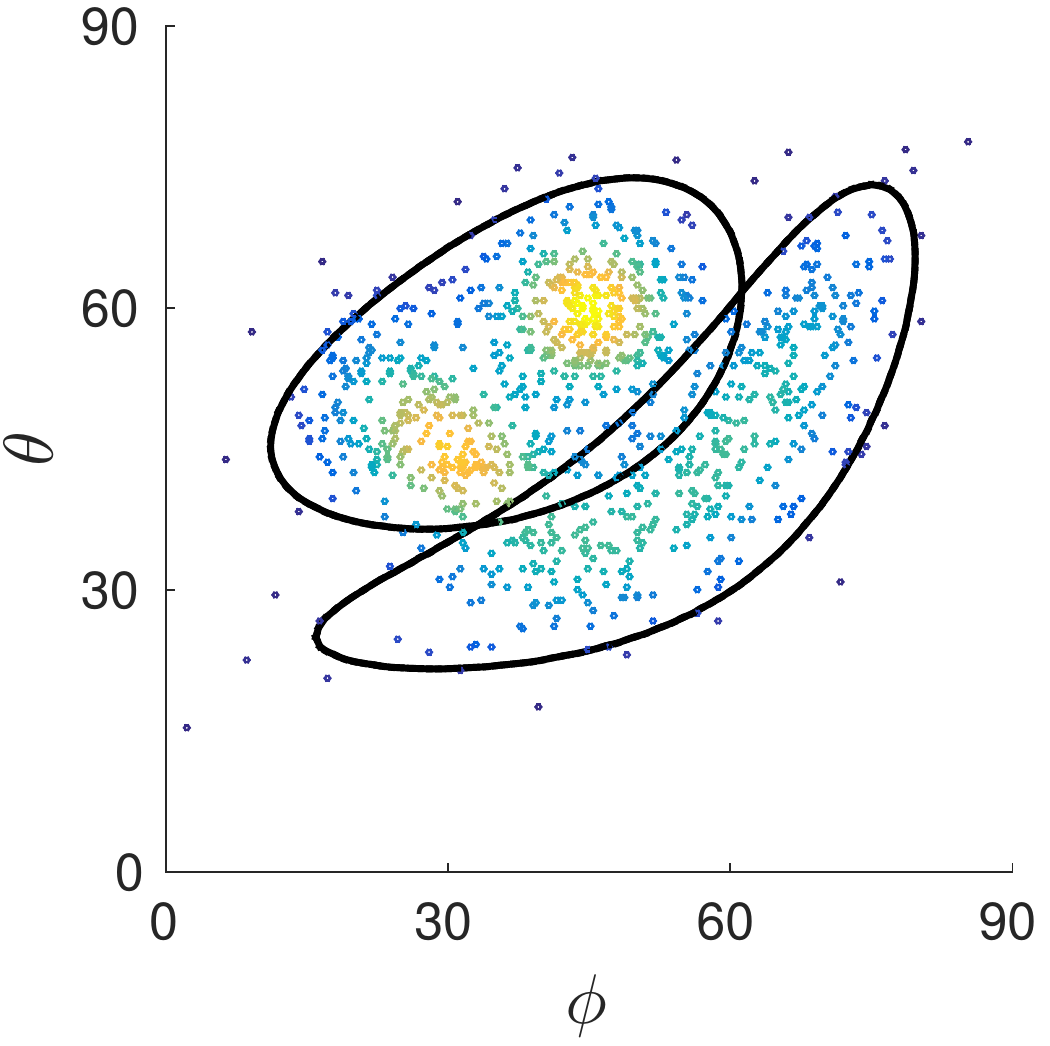}}\\
\subfloat[]{\includegraphics[width=0.33\textwidth]{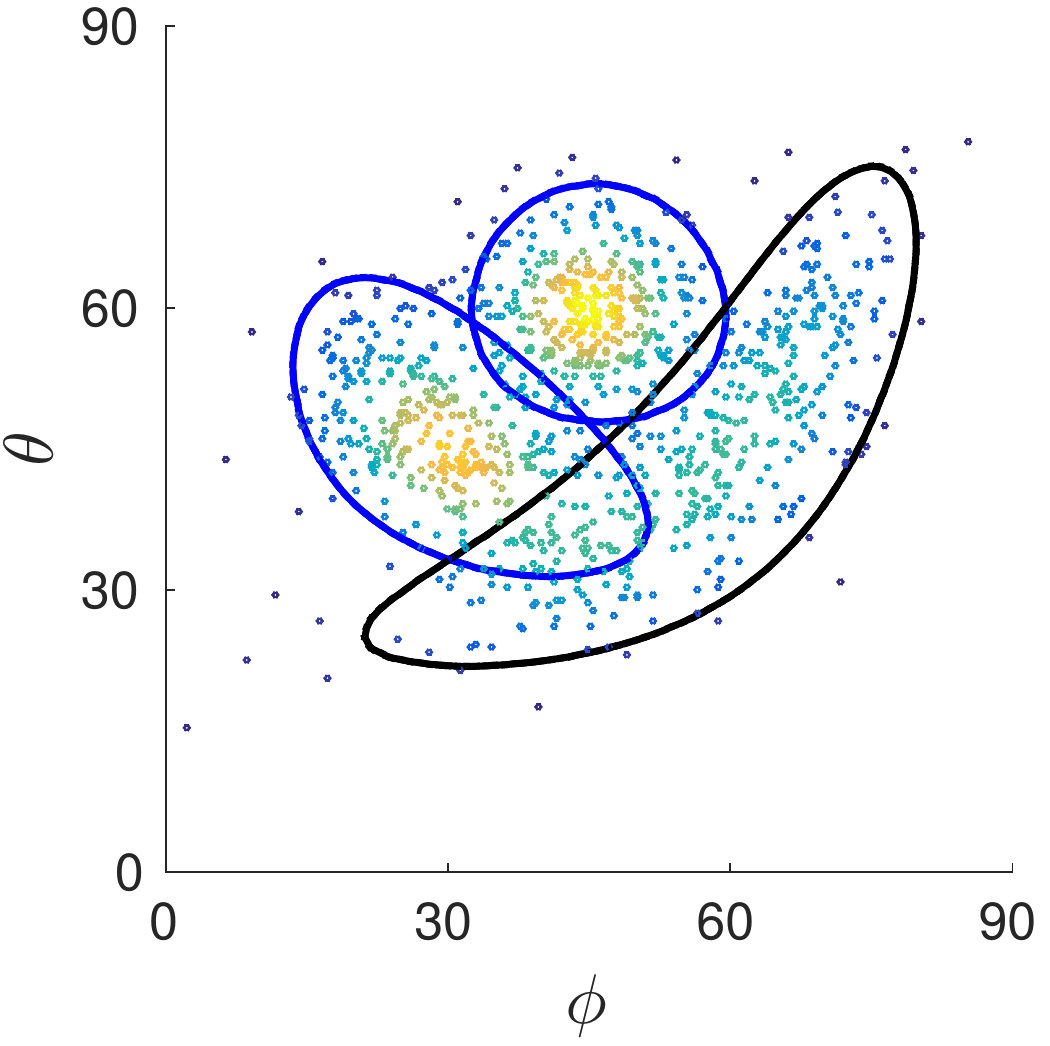}}
\subfloat[Before optimizing ($I=19244$)]{\includegraphics[width=0.33\textwidth]{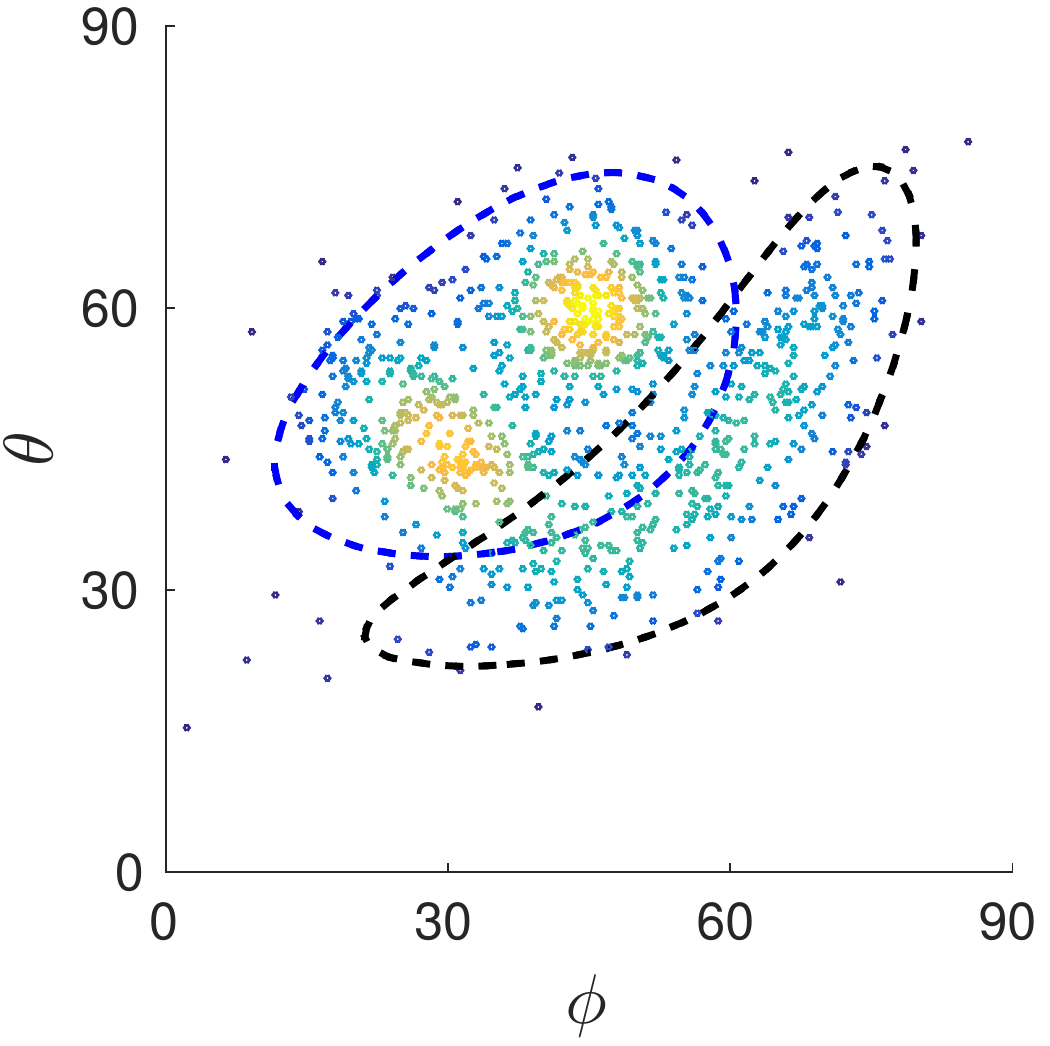}}
\subfloat[post-EM ($I=19236$)]{\includegraphics[width=0.33\textwidth]{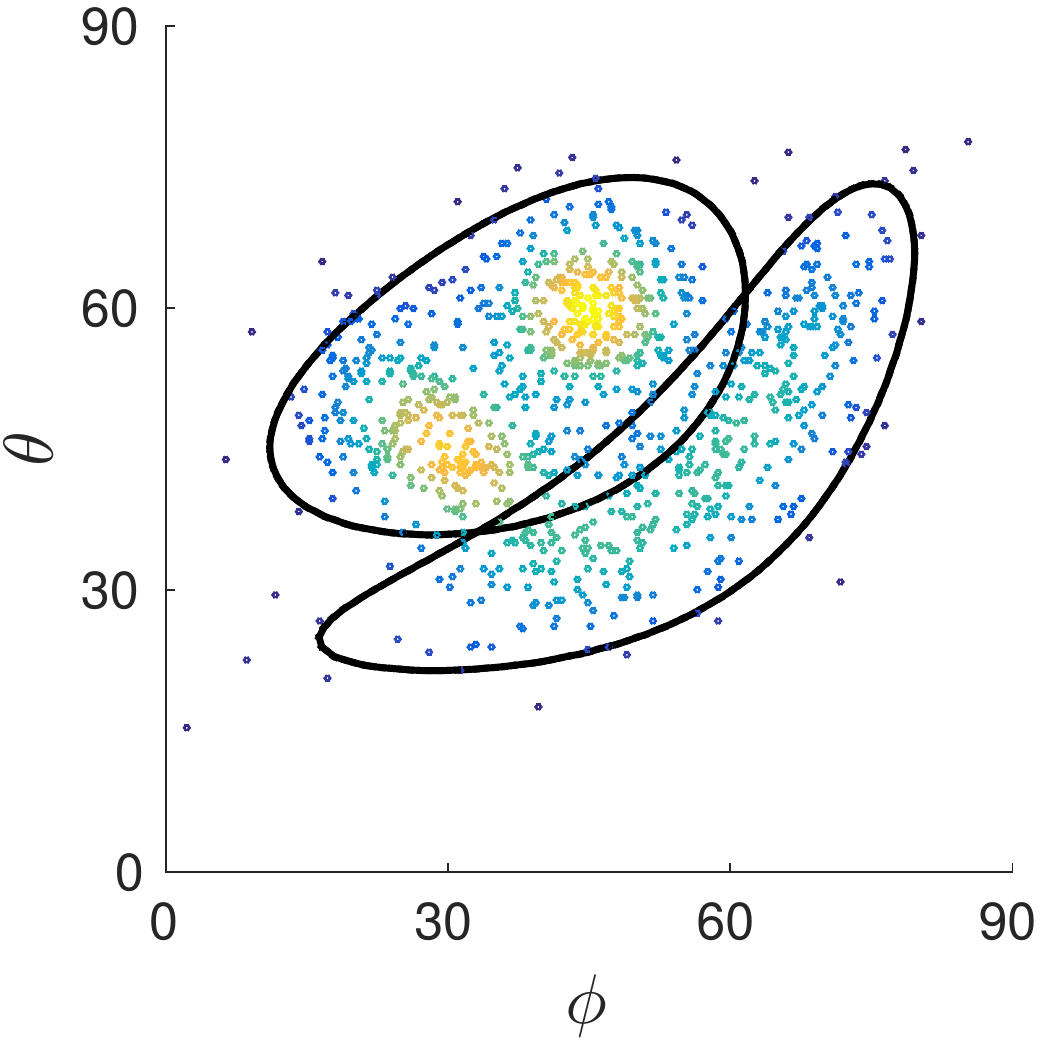}}
\caption{Iteration 3 -- perturbations of $P_2$ 
(a)-(c) splitting,
(d)-(f) deletion, 
(g)-(i) merging
} 
\label{fig:mix_iter3_c2}
\end{figure}

\begin{figure}[ht]
\centering
\subfloat[Initialize means of children]{\includegraphics[width=0.33\textwidth]{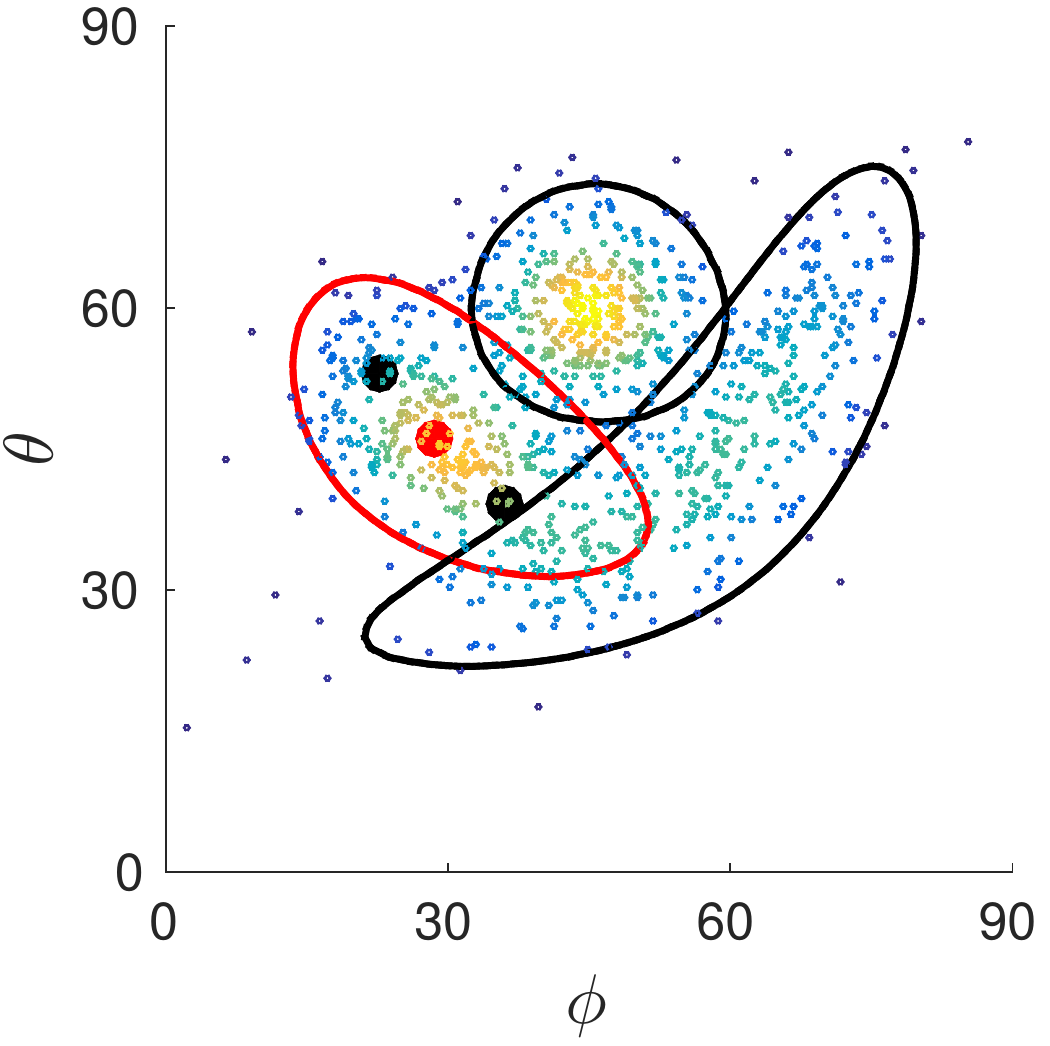}}
\subfloat[Optimized children ($I=19174$)]{\includegraphics[width=0.33\textwidth]{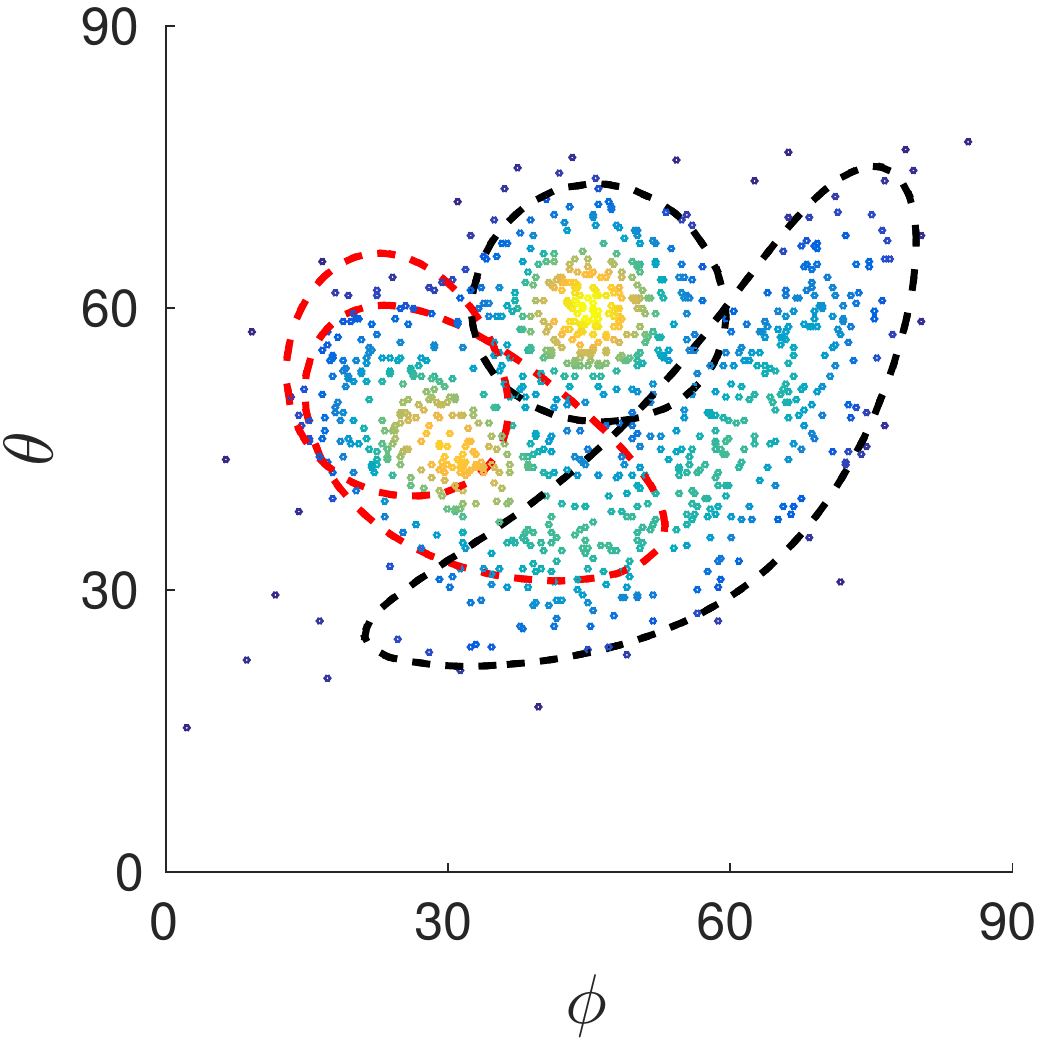}}
\subfloat[post-EM ($I=19168$)]{\includegraphics[width=0.33\textwidth]{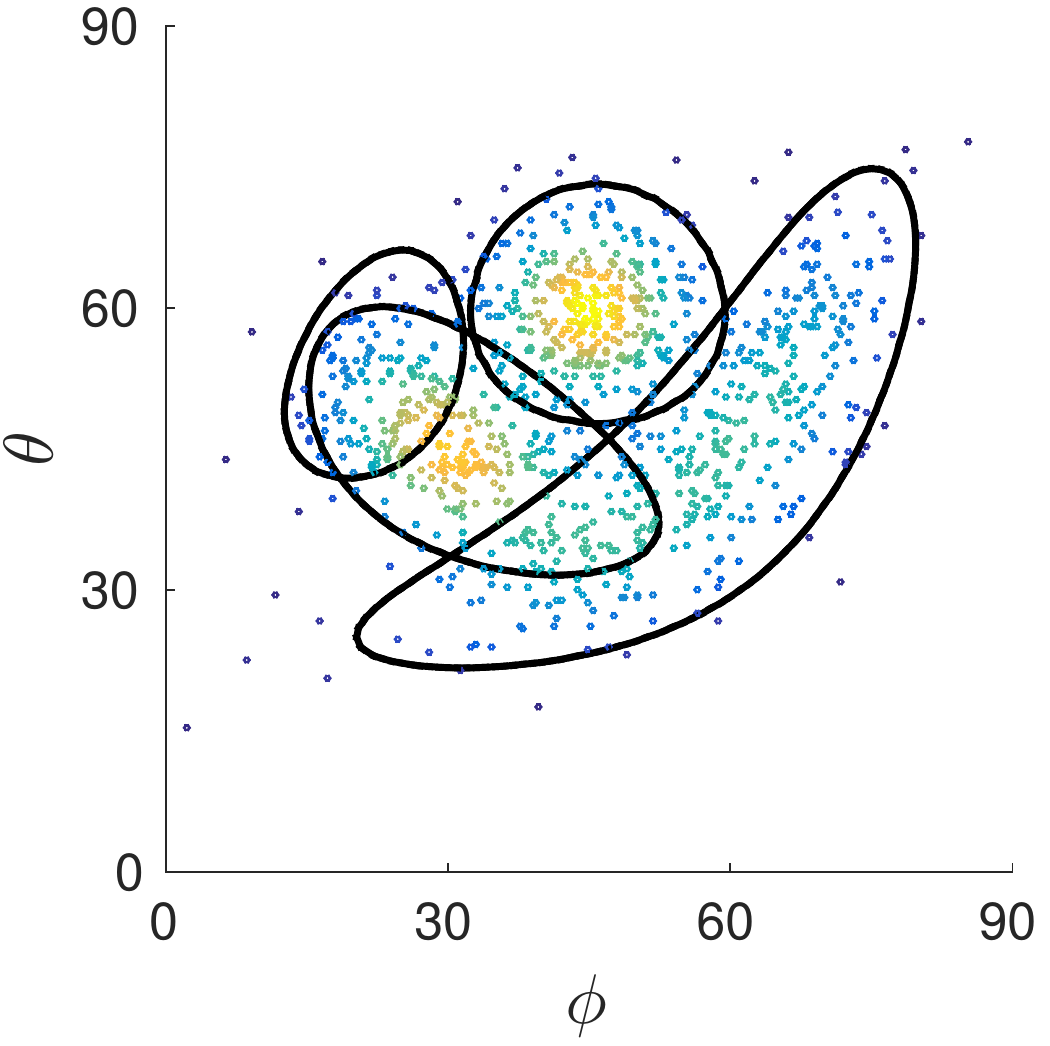}}\\
\subfloat[]{\includegraphics[width=0.33\textwidth]{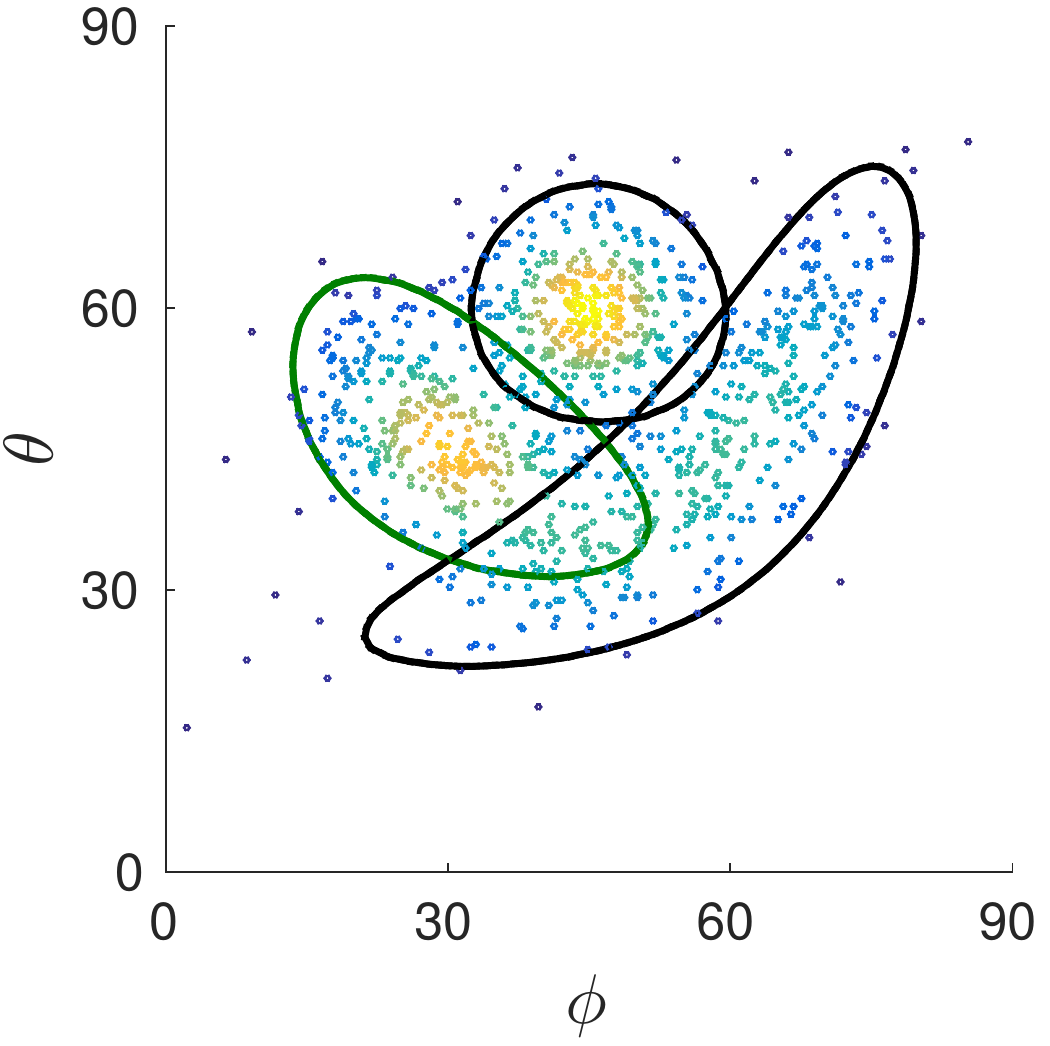}}
\subfloat[Before optimizing ($I=20743$)]{\includegraphics[width=0.33\textwidth]{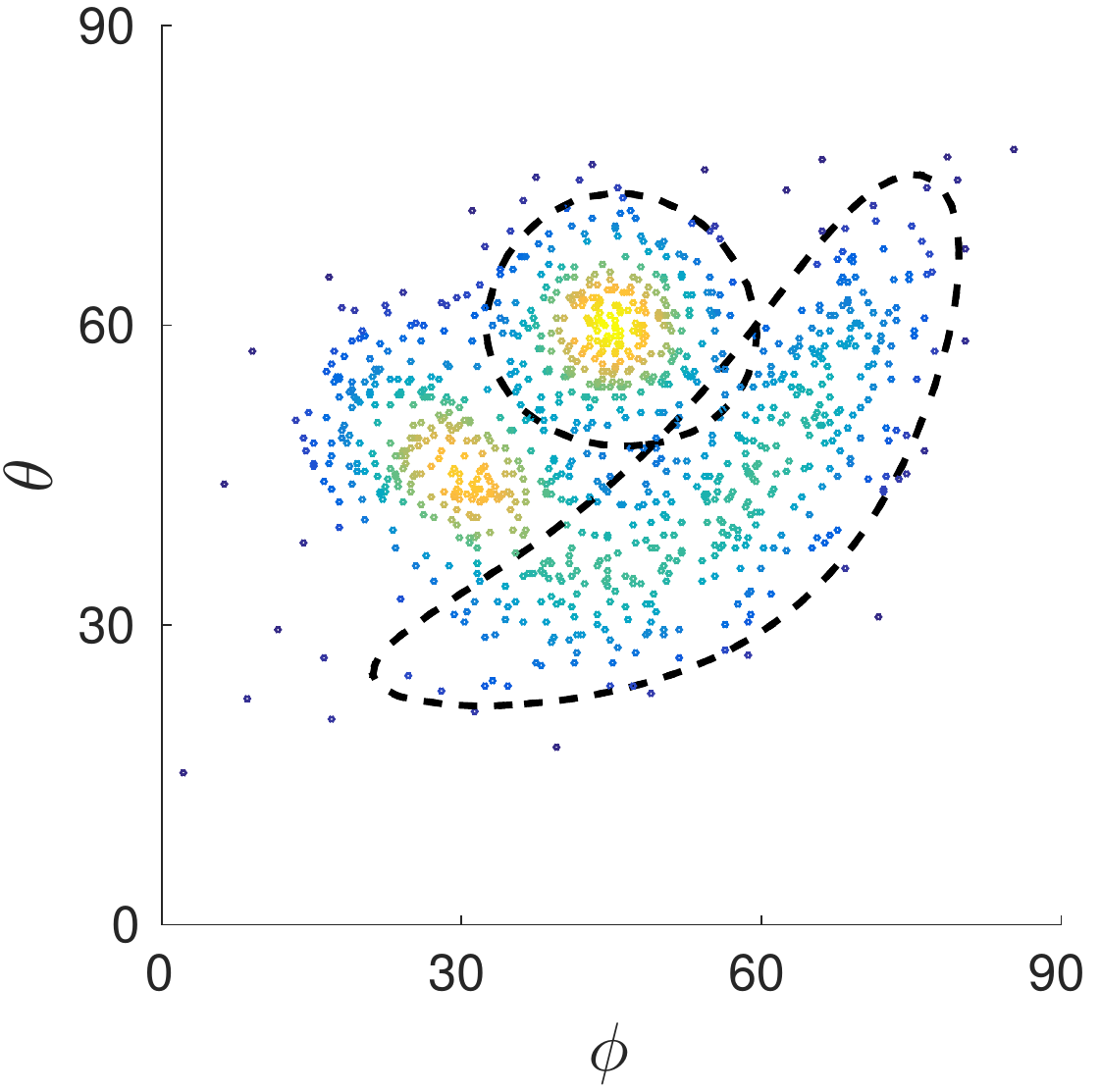}}
\subfloat[post-EM ($I=19237$)]{\includegraphics[width=0.33\textwidth]{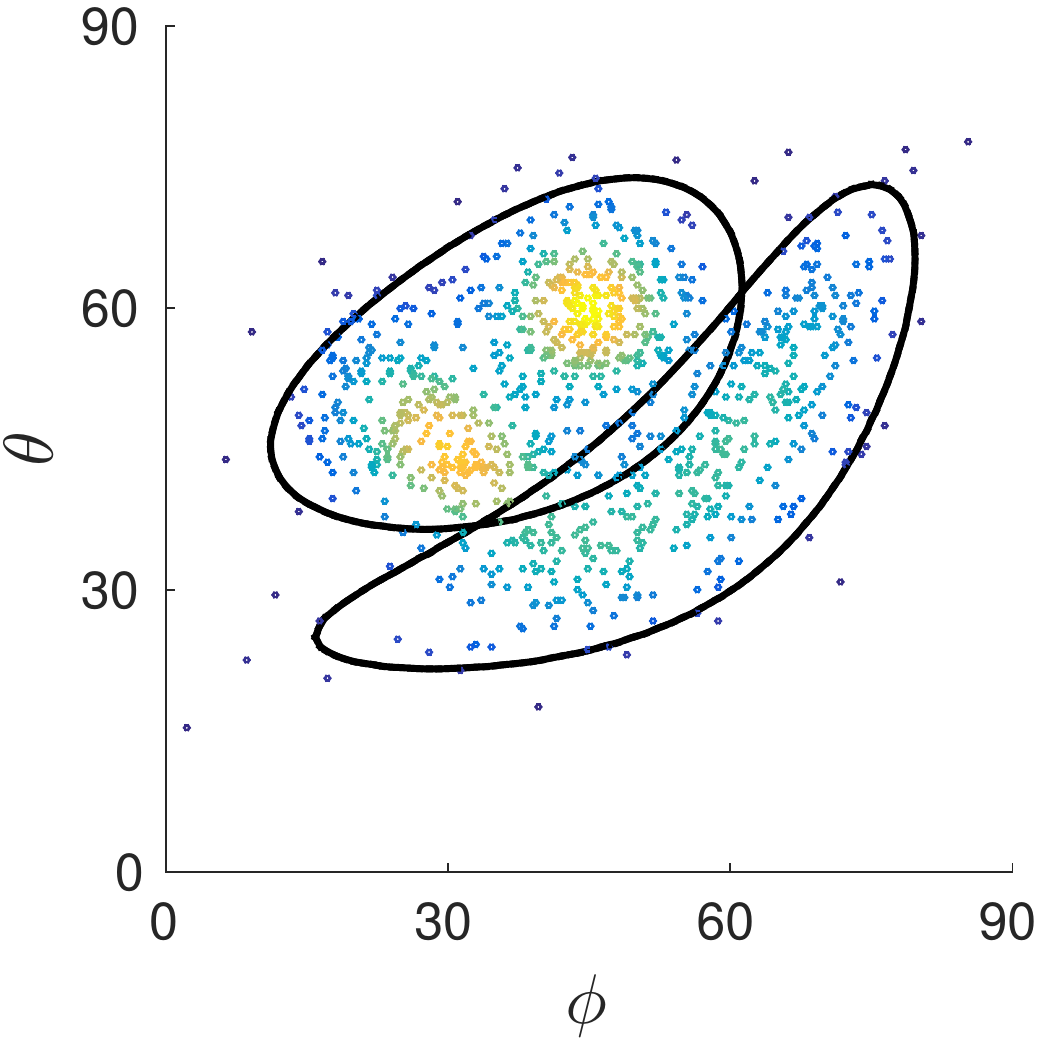}}\\
\subfloat[]{\includegraphics[width=0.33\textwidth]{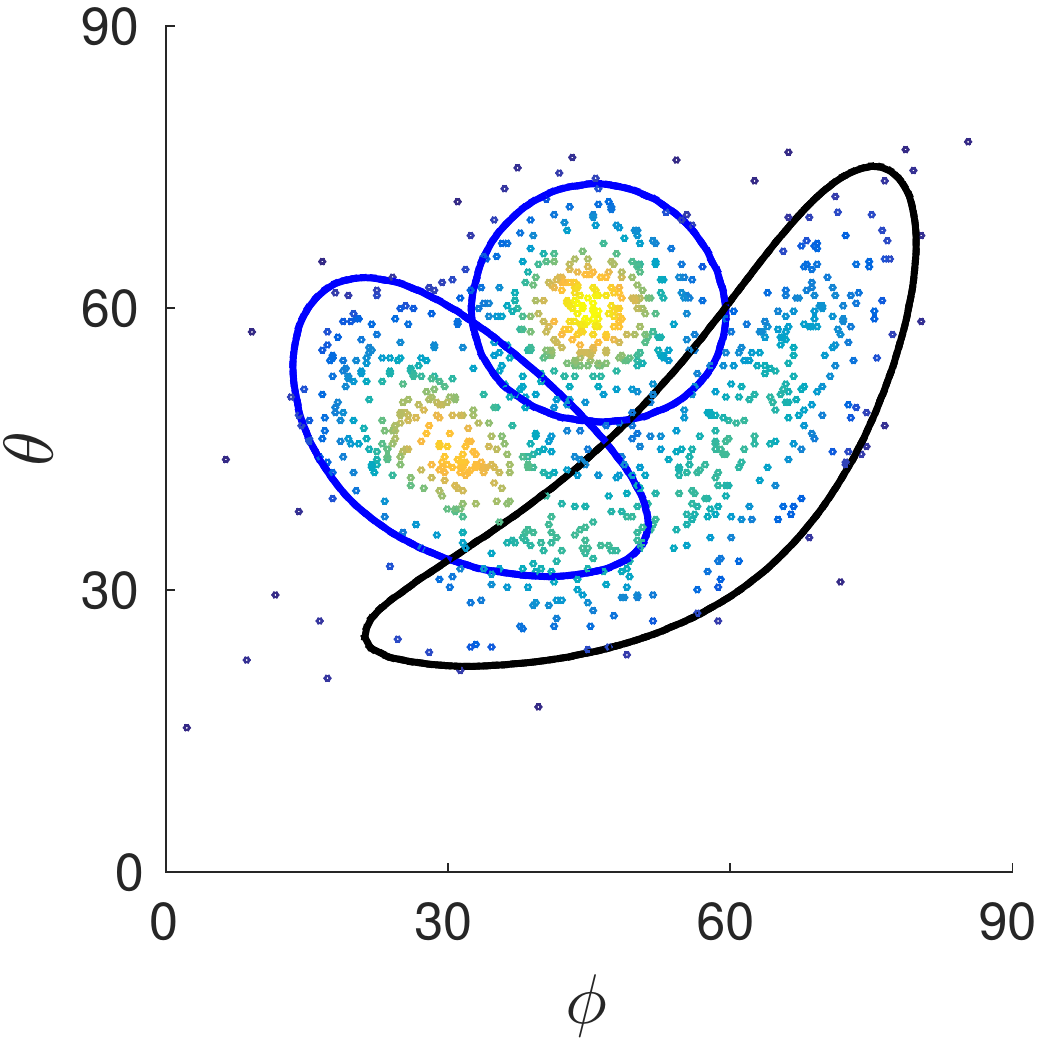}}
\subfloat[Before optimizing ($I=19244$)]{\includegraphics[width=0.33\textwidth]{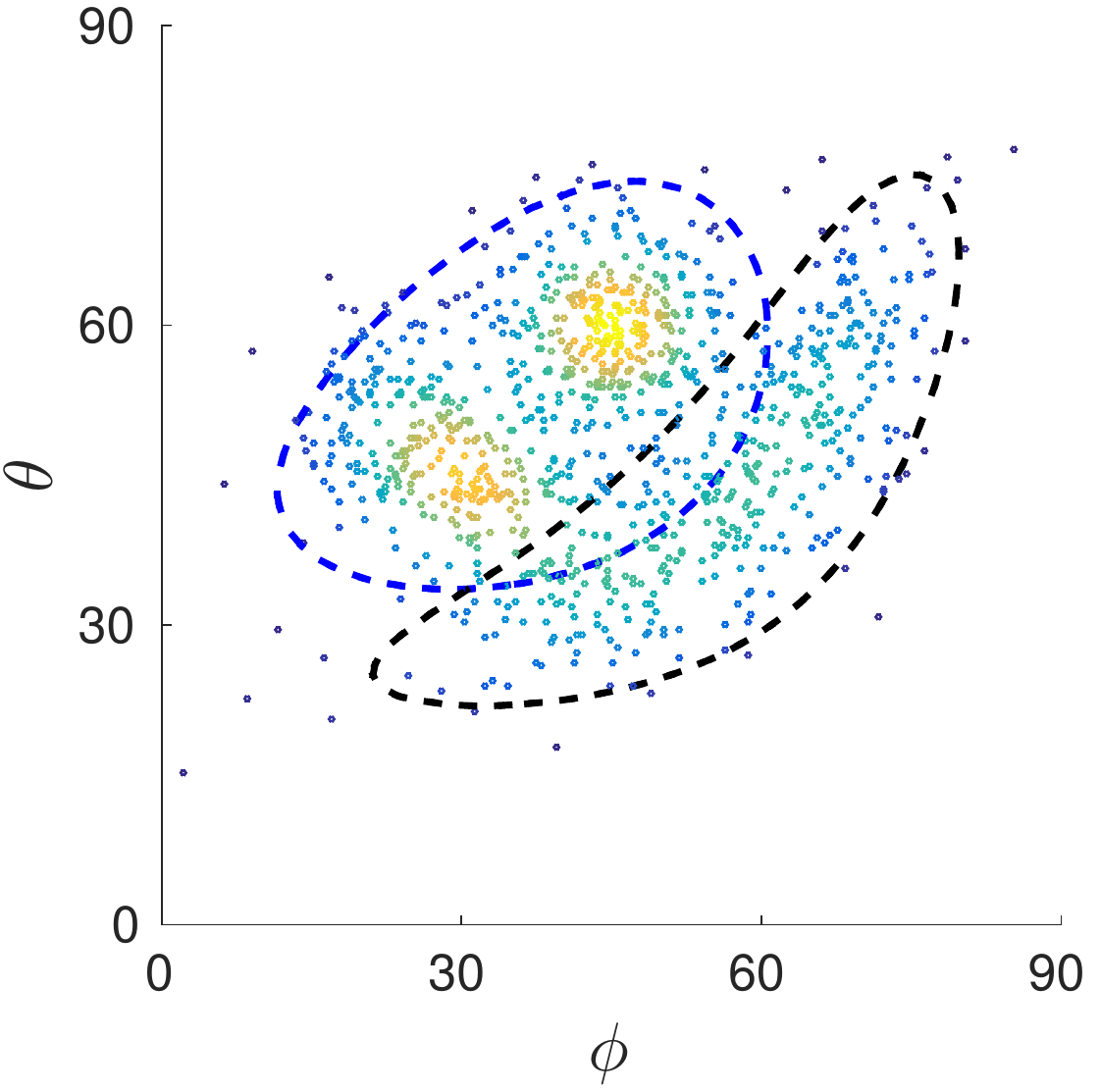}}
\subfloat[post-EM ($I=19236$)]{\includegraphics[width=0.33\textwidth]{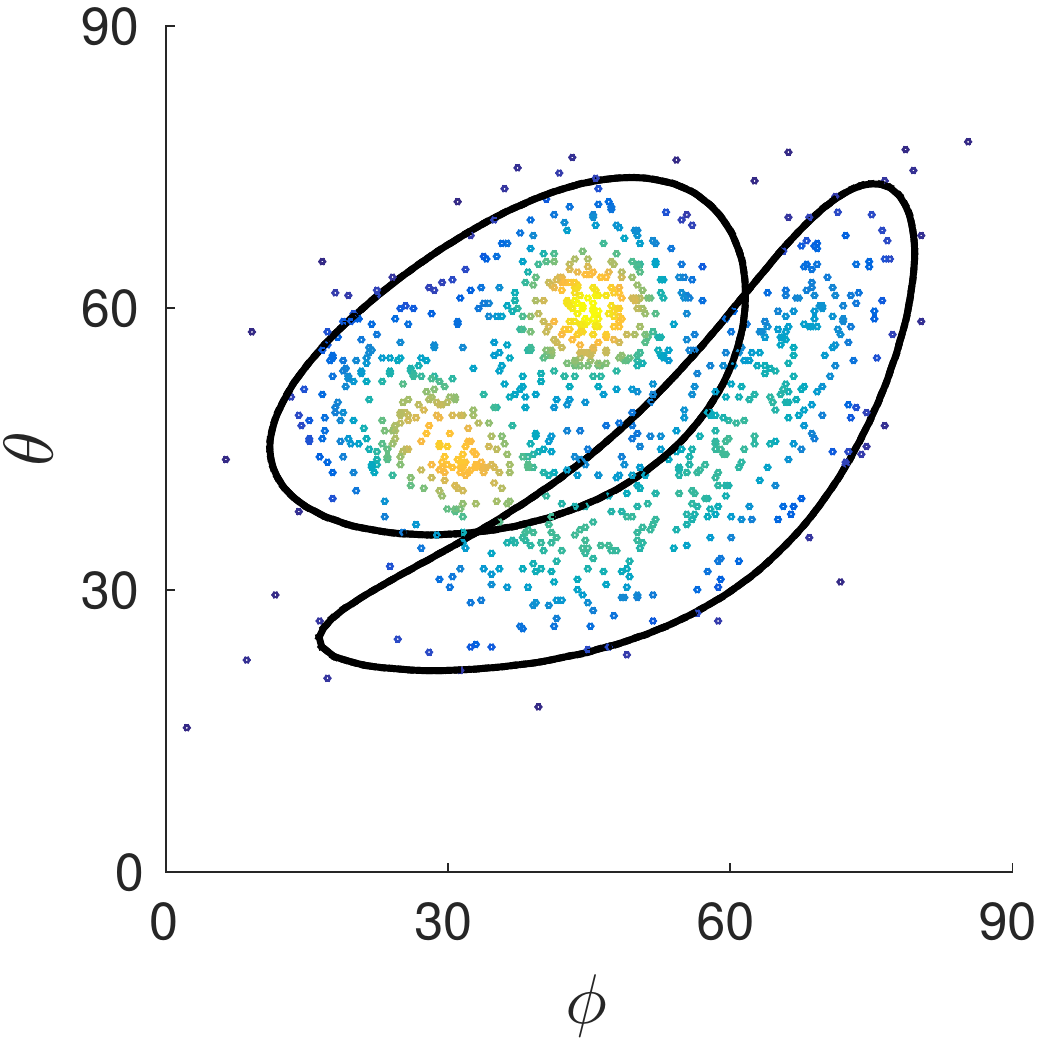}}
\caption{Iteration 3 -- perturbations of $P_3$ 
(a)-(c) splitting,
(d)-(f) deletion,
(g)-(i) merging
} 
\label{fig:mix_iter3_c3}
\end{figure}

The illustrations presented here are continuation of the example 
discussed in Section~\ref{subsubsec:search_method_explained}.
Figure~\ref{fig:mix_iter3_c2} illustrates the perturbations
carried out on the component $P_2$ in the mixture $\fancym_3$.
None of the split, delete, and merge operations involving $P_2$ result in improved
mixtures. The same is the case with component $P_3$ (depicted in
Figure~\ref{fig:mix_iter3_c3}).
It is interesting to note the different mixtures obtained by splitting 
$P_2$ and $P_3$ in Figure~\ref{fig:mix_iter3_c2}(c) and Figure~\ref{fig:mix_iter3_c3}(c)
respectively. These 4-component mixtures are different to the mixture
obtained by splitting $P_1$ (Figure~\ref{fig:mix_iter3_c1}c).
Also, the mixtures resulting from deleting and merging of $P_2$ and $P_3$
are different when compared to the mixtures obtained by the same operations on $P_1$.
This example demonstrates how the search method evaluates various
competing mixtures and selects the one which has the least overall
message length.

\end{document}